\documentclass[A4, 12pt]{article}
\usepackage[margin=1in]{geometry}

\usepackage{amsmath, amssymb, amsthm, amsbsy, amsfonts,  wrapfig,graphicx, comment, enumitem, relsize, indentfirst, colortbl, float} 
\usepackage[compact]{titlesec}
\usepackage[linesnumbered,ruled,vlined]{algorithm2e}
\usepackage[colorinlistoftodos]{todonotes}
\usepackage[colorlinks=true, allcolors=blue, pdfencoding=auto, psdextra]{hyperref}
\usepackage[square,comma, numbers,sort&compress]{natbib}
\titlespacing\section{0pt}{6pt plus 3pt minus 3pt}{0pt plus 2pt minus 2pt}
\titlespacing\subsection{0pt}{6pt plus 3pt minus 3pt}{0pt plus 2pt minus 2pt}

\newcommand{\dd}{\mathbf{d}}
\newcommand{\D}{\mathbf{D}}
\newcommand{\DD}{\mathcal{D}}

\newcommand{\x}{\mathbf{x}}

\newcommand{\e}{\mathbf{e}}

\newcommand{\V}{\mathbb{V}}

\newcommand{\bb}{\mathbf{b}}
\newcommand{\bs}{\boldsymbol}

\newcommand{\0}{\mathbf{0}}
\newcommand{\diag}{\mbox{diag}}

\allowdisplaybreaks 

\newcommand{\E}{\mathbb{E}}
\newcommand{\R}{\mathcal{R}}

\newcommand{\RN}[1]{\textup{\uppercase\expandafter{\romannumeral#1}}}

\theoremstyle{plain}
\newtheorem{thm}{Theorem}
\newtheorem{defn}{Definition}

\newtheorem{ass}{Assumption}
\newtheorem{lem}[thm]{Lemma}
\newtheorem{cor}[thm]{Corollary}
\newtheorem{pro}[thm]{Proposition}
\newtheorem{cla}[thm]{Claim}

\begin{document}
\title{Noise-Augmented Privacy-Preserving Empirical Risk Minimization with Dual-purpose Regularizer and Privacy Budget Retrieval and Recycling\footnote{The work was supported by NSF award \#1717417.}}
\author{\normalsize{Yinan Li and Fang Liu}\\ 
\normalsize{Applied and Computational Mathematics and Statistics}\\
\normalsize{University of Notre Dame, Notre Dame, Indiana, USA}\\
\normalsize{fliu2@nd.edu}}
\date{}

\maketitle

\begin{abstract}
We propose Noise-Augmented Privacy-Preserving Empirical Risk Minimization (NAPP-ERM) that solves ERM with differential privacy guarantees. Existing privacy-preserving ERM approaches may be subject to over-regularization with the employment of a $l_2$ term to achieve strong convexity on top of the target regularization. NAPP-ERM improves over the current approaches and  mitigates over-regularization by iteratively realizing target regularization through appropriately designed augmented data and delivering strong convexity via a single adaptively weighted dual-purpose $l_2$ regularizer. When the target regularization is for variable selection, we propose a new ``regularizer'' that achieves both privacy and sparsity guarantees simultaneously. Finally, we propose a strategy to retrieve privacy budget when the strong convexity requirement is met, which can be returned to users such that the DP of ERM is guaranteed at a lower privacy cost than originally planned, or be recycled to the ERM optimization procedure to reduce the injected DP noise and improve the utility of  DP-ERM. From an implementation perspective, NAPP-ERM can be achieved by optimizing a non-perturbed object function given noise-augmented data and can thus leverage existing tools for non-private ERM optimization.  We illustrate through extensive experiments the mitigation effect of the over-regularization and private budget retrieval by NAPP-ERM on  variable selection and outcome prediction. 

\vspace{12pt}
\noindent\textbf{keywords}: differential privacy,  dual-purpose, empirical risk minimization, noise augmentation, over-regularization,  privacy  budget retrieval and recycle,  utility analysis
\end{abstract}

\section{Introduction}\label{sec:intro}\vspace{-3pt}
 \subsection{Background}
Empirical risk minimization (ERM) is a principle in statistical learning. Through ERM, we can measure the performance of a family of learning algorithms based on a set of observed training data empirically without knowing the true distribution of the data and derive theoretical bounds on the performance. ERM is routinely applied in a wide range of learning problems such as regression, classification, and clustering. In recent years, with the increasing popularity in privacy-preserving machine learning that satisfies formal privacy guarantees such as differential privacy (DP) \cite{dwork2006our}, the topic of privacy-preserving ERM has also been investigated. Generally speaking, differentially private empirical risk minimization (DP-ERM) can be realized by perturbing the output (estimation or prediction), the objective function (input), or iteratively during  the algorithmic optimization, given  an ERM problem. For output perturbation, randomization mechanisms need to  be applied every time a new output is released; for iterative algorithmic perturbation, each iteration incurs a privacy loss, careful planning and implementation of privacy accounting methods to minimize the overall privacy loss is critical. In this paper, we focus on differentially private  perturbation of objective functions. Once an objective function is perturbed, the subsequent optimization does not incur additional privacy loss and all outputs generated from the optimization are also differentially private.

\vspace{-9pt}\subsection{Related Work}\vspace{-3pt} 
Examples on output perturbation in DP-ERM include classification via logistic regression that satisfies $\epsilon$-DP  \cite{dplogistic} and variable selection in lasso-regularized linear regression that satisfies $(\epsilon, \delta)$-DP and also achieves  near-optimal bounds for the excess risk with weaker assumptions than previous work  \cite{dplasso}. A functional mechanism that perturbs the coefficients of a polynomial representation of the original loss function is proposed in \cite{dpfun} that loosens the requirement on the normalization of predictors just for the purposes of satisfying certain assumptions.
Privacy-preserving ERM with strongly convex regularizers for classification problems  was first examined in \cite{dperm} with $\epsilon$-DP. The framework is subsequently extended to admitting convex regularization in general, variable selection and outcome prediction included,  with $(\epsilon, \delta)$-DP in high-dimensional settings \cite{dpermhd}. 
Privacy-preserving kernelized learning proposed in \cite{dperm} is extended to more general RKHS settings with a dimensionless bound for the excess risk of kernel functions in  \cite{dpkernel}. Dimension-independent expected excess risk bounds for $l_2$-regularized generalized linear models (GLMs) are established \cite{dpglmbound}.  The worst-case excess risk bound is further improved \cite{worsterb}.  Another line of work for privacy-preserving machine learning in general, ERM included, is through the iterative algorithmic perturbation  \cite{gradient2010,gradient2015,dplassobound,abadi2016deep, gradient2017AAAI,gradient2017NIPS,privacyamp,adaptdp,dpermbound, bu2020deep}. A fast stochastic gradient descent algorithm is developed in \cite{dpermbound} that improves the asymptotic excess risk bound with Lipschitz loss functions and bounded optimization domain.  In addition to the theoretical and algorithmic development, DP-ERM has been applied to online learning \cite{dponline, dponline1} and GWAS databases in the setting of elastic-net regularized logistic regression  \cite{dplogisticen}, among others.

\vspace{-6pt}\subsection{Our Contributions}\vspace{-3pt}
Despite the extensive research on objective function perturbation with DP in ERM, there still exists room for improvement. First, to ensure DP, the current framework requires strong convexity for a perturbed objective function, which is often achieved by including an extra $l_2$ term on parameters in addition to the target regularization term and the DP term. This may lead to \emph{over-regularization}, deviating parameter estimates further away from those obtained in the non-private setting on top of the deviation due to the DP term. The extra $l_2$ term may also lead to \emph{over-protection} for privacy as it introduces additional perturbation to the objective function on top of the DP term; in other words, the actual privacy cost can be smaller than the privacy budget pre-set by the DP term.  Second, when the DP noise term is large, it would dwarf the target regularizer, especially when the sample size is relatively small, and outputs from the DP-ERM would deviate significantly from those obtained in the non-private setting. 

We aim at overcoming the above limitations of the current DP-ERM framework. Toward that end, we propose a dual-purpose regularizer that realizes the target regularization (differentiable or not) and strong-convexity simultaneously through a single iterative adaptively weighted  $l_2$  regularizer. We name the procedure \emph{Noise Augmented  Privacy-Preserving ERM (NAPP-ERM)}. NAPP-ERM can be achieved through optimization of a non-perturbed objective function but constructed with noise-augmented observed data. When the aim of ERM is variable selection, we propose a new noise term to perturb the objective function to achieve privacy guarantees and sparsity in parameter estimation simultaneously. Finally, we propose a strategy to retrieve privacy budget when the strong convexity requirement is met. The retrieved privacy budget can be returned to users such that the DP of ERM is guaranteed at a lower privacy cost that originally planned, or it can be recycled to the ERM optimization procedure so to reduce the injected DP noise and improve the utility of  DP-ERM at the pre-set privacy cost. From an implementation perspective, since NAPP-ERM can be achieved by optimizing a non-perturbed object function given noise-augmented data, we can leverage existing software or tools for non-private ERM optimization.


\vspace{-3pt}\section{Preliminaries}\label{sec:prelim}
\begin{defn}[\textbf{$\epsilon$-DP} \cite{dpdef1}]\label{defn:dp} A randomized mechanism $\mathcal{R}$ satisfies $\epsilon$-differential privacy if for all data sets $\D,\D'$ differing by one entry and all result subsets $S$ to query $q$, $\Pr[\mathcal{R}( q,\D)\in S]\le e^{\epsilon}\Pr[\mathcal{R}(q,\D')\in S]$.
\end{defn}\vspace{-3pt}
$\epsilon$  is the pre-specified privacy budget or loss (parameter). The smaller $\epsilon$ is, the more privacy protection is imposed on the individuals in the data, in the sense that the probability of getting the same sanitized query results via $\mathcal{R}$ for $\D$ and $\D'$ is higher. The concept of DP is robust as it does not impose any assumptions about the behaviors or the background knowledge of data intruders. Besides the pure $\epsilon$-DP in Definition \ref{defn:dp}, there are also relaxed or extended versions of DP, such as $(\epsilon,\delta)$-DP \cite{dwork2006our}, and $(\epsilon,\delta)$-probabilistic DP \cite{probdp},  $(\epsilon,\delta,\gamma)$-random DP \cite{rdp}, and $(\mu, \tau)$-concentrated DP \cite{concentrateddp,cdp2016}, Gaussian DP\cite{dong2019gaussian}. Presented below is $(\epsilon,\delta)$-DP, which is employed in latter sections of this paper along with $\epsilon$-DP.  $\epsilon$-DP is a special case of $(\epsilon,\delta)$-DP with $\delta=0$. Many of the relaxed versions of DP can be converted to $(\epsilon,\delta)$-DP via some deterministic relationships between the privacy parameters in the former and $\epsilon$ in the latter given a $\delta>0$.
\begin{defn}[\textbf{$(\epsilon,\delta)$-DP} \cite{dwork2006our}]\label{def:eddp}
Let $\D,\D'$ be two data sets that differ by  only one entry. A randomized algorithm $\R$ is $(\epsilon, \delta)$-DP if for $\forall (\D,\D')$ and all result subsets $S$ to query $q$,  $\Pr(\R(q,\D))\in S)\leq e^{\epsilon}\Pr(\R(q,\D'))\in S)+\delta$ holds.
\end{defn}

Denote the observed data by $\D=\{\dd_1,\ldots,\dd_n\}$, where $\dd_i=(\x_i,y_i)$  is an i.i.d sample from the underlying distribution $\DD$ with a fixed domain $\mathcal{T}$, $\x_i$ is a $p$-dimensional feature/predictor vector and $y_i$ is the outcome. We consider the following ERM problem.
\begin{align}\label{eqn:erm}
\hat{\bs\theta}=&\arg\min\limits_{\bs\theta\in\Theta} J(\bs\theta|\D) 
=\arg\min\limits_{\bs\theta\in\Theta} n^{-1}\!\left(\textstyle\sum_{i=1}^{n}l(\bs\theta|\dd_i)+\Lambda R(\bs\theta)\right)\notag\\
=&\arg\min\limits_{\bs\theta\in\Theta} n^{-1}\!\left(l(\bs\theta|\D)+\Lambda R(\bs\theta)\right),
\end{align}
where  $\Theta\subseteq R^p$ is a closed convex set, loss function $l(\bs\theta|\D)$ and regularizer $R(\bs\theta)$ are both convex in $\bs\theta$, and $\Lambda\ge0$ is a tuning parameter. 

The goal of DP-ERM is to conduct privacy-preserving parameter estimation of the optimization problem in Eq (\ref{eqn:erm}). 
The DP-ERM framework with objective function perturbation considered in \cite{dperm} and \cite{dpermhd}  requires 
strong convexity\footnote{A function $f(\bs\theta)$ is $2\Lambda_0$-strongly convex if 
$f(\alpha\bs\theta_1\!+\!(1\!-\!\alpha)\bs\theta_2 )\! \le\!
\alpha f( \bs\theta_1 )\!+\! (1-\alpha)f(\bs\theta_2)-\Lambda_0\alpha(1-\alpha)\|\bs\theta_1 -\bs\theta_2 \|_2^2$ for $\forall\ \alpha\in(0,1)$ and $\bs\theta_1, \bs\theta_2$ in the domain of $f$.} on the objective function $J(\bs\theta|\D)$ 
\begin{align}\label{eqn:dperm}
J^{\text{priv}}(\bs\theta|\D)\!=\!n^{-1}\left(\textstyle l(\bs\theta|\D)
\!+\!\Lambda R(\bs\theta)
\!+\!\bb^T\bs\theta \!+\!\Lambda_0\|\bs\theta\|_2^2\right),  
\end{align}
where $\Lambda_0\geq \zeta_3/\epsilon$, and some regularity conditions as listed below in Assumption \ref{ass}.
\begin{ass}\label{ass} a) $\Theta$ is a closed convex set in an orthant of $R^p$; b) loss function $l(\bs\theta|\D)$ and regularizer $R(\bs\theta)$ are both convex in $\bs\theta$; c) $l(\bs\theta|\D)=\sum_{i=1}^{n}l(\bs\theta|\dd_i)$ has continuous gradient $\nabla l(\bs\theta|\D)$ and Hessian  $\nabla^2 l(\bs\theta|\D)$; d) for $\forall\ \dd_i\in\D$ and $\bs\theta$, $\|\x_i\|_2\leq\zeta_1, \|\nabla l(\bs\theta|\dd_i)\|_2<\zeta_2$, and 
$\max\{\mbox{eigen}( \nabla^2 l(\bs\theta|\dd_i))\}=\|\nabla^2 l(\bs\theta|\dd_i) \|_2<\zeta_3$. \footnote{$\nabla^2 l(\bs\theta|\dd_i)$ is of rank 1 with one sample $\dd_i$; therefore its maximum eigenvalue equals to its its $l_2$ norm.}  
\end{ass}
Compared to Eq (\ref{eqn:erm}),  Eq (\ref{eqn:dperm}) has an additional term $\bb^T\bs\theta$ to ensure DP, where $\bb$ is a noise term drawn from either spherical Laplace or multivariate Gaussian distributions, as well as an additional  $\|\bs\theta\|_2^2$  term to guarantee the strong convexity of $J^{\text{priv}}(\bs\theta|\D)$.  The regularizer $R(\bs\theta)$ in Eq (\ref{eqn:dperm}) can be non-differentiable, such as lasso or elastic net. Under Assumption \ref{ass}, the privacy-preserving parameter estimate $\hat{\bs\theta}^{\text{priv}}$ that minimizes the private loss function in Eq (\ref{eqn:dperm}), that is, $\hat{\bs\theta}^{\text{priv}}\!\!\!=\arg\min\limits_{\bs\theta\in\Theta} J^{\text{priv}}(\bs\theta|\D)$,  satisfies $\epsilon$-DP \cite{dpermhd}. 

As briefly discussed in Sec \ref{sec:intro}, there are several limitations in the existing DP-ERM framework in Eq (\ref{eqn:dperm}). The extra $l_2$ term that brings strong convexity to Eq (\ref{eqn:dperm}) can lead to additional regularization on parameters on top of the target regularizer $R(\bs{\theta})$. We refer to this phenomenon as \emph{over-regularization}.  Second, the additional $l_2$ term might also be associated with \emph{over-protection} from a privacy perspective as it introduces another source of perturbation to the objective function on top of the formal DP term $\bb^T\bs\theta$.  In other words, the actual privacy loss may be smaller than the pre-set privacy budget with Eq (\ref{eqn:dperm}).  Lastly, if $\bb^T\bs\theta$ is large, it would dwarf  $\Lambda R(\bs{\theta})$, especially when the sample size is relatively small. As a result, $\hat{\bs\theta}^{\text{priv}}$ may deviate significantly from its non-private counterpart. In the case of variable selection with sparsity-promoting  $R(\bs{\theta})$, it is possible that no entries in $\hat{\bs\theta}^{\text{priv}}$ are zeros.  On the other hand, we need the $\bb^T\bs\theta$ term to provide formal privacy guarantee, neither can we simply get rid of $\Lambda_0\|\theta\|_2^2$ due to the strong convexity requirement (even when the target regularizer $\Lambda R(\bs\theta)$ itself is strongly convex, such as $l_2$, as $\Lambda$ alone might not be able to meet the required level for strong convexity). 

\vspace{-6pt}\section{Noise-augmented Privacy-Preserving ERM}\label{sec:naperm}\vspace{-3pt}
We propose a new DP-ERM framework,  \emph{Noise-Augmented Privacy-Preserving (NAPP)} ERM, to resolve the above listed issues with the current DP-ERM framework.  We design and generate noisy data and attach them to observed data $\D$ to achieve the target (convex) regularization, strong convexity, and privacy guarantees simultaneously. Since NAPP-ERM uses a single dual-purpose weighted iterative regularizer to achieve the target regularization and delivers the required strong convexity  simultaneously, we can eliminate the need for an ad-hoc term just to bring strong convexity to the objective function and thus mitigate  over-regularization of the current DP-ERM framework. NAPP-ERM still guarantees privacy through the DP term $\bb^T\bs\theta$, but  the magnitude of $\bb$ can be reduced through a privacy budget retrieval and recycling scheme. We also propose a new type of DP term that targets specifically at variable selection with guaranteed sparsity and privacy protection. 

\vspace{-7pt}\subsection{\large Noise Augmentation Scheme}\label{sec:noise}
NAPP-ERM estimates $\bs\theta$ with DP and realizes the target regularization by iteratively solving an unregularized ERM problem with the combined observed data and augmented noisy data. Table \ref{tab:noise} depicts a schematic of how observed data are augmented with noisy data in iteration $t$ of the iterative NAPP-ERM procedure.
\begin{table}[!htp]\vspace{-7pt}
\caption{Noise augmentation in iteration $t$ of NAPP-ERM}\label{tab:noise}\vspace{-13pt}
\begin{center}
\resizebox{0.75\linewidth}{!}{
\begin{tabular}{l|l l l l| l}
\hline
observed& 
\cellcolor{gray!25}{$y_1$}&
\cellcolor{gray!25}{$x_{11}$}&
\cellcolor{gray!25}{$\cdots$}& 
\cellcolor{gray!25}{$x_{1p}$}& 
$e_{ij}^{(t)} = \tilde{e}_{ij}^{(t)} +e_j^*$ for
$i=1,\ldots, n_e$ and \\  
data & \cellcolor{gray!25}{$\vdots$} & \cellcolor{gray!25}{$\vdots$} & \cellcolor{gray!25}{$\cdots$} & \cellcolor{gray!25}{$\vdots$} 
&$j=1,\ldots,p$ is the augmented noise \\
&\cellcolor{gray!25}{$y_n$}&
\cellcolor{gray!25}{$x_{n1}$}& \cellcolor{gray!25}{$\cdots$} &
\cellcolor{gray!25}{$x_{np}$} & in iteration $t$. $\tilde{e}_{ij}^{(t)}$  realizes the target \\
\cline{1-5}
augmented &$e_{y1}$ & $e^{(t)}_{11}$ & $\cdots$ & $e^{(t)}_{1p}$ & regularization  upon convergence \\
noise  &$\vdots$  &$\vdots$ &  $\cdots$ &$\vdots$ & and  $e_j^*$ achieves DP. The value of $e_{y,i}$ \\
&$e_{yn_e}$ & $e_{n_e1}^{(t)}$ & $\cdots$ & $e_{n_ep}^{(t)}$ &    depends on the outcome type$^\dagger$.  \\
\hline
\end{tabular}}
\resizebox{0.75\linewidth}{!}{\begin{tabular}{l}
$^\dagger$ For example, we may set $e_{y,i}\!\equiv\!0$ for $i\!=\!1,\ldots,n$ for linear regression;  \\
\hspace{5pt} $e_{y,i}\!\equiv\!1$  for Poisson regression; $e_{y,i}\!=\!0$ for $i\!=\!1,\ldots,n_e/2$ and $e_{y,i}\!=\!1$ \\
\hspace{5pt}  for $i\!=\!n_e/2+1,\ldots,n_e$ for logistic regression.\\
\hline
\end{tabular}}\vspace{-11pt}
\end{center}\end{table}
The augmented data $e^{(t)}$ in iteration $t$ composes two components:  DP noise $e^*$ and regularization noise $\tilde{e}^{(t)}$. $e^*$ guarantees privacy and $\tilde{e}^{(t)}$ is designed to yield the target regularization upon convergence and changes with iteration.   $e^*$ is defined as 
\begin{align}
\!\!\!&\e_i^*=(e^*_{i1},\ldots,e^*_{ip}) =(n_e l'|_{\bs\eta=0})^{-1}\bb, \mbox{ where } \bb=(b_1,\ldots,b_p), \label{eqn:e*}\\
\!\!\!&\mbox{and }f(\bb)
\begin{cases}
\propto\exp\left( -(\zeta_1\zeta_2)^{-1}(r\epsilon)\|\bb\|_2\right)\mbox{ for $\epsilon$-DP} \label{eqn:b} \\
=\!N(\mathbf{0},2(r\epsilon)^{-2}\zeta_1^2\zeta_2^2\left(r\epsilon
\!-\!\log(\delta)\right) \mathbf{I}_p)\mbox{ for $(\delta,\epsilon)$-DP}\!\!\! 
\end{cases}
\end{align}
and is fixed throughout the iterations. $p$ in Eq (\ref{eqn:e*}) is the dimensionality of the predictor $\x$ and $l'$  is the first derivative of the one-data-point loss function with regard to  $\bs\eta=\x\bs\theta$ (i.e., the linear predictor); $\zeta_1$ and $\zeta_2$  in Eq (\ref{eqn:b}) are defined in Assumption \ref{ass}, $r\in(0,1)$ is the portion of total privacy budget $\epsilon$ associated with $f(\bb)$ directly, which is often set at $1/2$ \cite{dperm,dpermhd} (more detail is provided in Sec \ref{sec:DP} and around Eq (\ref{eqn:ratio})).  $\bb$ in Eq (\ref{eqn:e*})  is divided into $n_e$ portions because the total amount of DP noise is $\bb$ while there are $n_e$ noise terms so each noise term receives $1/n_e$ of $\bb$.  $f(\bb)$ in Eq (\ref{eqn:b}) that yields $\epsilon$-DP is the \emph{spherical Laplace  distribution} with  $\E(b_j)=\mathbf{0}, \V(b_j) =2\zeta_1^2\zeta_2^2(r\epsilon)^{-2}$ for $j=1,\ldots,p$, and $\mbox{Cov}(b_jb_{j'})=0$ for $j\ne j'$.  When $p=1$, it  reduces to the Laplace distribution.  To sample from the spherical Laplace  distribution, we may first sample $\|\bb\|_2$, which follows gamma distribution with shape$=1$ and scale$=2\zeta_1\zeta_2(r\epsilon)^{-1}$, and then draw from the $p$-dimensional sphere, the radius of which is $\|\bb\|_2$ from the first  step.

The variance of the target regularization noise $\tilde e_{ij}^{(t)}$ for $j\!=\!1,\ldots,p$ is adaptive to the $\hat{\bs\theta}^{*(t-1)}$ estimate during iterations and
\begin{equation}\label{eqn:tildee}
\tilde e_{ij}^{(t)}
\begin{cases}
\sim N(0,\V(\bs\theta^{(t-1)},\Lambda)) &\mbox{ for } i=1,\ldots,n_e/2\\
=- \tilde\e_{i-n_e/2}^{(t)} &\mbox{ for } i=n_e/2+1,\ldots,n_e
\end{cases},
\end{equation}
where $\Lambda$ contains tuning parameters. $\mbox{V}(\bs\theta,\Lambda)$ is designed in such a way so that $\tilde \e_{i}^{(t)}=(\tilde e_{i1},\ldots,\tilde e_{ij})$ will result in the target regularization while providing the required strong convexity quantified by $\Lambda_0$. By design, the sum  $\tilde{\e}_{.j}$ across $i=1,\ldots,n_e$ is 0 for each $j$ so that the linear term of the Taylor expansion of the noise-augmented loss function is used to realize DP and its quadratic term realizes the target regularization; more detail is provided in Sec \ref{sec:nap}. If $n_e$ is very large, the sum of $\tilde{\e}$ over $i$ would be very close to 0 anyway, though not exactly 0 guaranteed by Eq (\ref{eqn:tildee}).

\vspace{-6pt}\subsection{\large Noise Augmented Private ERM}\label{sec:nap}\vspace{-3pt}
Assume the loss function of the original ERM problem in Eq (\ref{eqn:erm}) has non-zero, finite, and continuous gradient and Hessian in the neighborhood of $\bs\eta=\x\bs\theta=0$. \footnote{The regularity condition $\bs\eta=0$ is  satisfied in some common ERM problems such as $l_2$ loss, GLMs, and SVMs with the smoothed Huber loss (refer to Sec \ref{sec:discuss} for more discussion).} The optimization problem in iteration $t$ of the NAPP-ERM procedure is
\begin{align}
\hat{\bs\theta}^{*(t)}
&=\arg\min\limits_{\bs\theta\in\Theta} J^{(t)\mbox{priv}}_p(\bs\theta|\D,\e^{(t)})
=\arg\min\limits_{\bs\theta\in\Theta} n^{-1}\!\textstyle\left(\sum_{i=1}^{n}l(\bs\theta|\dd_i)+\sum_{i=1}^{n_e}l(\bs\theta|\e^{(t)}_i)\right),\label{eqn:perm}
\end{align}
where $\e_i$ for $i=1,\ldots,n_e$ is the augmented noise in Table \ref{tab:noise}.
\begin{pro}\label{prop:regularization}
The optimization problem in NAPP-ERM in Eq (\ref{eqn:perm}) is second-order equivalent to
\begin{align}\label{eqn:napp}
\hat{\bs\theta}^{*(t)}
\!=&\textstyle \arg\min\limits_{\bs\theta\in\Theta} n^{-1}\!\left(\sum_{i=1}^{n}l(\bs\theta|\dd_i)\!+\!\sum_{j=1}^{p}b_j\theta_j + R^{(t)}(\bs\theta)\right) \mbox{ as $n_e\!\rightarrow\!\infty$}, 
\end{align}
where $R^{(t)}(\bs\theta)\!=\!2^{-1}n_el''|_{\bs\eta=\0}\sum_{j=1}^p\!\V\left(\tilde{e}_{ij}^{(t)}\right)\!\theta_j^2$. 
\end{pro}
The proof is provided in the supplementary materials.  $R^{(t)}(\bs\theta)$ is the  regularizer in iteration $t$ that realizes strong convexity and helps achieve the target regularization upon convergence if the estimation of $\bs\theta$ is not over-regularized. For different target regularizers, $\mbox{V}(\bs\theta,\Lambda)$ takes different forms. Some examples on $\mbox{V}(\bs\theta,\Lambda)$ are given in Table \ref{tab:noisereg}.  In all three examples, the variance at $t=1$ is independent of $\bs\theta$ , meaning  that NAPP-ERM always starts with $l_2$ regularization to obtain initial parameter estimates and set the stage for noise generation in subsequent iterations. On the other hand, if the target regularization $R(\bs{\theta})$ is ridge, it can be realized at $t=1$, along with strong convexity, by choosing the maximum of $(\Lambda,\Lambda_0)$ as the tuning parameter as long as $n_e$ is large.  For bridge and elastic net regularization, $\mbox{V}(\bs\theta,\Lambda)$ depends on the most updated $\theta$ estimate after $t>1$. 
\vspace{-6pt}
\begin{table}[!htp]
\caption{Some examples of $\mbox{V}(\theta_j,\Lambda)$ in Eq (\ref{eqn:tildee})}\label{tab:noisereg}\vspace{-15pt}
\begin{center}
\resizebox{0.9\columnwidth}{!}{
\begin{tabular}{@{}lll@{}}
\hline
regularization &$t=1$& $t\geq2$  \\
\hline
ridge&
$2(n_el''|_{\bs\eta=\0})^{-1}\!\max\!\left\{\Lambda,\Lambda_0 \right\}$ & $2(n_el''|_{\bs\eta=\0})^{-1}\!\max\!\left\{\Lambda,\Lambda_0 \right\}$\\	
bridge $l_{2-\gamma}$ &
$2(n_el''|_{\bs\eta=\0})^{-1}\Lambda_0 $ & 
$2(n_el''|_{\bs\eta=\0})^{-1}\!\max\! \left\{\!\Lambda|\hat{\theta}_j^{(t-1)}|^{-\gamma},\Lambda_0\right\}$  \\
elastic net  &
$2(n_el''|_{\bs\eta=\0})^{-1}\!\max\!\left\{\!\Lambda\kappa,\Lambda_0 \right\}$ & 
$2(n_el''|_{\bs\eta=\0})^{-1}\!\max\! \left\{\!\Lambda|\hat{\theta}_j^{(t-1)}|^{-1}\!+\!\Lambda\kappa,\Lambda_0\right\}$  \\
\hline
\end{tabular}}
\resizebox{0.9\columnwidth}{!}{
\begin{tabular}{l}
$\gamma\in[0,2); \kappa\in(0,1)$\\
$l''$ is the 2nd-order derivative of then one-data-point loss function with respect to  $\bs\eta\!=\!\bs\theta^T\x$.\\
\hline
\end{tabular}}
\vspace{-24pt}\end{center}
\end{table}
\begin{cla}\label{cla:napp4dp}
The NAPP-ERM procedure can be applied to solve the existing DP-ERM problem in Eq (\ref{eqn:dperm}). Rather than using the maximum of  $\Lambda_0$ and the term that involves $\Lambda$ as shown in Table \ref{tab:noisereg}, the sum of the two are used except when $R(\bs\theta)$ is ridge. Specifically, for ridge,  $\mbox{V}(\theta_j,\Lambda)=2(n_el''|_{\bs\eta=\0})^{-1}\!\max\{\Lambda,\Lambda_0\}$ as in  Table \ref{tab:noisereg}; for bridge and elastic net regularizations,
$\mbox{V}(\theta_j,\Lambda)=2(n_el''|_{\bs\eta=\0})^{-1}\Lambda_0$ and $2(n_el''|_{\bs\eta=\0})^{-1}(\Lambda\kappa\!+\!\Lambda_0)$ at $t=1$; $2(n_el''|_{\bs\eta=\0})^{-1}\! \left(\!\Lambda|\hat{\theta}_j^{*(t-1)}|^{-\gamma}\!+\!\Lambda_0\!\right)$ and $2(n_el''|_{\bs\eta=\0})^{-1}\! \left(\!\Lambda|\hat{\theta}_j^{*(t-1)}|^{-1}\!+\!\Lambda\kappa\!+\!\Lambda_0\!\right)$ for $t\ge2$, respectively  .  
\end{cla}

\vspace{-12pt}
\subsection{\large Dual-purpose Regularization and Mitigation of Over-regularization (MOOR) through Iterative Weighted $l_2$} \label{sec:MOOR}
\begin{figure}[!htb]
\vspace{-12pt}\centering
\includegraphics[width=0.32\linewidth]{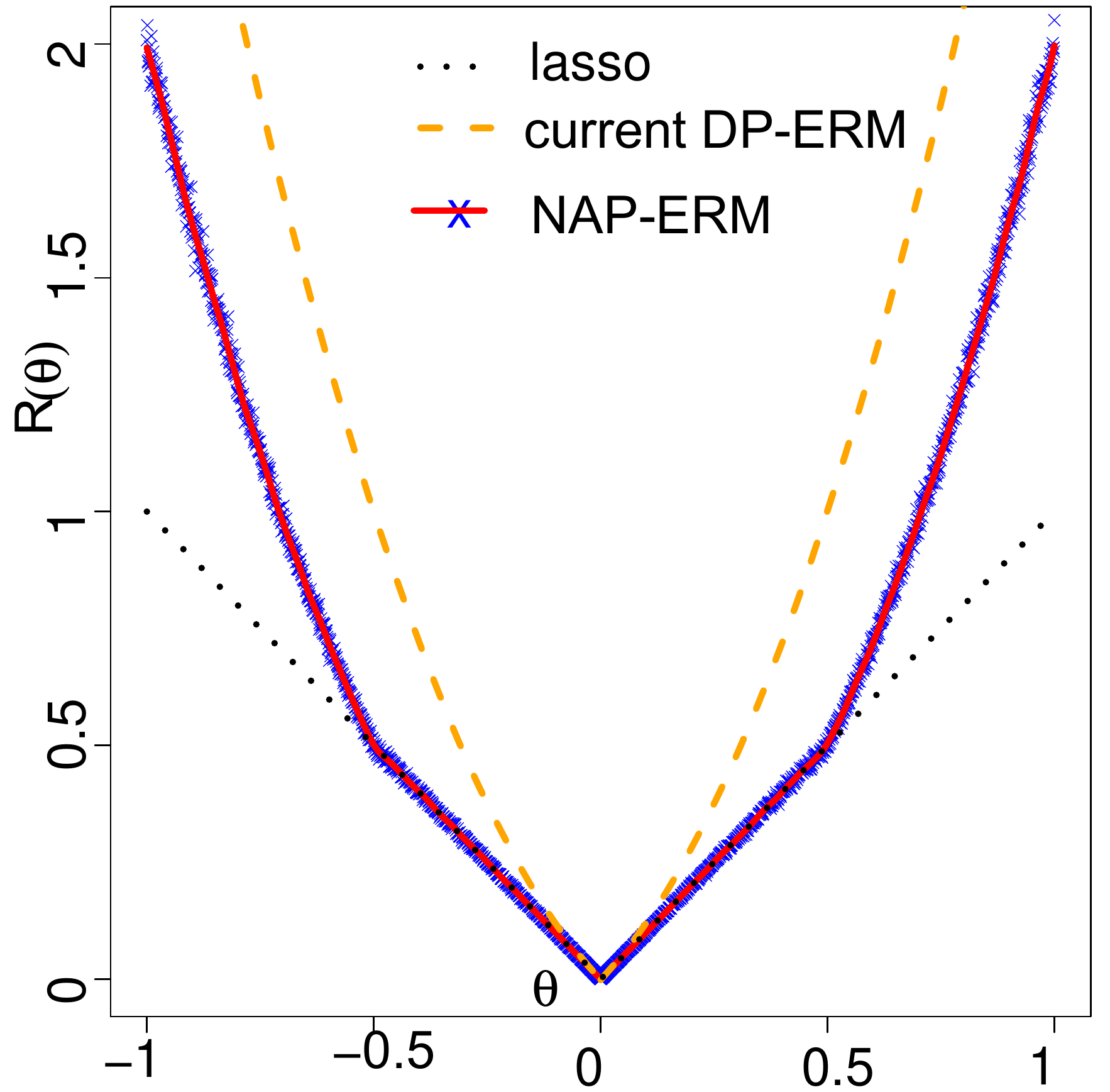}
\includegraphics[width=0.32\linewidth]{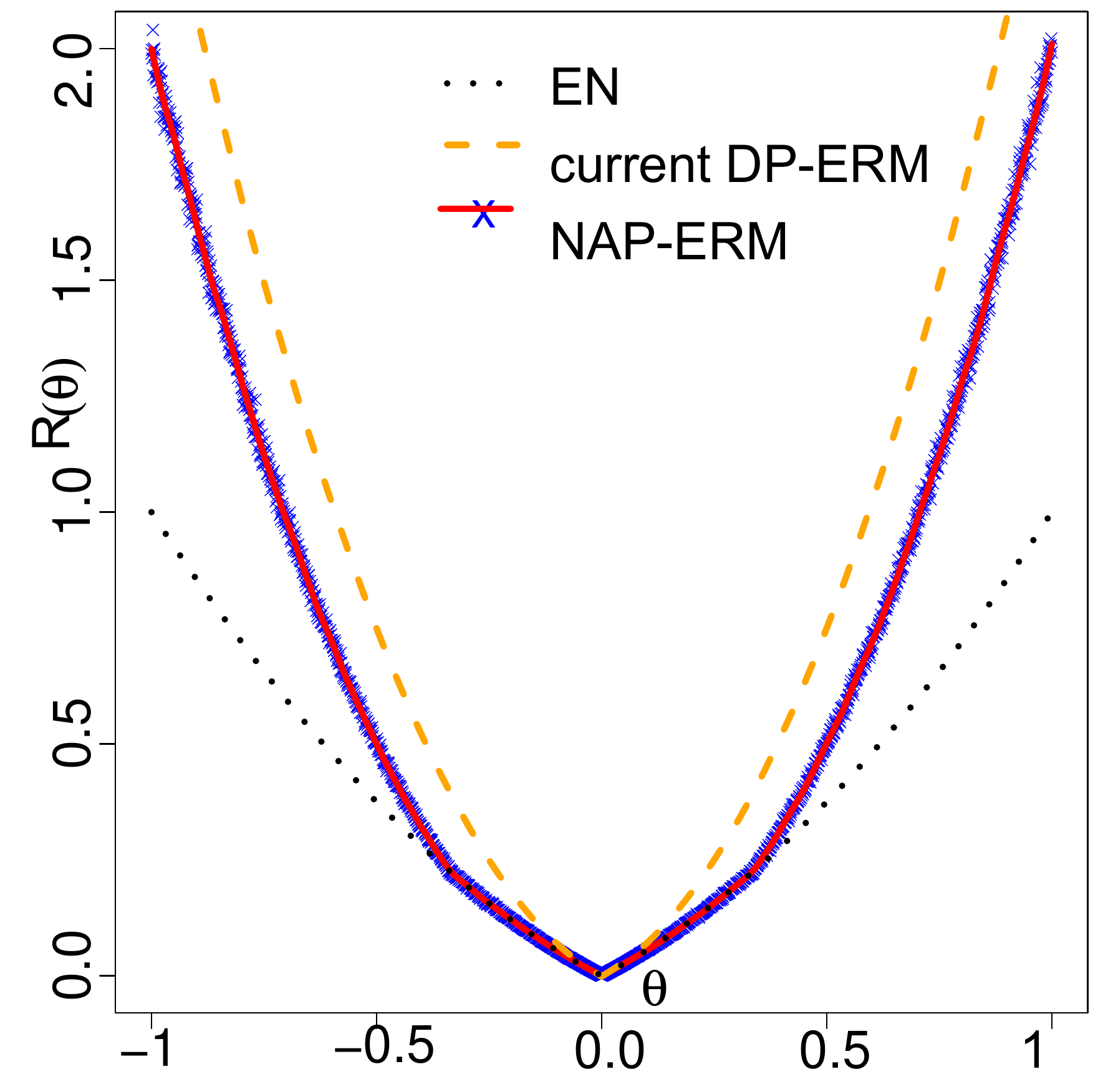}
\includegraphics[width=0.32\linewidth]{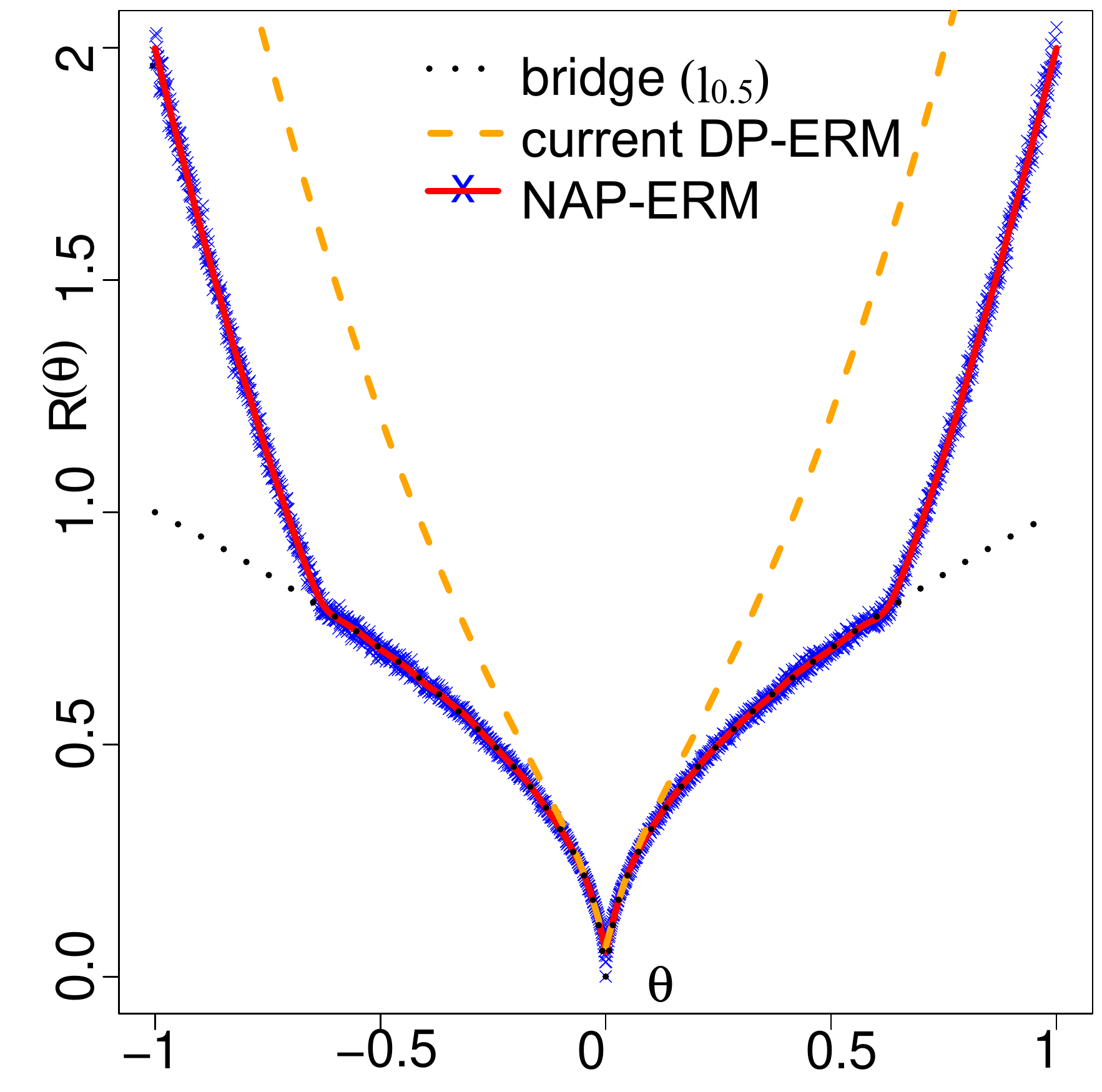}\\
\includegraphics[width=0.32\linewidth]{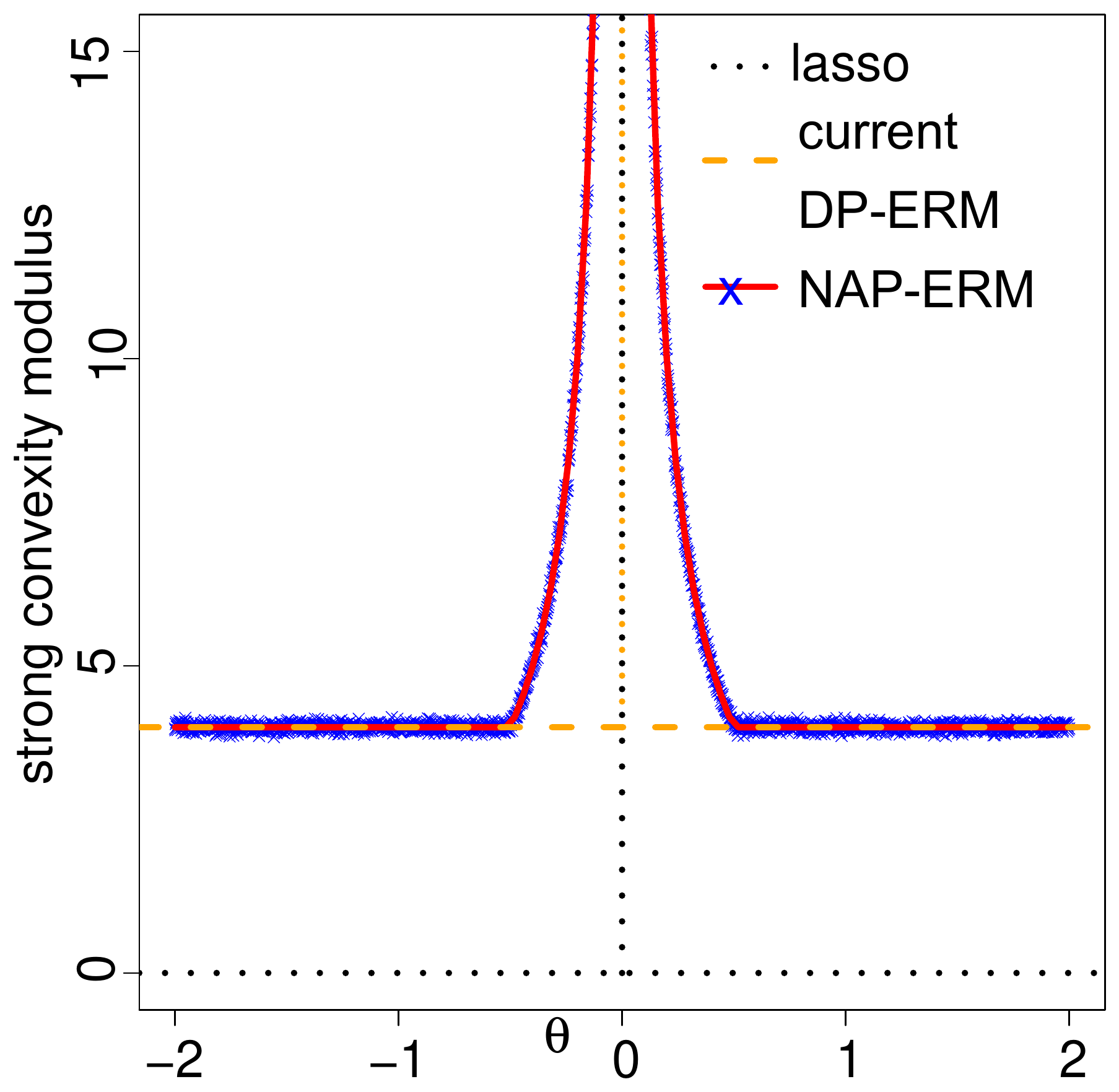}
\includegraphics[width=0.32\linewidth]{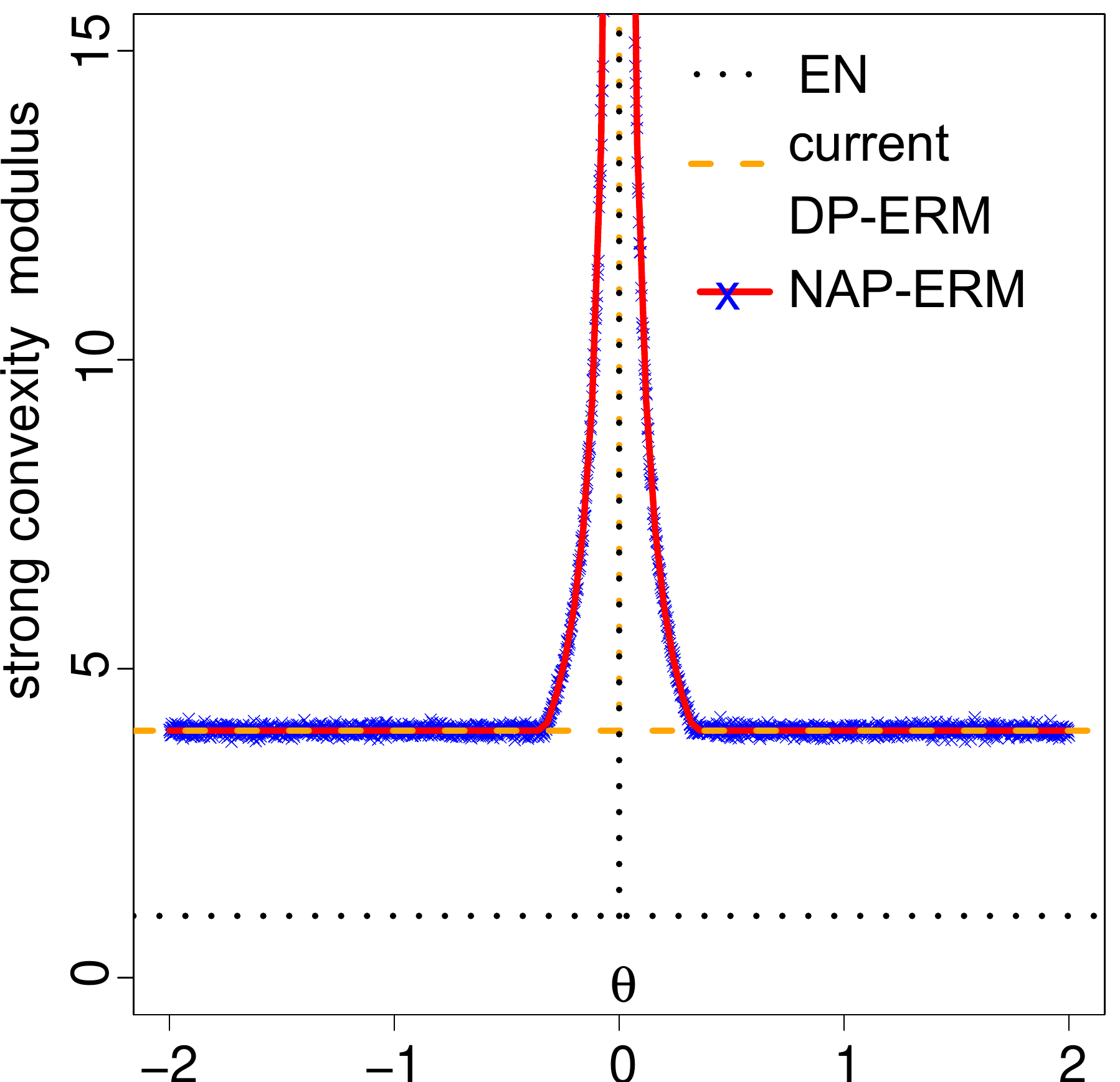}
\includegraphics[width=0.32\linewidth]{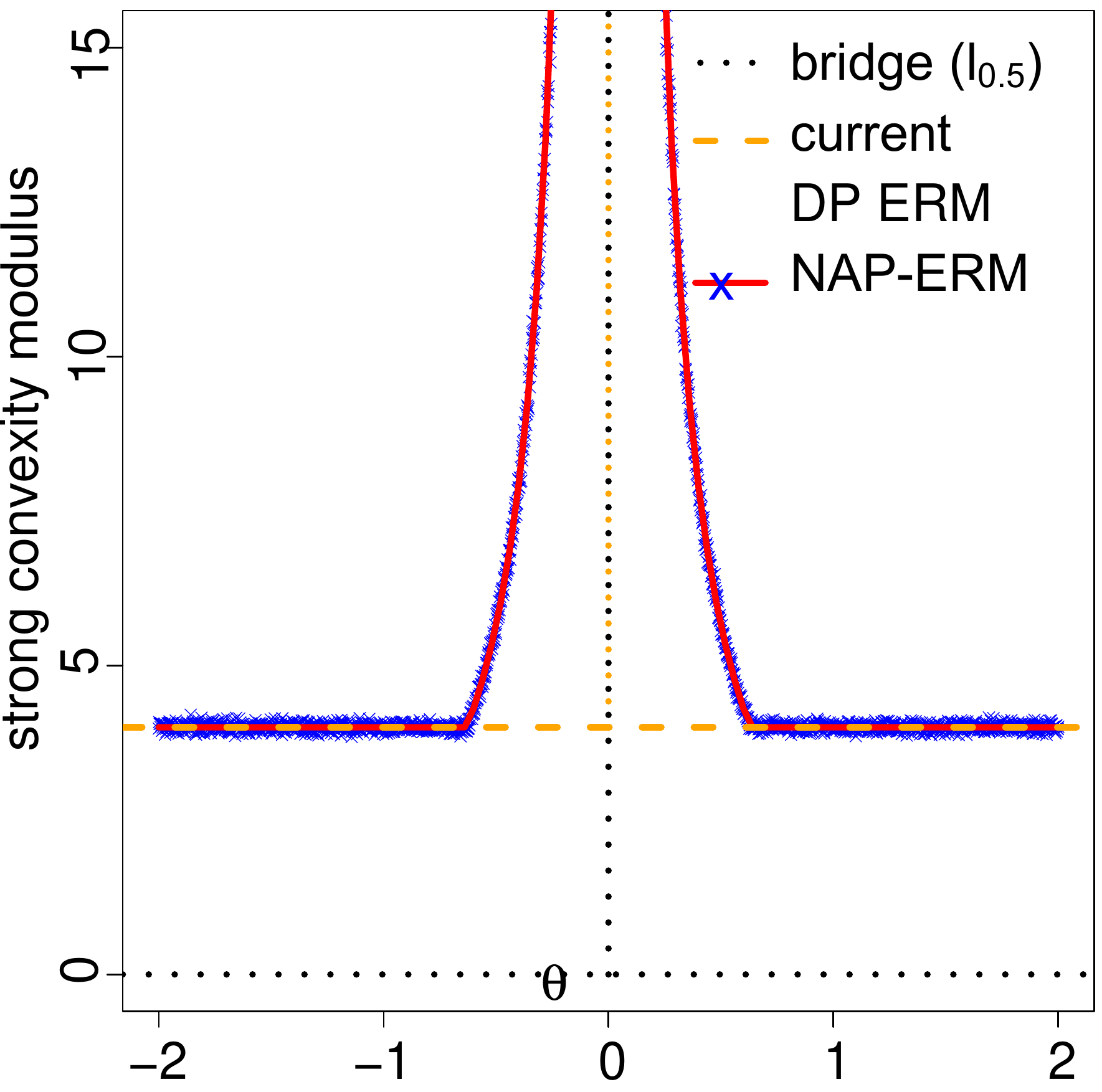}
\vspace{-6pt}\caption{Realized regularizer  $R(\theta)$ (top) and the corresponding modulus of strong convexity (bottom) when the target regularizer (dotted black lines) is lasso, elastic net (EN) and $l_{0.5}$, respectively.  Solid red and dashed orange lines are the analytical realized  regularization and modulus for NAPP-ERM and existing DP-ERM frameworks, respectively; blue crosses represent the empirically realized regularization and modulus at $n_e=10^4$ through NAPP-ERM.}  \label{fig:regeffect}\vspace{-12pt}
\end{figure}

Proposition \ref{prop:regularization} suggests that the  regularization in each iteration is a weighted $l_2$ regularization with weight $2^{-1}n_e l''|_{\bs\eta=\0}\mbox{V}\!\left(\tilde{e}_{j}^{(t)}\right)$ for $\theta^2_j$. If the weight is large enough to also achieve $2\Lambda_0$ strong convexity, there will be no need for an additional ad-hoc  $l_2$ term for strong convexity guarantees as adopted by the current DP-ERM practice (Eq (\ref{eqn:dperm})). Therefore, the NAPP-ERM framework offers an opportunity to mitigate over-regularization in the current DP-ERM framework and improve the utility of the private estimates of $\bs{\theta}$.
\begin{pro}[\textbf{Dual-purpose Regularization}]\label{cla:dual}
NAPP-ERM guarantees strongly convexity with modulus $n_e l''|_{\bs\eta=\0}\min\limits_{j=1,\ldots, p}\!\!\! \mbox{V}\!\left(\tilde{e}_{ij}^{(t)}\right)$ for the DP-ERM problem in each iteration $t$, while achieving the target regularization upon convergence.
\end{pro}
The proof of Proposition \ref{cla:dual} is straightforward per the formulation of $V(\tilde{e})$ which uses the larger modulus of the two $l_2$ terms ($\Lambda_0$ and the one leads to the target regularization upon convergence). Fig \ref{fig:regeffect} illustrates the strong convexity guarantees and the realized  regularization through NAPP-ERM and the existing DP-ERM framework when the target  regularizer is lasso, elastic net, and $l_{0.5}$ (none of which is strongly convex), respectively. Both the realized regularizations of NAPP-ERM and the existing DP-ERM  deviate from their targets, but the former approximates the targets significantly better than the latter, especially when $\theta$ is in the neighborhood of 0.

\subsection{Computational Algorithm}\label{sec:algorithm}\vspace{-3pt}
The algorithmic steps for solving the NAPP-ERM problem is given in Algorithm \ref{alg:NAPP-ERM}. The algorithm starts with data augmented with noises generated from Gaussian distribution the variance of which is independent of $\bs\theta$ (see Table \ref{tab:noisereg} for examples); therefore,  users do not need to supply an initial value for $\bs\theta$. 

\begin{algorithm}[!htb]
\caption{NAPP-ERM}\label{alg:NAPP-ERM}
\SetAlgoLined
\SetKwInOut{Input}{input}
\SetKwInOut{Output}{output}
\Input{ Observed data $\D=\{\dd_1,\ldots,\dd_n\}$ and loss function $l(\bs\theta|\D)$ that satisfy Assumption \ref{ass}; number of iterations $T$; size of injected noisy data $n_e$, overall privacy budget $\epsilon$ (and $\delta$ if $(\epsilon,\delta)$-DP); portion $r\in(0,1)$ of the overall budget $\epsilon$ allocated to bounding ratio $r_1$ in Eq (\ref{eqn:ratio}); tuning parameter $\Lambda$ for the target regularization.}
\Output{Private estimate $\hat{\bs\theta}^*$
Set $\Lambda_0\ge\zeta_3/(2(1-r)\epsilon)$}
Draw  $\bb$ per Eq (\ref{eqn:b}) and  calculate $\e_i^*$ in  Eq (\ref{eqn:e*}) for $i\!=\!1,\ldots,n_e$\;
$t\leftarrow1$ and convergence $\leftarrow0$\;
\While{convergence $=0$ \textbf{and} $t<T$}{
Draw noises $\tilde{\e}_i^{(t)}$ per Eq (\ref{eqn:tildee}) and obtain $\e_i^{(t)}=\tilde{\e}_i^{(t)}+\e_i^*$  for $i=1,\ldots,n_e$\;
Augment $\D$ with $\e_{1:n_e}\!$ and solve $\hat{\bs\theta}^{(t)}\!\!=\!\arg\min\limits_{\bs\theta\in\Theta} J_p^{(t)}(\bs\theta|\D,\e^{(t)})$ in Eq (\ref{eqn:perm})\;
 convergence $\leftarrow1$ if the algorithm converges\;
$t\leftarrow t+1$}
\end{algorithm}

Algorithm \ref{alg:NAPP-ERM} may stop based on several criteria, similar to other noise augmentation approaches \cite{pandaGLM}. For example, we may eyeball the trace plot of $l(\hat\theta^{(t)}|\D,\e^{(t)})$ over a period of iterations to see whether it has stabilized, or calculate the percentage change in $l(\hat\theta^{(t)}|\D,\e^{(t)})$ between two consecutive iterations and see if it is below a pre-specified threshold. 

Regarding the choice of $n_e$ in Algorithm \ref{alg:NAPP-ERM}, since NAPP-ERM realizes the designated regularization and guarantees DP based on the second order Taylor expansion and the higher-order terms are on the order of $O(n_e^{-1/2})$, better approximation will be achieved if $n_e$ is set at a large value. In the experiments in Sec \ref{sec:examples}, we used $n_e=10^4$. Regarding the specification of  $\Lambda_0$, it needs to be $\ge \zeta_3/(2(1-r)\epsilon)$ to guarantee the required strong convexity and bound the Jacobian  ratio $r_2$ in Eq (\ref{eqn:ratio}) for DP guarantees (Sec \ref{sec:DP}). $1-r$ is the assigned portion out of the total budget $\epsilon$ to bound $r_2$ and $r=1/2$ is used in \cite{dperm,dpermhd}.  While other values of $r$ can be specified, we recommend  $r=1/2$ as $\Lambda_0=\zeta_3/\epsilon$ might offer the best trade-off between the amount of DP noise and guarantees of strong convexity. More discussion is provided in Sec \ref{sec:DP} after the establishment of DP. Regarding the specification of $\bf{\Lambda}$ -- a user-specified hyperparameter for the target regularization,   we recommend the two strategies given in \cite{dperm}, both of which apply to the NAPP-ERM algorithm.  

As for solving $\hat{\bs{\theta}}$ from Eq (\ref{eqn:perm}) during the iterations on the augmented data $(\D,\e^{(t)})$,  it is basically just solving an unregularized and unperturbed objective function $l$ given  $(\D,\e^{(t)})$ and there are many existing tools and software for minimizing $l$ in various types of regression. For example, for GLMs, the loss function is the negative log-likelihood. With  a large $n_e$ so that $n+n_e>p$, and the augmented data $(\D,\e)$ can be fed to any software that can run a regular GLM (e.g. the \texttt{glm} function in R; \texttt{tfp.glm.ExponentialFamily} in TensorFlow). This is the same idea as the PANDA technique \cite{pandaGLM} for regularzing undirected graphic models.

\vspace{-3pt}\subsection{NAPP-ERM for Variable Selection}\label{sec:NAPP-ERM-vs}
Though the current DP-ERM formulation in Eq (\ref{eqn:dperm}) can accommodate variable selection by employing a sparsity regularizer $R(\bs\theta)$, the DP noise term $\bb^T\bs\theta$ can trump the variable selection goal, resulting in non-sparsity, especially when $\epsilon$ or $n$  is relatively small. To improve on the current approach for differentially private variable selection for ERM problems in general, we propose a new type of DP noise term that guarantees to lead to some level of sparsity.

\begin{equation}\label{eqn:dpermvs}
\hat{\bs\theta}^{\text{priv}}=\arg\min\limits_{\bs\theta\in\Theta} n^{-1}\left(l(\bs\theta|\D)+\Lambda R(\bs\theta)+\Lambda_0\|\bs\theta\|_2^2+\bb^T|\bs\theta|\right), 
\end{equation}

Compared to the DP-ERM problem in Eq (\ref{eqn:dperm}), the DP term in Eq (\ref{eqn:dpermvs}) is formulated as $\bb^T|\bs\theta|$ instead of $\bb^T\bs\theta$, which can be regarded as a \emph{random} version of the lasso where the ``turning parameter'' $b_j$ is randomly sampled and differs by $|\theta_j|$. To reduce the likelihood of sampling negative $b_j$, rather than sampling directly from the spherical Laplace or Gaussian distribution from Eq (\ref{eqn:b}), the left truncated spherical Laplace distribution or the left Gaussian distribution can be used, defined below.
\begin{align}
\mbox{$\epsilon$-DP: } & f(\bb) \propto I(\bb>c)\exp\left( -(\zeta_1\zeta_2)^{-1}(r\epsilon)\|\bb\|_2\right)\label{eqn:LaplaceT}\\
\!\!\!\mbox{$(\epsilon,\delta)$-DP: } &
f(\bb)\!\sim\! I(\bb\!>\!c)\frac{N(0,2(r\epsilon)^{-2}\zeta_1^2\zeta_2^2\left(-\log(\delta)\!+\!r\epsilon \right)\mathbf{I}_p}{
\Pr(b_j>c\;\forall j=1,\ldots,p)},\!\label{eqn:GaussianT}
\end{align}
where $I(\bb>c)$ is a indicator function that $b_j\!>\!c\;\forall j\!=\!1,\ldots,p$. The truncation point $c\le0$ can be set the same for all $j=1,\ldots,p$, especially after taking into account the privacy considerations. When $c=0$, referred to as \emph{NAPP-VS+} hereafter, the distribution of $\bb$ becomes the half spherical or half Gaussian distribution, and the DP term $\bb^T|\bs\theta|$ in Eq (\ref{eqn:dpermvs}) becomes a ``weighted'' lasso term, which will always lead to sparsity in $\hat{\bs\theta}^{\text{priv}}$. 
When $c<0$, a subset of $\theta$ will be subject to the weighted lasso regularization, but the formation of this subset is completely random. 

The data augmentation scheme for NAPP-VS is the same as that for the general NAPP-ERM and the computational steps  are the same as Algorithm \ref{alg:NAPP-ERM} except for the adjustment for the sign of DP noise $\e^*$ in each iteration of $t\!\ge\!2$; i.e., $e_j^{*}\!\leftarrow\!e_j^*\mbox{sgn}\big(\hat{\theta}_j^{(t-1)}\big)$ for $j\!=\!1,\ldots,p$. The adjustment of the sign does not constitute a threat for privacy because the sign is determined by the parameter estimate from the previous iteration,  which already satisfies DP (Theorem \ref{thm:privacyguarantee} in Sec \ref{sec:DP}). Also noted is that the sign of $\hat{\theta}^{(t)}_j$ will eventually stabilize if it is non-zero. If $\hat{\theta}^{(t)}_j$ fluctuates around 0 after convergence, its final estimate will be set at 0.

\vspace{-6pt}\subsection{Guarantees of DP in NAPP-ERM}\label{sec:DP}\vspace{-1pt}
Before we provide the formal proof on DP satisfaction by NAPP-ERM, it should be noted that NAPP-ERM guarantees DP through objective function perturbation with one-time DP noise injection. Though the NAPP-ERM is realized through an iterative procedure, this is only an algorithmic artifact to leverage existing tools for solving non-regularized problems to achieve the target regularization effect. In other words, the NAPP-ERM algorithm only queries the original observed  data once to output one final estimate $\hat{\bs\theta}^*$ rather than querying the data multiple times to output multiple statistics or multiple versions of  sanitized $\hat{\bs\theta}^*$. Therefore, we only need to show the per-iteration privacy guarantees and there is no need to perform privacy accounting over iterations. 
\begin{thm}[\textbf{DP guarantees}] \label{thm:privacyguarantee}
Under Assumption \ref{ass}, the NAPP-ERM  procedure in Algorithm \ref{alg:NAPP-ERM} and the NAPP-ERM  procedure for variable selection  satisfy DP.
\end{thm}
The proof  for $\epsilon$-DP and $(\epsilon,\delta)$-DP is provided in the supplementary materials.
The key step in the proof is to bound the ratio $\frac{f\left(\hat{\bs\theta}^{(t)}|\D\right)}{f\left(\hat{\bs\theta}^{(t)}|\D'\right)}$ by $\e^\epsilon$ for two data sets $\D$ and $\D'$ differing by 1. This  is achieved by bounding ratios $r_1$ and $r_2$, separately; that is, 
\begin{equation}\label{eqn:ratio}
\frac{f\left(\hat{\bs\theta}^{(t)}|\D\right)}{f\left(\hat{\bs\theta}^{(t)}|\D'\right)}=\frac{f_{\bb}(\bb^{-1}(\hat{\bs\theta}^{(t)}|\D))} {f_{\bb}(\bb^{-1}(\hat{\bs\theta}^{(t)}|\D'))}\times\frac{|\det(J_{\bb}(\hat{\bs\theta}^{(t)}|\D'))|}{|\det(J_{\bb}(\hat{\bs\theta}^{(t)}|\D))|} =r_1r_2,
\end{equation}
$r_1$ relates directly to the amount of DP noise $\bb$, but the Jacobian ratio $r_2$ is not. On the other hand, $r_2$ still costs privacy per Eq (\ref{eqn:ratio}) due to the change of variable in the distribution from $\bb$ to $\hat{\bs\theta}^{(t)}$. A by-product of the proof  of $(\epsilon,\delta)$-DP  is a lower bound on  $\sigma^2$ for the Gaussian distributions in Eqs (\ref{eqn:b}) and (\ref{eqn:GaussianT}) as given in Corollary \ref{cor:boundsigma2}. The proof is provided in the supplementary materials. 
\begin{cor}\label{cor:boundsigma2}
The lower bound on the variance $\sigma^2$  of the Gaussian noise $\bb$ that leads to $(\epsilon,\delta)$-DP is
\begin{equation}\label{eqn:boundsigma2}
\sigma\geq\epsilon^{-1}\zeta_1\zeta_2\left(\sqrt{-2\log(\delta)+\epsilon}+ \sqrt{-2\log(\delta)}\right),
\end{equation}
If $\epsilon$ is small compared to $-2\log(\delta)$, the bound can be simplified to 
$$\sigma\geq2\epsilon^{-1}\zeta_1\zeta_2\sqrt{-2\log(\delta)+\epsilon}.$$
\end{cor}

\vspace{-9pt}\subsection{Privacy Budget Retrieval}\label{sec:retrive}
The proof of Theorem \ref{thm:privacyguarantee} bounds the two ratios in Eq (\ref{eqn:ratio}) separately to achieve  $\epsilon$-DP and $(\epsilon,\delta)$-DP. Although only $r_1$ determines the amount of the DP noises $\bb$, $r_2$  has to be bounded due to the ``change of variable'' from the distribution of $\bb$ to that of $\hat{\bs\theta}$, consuming a portion  of the total budget $\epsilon$. This section explores whether it is possible to cut back on the spending of the budget on bounding $r_2$ and re-allocate it to $r_1$ so to reduce the inject the level of DP noise and achieve better utility for the estimated private $\hat{\bs\theta}\;\!\!^{(t)}$, while still maintaining the overall budget at $\epsilon$.

Denote the proportions of $\epsilon$ allocated to bounding $r_1$ and $r_2$ are $r$ and $1-r$, respectively. The modulus $2\Lambda_0$ associated with the strong convexity term $\|\bs\theta\|_2$ is $\propto((1-r)\epsilon)^{-1}$.  If $1-r$ is large, then $\Lambda_0$ is small and the  dual-purpose weighted  $l_2$ regularization  is positioned to cover  the required  strong convexity with modulus $2\Lambda_0$. However, since $r_1$ receives only a small budget $r\epsilon$ and the generated DP noise $e^*$ and thus the overall augmented noise $e$ will be large in magnitude, overshadowing the targeted regularization realized by  dual-purpose weighted  $l_2$ and resulting in parameter estimates that  deviate significantly from what would be obtained without the DP noise. If we set $r$ at a large value and leave only a small portion $1-r$ of $\epsilon$ to $r_2$. Though the generated DP noise $e^*$ would not be a big distraction to the targeted regularization, NAPP can still be subject to severe over-regularization as  $\Lambda_0$ is large, resulting in a higher  strong convexity requirement than that carried by the dual-purpose weighted $l_2$ term. In other words, the variance of $\tilde{e}$ will be determined by the larger $\Lambda_0$ value, leading to over-regularization. All taken together, a good choice might be to start somewhere in the middle around $r=1/2$, leading some opportunity to re-allocate $\epsilon$ with a re-run of the NAPP-ERM algorithms if $1-r$ is a bit too large. This motivates the \emph{privacy budget retrieval} strategy stated in Proposition \ref{pro:retrieve}. The proof is provided in the supplementary materials.
\begin{pro}[\textbf{privacy budget retrieval via NAPP-ERM}]\label{pro:retrieve}
Let $(1-r)\epsilon$ be the privacy budget  allocated to bounding Jacobian ratio $r_2$ in Eq (\ref{eqn:ratio}), and $T_0$ be the iteration when the NAPP-ERM algorithm converges. The retrievable budget out of $(1-r)\epsilon$ upon convergence is
\begin{equation}\label{eqn:retrieve}
\Delta_\epsilon=\min\left\{0, (1-r)\epsilon\left(1-\Lambda_0\big(n_e l''|_{\bs\eta=\0}V_{(1)}^{(t)}\big)^{-1} \right)\right\},
\end{equation}
where $V^{(t)}_{(1)}\!=\!\!\!\min\limits_{j=1,\ldots, p}\!\! \mbox{V}\left(\tilde{e}_{ij}^{(t)}\right)$. As long as $\Delta_\epsilon\!>\!0$, we can retrieve some privacy budget originally allocated to $r_2$ upon convergence.  The retrieved budget $\Delta_\epsilon$ can be returned to the user, or recycled back to the NAPP algorithm for a re-run with an updated distribution of DP noise $\bb$ given the re-allocated budget  $r\epsilon+\Delta_\epsilon$, the sum of the originally allocated budget to ratio $r_1$ and the retrieved budget.
\end{pro}
Eq (\ref{eqn:retrieve}) suggests that budget can only be retrieved when the required strong convexity is automatically fulfilled by the target regularization via the dual-purpose weighted $l_2$ regularization in NAPP-ERM; that is, $\Lambda_0\!<\!n_e l''|_{\bs\eta=\0}V_{(1)}^{(t)}$. For example,  in the bridge regularization (Table \ref{tab:noisereg}), $\mbox{V}_{(1)}^{(t)}\!=\!2(n_el''|_{\bs\eta=\0})^{-1}\!\!\min\limits_{j=1,\ldots, p}\!\big\{ \max\!\big\{\!\Lambda|\hat{\theta}_j^{(t-1)}|^{-\gamma},\Lambda_0\big\}\!\big\}$ and thus $\Delta_{\epsilon}\!\!=\!\min\!\big\{0,(1-r)\epsilon \big(\!1-\Lambda_0\big(\!\!\min\limits_{j=1,\ldots, p}\!\!\big\{\!\max\!\big\{\!\Lambda|\hat{\theta}_j^{(T-1)}|^{-\gamma},\Lambda_0\big\}\!\big\}\!\big)^{\!-1}\big)\!\big\}$ in Eq (\ref{eqn:retrieve}). If $\Lambda|\hat{\theta}_j^{(T-1)}|^{-\gamma}\!<\!\Lambda_0$ for all $j\!=\!1,\ldots,p$, then  $\Delta_{\epsilon}=0$ and there is no retrieval  budget; otherwise, the retrieved budget is $\Delta_{\epsilon}\!=\!(1-r)\epsilon\!\left(\!1\!-\!\frac{\Lambda_0}{\Lambda}\big(\min\limits_{j\in\mathcal{J}}|\hat{\theta}_j^{(T-1)}|^{-\gamma}\big)^{-1}\!\right)$, where $\mathcal{J}\!=\!\{j:\Lambda|\hat{\theta}_j^{(T-1)}|^{-\gamma}\!>\!\Lambda_0\}$. In  other words, when $\Lambda$ or  $\min\limits_{j\in\mathcal{J}}|\hat{\theta}_j^{(T-1)}|^{-\gamma}$ is large, there is a higher chance to retrieve privacy budget.

The retrieved privacy budget $\Delta_\epsilon$ can be used in two ways. First, it can be returned to users so that the actual privacy cost is  $\epsilon-\Delta_\epsilon$ lower than the original planned costs $\epsilon$, meaning that the released results enjoy a higher level of privacy. Second, it  can be re-allocated to bounding $r_1$ in Eq (\ref{eqn:ratio}) that directly relates to the scale of DP noise $\bb$ so that less DP noise is injected and higher utility of the private $\hat{\bs\theta}^*$ can be achieved. Specifically, the distributions of $\bb$ are updated to 
\begin{align}
f(\bb) 
\begin{cases}
\propto\exp\left( -(r\epsilon+\Delta_\epsilon)(\zeta_1\zeta_2)^{-1}\|\bb\|_2\right)\mbox{ for $\epsilon$-DP} \\
= N(\mathbf{0},2(r\epsilon+\Delta_\epsilon)^{-2}\zeta_1^2\zeta_2^2\left(-\log(\delta_1)+r\epsilon+\Delta_\epsilon \right) \mathbf{I}_p)
\mbox{\hspace{9pt} for $(\delta,\epsilon)$-DP}
\end{cases},\label{eqn:b1} 
\end{align}
and the Gaussian distribution from which $\tilde{\e}$ is sampled is also updated with a new variance term $\V(\tilde{e}_{ij})$. For example, when the target regularization is bridge and elastic net, 
\begin{align}
\V(\tilde{e}_{ij})&\!=\!
2(n_el''|_{\bs\eta=\0})^{-1}\max\! \left\{\!\Lambda|\hat{\theta}_j^{(T_0)}|^{-\gamma},\Lambda|\hat{\theta}_{j^{(\tau)}}^{(T_0)}|^{-\gamma},\Lambda_0\right\},\label{eqn:V}\\
\V(\tilde{e}_{ij})&\!=\!2(n_el''|_{\bs\eta=\0})^{-1}\max\! \left\{\!\Lambda|\hat{\theta}_j^{(T_0)}|^{-1}\!+\!\Lambda\kappa,
\Lambda|\hat{\theta}_{j^{(T_0)}}^{(T_0)}|^{-1}\!+\!\Lambda\kappa,\Lambda_0\right\}\notag
\end{align}
respectively, where  $j^{(T_0)}\!\triangleq\!\arg\min_j\!\mbox{V}\!\big(\tilde{e}_{ij}^{(T_0)}\big)$. In other words, the updated variance of $\tilde{e}_{ij}$ ensures the strong convexity by choosing the largest modulus out of the following three: that associated with the weighted $l_2$ term based on the parameter estimate $\hat{\theta}_j^{(T_0)}$, the ``old'' modulus $2\Lambda_0$, and the new modulus $2\Lambda|\hat{\theta}_{j^{(T_0)}}^{(T_0)}|^{-1}$ (bridge) or $2\Lambda|\hat{\theta}_{j^{(T_0)}}^{(T_0)}|^{-1}\!+\!2\Lambda\kappa$ (elastic net).

Algorithm \ref{alg:retrieve} lists the steps that incorporate the  privacy budget retrieval in Algorithm \ref{alg:NAPP-ERM} and the NAPP-ERM variable selection procedure.  Algorithm \ref{alg:retrieve} may be applied multiple times if the user chooses to ``recycle''. Specifically, after a round of budget retrieval and recycling and a re-run of the NAPP algorithm,  a new set of parameter estimates  $\hat{\bs\theta}^*$ will be obtained upon convergence. Further budget retrieval is possible through another re-run as long as there is retrievable budget, and each re-run will allocate more budget to the sampling of DP noise. Though the DP noise will keep decreasing through the recycling, each additional round of  budget retrieval/recycling also means that the required strong convexity is getting stronger each time until eventually stabilizing. Our empirical results suggest significant retrieval often occurs just once, and the amount of retrieved budget in later rounds is often minimal.

\begin{algorithm}[!htb]
\caption{Privacy budget retrieval through NAPP-ERM}\label{alg:retrieve}
\SetKwInOut{Input}{input}
\SetKwInOut{Output}{output}
\Input{choice $\mathcal{A}$: return or recycle the retrieved budget?}
\Output{$\Delta_\epsilon$ and  parameter estimate  $\hat{\bs{\theta}}^*$ if $\mathcal{A}=$ ``return''; parameter estimate  $\hat{\bs{\theta}}^*$ if $\mathcal{A}=$ ``recycle''.}
Run the NAPP algorithm (e.g. Algorithm \ref{alg:NAPP-ERM}), calculate retrievable privacy budget $\Delta_\epsilon$ per Eq (\ref{eqn:retrieve}) given parameter estimate $\hat{\bs{\theta}}^{(T_0)}\!$ upon convergence at iteration $T_0$ \;
\If{$\mathcal{A}=$ ``recycle''}{
Let  $\Delta^{\text{cum}}_\epsilon\leftarrow0$\;
\While{$\Delta_\epsilon>0$}{
$\Delta^{\text{cum}}_\epsilon\leftarrow\Delta^{\text{cum}}_\epsilon+\Delta_\epsilon$\;
Rescale $\e^*\leftarrow\e^*(r\epsilon)/(r\epsilon +\Delta^{\text{cum}}_\epsilon)$\;
Let $\Lambda_0^{\text{old}}\leftarrow \Lambda_0$\;
Draw $\tilde{\e}$ from the Gaussian distribution with the updated $\Lambda_0$ (e.g., Eq (\ref{eqn:V}))\;
Run the NAPP Algorithm  with the updated augmented noisy data $\e^*+\tilde{\e}$ till convergence. Denote the  iteration at the convergence by $T^{(0)}$\;
$\Lambda_0\leftarrow n_e l''|_{\bs\eta=\0}V_{(1)}^{(T^{(0)})}$\;
Calculate retrievable privacy budget given the updated parameter estimate $\hat{\bs{\theta}}^{(T^{(0)})}$: 
$\Delta_\epsilon\!=\!\min\!\bigg\{0,((1\!-\!r)\epsilon\!-\!\Delta^{\text{cum}}_\epsilon)\!\left(\!1\!-\!\max\{\Lambda^{\text{old}}_0,\Lambda_0\}\!\left(n_e l''|_{\bs\eta=\0}V_{(1)}^{(T_0)}\right)^{\!-1}\!\right)\!\bigg\}$\;
}}
\vspace{-6pt}
\end{algorithm}

\subsection{Summary on NAPP-ERM}
We end Sec \ref{sec:naperm} by presenting Fig 2, which summarizes the ideas behind NAPP-ERM  and compares it with the existing DP-ERM framework and  illustrates  how and why NAPP-ERM works with its dual-purpose iterative weighted $l_2$ regularization term, and when privacy budget can be retrieved. 
\begin{figure}[!htb]\label{fig:Diagram}
\vspace{-12pt} \centering
\includegraphics[width=0.8\linewidth]{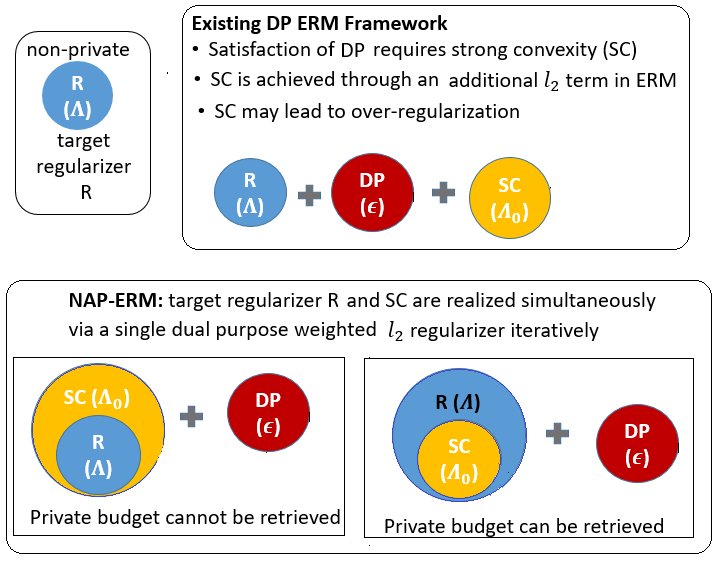}\vspace{-12pt}
\caption{Illustration of dual-purpose regularization and privacy budget retrieval in NAPP-ERM}\vspace{-12pt}
\end{figure}

\section{Utility Analysis}\label{sec:utility}
In this section, we consider the utility of the estimated $\hat{\bs\theta}$ via NAPP-ERM in two aspects: excess risk bound and sample complexity.  WLOG, we demonstrate the utility of NAPP-ERM without privacy budget retrieval.  The steps of deriving the theoretical bounds and sample complexity when there is budget retrieval are similar to what's given below, though the final mathematical results will be different. 

To start, we first define several types of loss functions (Table \ref{tab:loss}).  The noise-augment but non-private loss $J_p^{(t)}(\bs\theta|\D)$ can be regarded as a special case of $J_p^{(t)*\mbox{priv}}(\bs\theta|\D)$ with $\bb=0;\e^*=\0$, and  $R^{(t)}(\bs\theta)$ is expected to converge to $\max\{\Lambda R(\bs\theta),\Lambda_0\|\bs\theta\|_2\}$ through noise augmentation for $n_e\rightarrow\infty$.  When  strong convexity with modulus $2\Lambda_0$ is simultaneously realized by the target regularization, then  $R^{(t)}(\bs\theta)$ converges to $\Lambda R(\bs\theta)$. 
\begin{table}[!htp] \vspace{-9pt}
\caption{Loss Functions and Minimizers}\label{tab:loss}
\begin{center}\vspace{-12pt}
\resizebox{.9\textwidth}{!}{
\begin{tabular}{llc}
\hline
\multicolumn{2}{c}{loss function} &  minimizer\\
\hline
Expected loss $L(\bs\theta)$& $\E_{\dd}\left(n^{-1}\sum_{i=1}^{n}l(\bs\theta|\dd_i)\right)$ & ${\bs\theta}^0$\\
Regularized empirical\\		
loss $J(\bs\theta|\D)$  & $n^{-1}\!\left(\sum_{i=1}^nl(\bs\theta|\dd_i)\!+ \Lambda R(\bs\theta)\right)$& $\hat{\bs\theta}$\\
Noise-augment empirical  \\	
loss in iteration $t$:  $J_p^{(t)}\!(\bs\theta|\D)$& $n^{-1}\left(\!\sum_{i=1}^{n}l(\bs\theta|\dd_i)+ R^{(t)}(\bs\theta)\right)$& $\hat{\bs\theta}^{(t)}$ \\
Expected NA empirical \\
loss in iteration $t$: $\bar{J}_p^{(t)}(\bs\theta)$ & $\E_{\dd} \left(J_p^{(t)}(\bs\theta|\D)\right)$& $\bar{\bs\theta}^{(t)}$\\
NAPP empirical loss \\	
in iteration $t$: $J_p^{(t)\mbox{priv}}\!(\bs\theta|\D)$  & $n^{-1}\!\left(\!\sum_{i=1}^{n}l(\bs\theta|\dd_i)\!+\!\sum_{j=1}^{p}b_j\theta_j \!+\! R^{(t)}(\bs\theta)\right)$& $\hat{\bs\theta}^{(t)*}$\\
\hline
\end{tabular}}
\vspace{-12pt}\end{center}
\end{table}

\vspace{-6pt}\subsection{Excess Risk Bound}\label{sec:excessrisk}
The excess risk in this context is defined as the expected difference  between $\!J\left(\hat{\bs\theta}^{\!(t)*}|\D\right)$, where $\hat{\bs\theta}^{\!(t)*}$ is the  private estimate of $\bs\theta$ in the $t$-th iteration of the NAPP-ERM algorithm,  and $J(\hat{\bs\theta}|\D)$, where the $\hat{\bs\theta}$ is the non-private minimizer of $J(\hat{\bs\theta}|\D)$, over the distribution of DP noise $\bb$. Before we present the main results in Theorem \ref{thm:excessriskbound}, we first derive an upper bound for $\!J\left(\hat{\bs\theta}^{(t)*}|\D\right)-J(\hat{\bs\theta}|\D)$ in Lemma  \ref{lem:lossdiff}, based which bound the expected difference. 
\begin{lem}[\textbf{empirical risk bound}]\label{lem:lossdiff}
\vspace{-6pt} Under Assumption \ref{ass},  for $\hat{\bs\theta}^{(t)*}$  obtained in iteration $t$ of the NAPP-ERM algorithm 
\begin{align}\label{eqn:lossdiff}
J\left(\!\hat{\bs\theta}^{(t)*}|\D\!\right)\!-\!J\!\left(\hat{\bs\theta}|\D\!\right)\leq n^{-1}\left( \|\bb\|_2^2\textstyle\left(\!n_e l''|_{\bs\eta=\0}\mbox{V}_{(1)}\!\left(\tilde{e}_{ij}^{(t)}\!\right) \right)^{-1}\!\!\!\!+
\big( R^{(t)}(\hat{\bs\theta})\!-\!R(\hat{\bs\theta})\big)\right).
\end{align}
\end{lem}
The proof of Lemma \ref{lem:lossdiff} is provided in the supplementary materials. Eq \ref{eqn:lossdiff} suggests that the upper bound of the empirical risk at iteration $t$ decreases at a rate of $O(n^{-1})$, and is proportion to the squared $l_2$ norm of DP noise $\bb$ and the difference between $R^{(t)}(\hat{\bs\theta})\!-\!R(\hat{\bs\theta})$. 

Based on the results in Lemma \ref{lem:lossdiff}, we obtain Theorem \ref{thm:excessriskbound}, the proof of which is provided in the supplementary materials. 
\begin{thm}[\textbf{excess risk bound}]\label{thm:excessriskbound}
\vspace{-6pt}
For $\hat{\bs\theta}^{(t)*}$  obtained in iteration $t$ of the NAPP-ERM algorithm,  with probability $\ge 1-\pi$,  
\begin{align}\label{eqn:excessrisk}
\E_\bb\left(J\big(\hat{\bs\theta}^{(t)*}|\D\big) \!-\!J\big(\hat{\bs\theta}|\D\big)\right)
=\begin{cases}
O\big(B_1(\hat{\bs\theta},n,p,\Lambda_0,\zeta_1,\zeta_2,\epsilon,\pi)\big) \mbox{ for $\epsilon$-DP}\\
O\big(B_2(\hat{\bs\theta},n,p,\Lambda_0,\zeta_1,\zeta_2,\epsilon,\delta,\pi)\big)\mbox{ for $(\epsilon,\delta)$-DP}
\end{cases},
\end{align}
 where
\begin{align}
B_1(\;)&\!=\!\big(p\zeta_1\zeta_2(r\epsilon)^{-1}\log(p\pi^{-1})\big)^2n^{-1}\textstyle\big[\big(n_el''|_{\bs\eta=\0}\mbox{V}^{(t)}_{(1)} \big)^{-1}\!\!+\! R^{(t)}(\hat{\bs\theta})\!-\!R(\hat{\bs\theta})\big] \label{eqn:B1}\\
&\!=\! O\big(n^{-1}p^2\log(p)\epsilon^{-2}\big);\notag\\
B_2(\;)&\!=\!
4p\zeta_1^2\zeta_2^2(r\epsilon)^{-2}(r\epsilon\!+\!\log(2\delta^{-1}))\!\log(\pi^{-1}) n^{-1}\big[\!\big(n_el''|_{\bs\eta=\0}\mbox{V}^{(t)}_{(1)}\big)^{-1}\!\!\!+\! R^{(t)}(\hat{\bs\theta})\!-\!R(\hat{\bs\theta})\big]\label{eqn:B2}\\
&\!=\! O\big(n^{-1}p(\epsilon^{-1}+\epsilon^{-2}\log(\delta^{-1)})\big).\notag
\end{align}
\end{thm}

The bounds  $B_1$ and $B_2$ in Eqs (\ref{eqn:B1}) and (\ref{eqn:B2}) are tighter than the bounds given by Theorem 26 in \cite{dpermhd}. This can be easily seen. First, the first terms in $B_1$ and $B_2$ in Eqs (\ref{eqn:B1}) and (\ref{eqn:B2}) are no larger than the first terms of the bounds in \cite{dpermhd}. Second, the second terms in  $B_1$ and $B_2$ are smaller than the second terms of the bounds in \cite{dpermhd}; that is,  $R^{(t)}(\hat{\bs\theta})-R(\hat{\bs\theta})< \Lambda_0\|{\bs\theta}\|_2^2$ (if strong convexity is realized by the target regularization in NAPP-ERM,  $R^{(t)}(\hat{\bs\theta})-R(\hat{\bs\theta})\rightarrow0$ as $n_e\rightarrow\infty$ with the MOOR effect of NAPP-ERM). Taken together, the NAPP-ERM excess risk bounds  $B_1$ and $B_2$  are tighter than the bounds given in \cite{dpermhd}.

\vspace{-6pt}\subsection{Sample Complexity}\label{sec:sc}
The  sample complexity of a machine learning algorithm is defined as the training data  size $n$ that is needed to upper-bound the excess risk evaluated at the learner over the optimal risk.  In our context, the optimal risk is the ideal loss $L$  evaluated at $\bs\theta^0$ (Table \ref{tab:loss}), and the excess risk is the difference between $L(\hat{\bs\theta}^{(t)*})$, the ideal loss $L$ evaluated at the private parameter estimates via NAPP-ERM, versus $L(\bs\theta^0)$. Before presenting the main results in Theorem \ref{thm:SC}, we first present Lemma \ref{lem:diffJ}, on which Theorem \ref{thm:SC} is based.
\vspace{-3pt} 
\begin{lem}\label{lem:diffJ} There exists $C'$ such that, with probability at least $1-\pi'\;\forall\; \pi'\in(0,1)$,
\begin{align}
\bar{J}_p^{(t)}\!\big(\hat{\bs\theta}^{(t)*}\big) \!-\!\bar{J}_p^{(t)}\!\big(\hat{\bs\theta}^{(t)} \big)\leq &2 \big({J}_p^{(t)}(\hat{\bs\theta}^{(t)*}|\D)\!-\!{J}_p^{(t)}(\hat{\bs\theta}^{(t)}|\D)\big)\!-\!C'\log(\pi')\big(2n_e l''|_{\bs\eta=\0}\mbox{V}^{(t)}_{(1)} \big)^{-1}.\label{eqn:samplecplx1}
\end{align}
\end{lem}
\begin{thm}[\textbf{sample complexity of NAPP-ERM}]\label{thm:SC}
For any given $\varrho >0$, when the training sample size 
\begin{align}\label{eqn:ss}
n>&\textstyle \big(\varrho+C' \log(\pi')\big(2 n_e l''|_{\bs\eta=\0}\mbox{V}^{(t)}_{(1)}\big)^{-1} \big)^{-1}\left(R^{(t)}\!\big(\bs\theta^0\big)\!+\!C\big( n_e l''|_{\bs\eta=\0}\mbox{V}^{(t)}_{(1)} \big)^{-1}\right),\\
\mbox{then\hspace{12pt} } &
\Pr\left(L(\hat{\bs\theta}^{(t)*})\leq  L(\bs\theta^{0})+\varrho \right)\geq 1-\pi'-\pi,
\end{align}
where  $C=2\big( p\zeta_1\zeta_2(r\epsilon)^{-1}\log(p\pi^{-1})\big)^2$ for $\epsilon$-DP and $4p\zeta_1^2\zeta_2^2(r\epsilon)^{-2}\big(r\epsilon +\! \log(2/\delta)\big)\log(\pi^{-1})$ for $(\epsilon,\delta)$-DP,$\pi'$ is defined in Lemma \ref{lem:diffJ}, and $\pi$ is defined in the same way as  in Theorem \ref{thm:excessriskbound}. 
\end{thm}
\vspace{-3pt} 
The proofs of  Lemma \ref{lem:diffJ} and Theorem \ref{thm:SC} are provided in the supplementary materials. Since the work in \cite{dpermhd} does not perform an analysis the sample complexity, we compare the NAPP sample complexity in Eq (\ref{eqn:ss}) with that in \cite{dperm} that focuses on $\epsilon$-DP and $R(\bs\theta)=2^{-1}\Lambda\|\bs\theta\|_2$. When $R(\bs\theta)$ is the $l_2$ regularization, the DP-ERM framework in \cite{dperm} adds an extra $l_2$ term only when needed, the sample complexity there turns out to be the same as in Eq (\ref{eqn:ss}), as shown below. Let  $r\!=\!1/2$ and plug $C$ for $\epsilon$-DP in Eq (\ref{eqn:ss}), we obtain
$L\!\big(\hat{\bs\theta}^{(t)*}\big)\!-\! L\big( \bs\theta^{0}\big)\!\leq\!\\ n^{-1}\! R^{(t)}\!\left(\bs\theta^0\right)\!+\!\left(\frac{8p^2\zeta_1^2\zeta_2^2\log^2(p/\pi)}{n\epsilon^2}\!-2^{-1}C' \log(\pi')\right)\!\left(n_e l''|_{\bs\eta=\0}\mbox{V}^{(t)}_{(1)} \right)^{-1}$.
Now replace $n^{-1}R^{(t)}$ with $2^{-1}\Lambda\|\bs\theta^0\|_2^2$ and $n_e l''|_{\bs\eta=\0}\mbox{V}^{(t)}_{(1)}$ with $2^{-1}n\Lambda$ above, we have $L\!\left(\hat{\bs\theta}^{(t)*}\right)\!-\! L\left( \bs\theta^{0}\right)\leq \frac{\Lambda}{2}\|\bs\theta^0\|_2^2\!+\!\left(\frac{8p^2\zeta_1^2\zeta_2^2\log^2(p/\pi)}{n\epsilon^2}\!-\!\frac{1}{2}C_1 \log(\pi')\right)\!\left(\frac{1}{2}n \Lambda \right)^{-1}$.
Follow the assumption that $\zeta_1\leq1,\zeta_2\leq1$ in \cite{dperm}, and re-expressing the $C_1$ term with the Big-$O$ expression, we get
\begin{align*}
L\!\left(\hat{\bs\theta}^{(t)*}\right)\!-\! L\left( \bs\theta^{0}\right)\!\leq\!\frac{\Lambda}{2}\|\bs\theta^0\|_2^2\!+\!\frac{16p^2\log^2(p/\pi)}{\epsilon^2n^2 \Lambda}\!+\!O\left(\frac{\log(1/\pi')}{n \Lambda}\right),
\end{align*}
which is the same bound as given in Eq (18) in the proof of Theorem 18 of \cite{dperm}.

\vspace{-3pt}\section{Experiments}\label{sec:examples}
We run  experiments to demonstrate the improvement of NAPP-ERM  over the current DP-ERM framework in the settings of linear, Poisson and logistic regressions with the lasso regularizer.  Specifically, Sec \ref{sec:moor} examines the effect of MOOR on model parameter estimation and outcome prediction accuracy; Sec \ref{sec:VS} shows the improvement made by the new NAPP-ERM variable selection framework in Sec \ref{sec:NAPP-ERM-vs} compared to the general NAPP-ERM procedure and the existing DP-ERM procedure in privacy-preserving variable selection; privacy budget retrieval and recycling in NAPP-ERM is demonstrated in Sec \ref{sec:retrieval}.

We use both simulated and real data to run the experiments. For the simulated data, we vary the training set size  $n$, privacy loss $(\epsilon,\delta)$, and tuning parameter $\Lambda$ to set up different simulation scenarios.  500 repeats were run in each scenario. The regression settings are summarized in Table \ref{tab:sim}. 
\begin{table}[!htp]\vspace{-9pt}
\caption{Simulation settings}\label{tab:sim}
\vspace{-15pt}\begin{center}
\resizebox{1\linewidth}{!}{
\begin{tabular}{llll}
\hline
& linear & Poisson & logistic\\
\hline
$\x_i\;(16
\times 1)$ &$\mbox{Unif}(-0.25,0.25)$&$\mbox{Unif}(-0.25,0.25)$&$\mbox{Unif}(-0.5,0.5)$ \\
8 nonzero $\bs\theta$ & $\{1-0.8k/7\}_{k=0,\ldots,7}$ &$\{4-0.5k/7\}_{k=0,\ldots,7}$ &$\{4-0.5k/7\}_{k=0,\ldots,7}$\\
$y_i$& $\x_i\bs\theta\!+\!\mbox{tN}(0,0.25^2,-0.5, 0.5)$ &Poisson$(\exp(\x_i\bs\theta))$&$\mbox{Bern}((1\!+\!\exp(\x_i\bs\theta))^{-1})$\\
$n$ &200, 500 &500, 1000 &500, 1000 \\
$\epsilon$& 0.4, 0.6, 0.8, 1.0 & 0.4, 0.6, 0.8, 1.0 &0.4, 0.6, 0.8, 1.0\\
$\delta$ & 0, 0.001 & 0, 0.001 & 0, 0.001\\
\hline
\end{tabular}}
\end{center}\vspace{-9pt}
\end{table}

We present the results for $(\epsilon,\delta=0.0001)$-DP in the main text and the results from $\epsilon$-DP in the supplementary materials, which are very similar to $(\epsilon,\delta)$-DP in all the experiments.

\vspace{-6pt}\subsection{Mitigation of Over-regularization (MOOR)}\label{sec:moor}
The goal of this experiment is to demonstrate the MOOR effect with with the dual-purposes regularization for achieving strong convexity and target regularization with a single iterative weighted $l_2$ term, as proposed and employed by the NAPP-ERM framework, in contrast to employment of two separate terms in the existing DP-ERM framework. Though the strong convexity requirement is a result of privacy guarantees, the MOOR effect is independent of privacy  As such, we set DP noise $\bb=0$ in Eqs (\ref{eqn:dperm}) and (\ref{eqn:napp}) and examine MOOR effect in the non-private setting, benchmarked against the regular lasso regression (without the requirement for strong convexity). In the implementation of the NAPP-ERM algorithm, $\e$ contains only the noise generated from the Gaussian distribution $\tilde \e$ from Eq (\ref{eqn:tildee}) and but not the DP noise component $\e^*$  from Eq (\ref{eqn:e*}). 
 
\begin{figure}[!htb]
\centering \vspace{-6pt}
\small{linear regression \hspace{0.7cm} 
logistic regression\hspace{0.5cm} Poisson regression}\\
\includegraphics[width=0.32\linewidth, trim=0cm 0.5cm 0.5cm 0cm, clip]{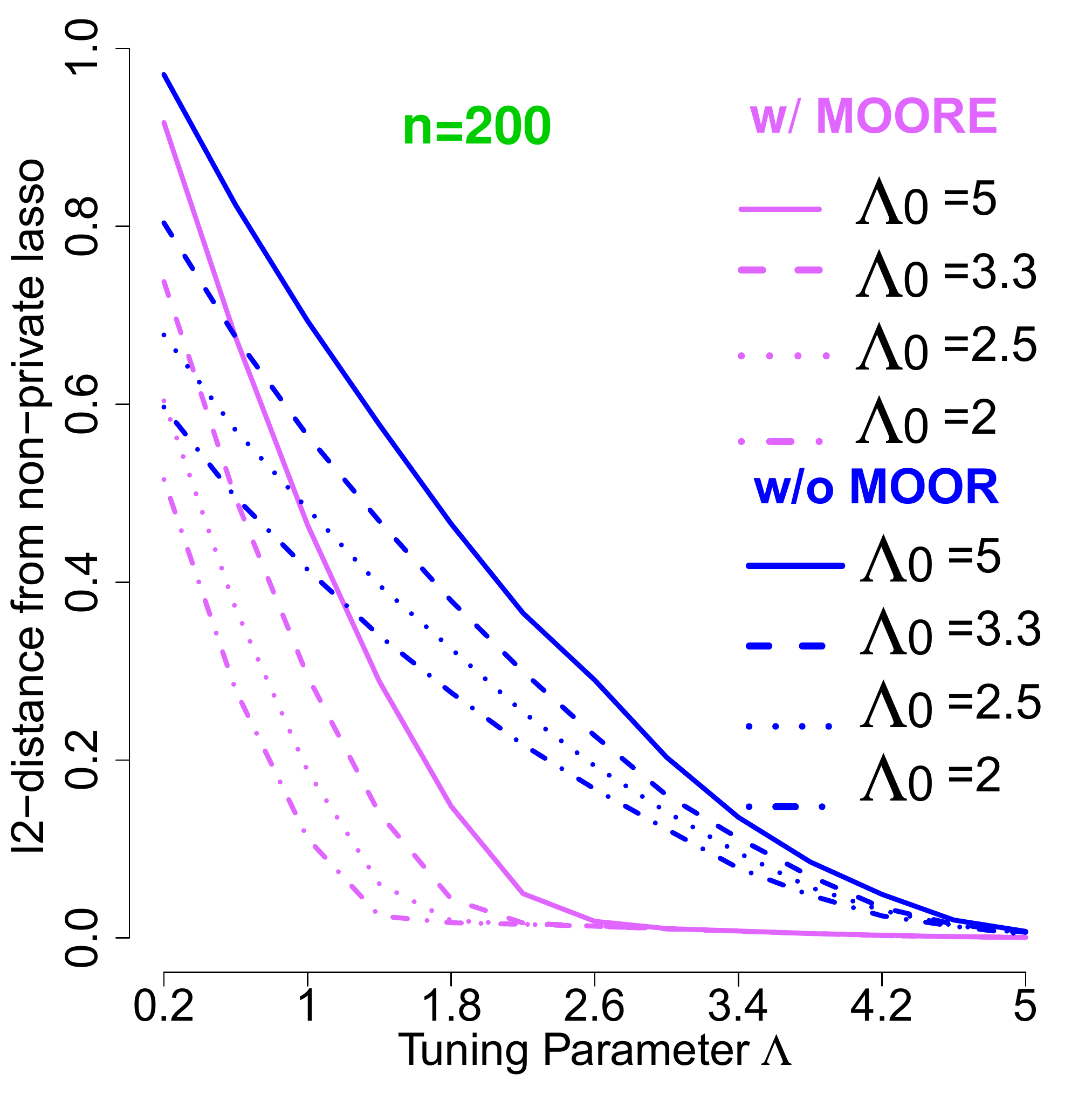}
\includegraphics[width=0.33\linewidth, trim=0cm 0.2cm 0.3cm 0cm, clip]{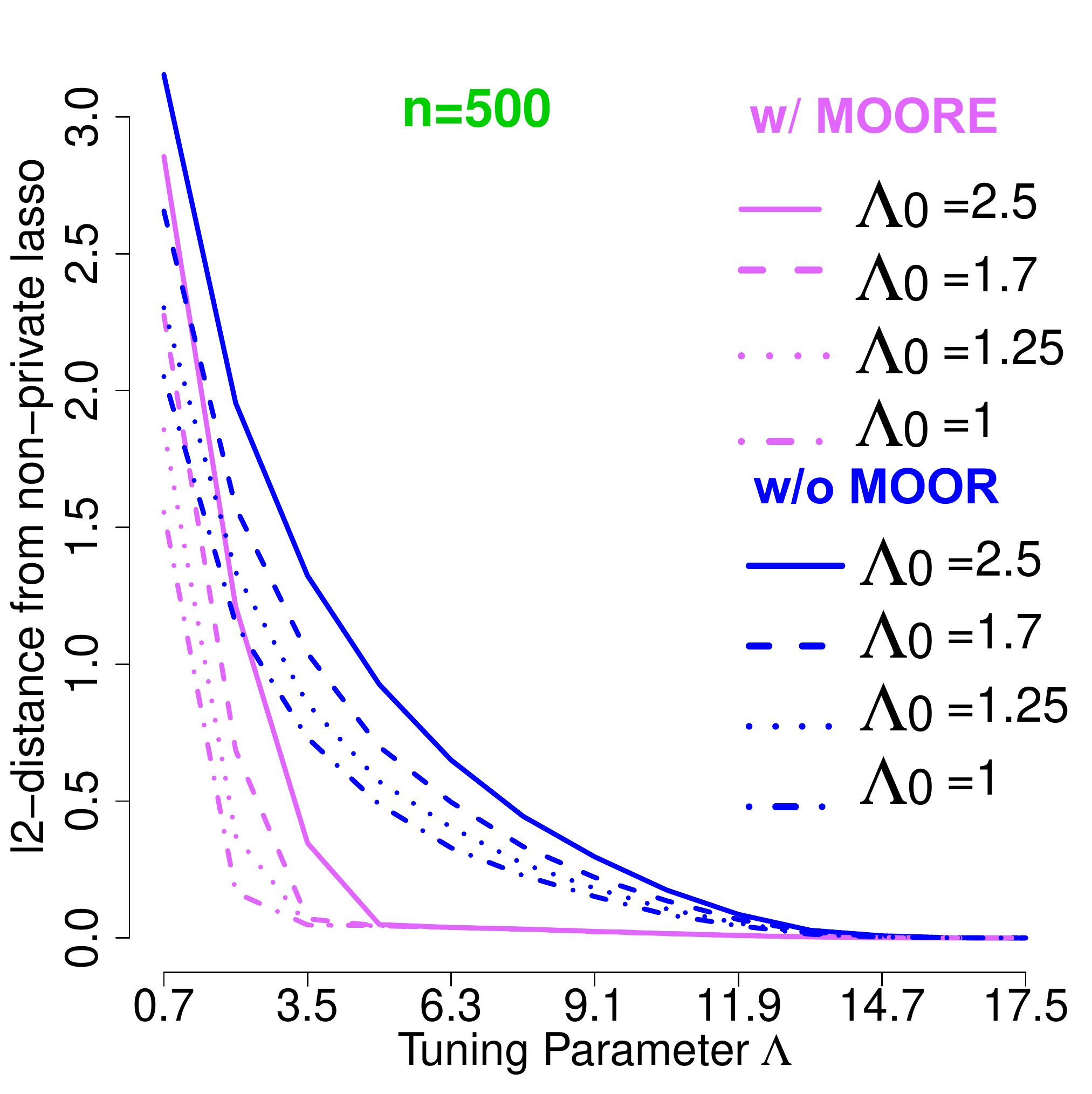}
\includegraphics[width=0.33\linewidth, trim=0cm 0.2cm 0.3cm 0cm, clip]{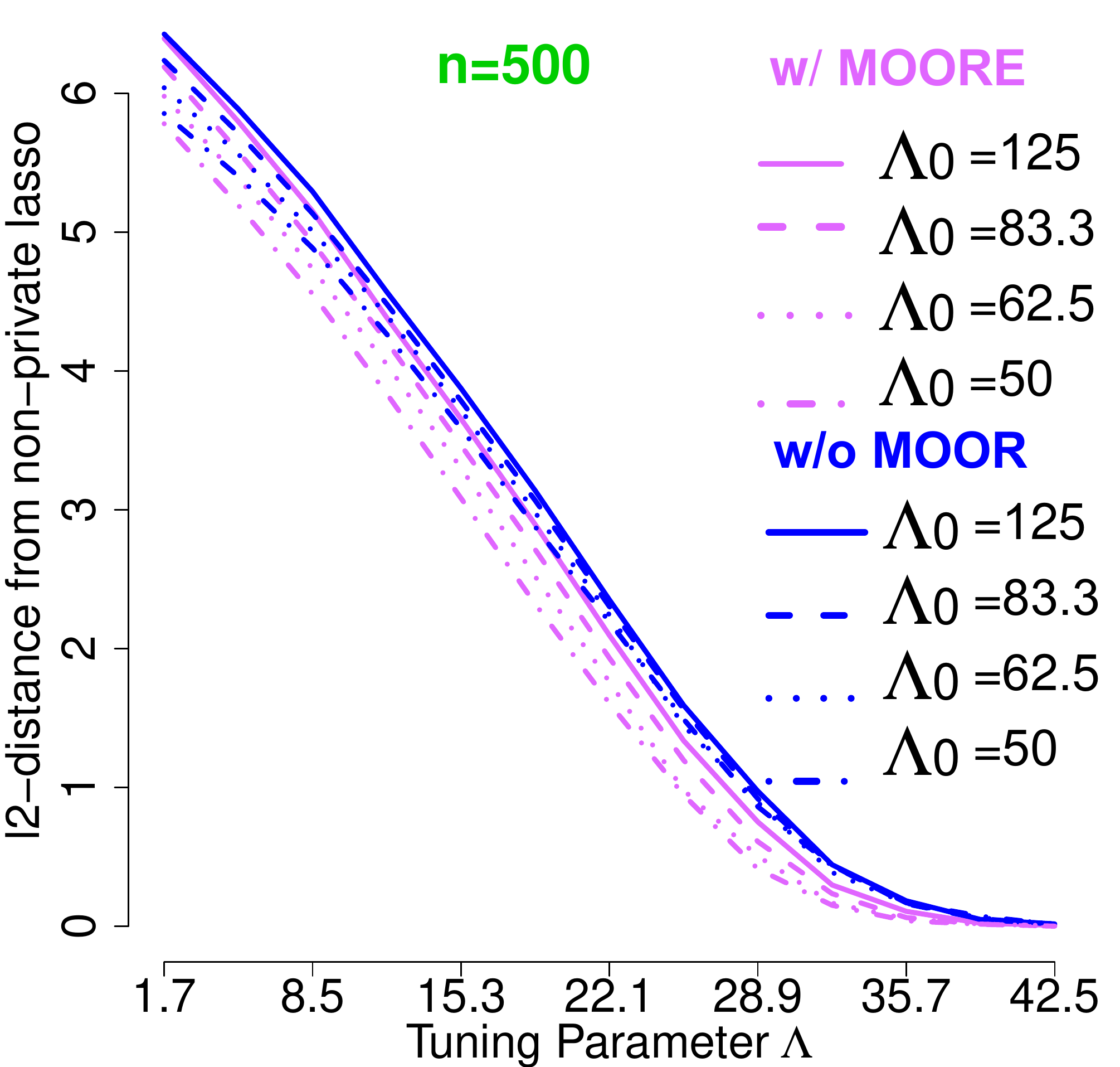}\\
\includegraphics[width=0.32\linewidth, trim=0cm 0.5cm 0.5cm 0cm, clip]{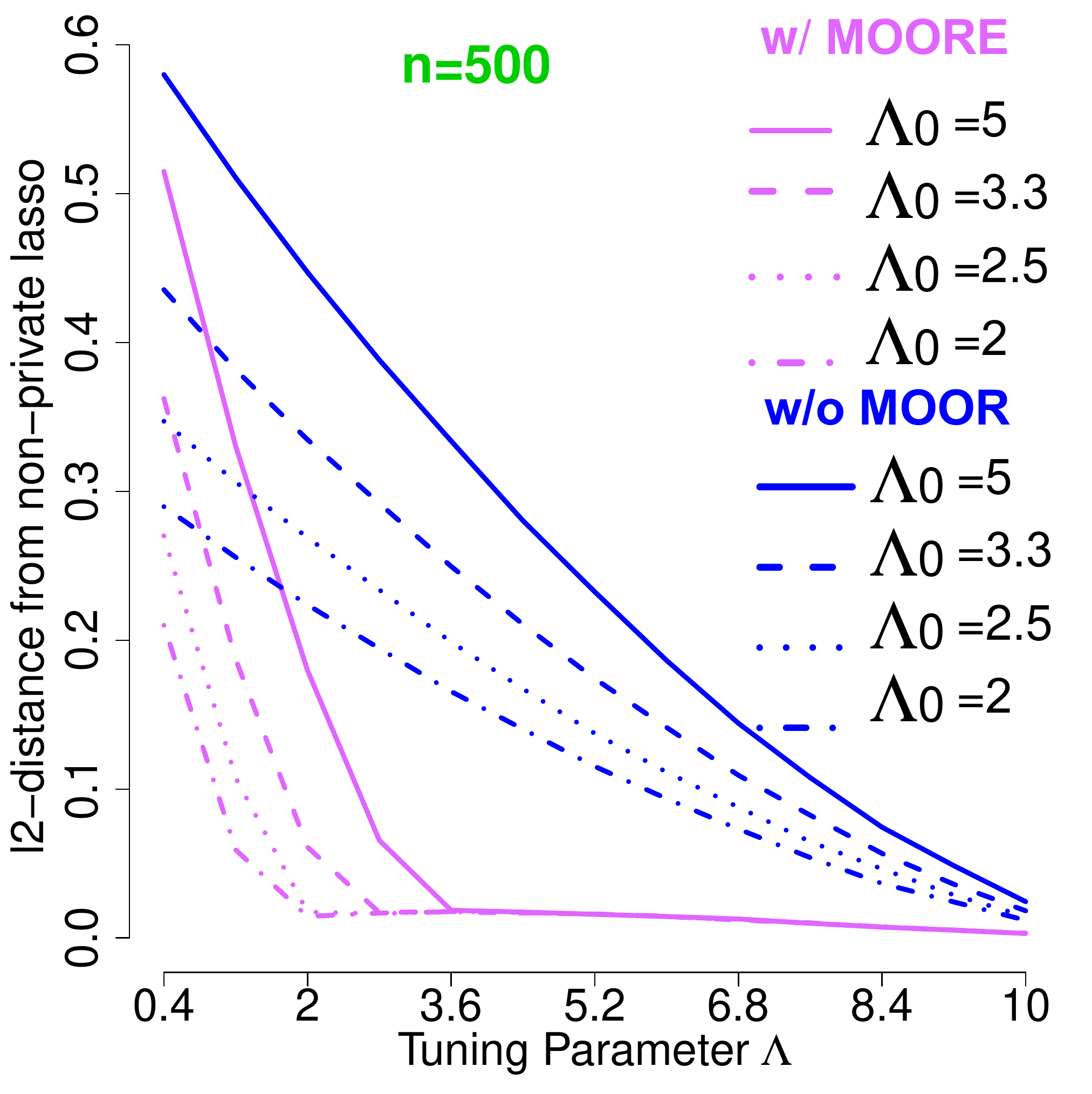}
\includegraphics[width=0.33\linewidth, trim=0cm 0.2cm 0.3cm 0cm, clip]{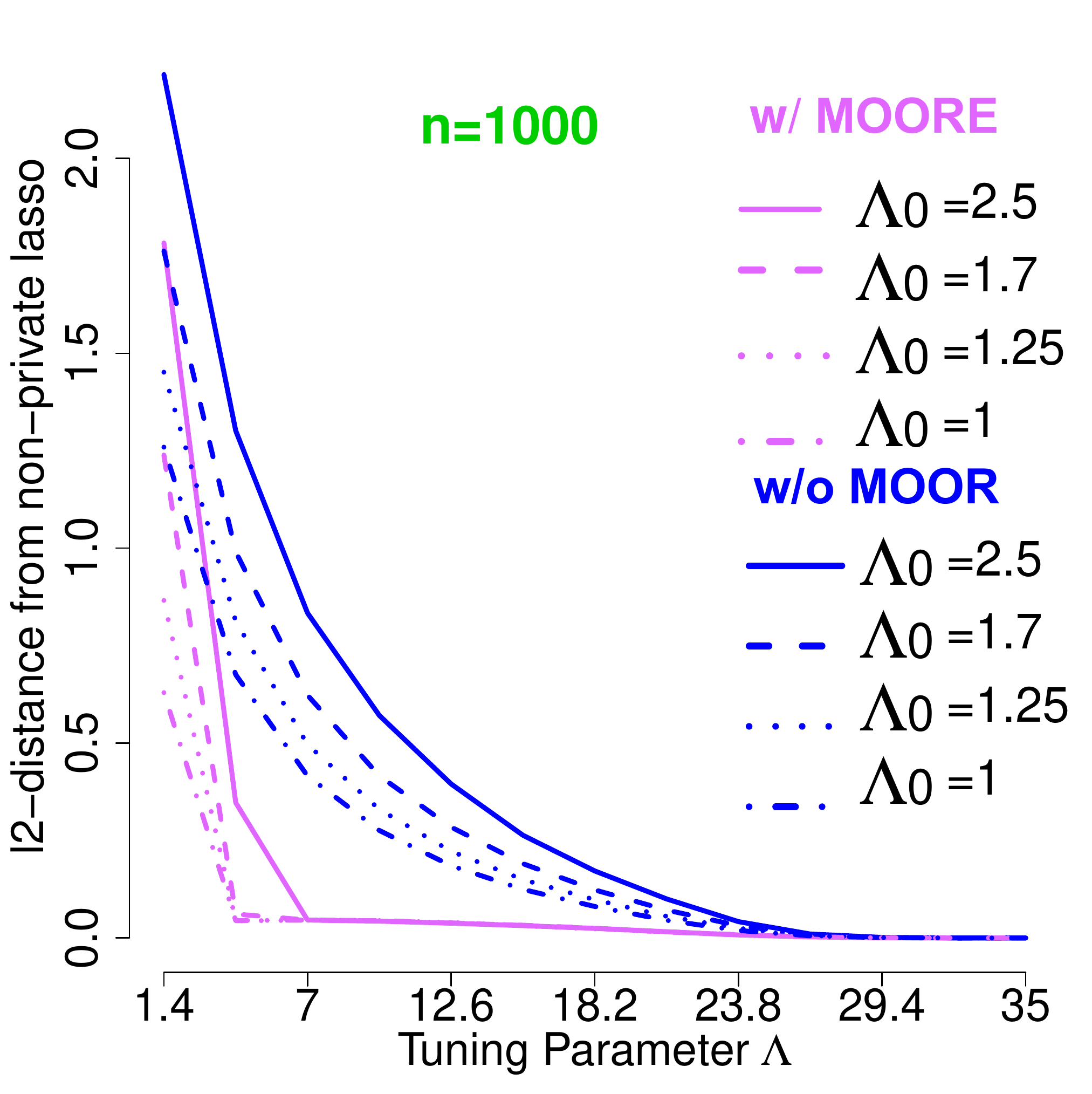}
\includegraphics[width=0.33\linewidth, trim=0cm 0.2cm 0.2cm 0cm, clip]{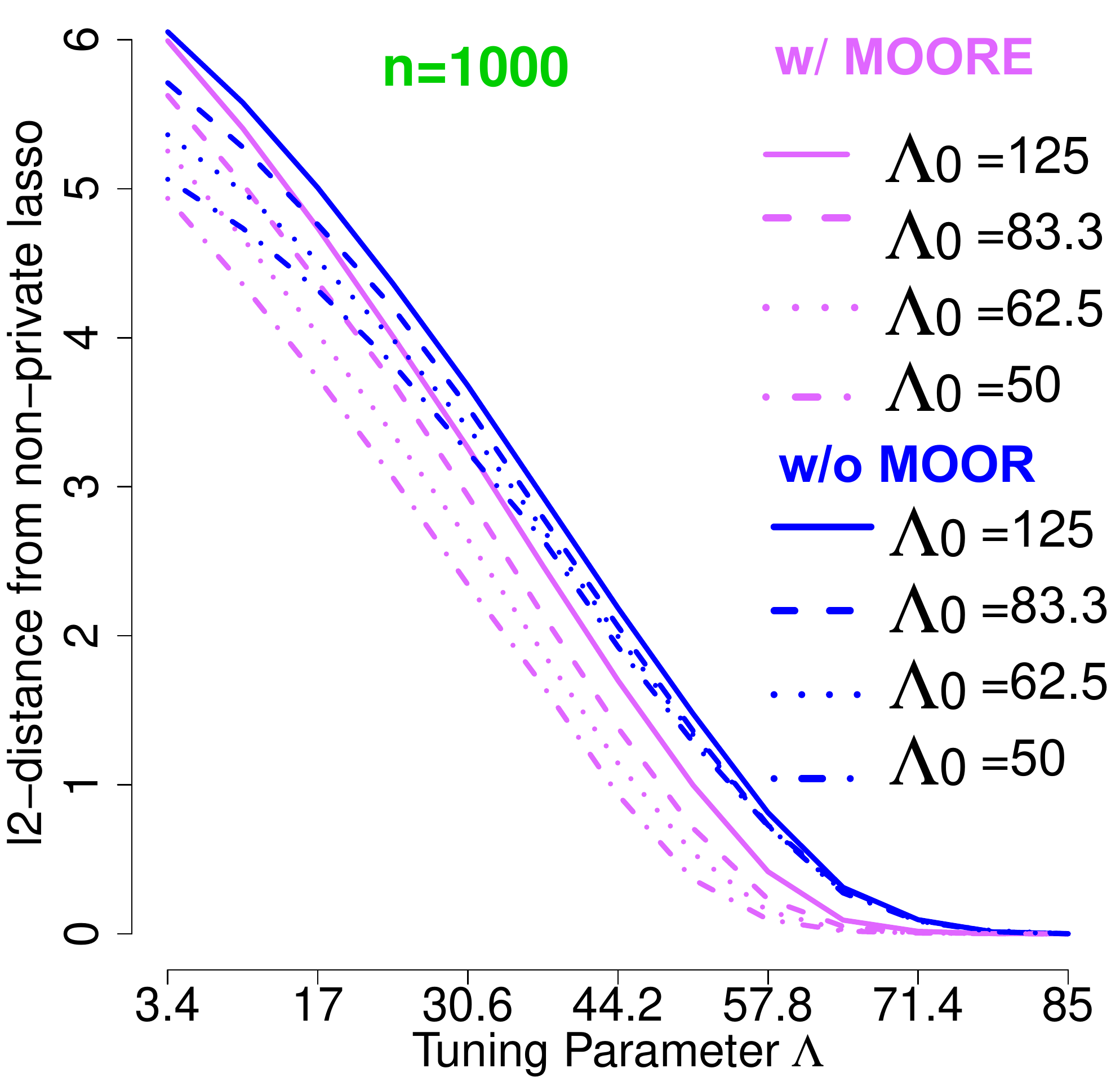}\vspace{-9pt}
\caption{$l_2$ distance of  $\bs\theta$ estimates obtained via non-private noise augmented ERM with MOOR vs without MOORE from the original lasso estimates} 
\label{fig:SIM1}\vspace{-9pt}
\end{figure}

Fig \ref{fig:SIM1} plots the $l_2$ distance between the regular lasso estimates and those via the NAPP-ERM algorithm (by setting $\bb$ and $\e^*$ at 0) with MOOR vs without MOOR.  
In the linear and logistic regressions, the increase in accuracy (smaller $l_2$ distance) with MOOR is significant for all the examined $\Lambda_0$ and $\Lambda$ values. There is also improvement in the Poisson regression though it is not as obvious as the other two because of the large $\zeta_3$ and $\Lambda_0$ values, leaving little room for NAPP-ERM to execute its MOOR power.

\vspace{-6pt}\subsection{Private Variable Selection and Outcome Prediction}\label{sec:VS}\vspace{-3pt}
In this experiment, we compare 5 approaches listed below in private variable selection and prediction accuracy with the lasso regularizer, benchmarked against the non-private regular lasso. 
\vspace{-9pt}
\begin{itemize}\setlength\itemsep{-3pt}
\item NAPP-VS+ without MOOR in lasso regression
\item NAPP-VS without MOOR in lasso regression
\item existing DP-ERM in lasso regression (without MOOR)
\item NAPP-VS+ with MOOR in lasso regression
\item NAPP-VS with MOOR in lasso regression
\item non-private lasso regression
\end{itemize}\vspace{-3pt}
For the NAPP-VS+ approaches above, $c\!=\!0$ in Eqs (\ref{eqn:LaplaceT}) and (\ref{eqn:GaussianT}); for NAPP-VS, $c\rightarrow -\infty$. For variable selection, we define ``positive'' as correct selection of a covariate associated with non-zero $\theta$ and plot the ROC curves. For  outcome prediction,  we examine the prediction mean squared error (MSE) on the outcome in independently simulated testing data ($n=10,000$) in linear and poisson regressions, and  the misclassification rate on the predicted outcome in logistic regression. 

\begin{figure}[!htb]\vspace{-12pt}
\centering
\footnotesize linear regression $n=200$ \\
\includegraphics[width=0.20\linewidth, trim=4pt 12pt 15pt 18pt,clip]{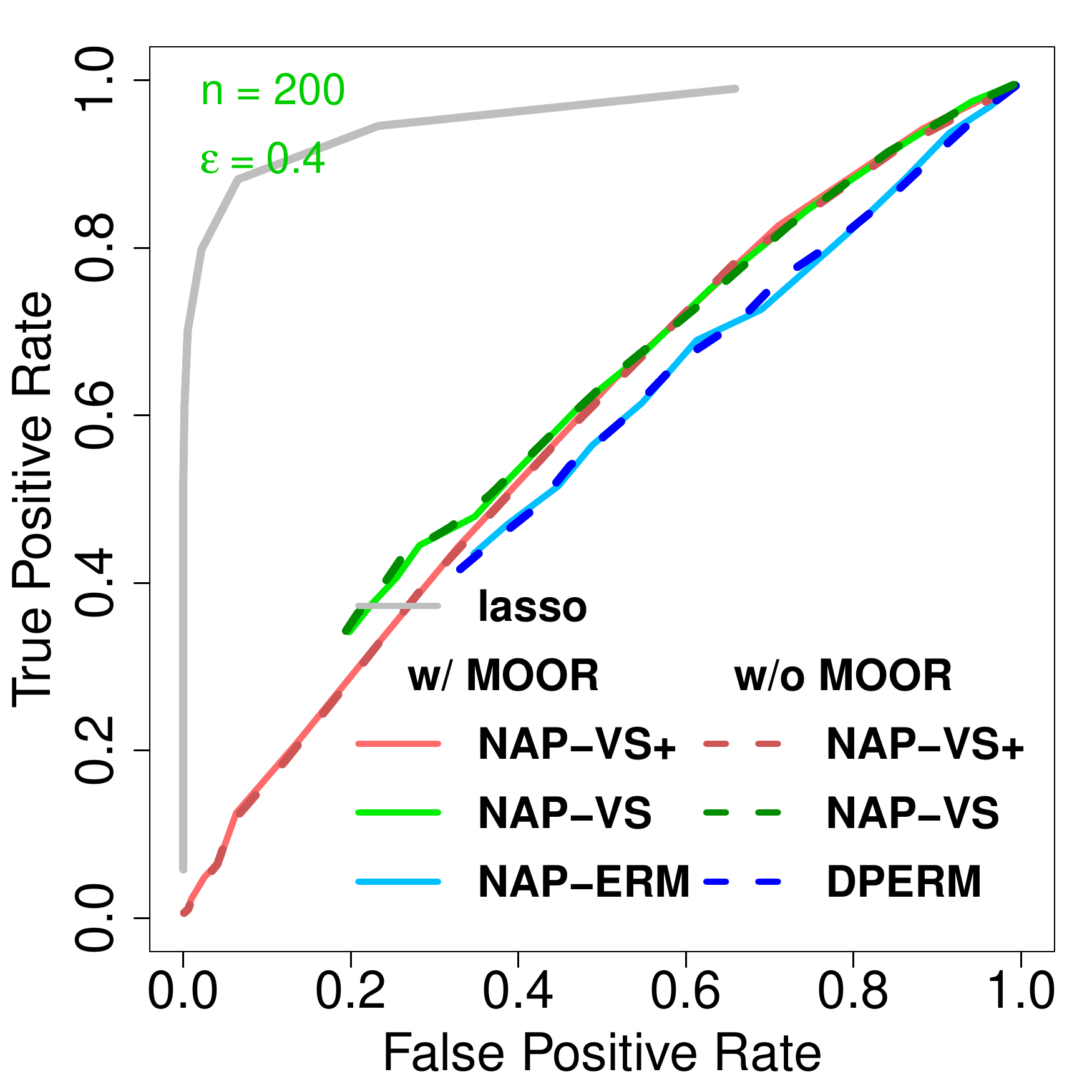}
\includegraphics[width=0.20\linewidth, trim=4pt 12pt 15pt 18pt,clip]{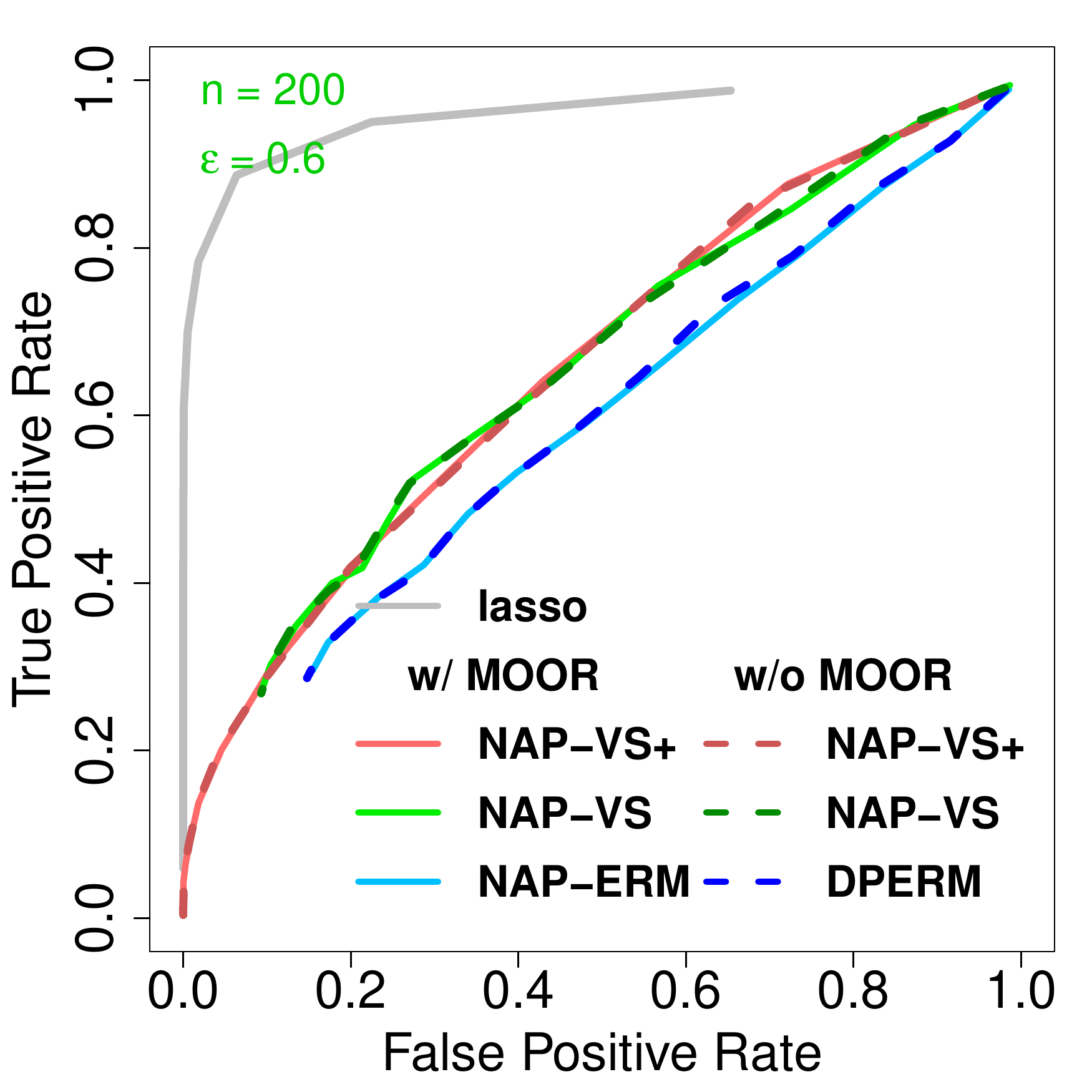}
\includegraphics[width=0.20\linewidth, trim=4pt 12pt 15pt 18pt,clip]{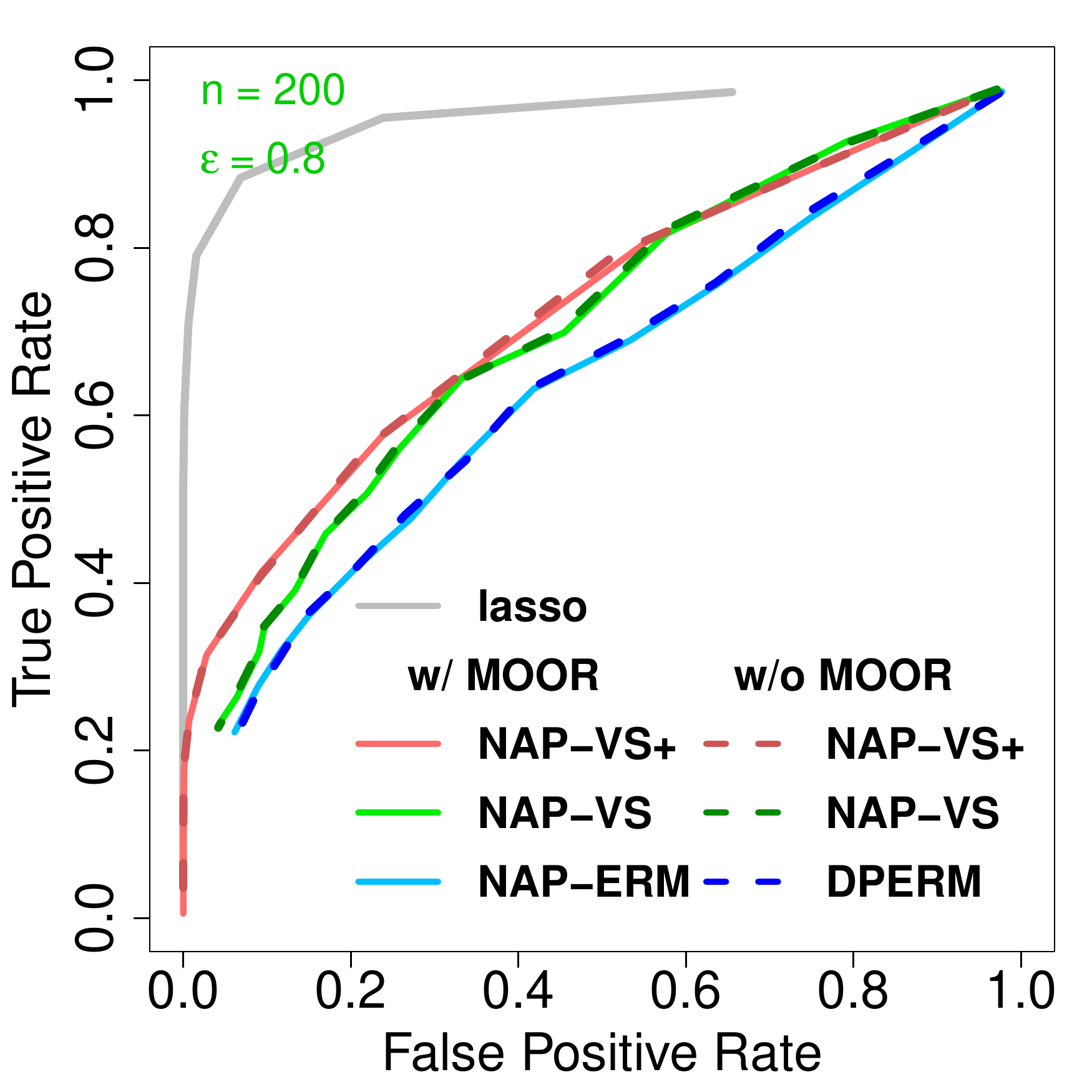}
\includegraphics[width=0.20\linewidth, trim=4pt 12pt 15pt 18pt,clip]{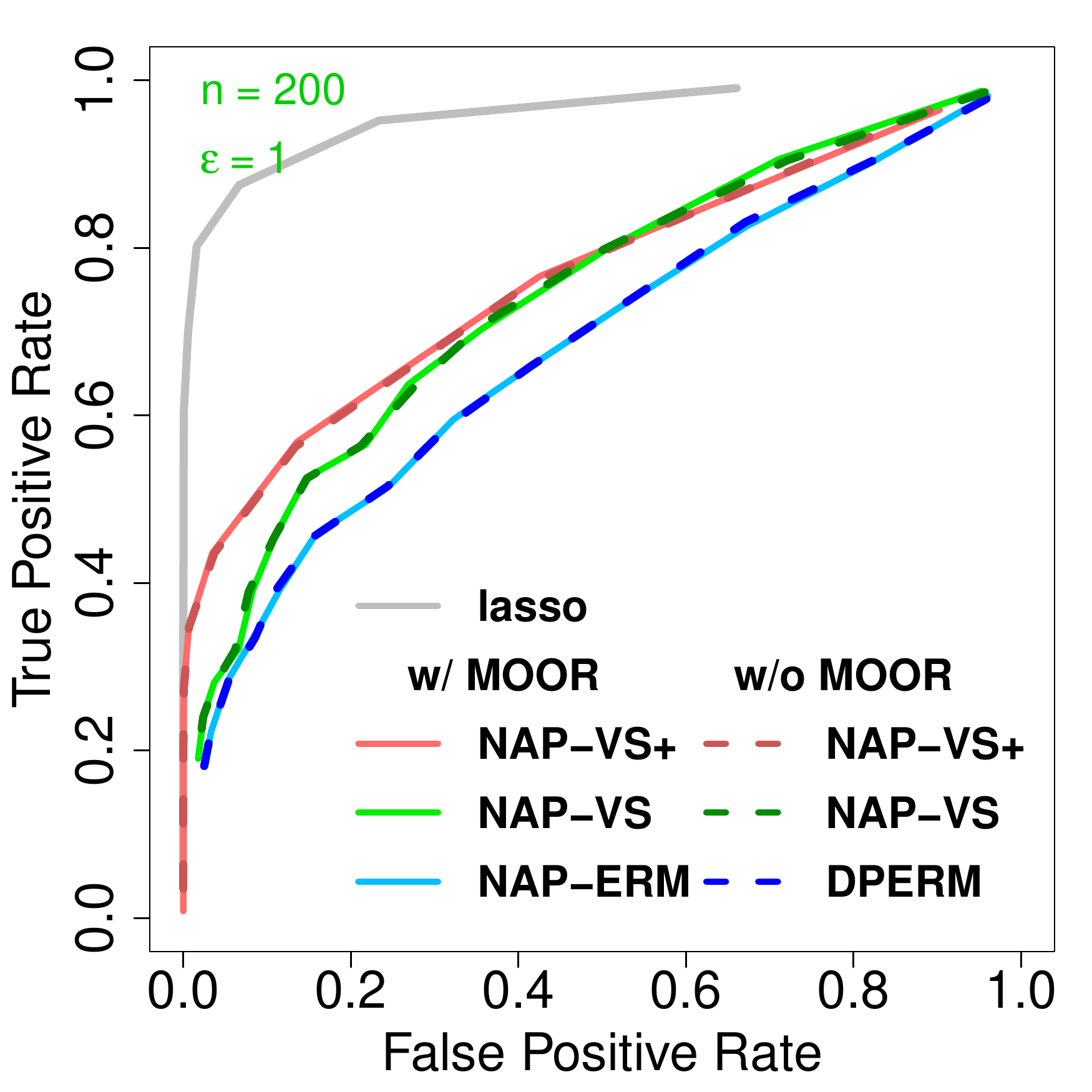}\\
\footnotesize linear regression $n=500$ \\
\includegraphics[width=0.20\linewidth, trim=4pt 12pt 15pt 18pt,clip]{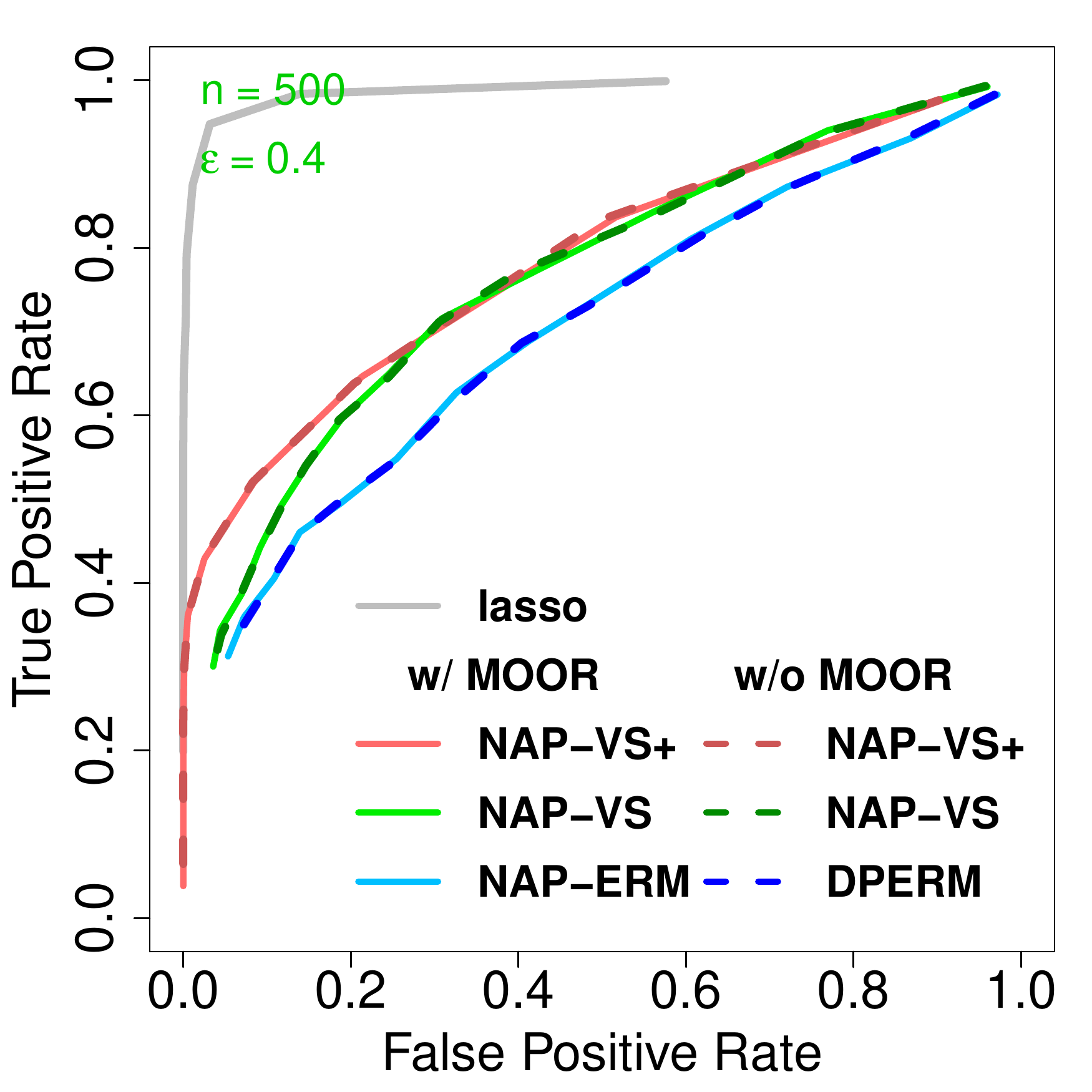}
\includegraphics[width=0.20\linewidth, trim=4pt 12pt 15pt 18pt,clip]{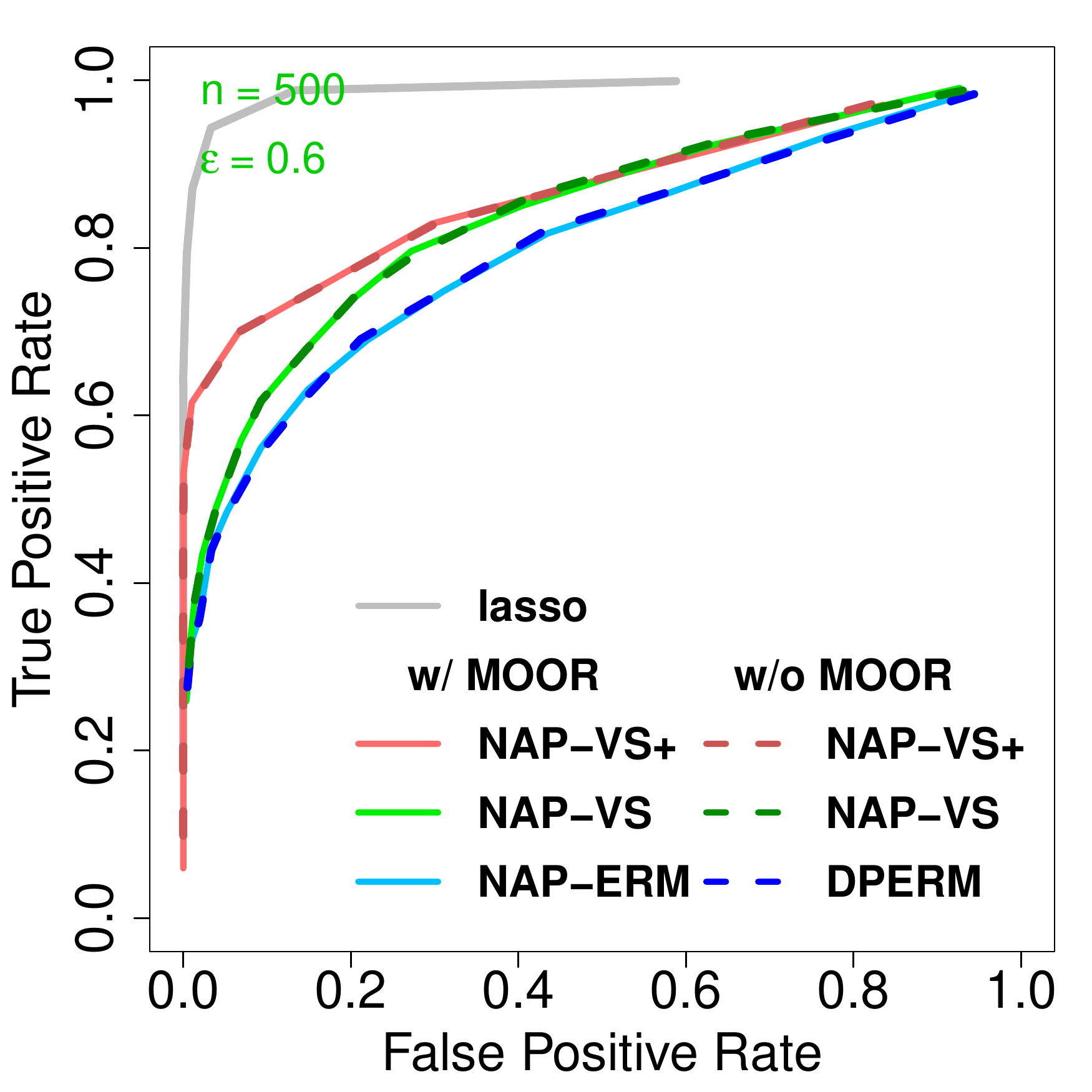}
\includegraphics[width=0.20\linewidth, trim=4pt 12pt 15pt 18pt,clip]{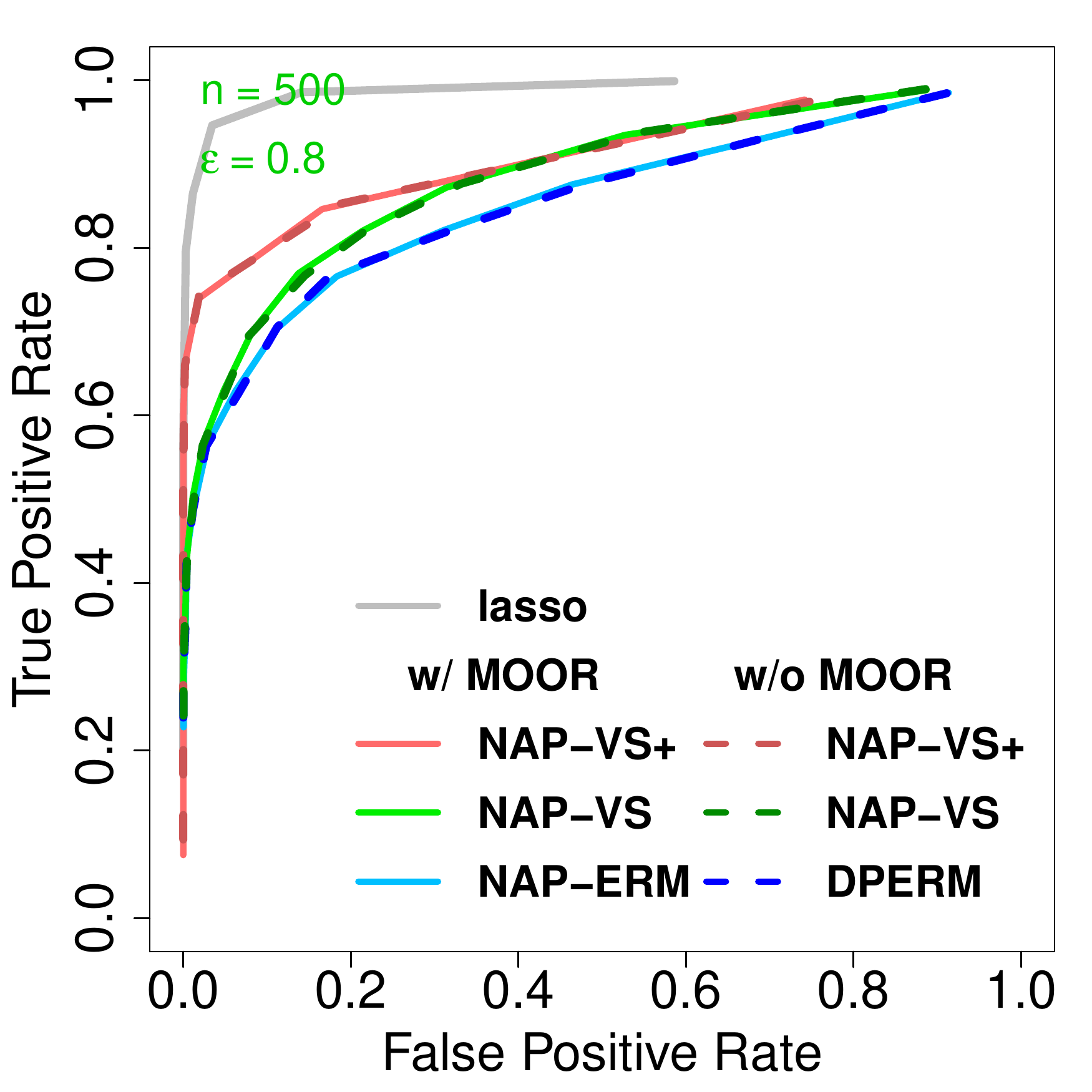}
\includegraphics[width=0.20\linewidth, trim=4pt 12pt 15pt 18pt,clip]{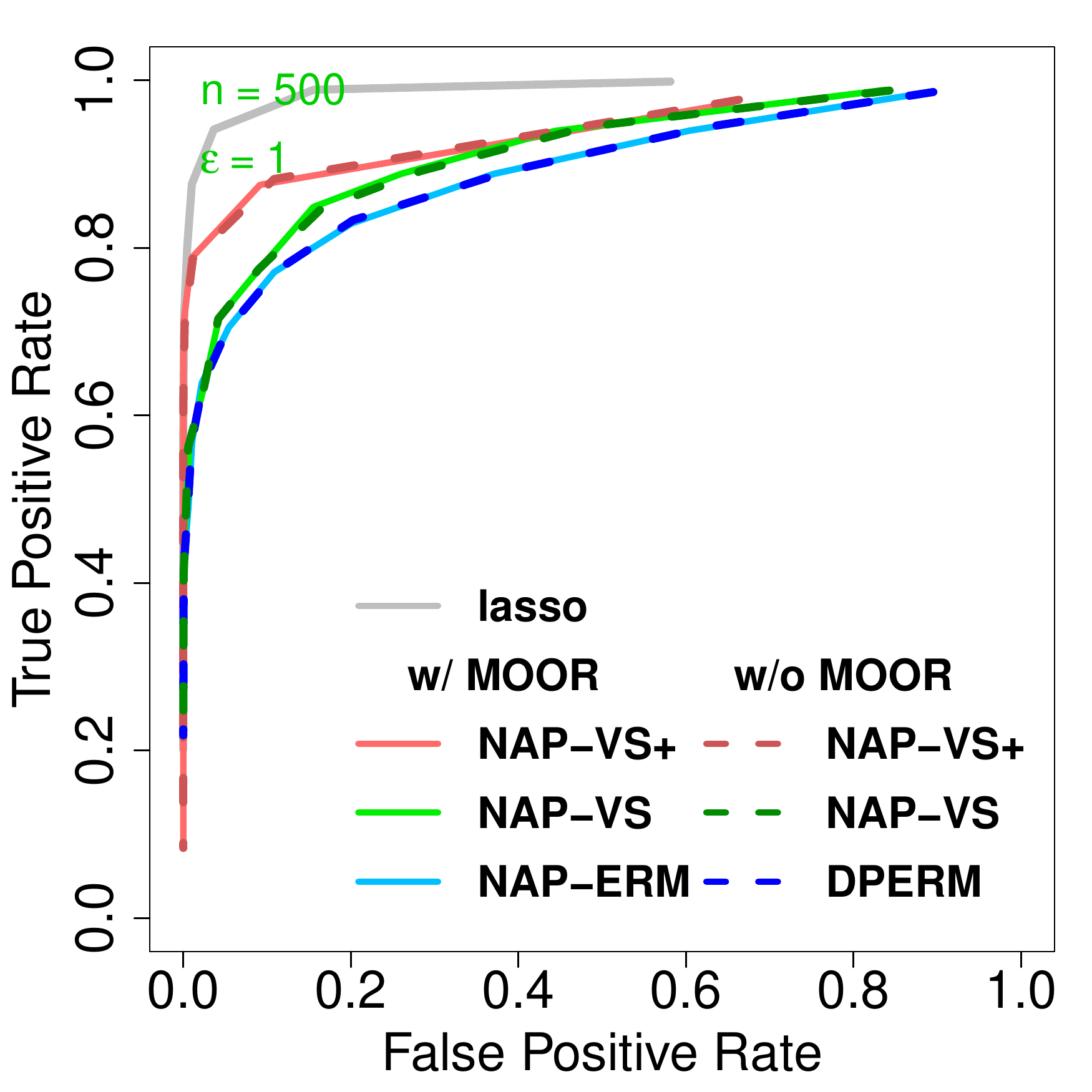}\\
\footnotesize logistic regression $n=500$ \\
\includegraphics[width=0.20\linewidth, trim=4pt 9pt 15pt 18pt,clip]{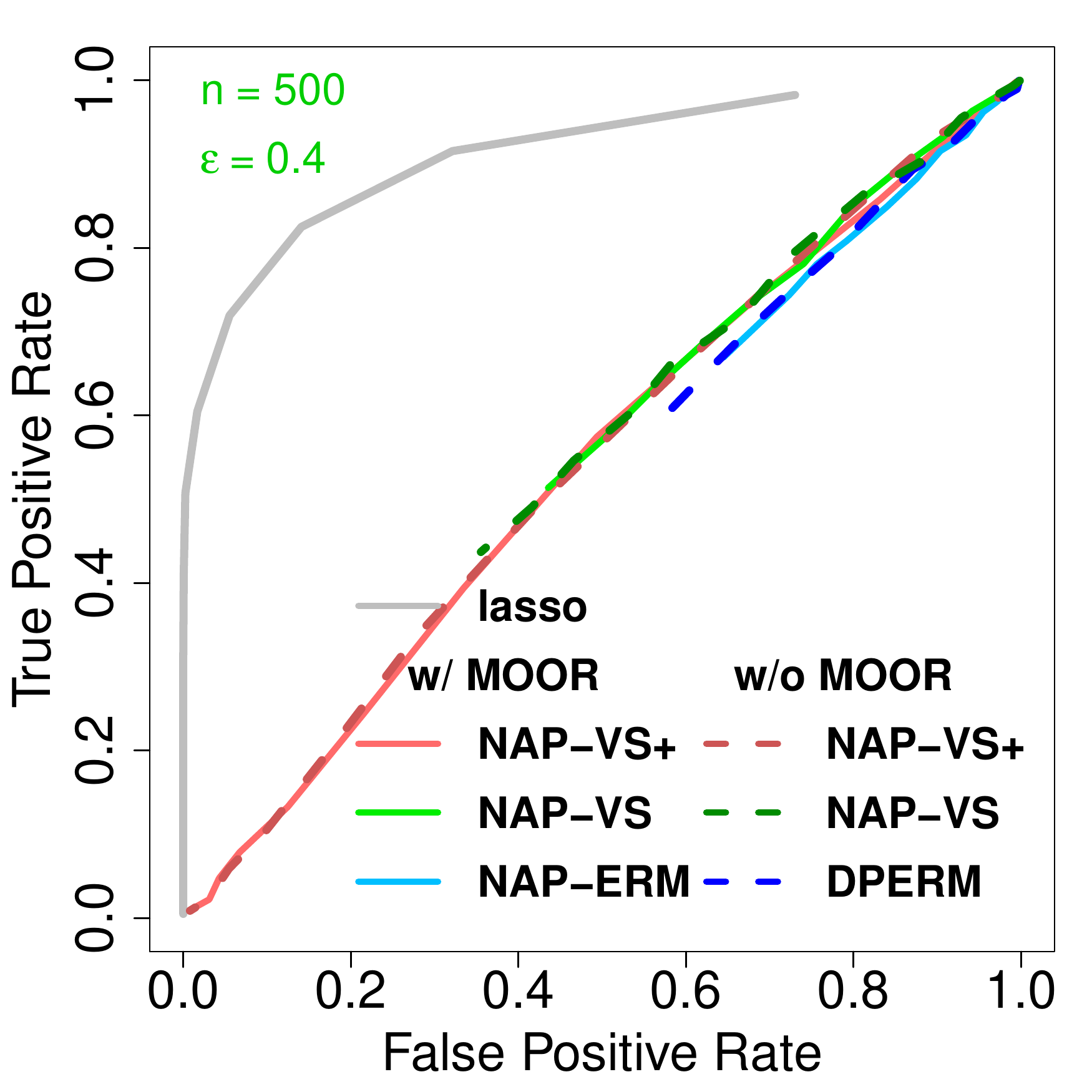}
\includegraphics[width=0.20\linewidth, trim=4pt 9pt 15pt 18pt,clip]{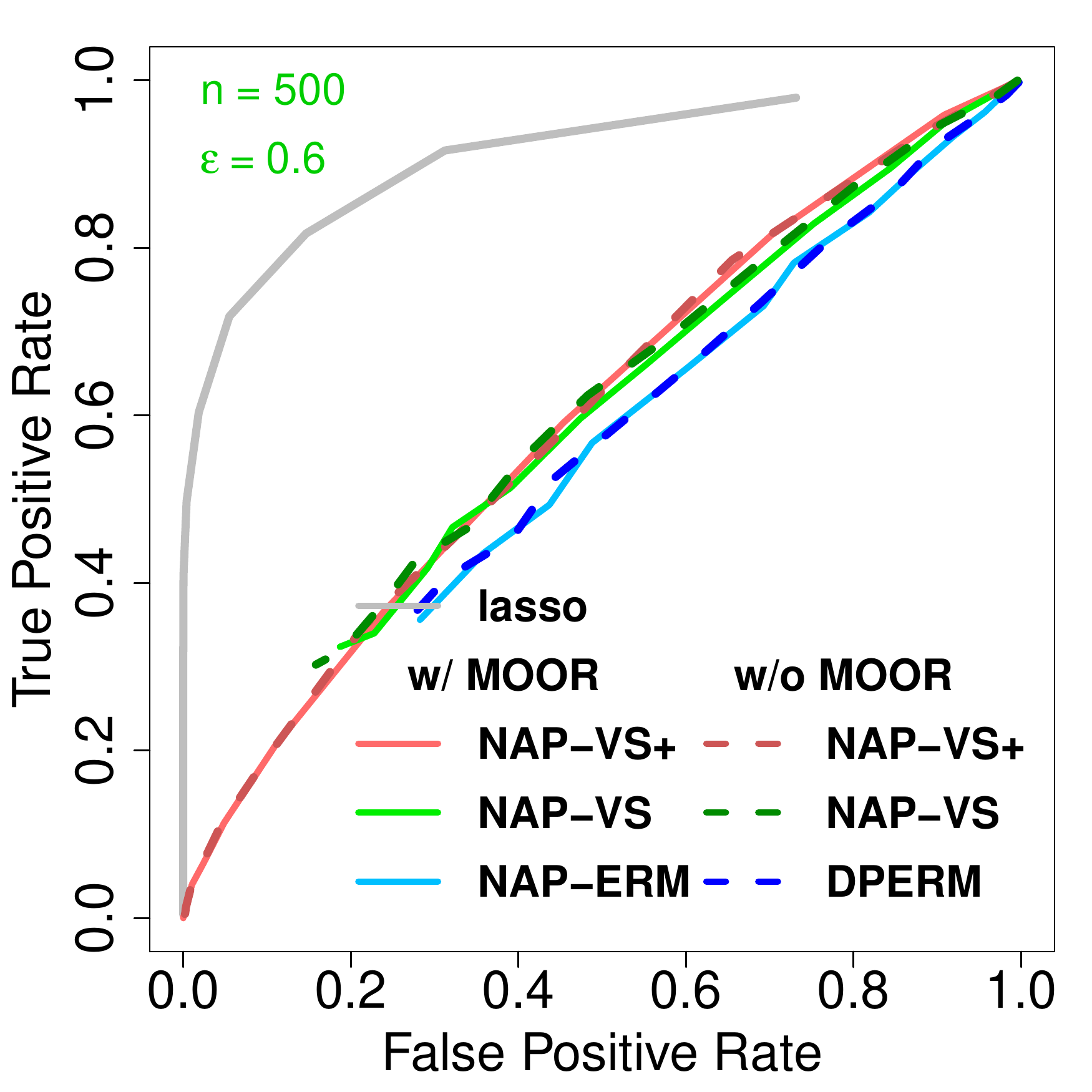}
\includegraphics[width=0.20\linewidth, trim=4pt 9pt 15pt 18pt,clip]{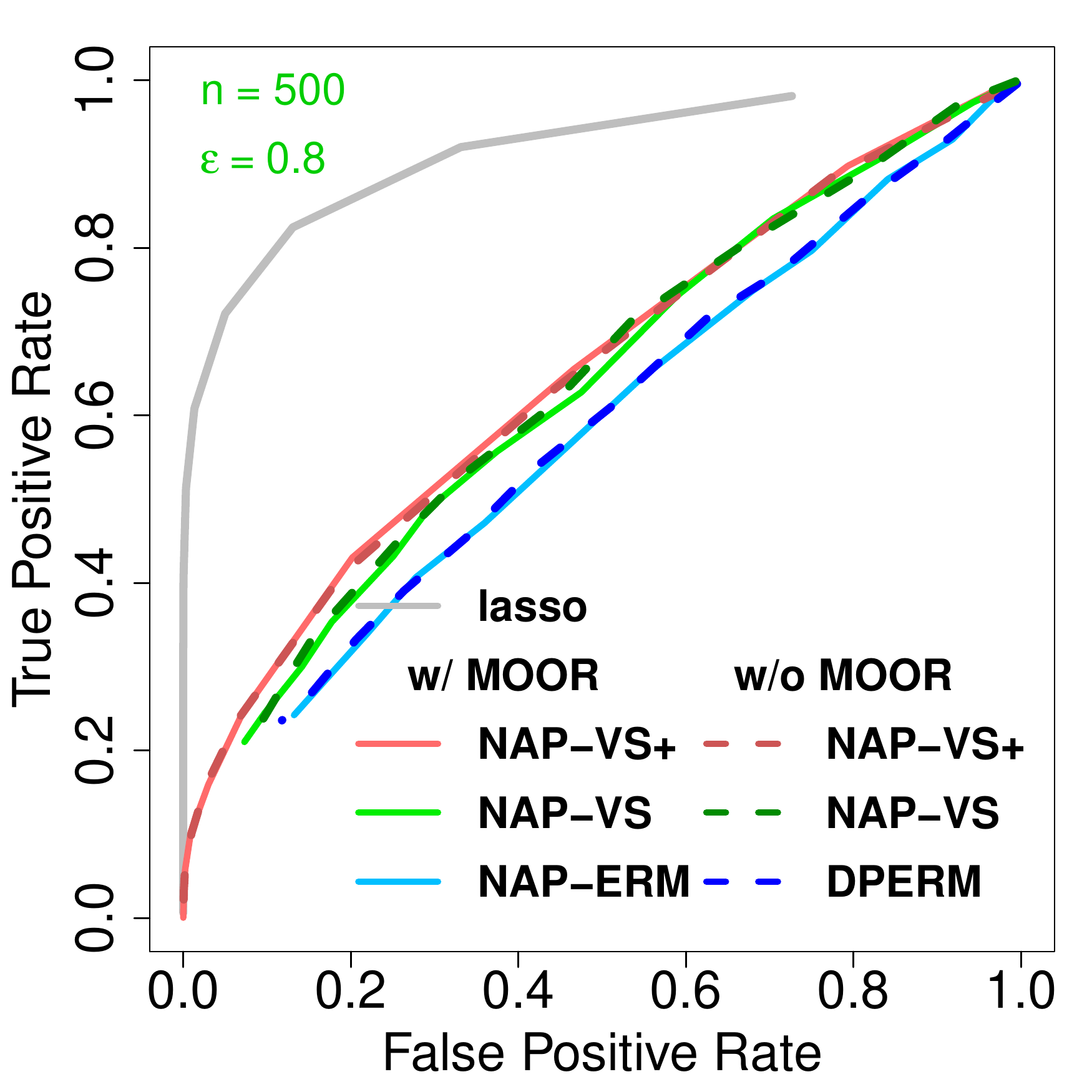}
\includegraphics[width=0.20\linewidth, trim=4pt 9pt 15pt 18pt,clip]{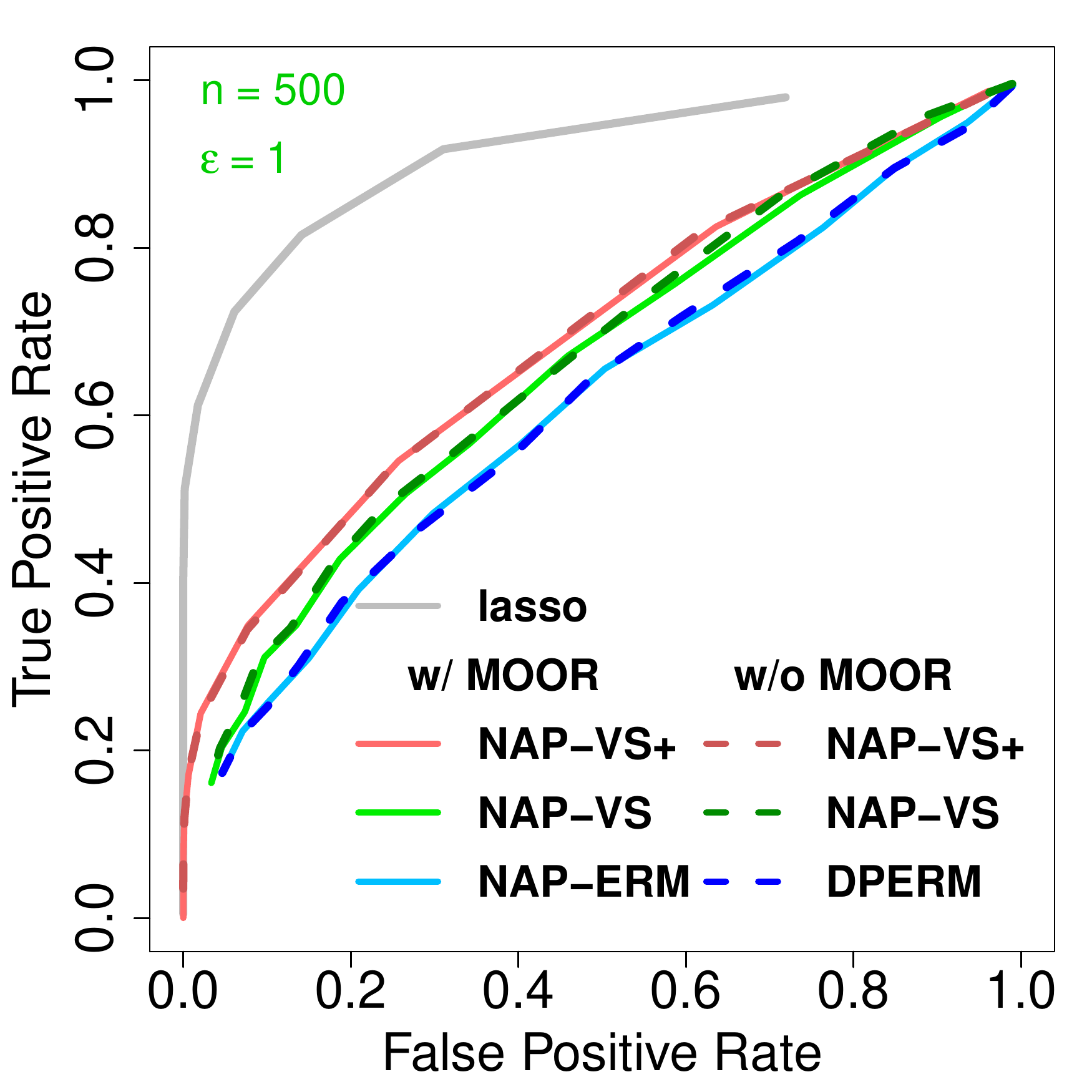}\\
\footnotesize logistic regression  $n=1000$ \\
\includegraphics[width=0.20\linewidth, trim=4pt 9pt 15pt 18pt,clip]{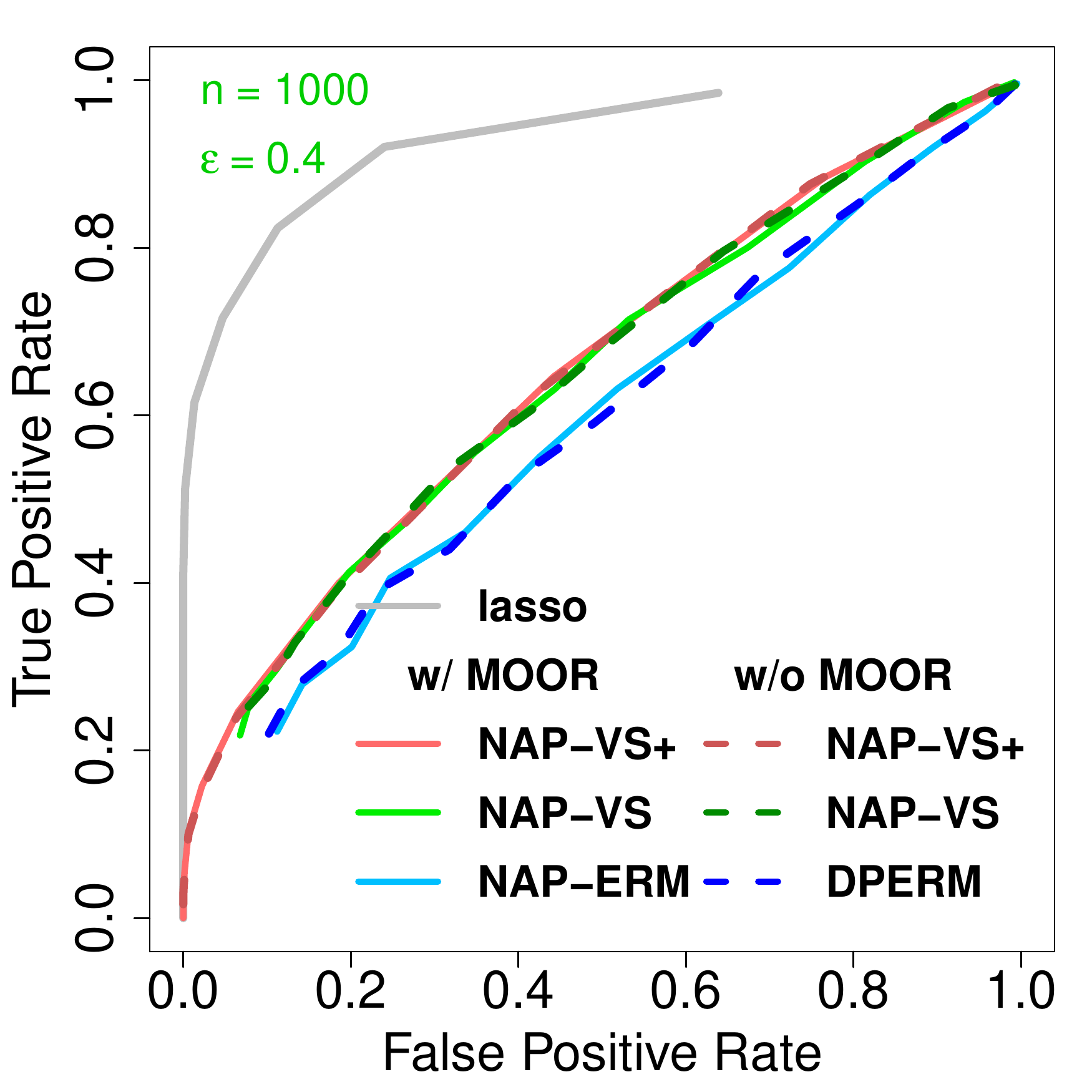}
\includegraphics[width=0.20\linewidth, trim=4pt 9pt 15pt 18pt,clip]{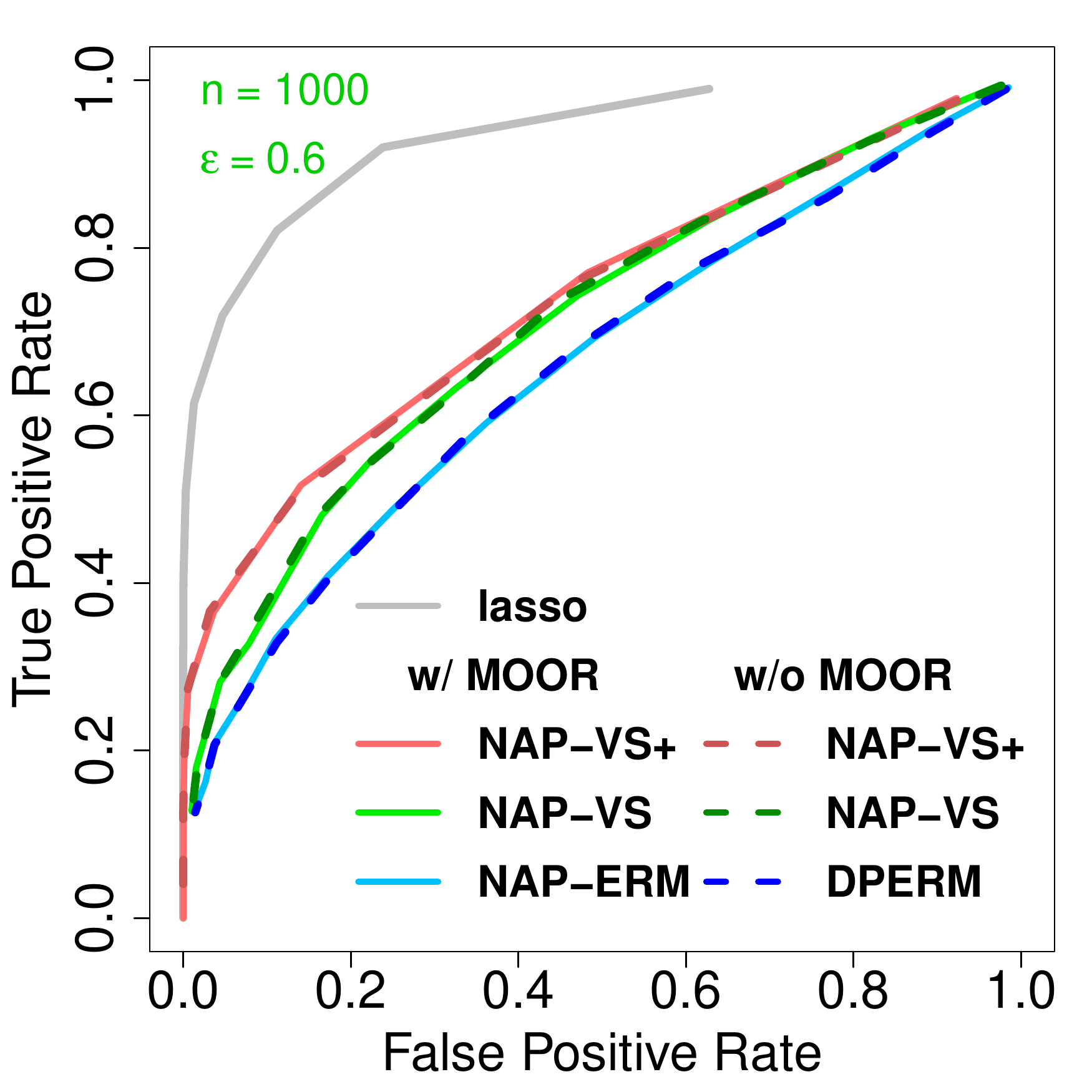}
\includegraphics[width=0.20\linewidth, trim=4pt 9pt 15pt 18pt,clip]{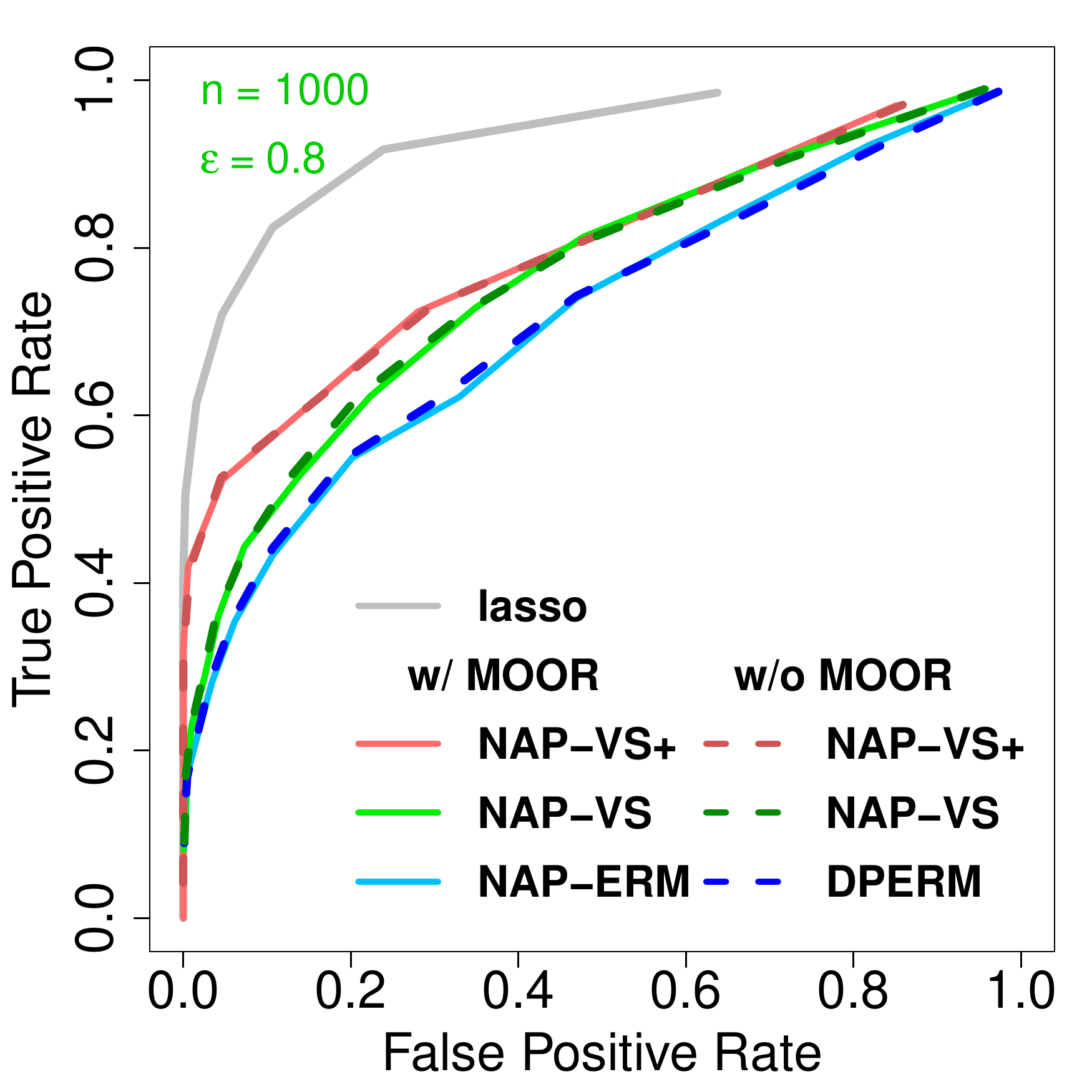}
\includegraphics[width=0.20\linewidth, trim=4pt 9pt 15pt 18pt,clip]{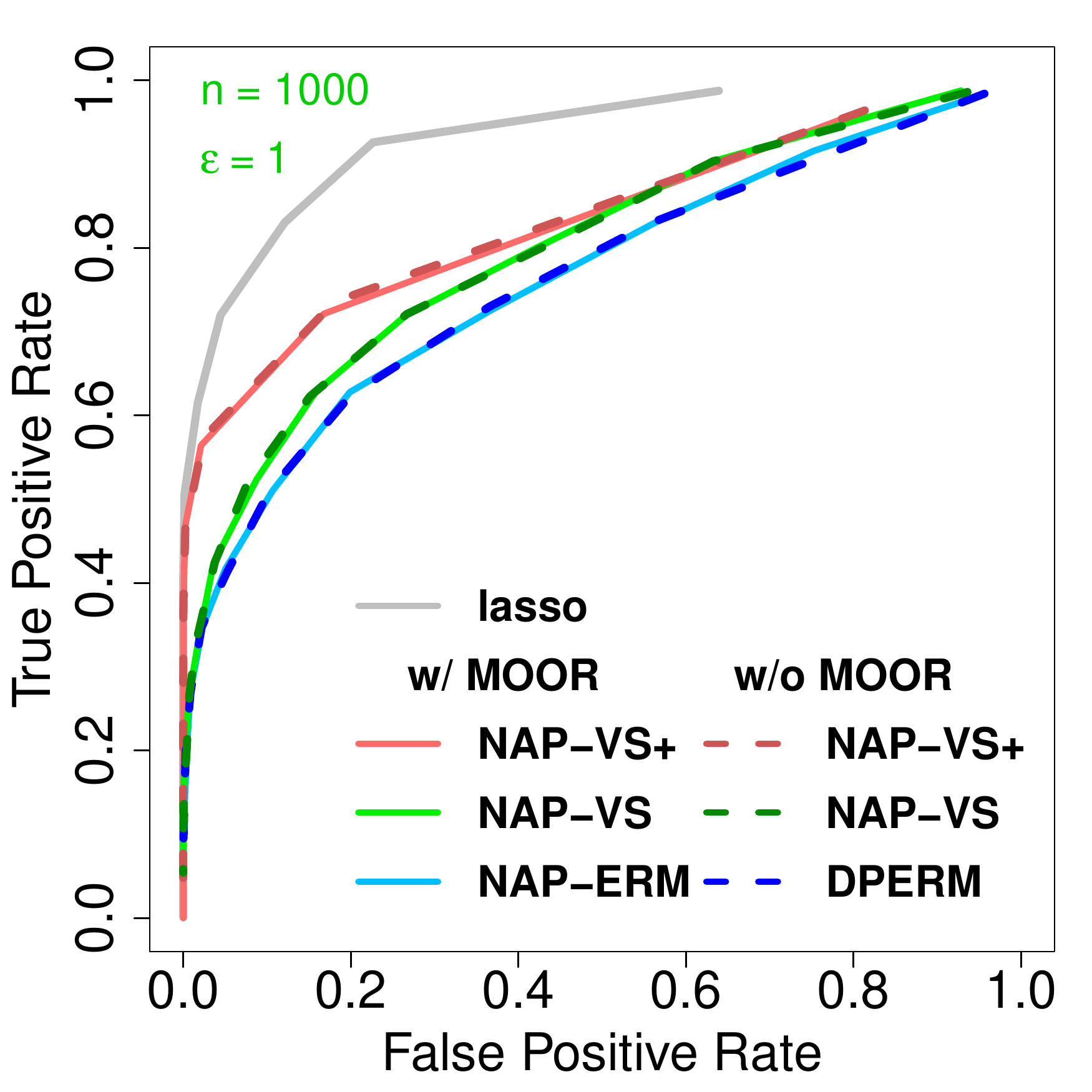}\\
\footnotesize Poisson regression $n=500$ \\
\includegraphics[width=0.20\linewidth, trim=4pt 9pt 15pt 18pt,clip]{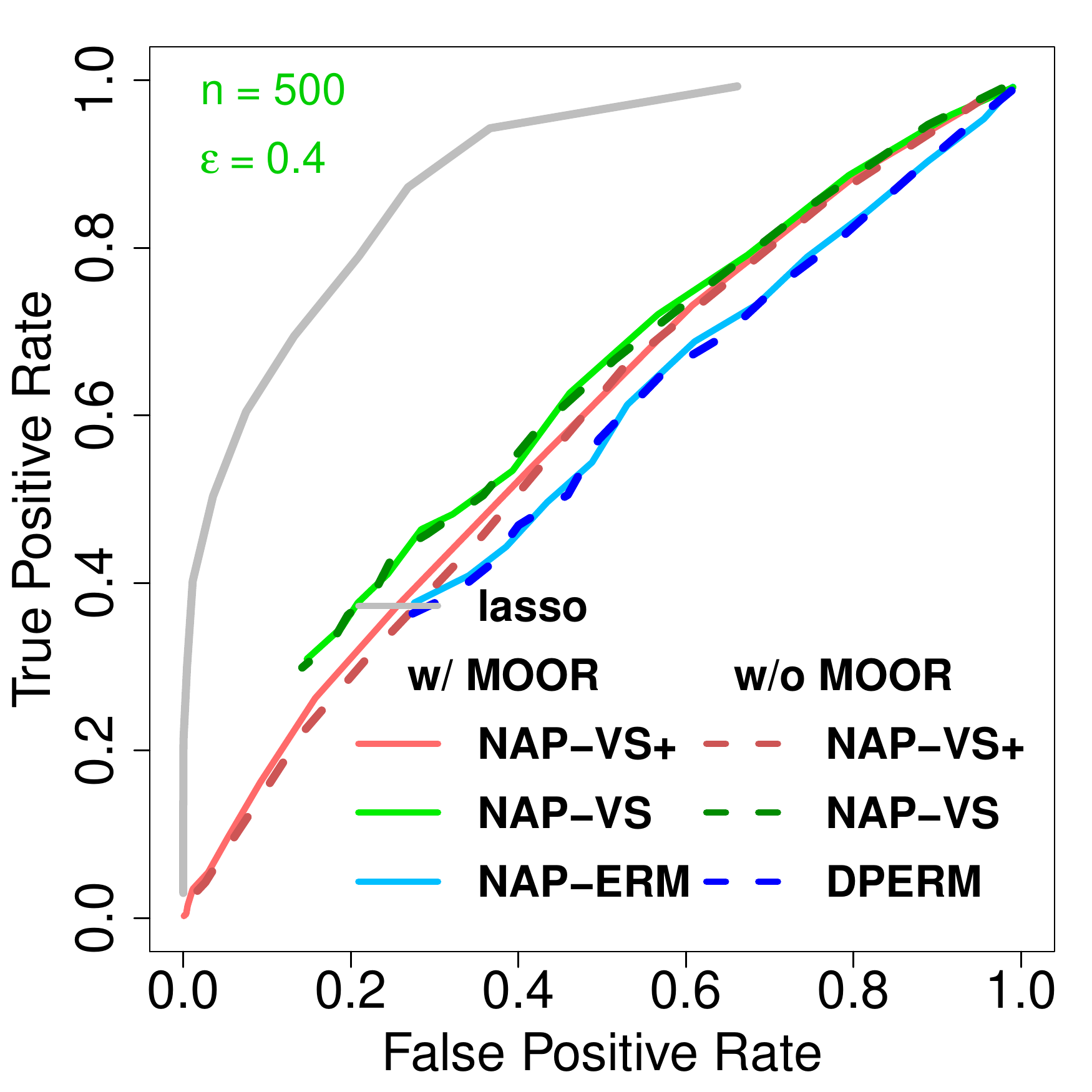}
\includegraphics[width=0.20\linewidth, trim=4pt 9pt 15pt 18pt,clip]{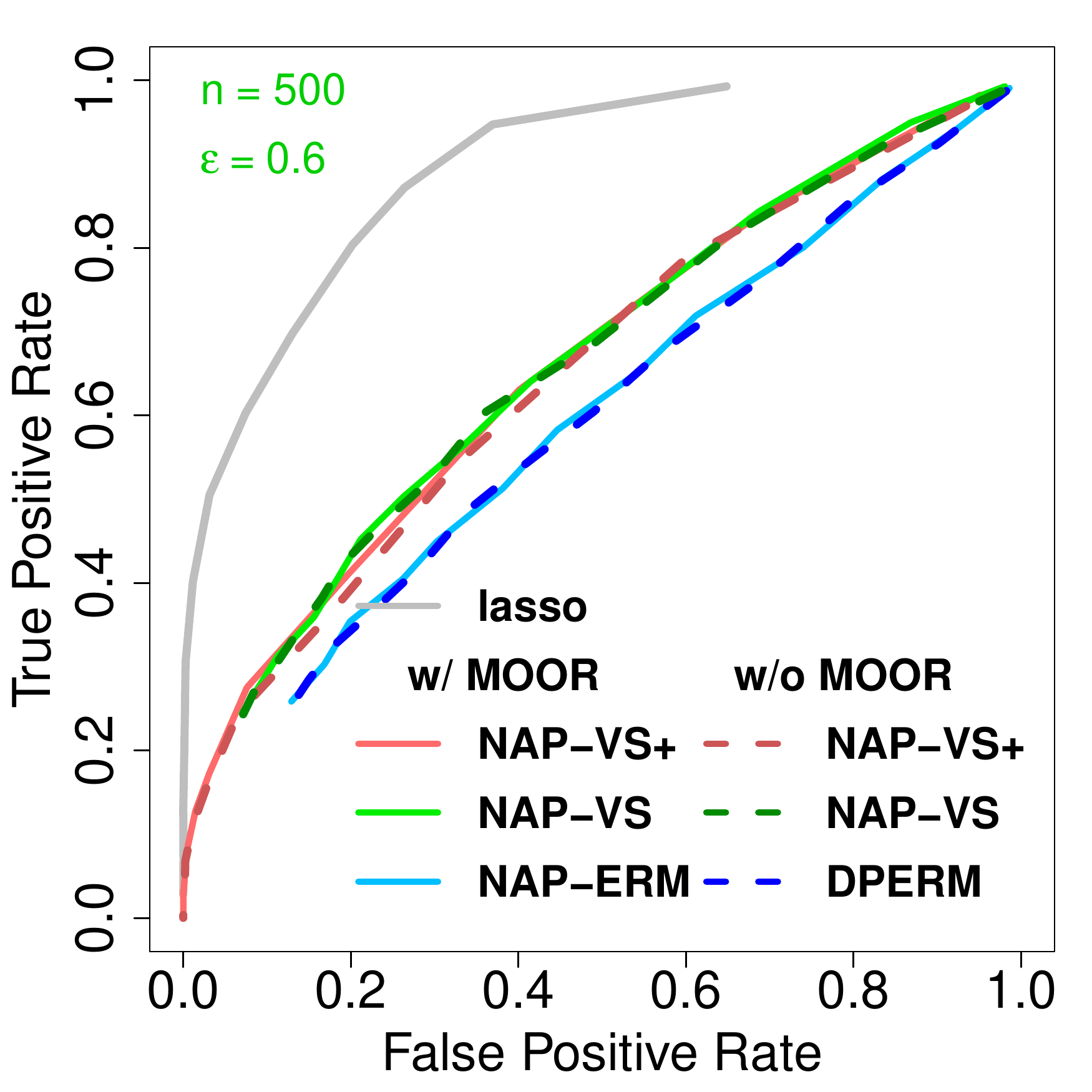}
\includegraphics[width=0.20\linewidth, trim=4pt 9pt 15pt 18pt,clip]{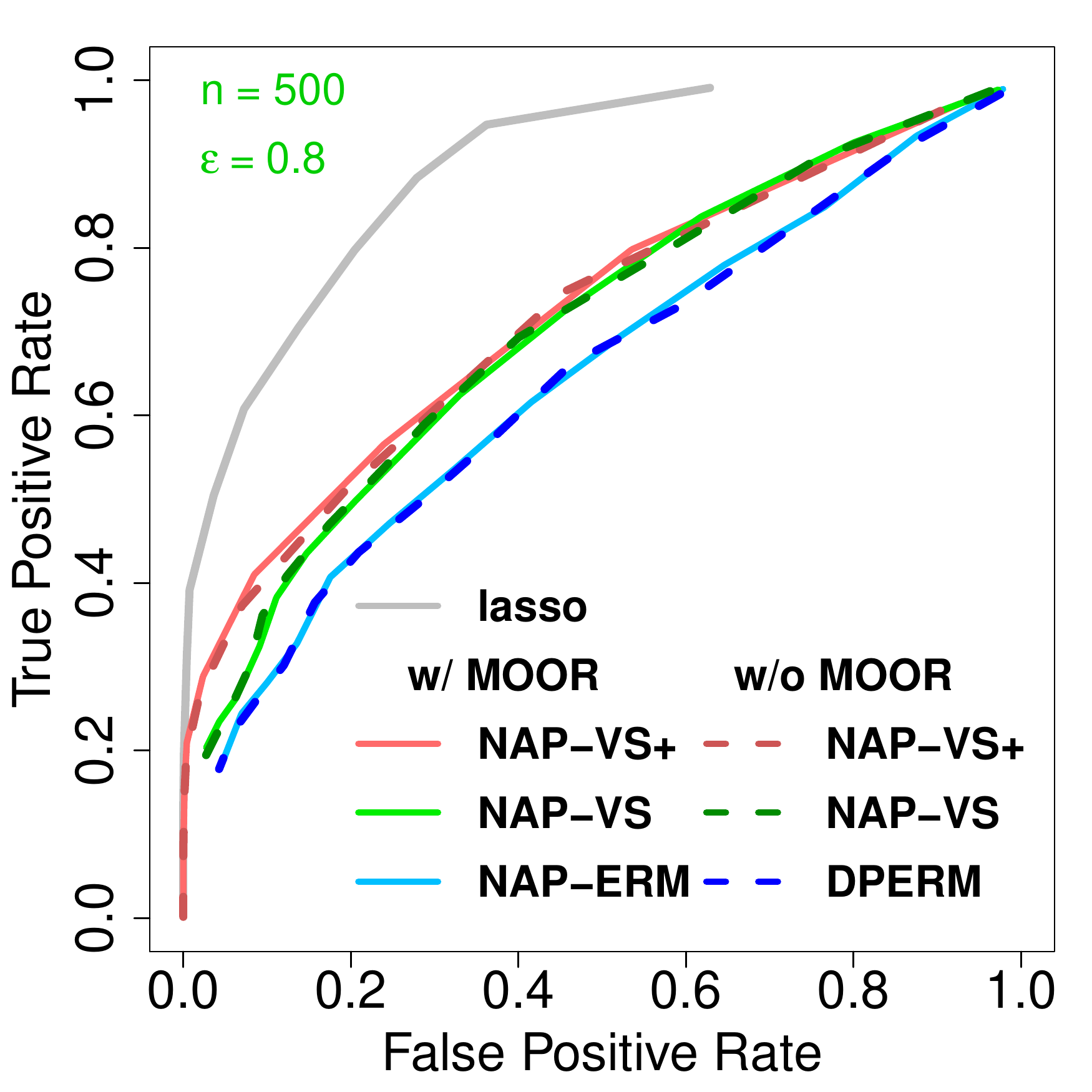}
\includegraphics[width=0.20\linewidth, trim=4pt 9pt 15pt 18pt,clip]{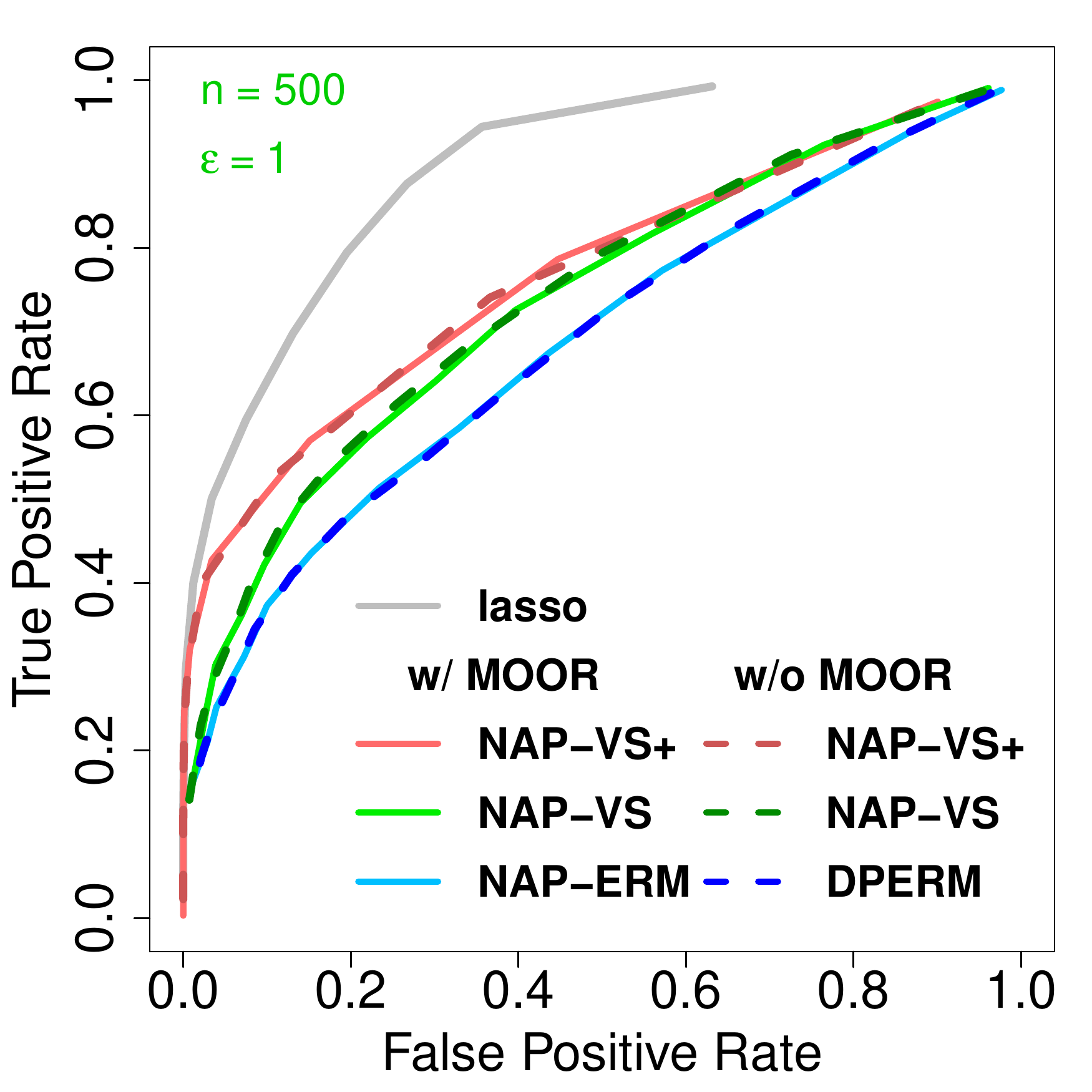}\\\vspace{-2pt}
\footnotesize Poisson regression $n=1000$\\
\includegraphics[width=0.20\linewidth, trim=4pt 9pt 15pt 18pt,clip]{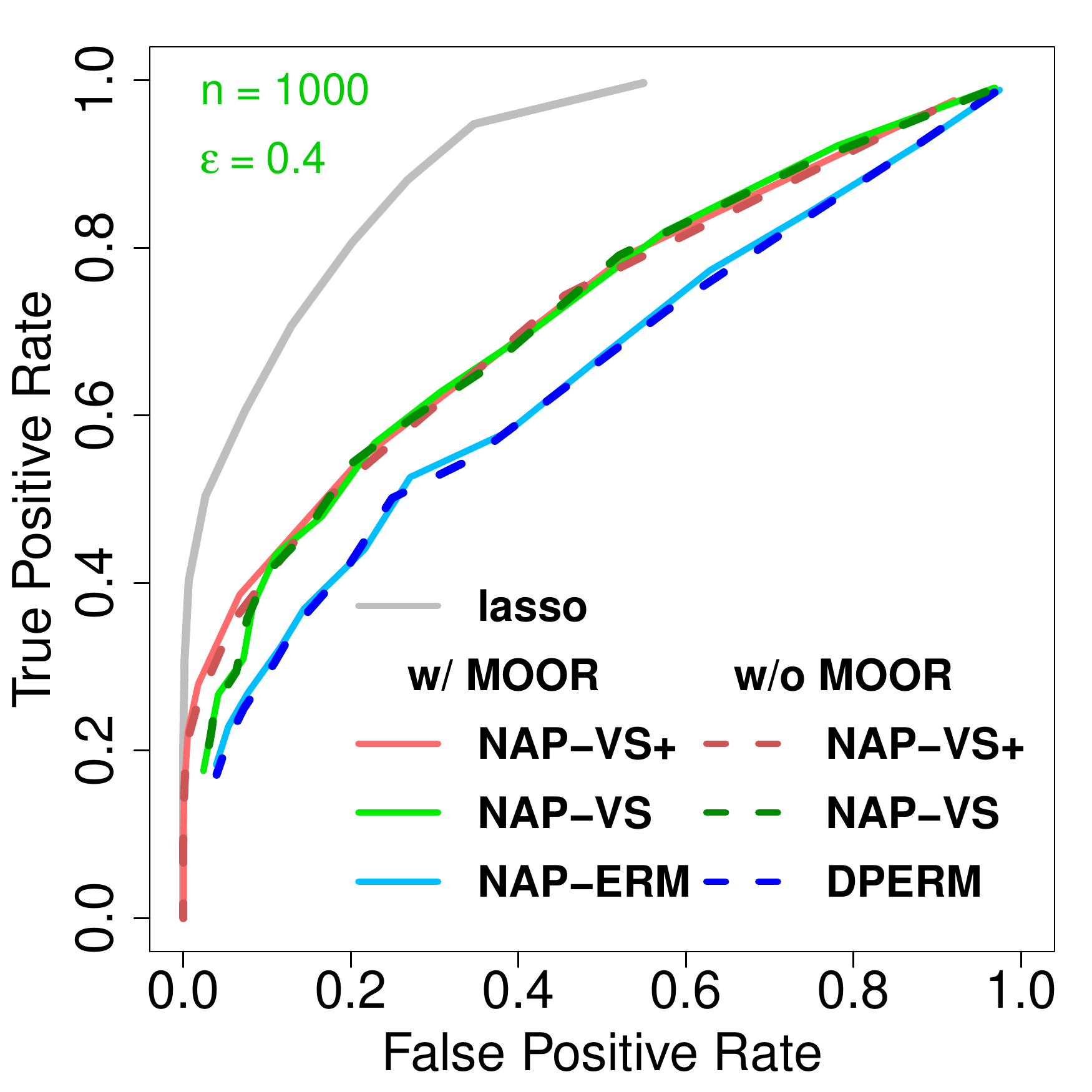}
\includegraphics[width=0.20\linewidth, trim=4pt 9pt 15pt 18pt,clip]{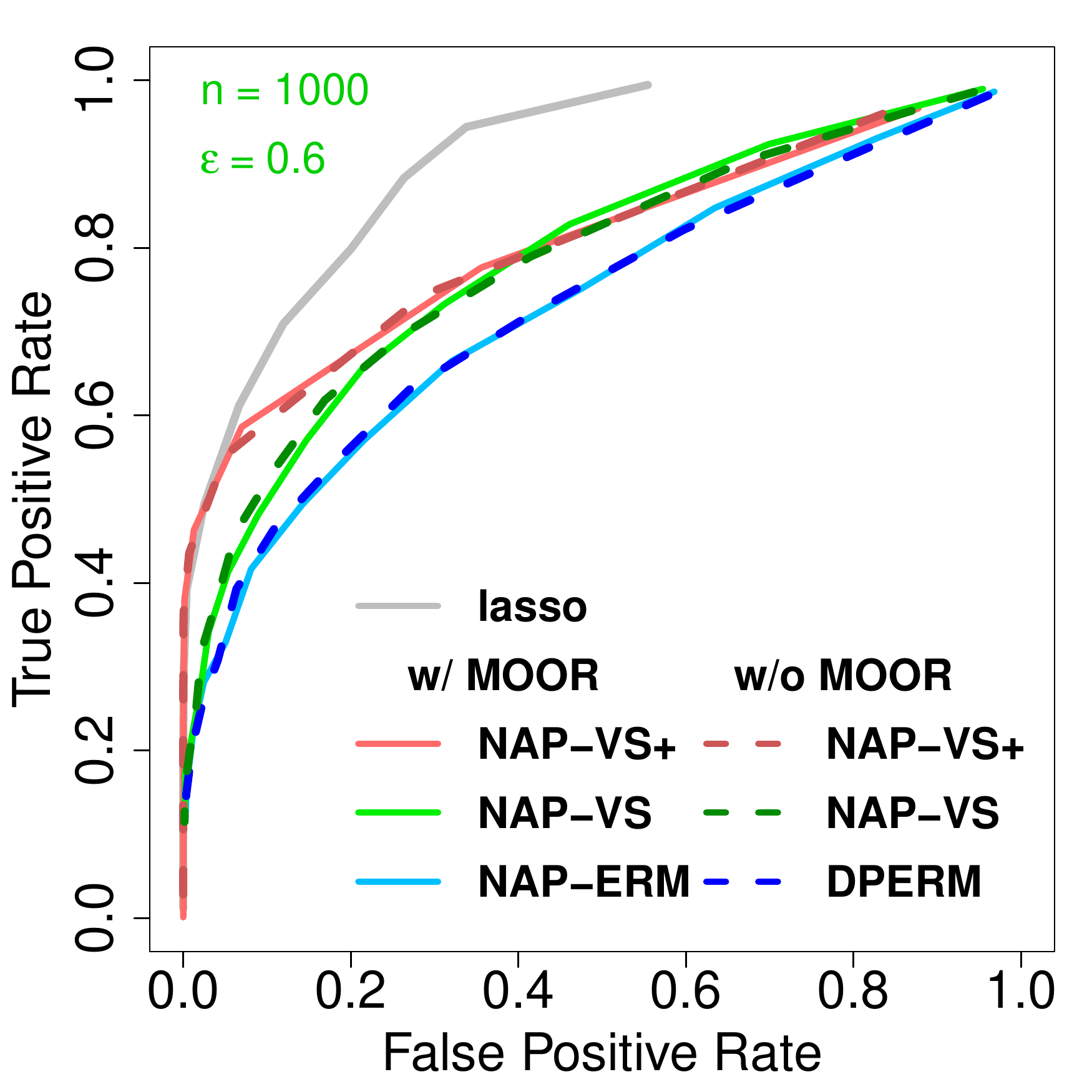}
\includegraphics[width=0.20\linewidth, trim=4pt 9pt 15pt 18pt,clip]{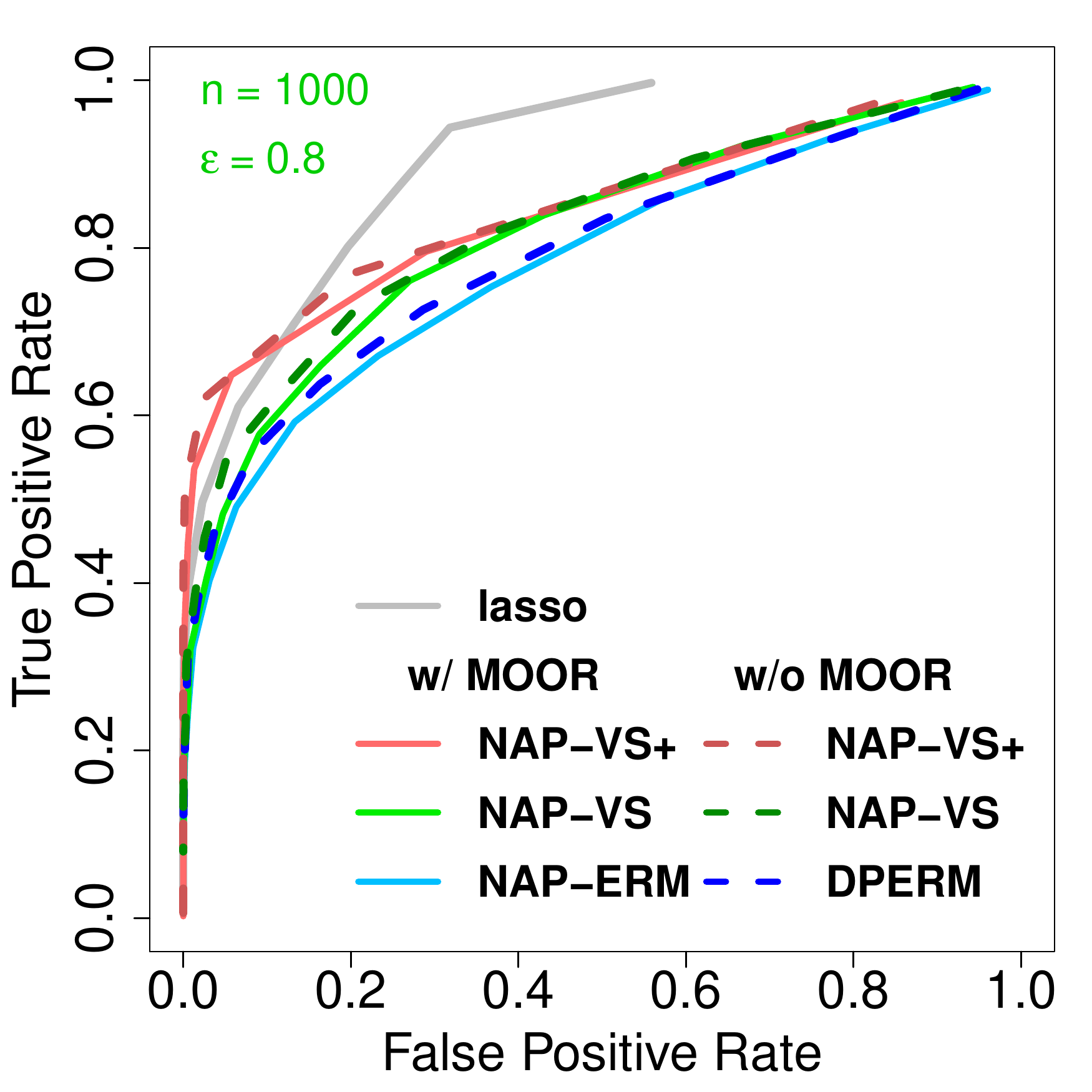}
\includegraphics[width=0.20\linewidth, trim=4pt 9pt 15pt 18pt,clip]{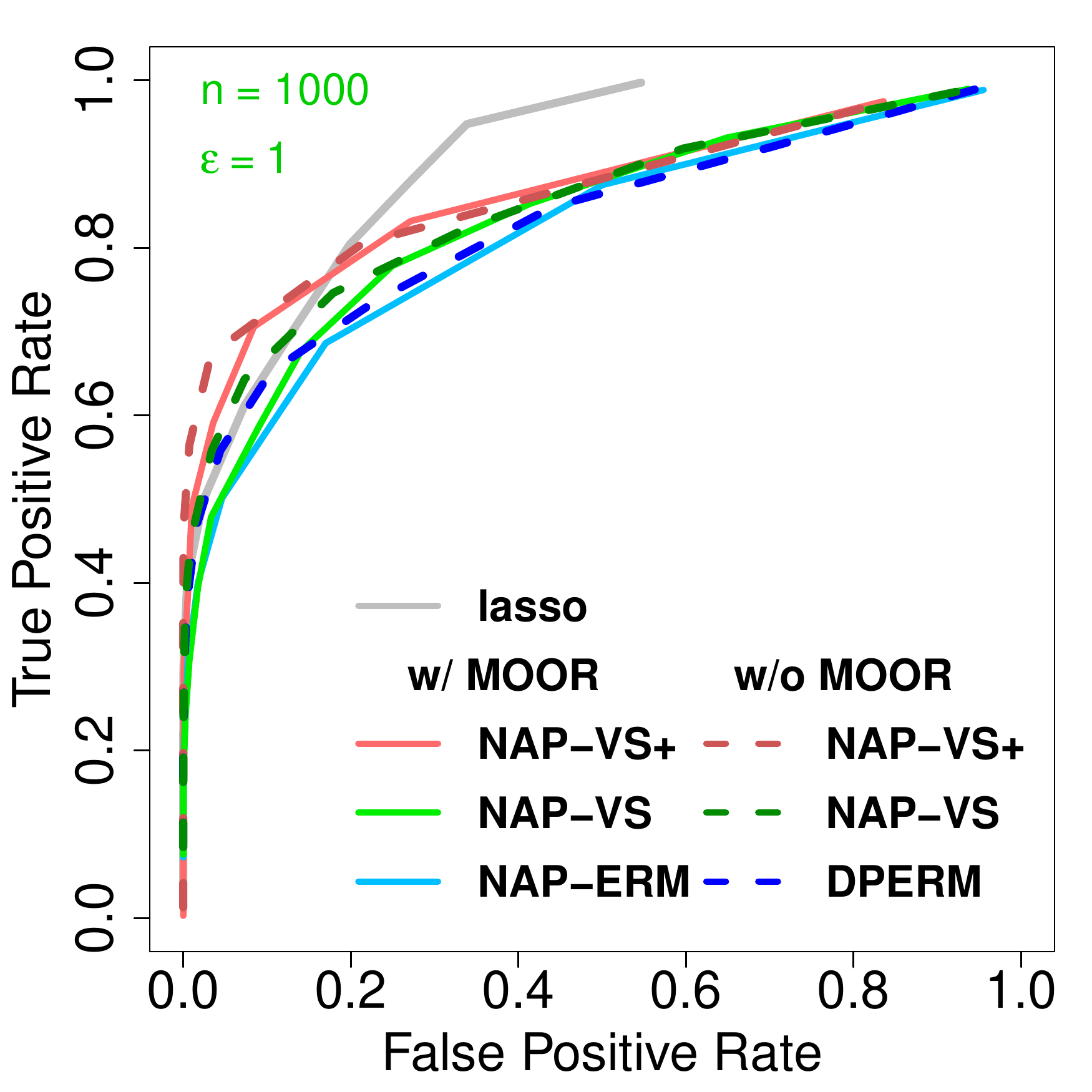}\\ \vspace{-12pt}
\caption{variable selection ROC by varying  $\Lambda$ at different $n$ and $\epsilon$} \label{fig:SIM2}\vspace{-12pt}
\end{figure}

The results on private variable selection are illustrated in Fig \ref{fig:SIM2} and  summarized below. 1) NAPP-VS+ performs better NAPP-VS, which is better than NAPP-ERM and the existing DP-ERM, when $n$ and $\epsilon$ are large. 2) NAPP-VS+ is the only private approach that leads to zero false positive rates (the corresponding ROC curves go through the origin $(0,0)$) when $n$ or $\epsilon$ is small, meaning that if all regression coefficients  are 0, NAPP-VS+ is able to find all of them when $\Lambda$ is large, whereas all the others (NAPP-VS, NAPP-ERM, current DP-ERM approach) will set at least coefficient at nonzero regardless of how large $\Lambda$ is. 3) There is not much a difference in variable selection with vs. without MOOR in this experiment (the dashed curves are similar to the solid curves of the same colors).

\begin{figure}[!htb]\centering
\footnotesize linear regression $n=200$\\\vspace{-1pt}
\includegraphics[width=0.19\linewidth, trim=4pt 9pt 15pt 18pt,clip]{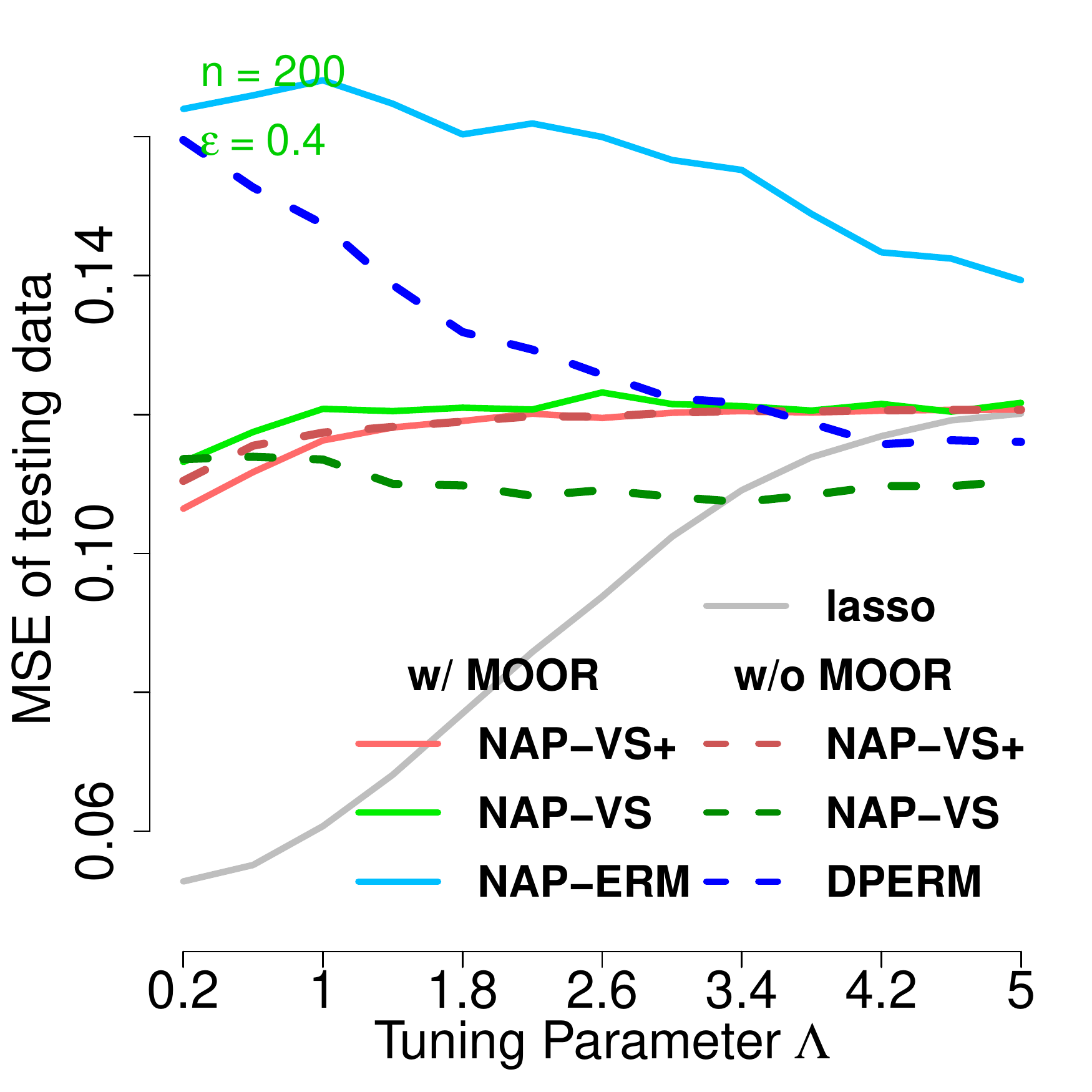}
\includegraphics[width=0.19\linewidth, trim=4pt 9pt 15pt 18pt,clip]{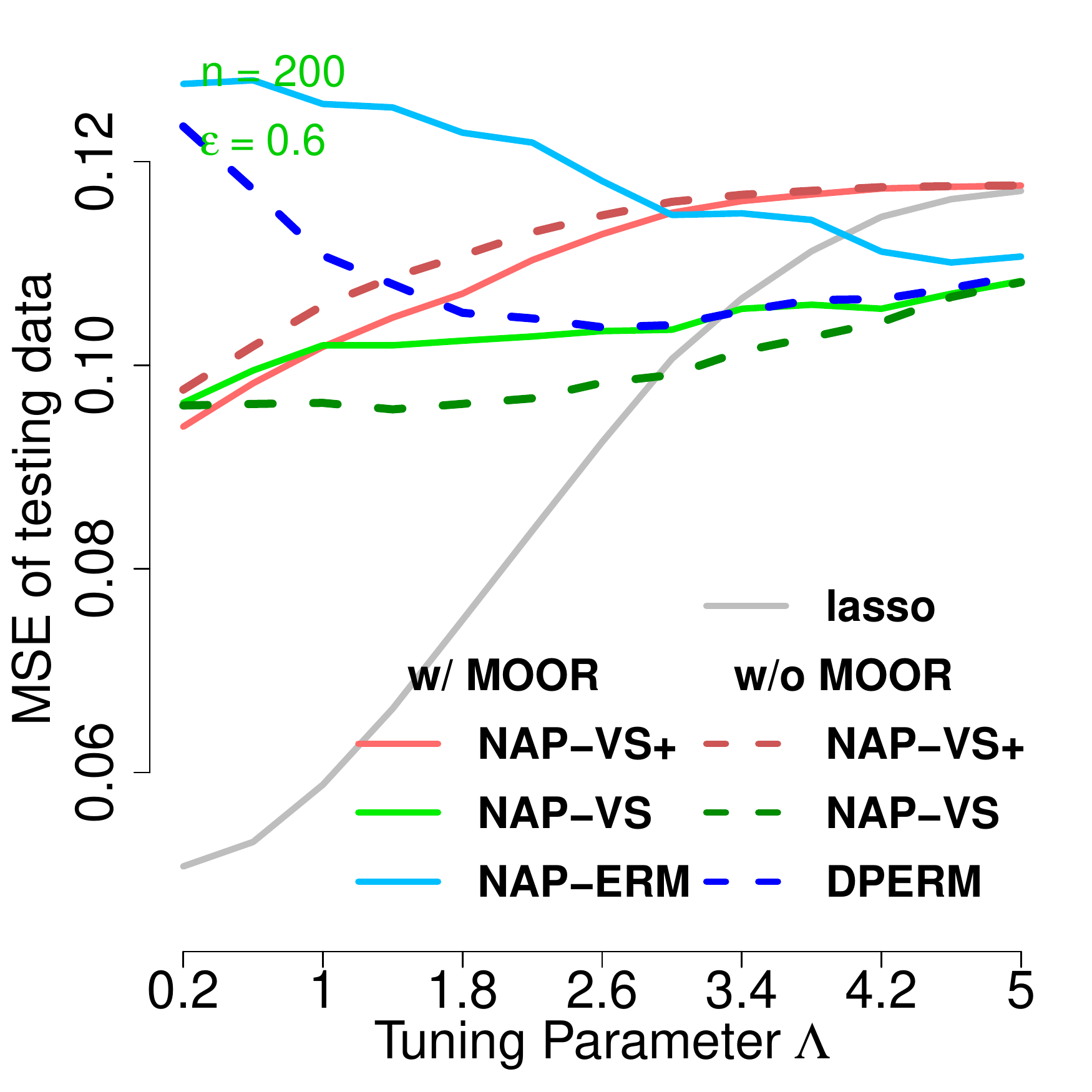}
\includegraphics[width=0.19\linewidth, trim=4pt 9pt 15pt 18pt,clip]{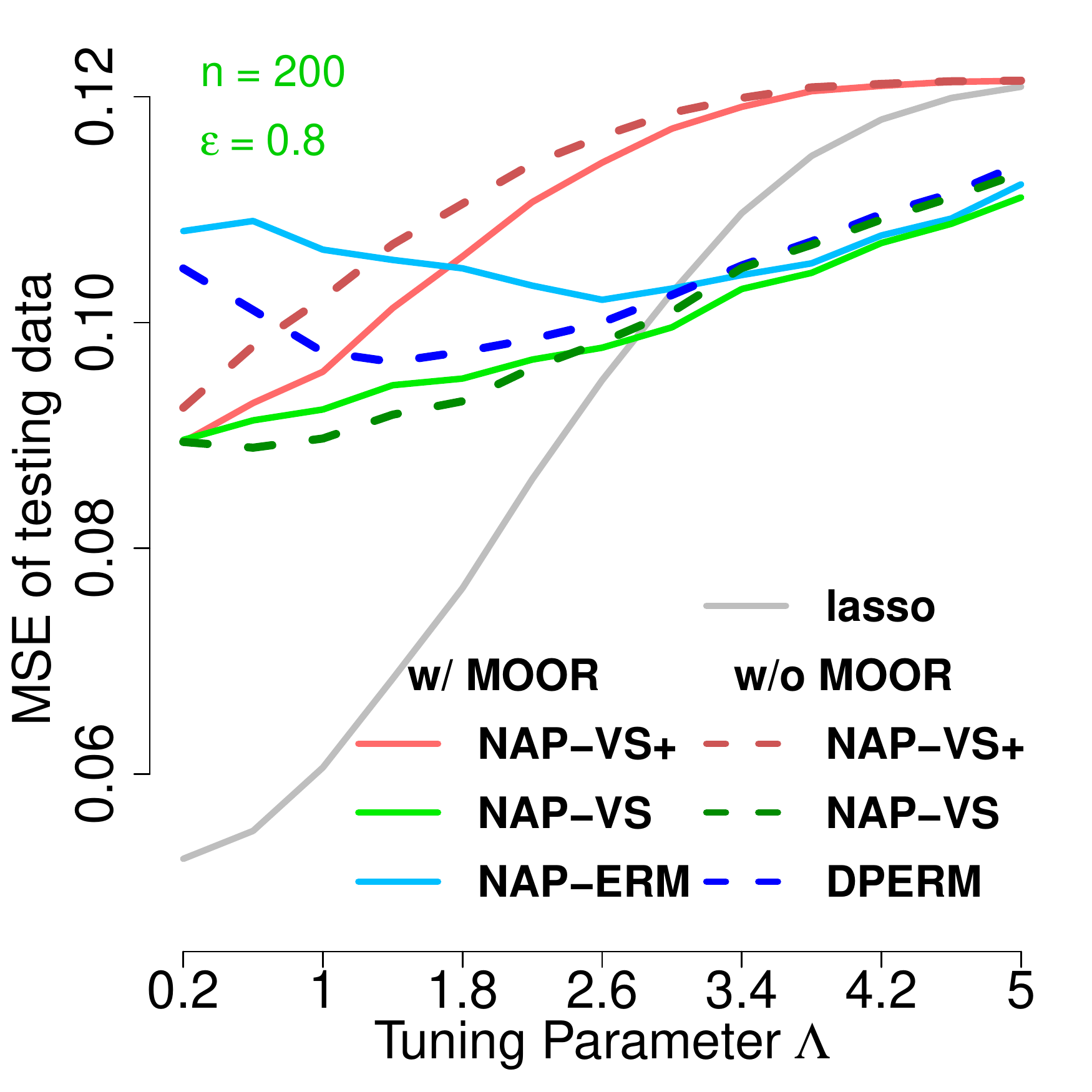}
\includegraphics[width=0.19\linewidth, trim=4pt 9pt 15pt 18pt,clip]{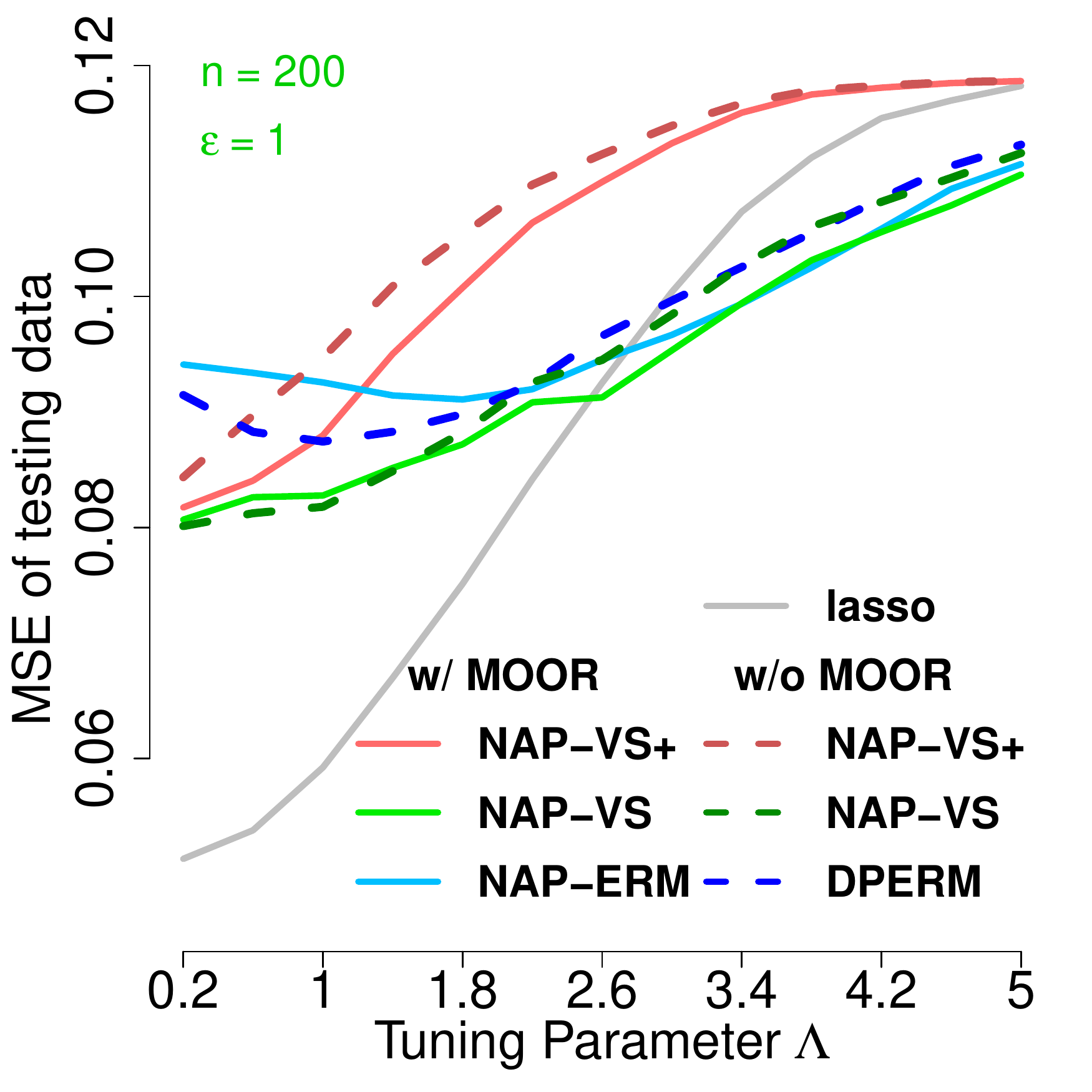}\\\vspace{-3pt}
\footnotesize linear regression $n=500$\\ \vspace{-1pt}
\includegraphics[width=0.19\linewidth, trim=4pt 9pt 15pt 18pt,clip]{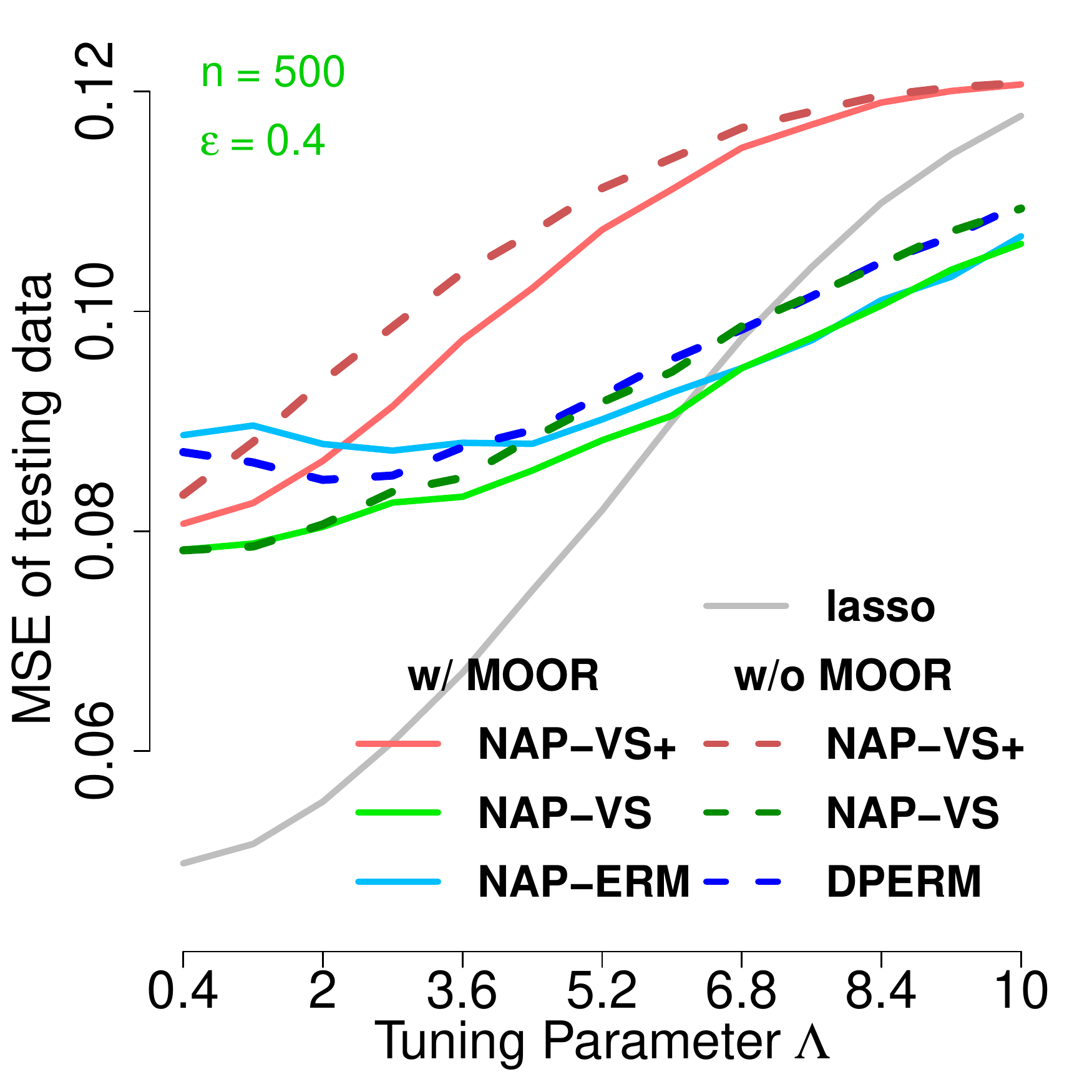}
\includegraphics[width=0.19\linewidth, trim=4pt 9pt 15pt 18pt,clip]{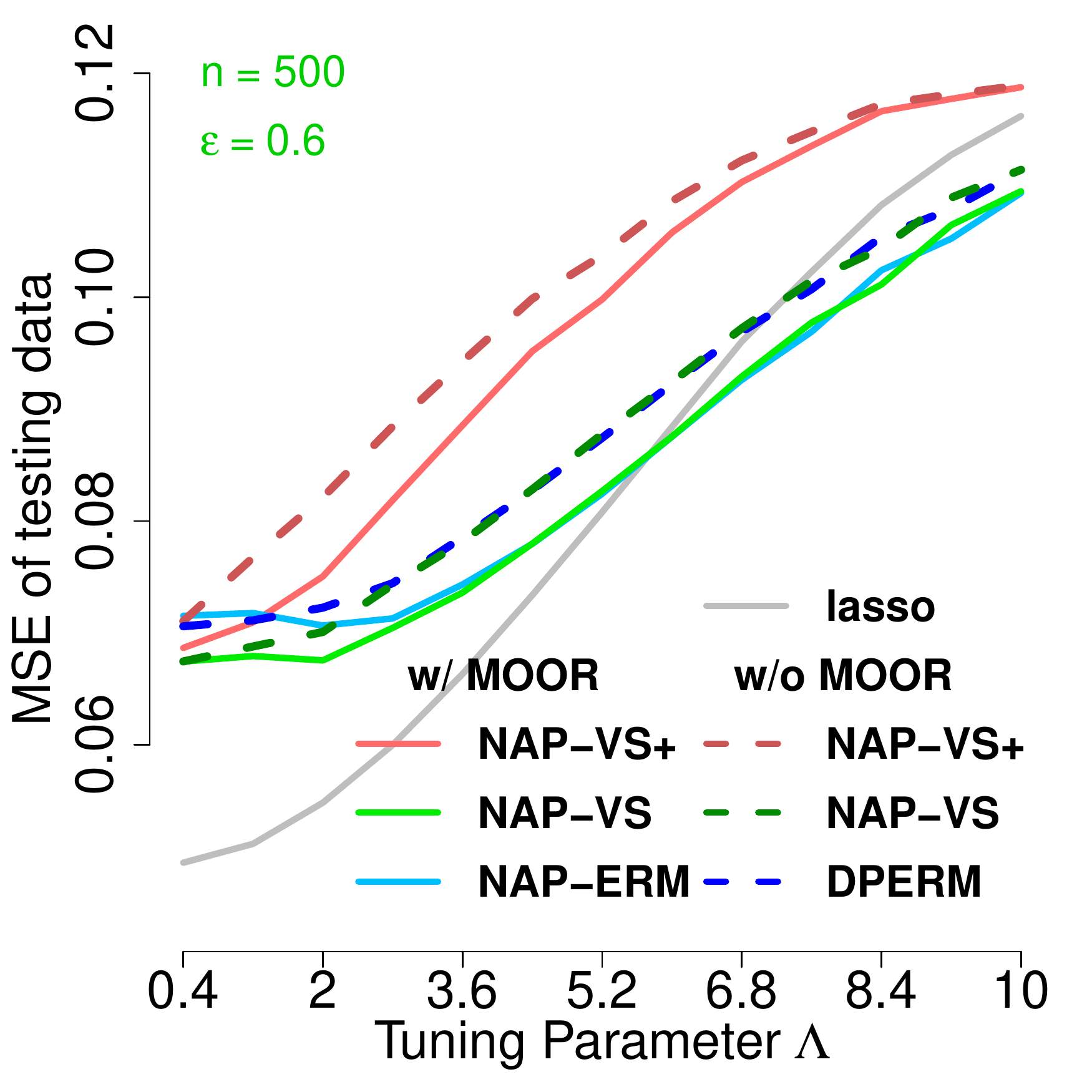}
\includegraphics[width=0.19\linewidth, trim=4pt 9pt 15pt 18pt,clip]{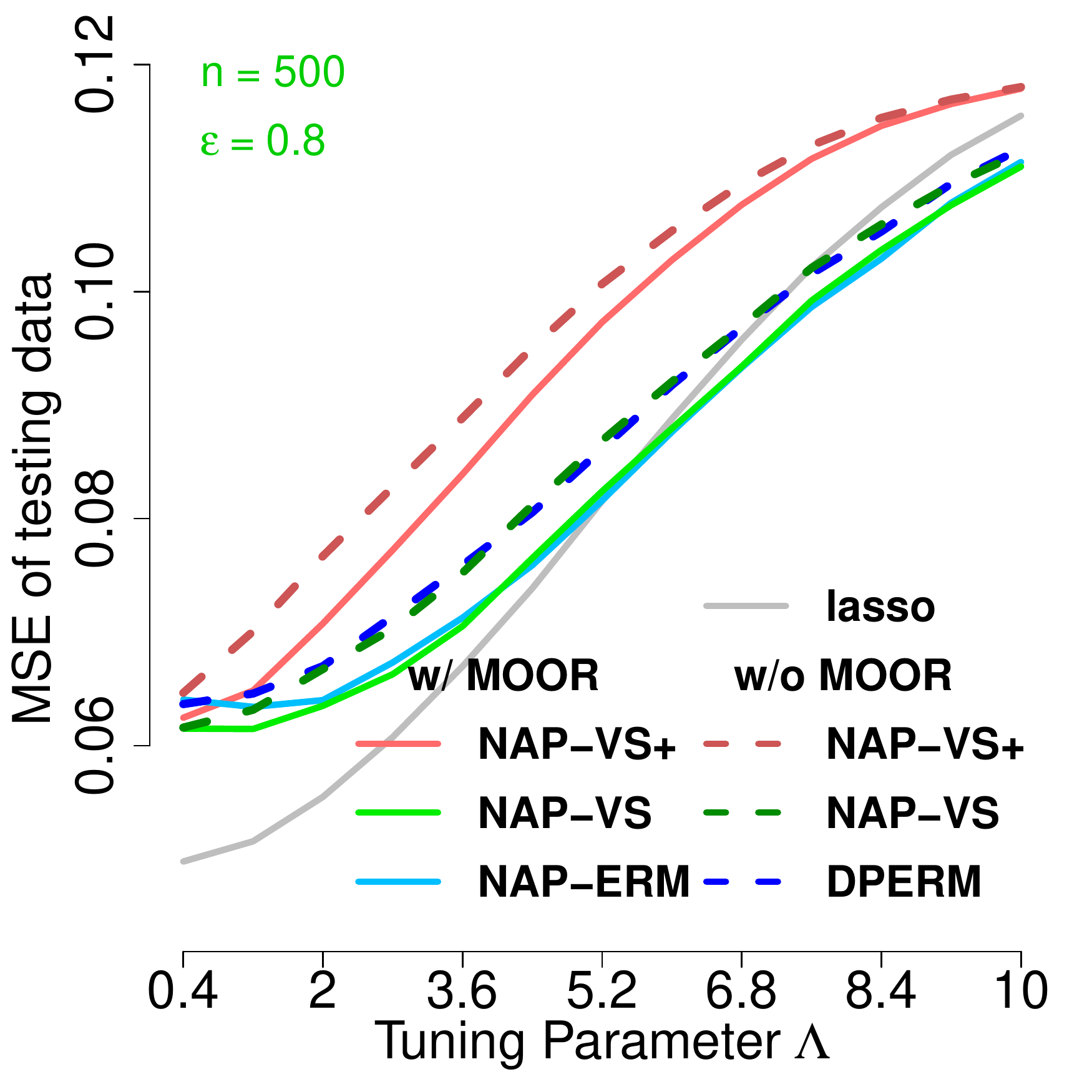}
\includegraphics[width=0.19\linewidth, trim=4pt 9pt 15pt 18pt,clip]{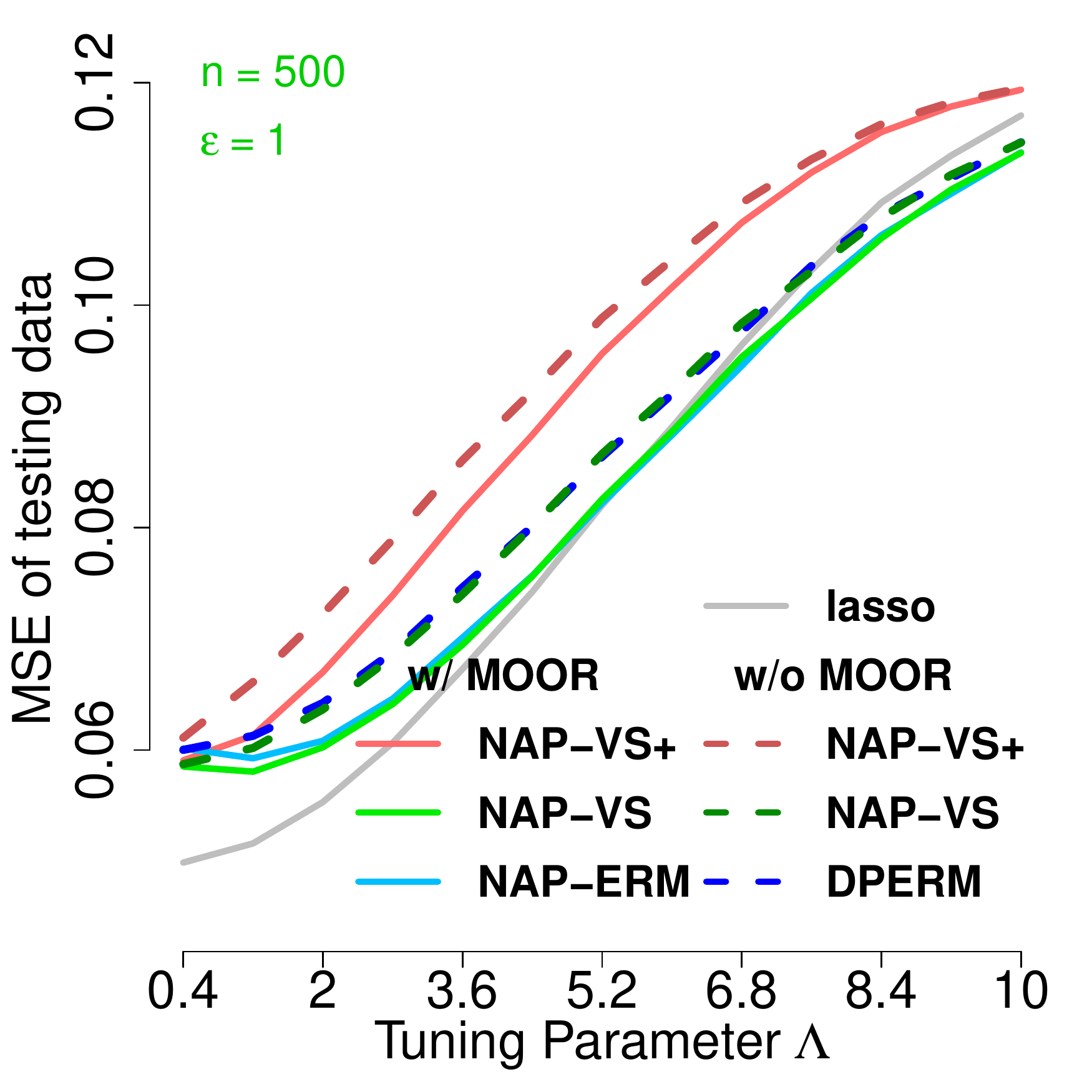}\\\vspace{-3pt}
\footnotesize Poisson regression $n=500$\\\vspace{-1pt}
\includegraphics[width=0.19\linewidth, trim=4pt 9pt 15pt 18pt,clip]{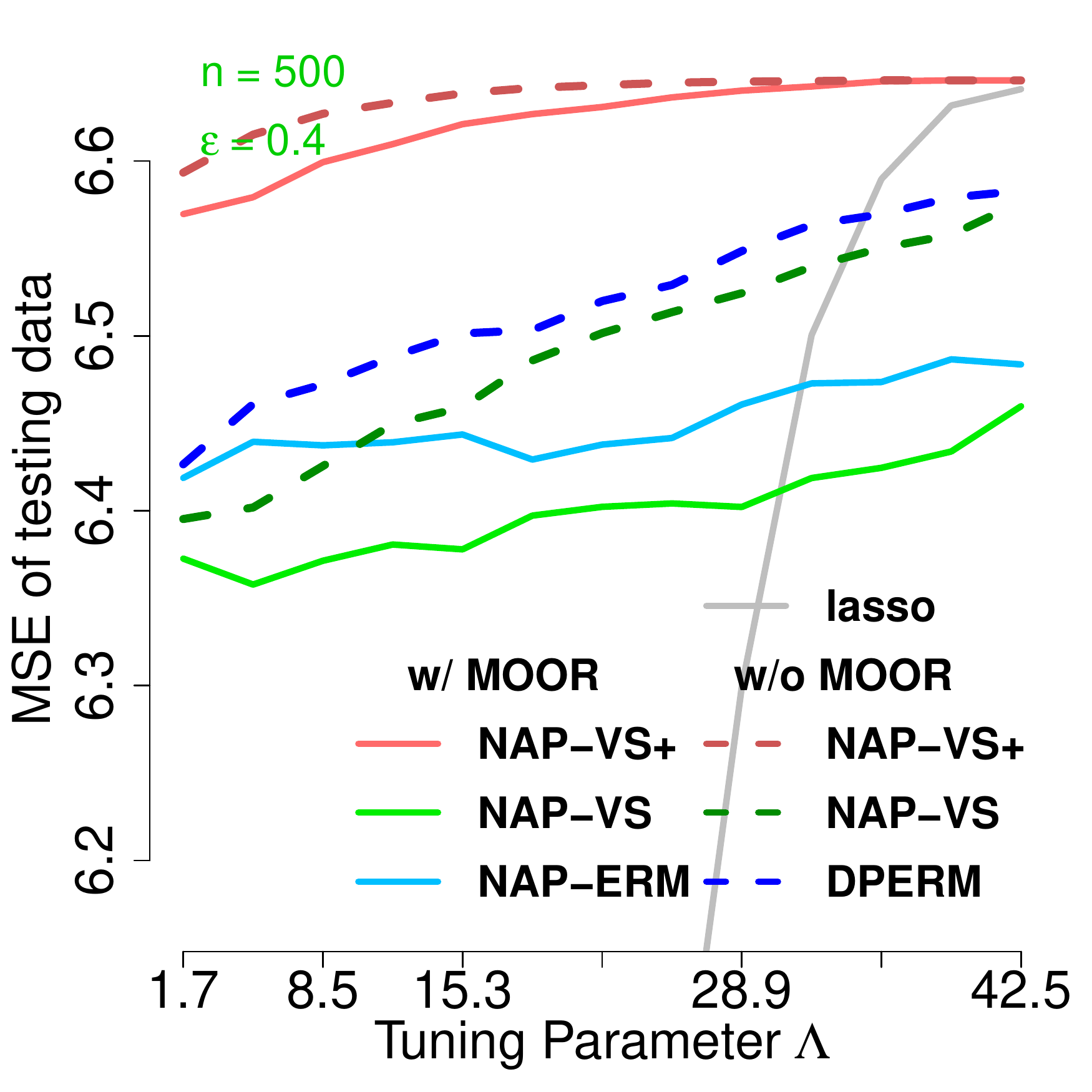}
\includegraphics[width=0.19\linewidth, trim=4pt 9pt 15pt 18pt,clip]{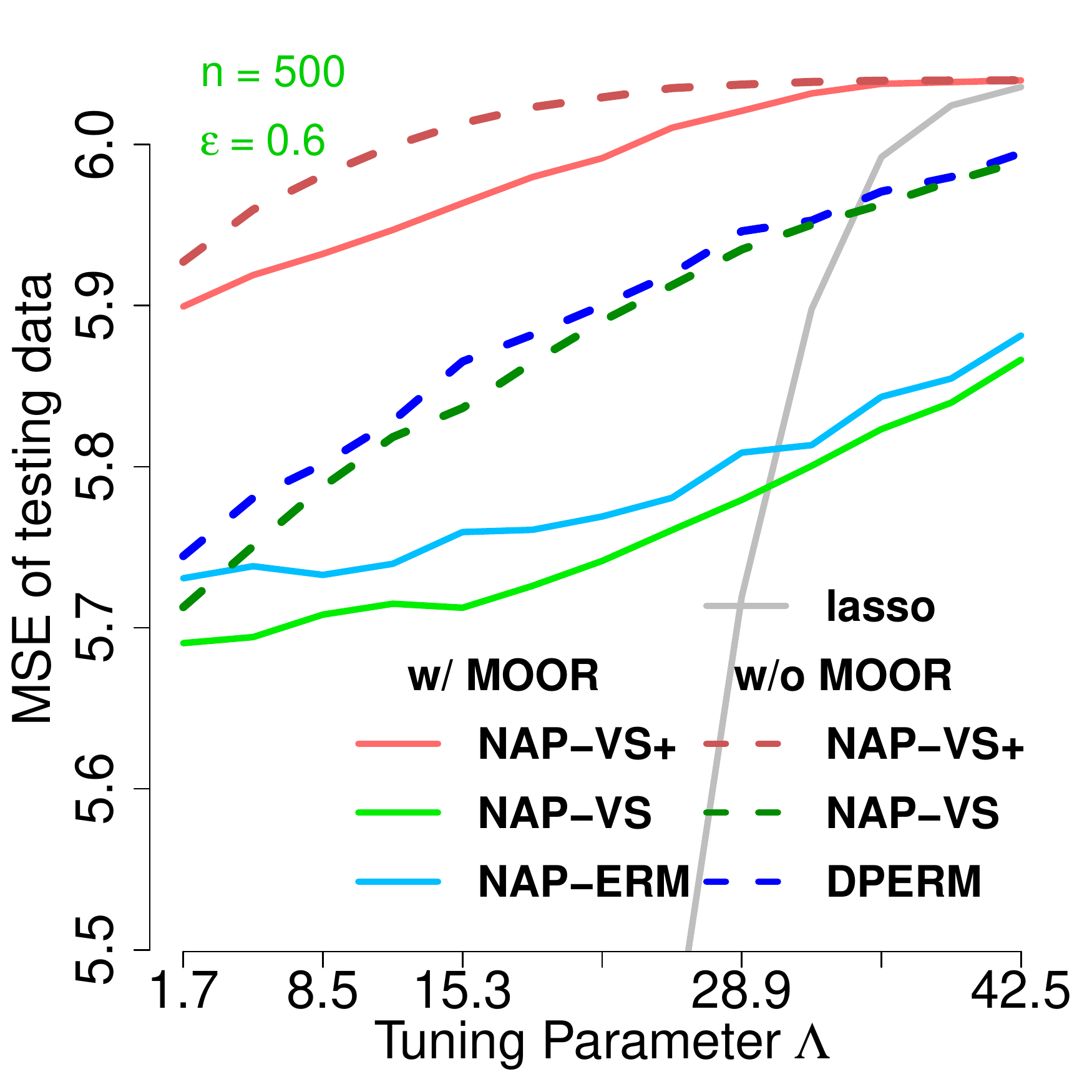}
\includegraphics[width=0.19\linewidth, trim=4pt 9pt 15pt 18pt,clip]{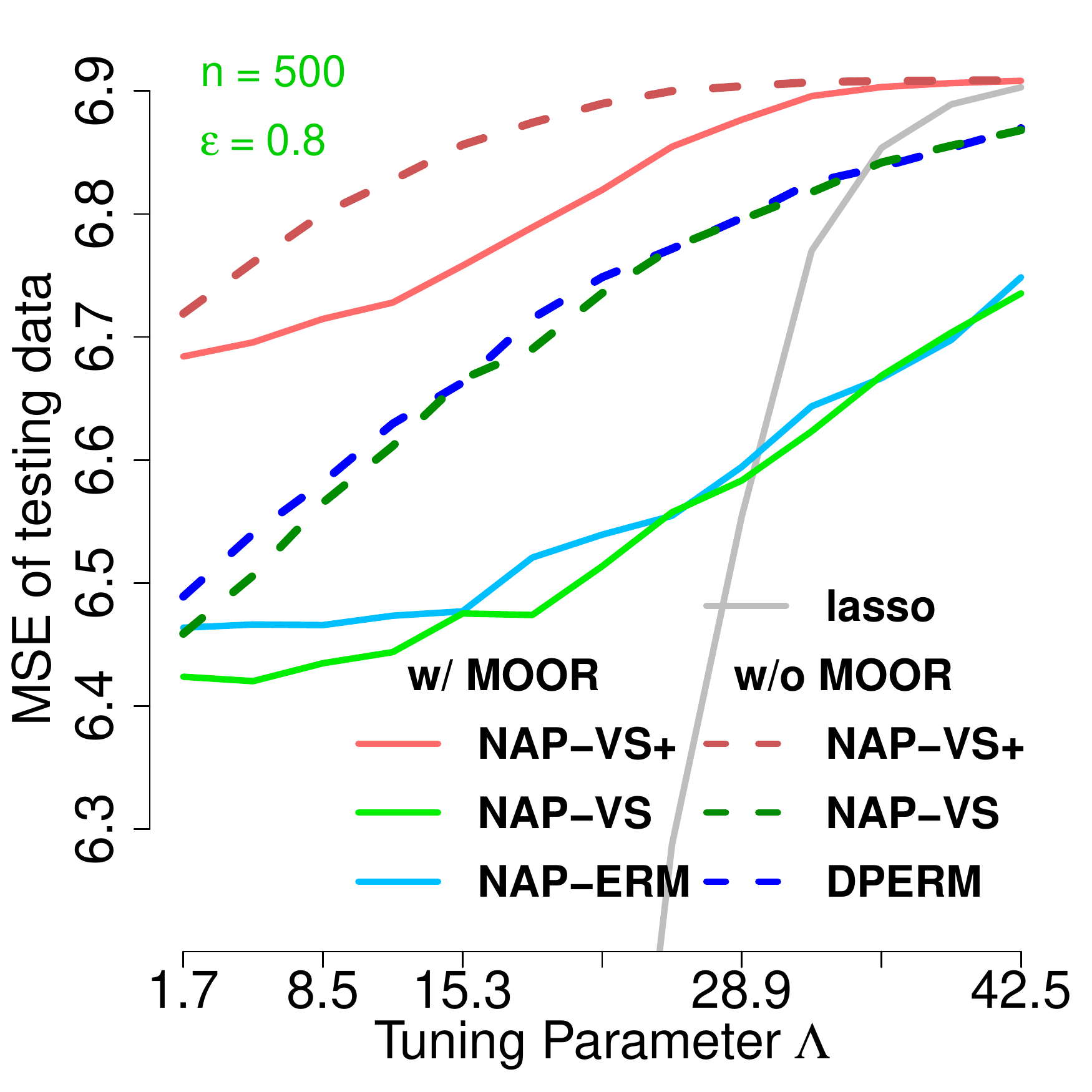}
\includegraphics[width=0.19\linewidth, trim=4pt 9pt 15pt 18pt,clip]{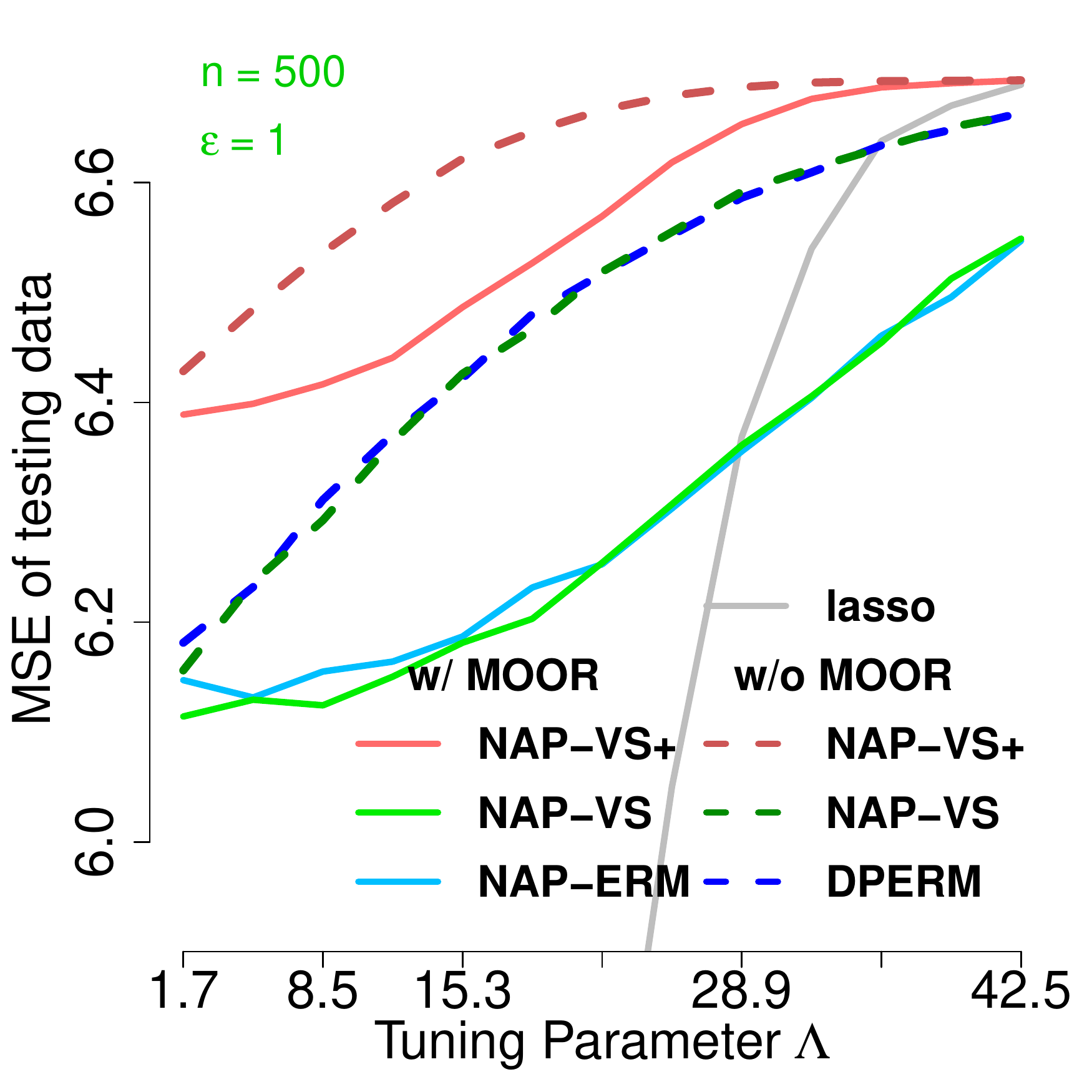}\\\vspace{-3pt}
\footnotesize  Poisson regression $n=1000$\\\vspace{-1pt}
\includegraphics[width=0.19\linewidth, trim=4pt 9pt 15pt 18pt,clip]{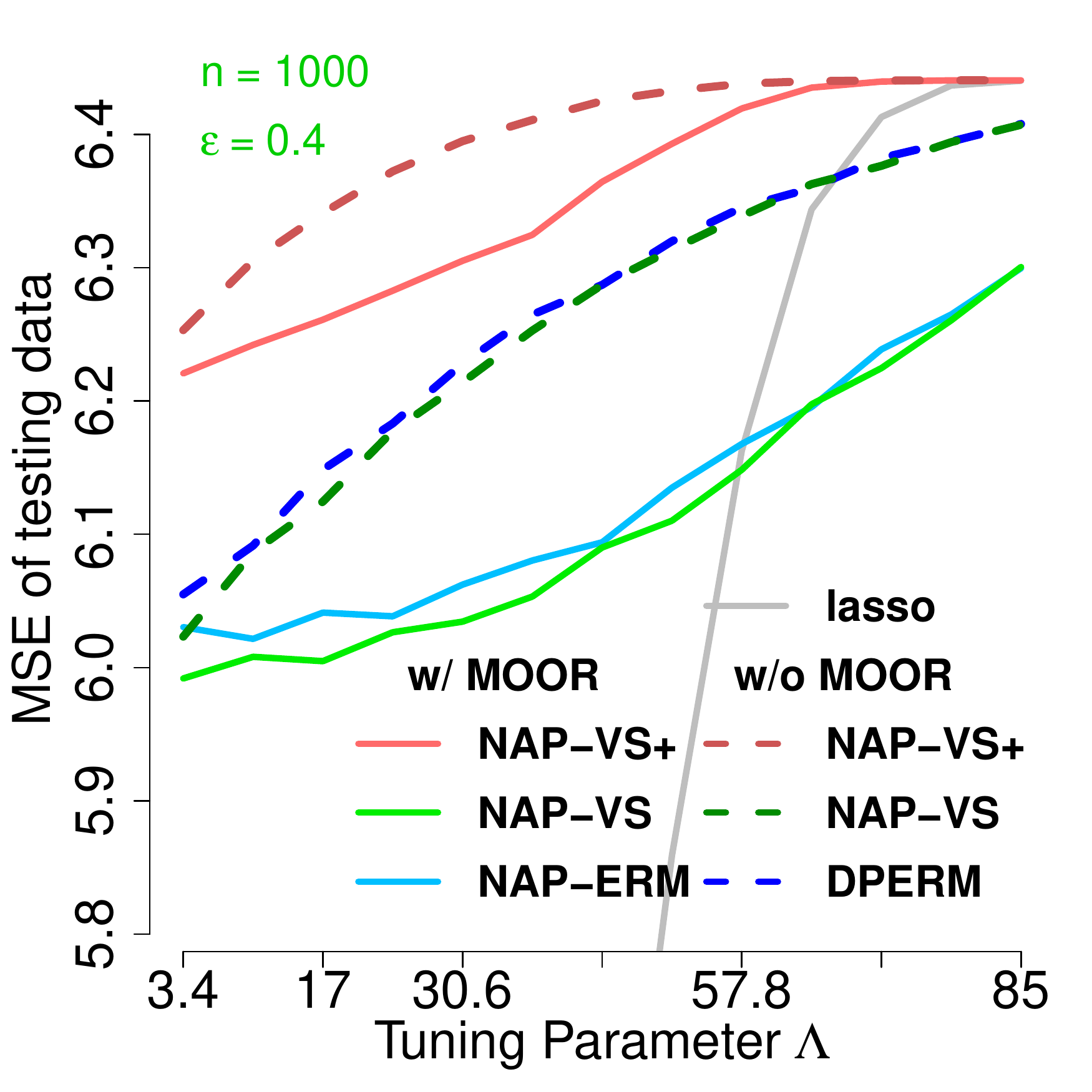}
\includegraphics[width=0.19\linewidth, trim=4pt 9pt 15pt 18pt,clip]{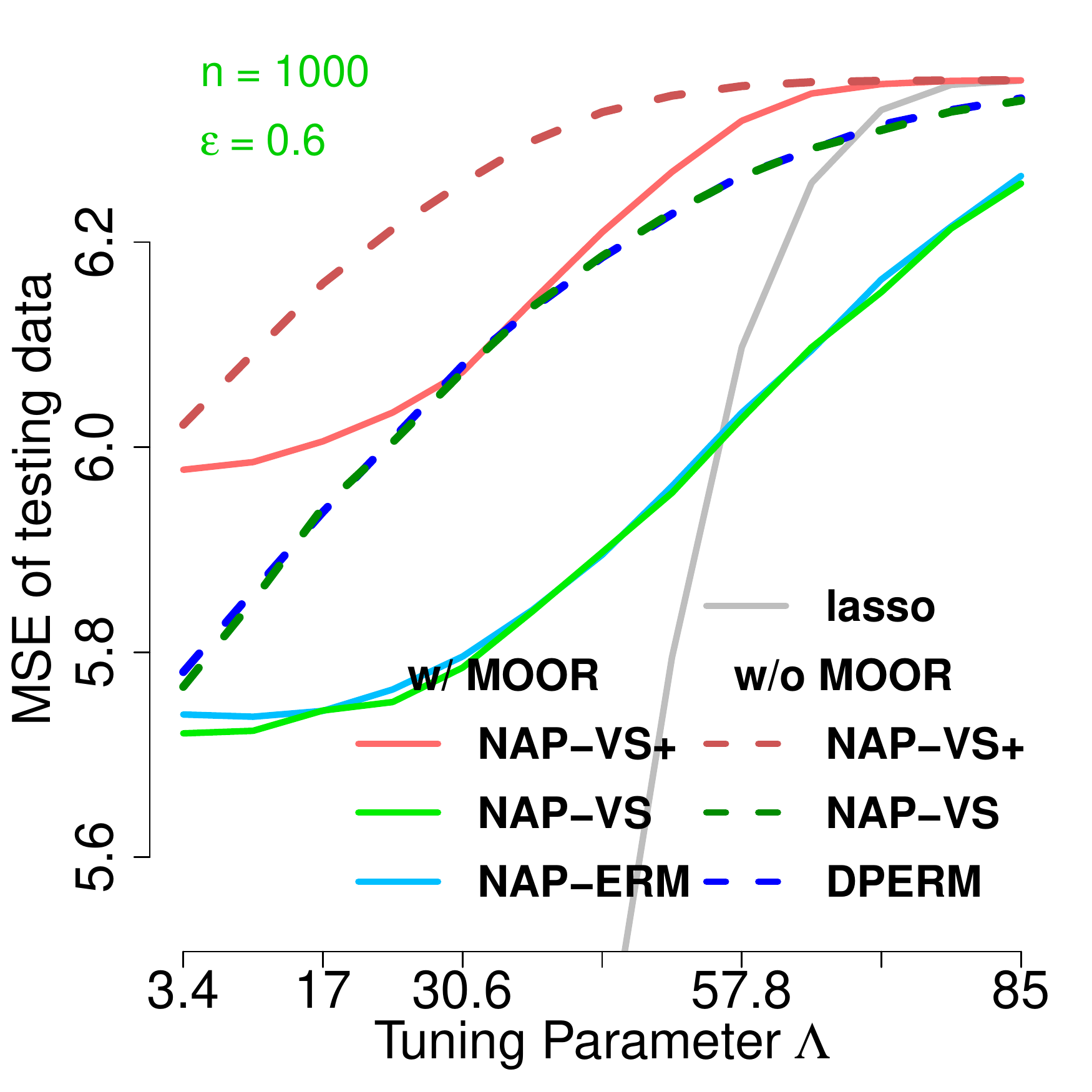}
\includegraphics[width=0.19\linewidth, trim=4pt 9pt 15pt 18pt,clip]{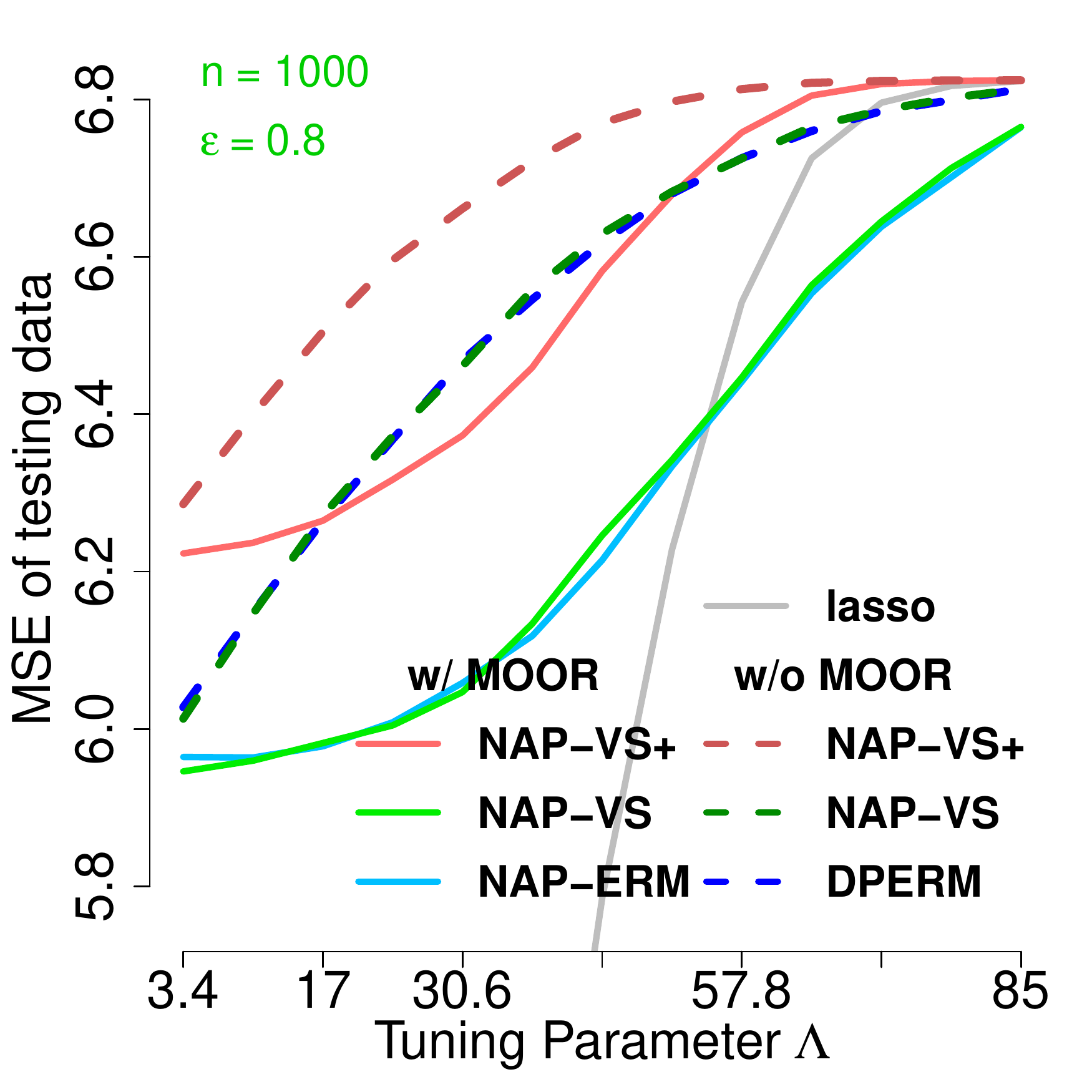}
\includegraphics[width=0.19\linewidth, trim=4pt 9pt 15pt 18pt,clip]{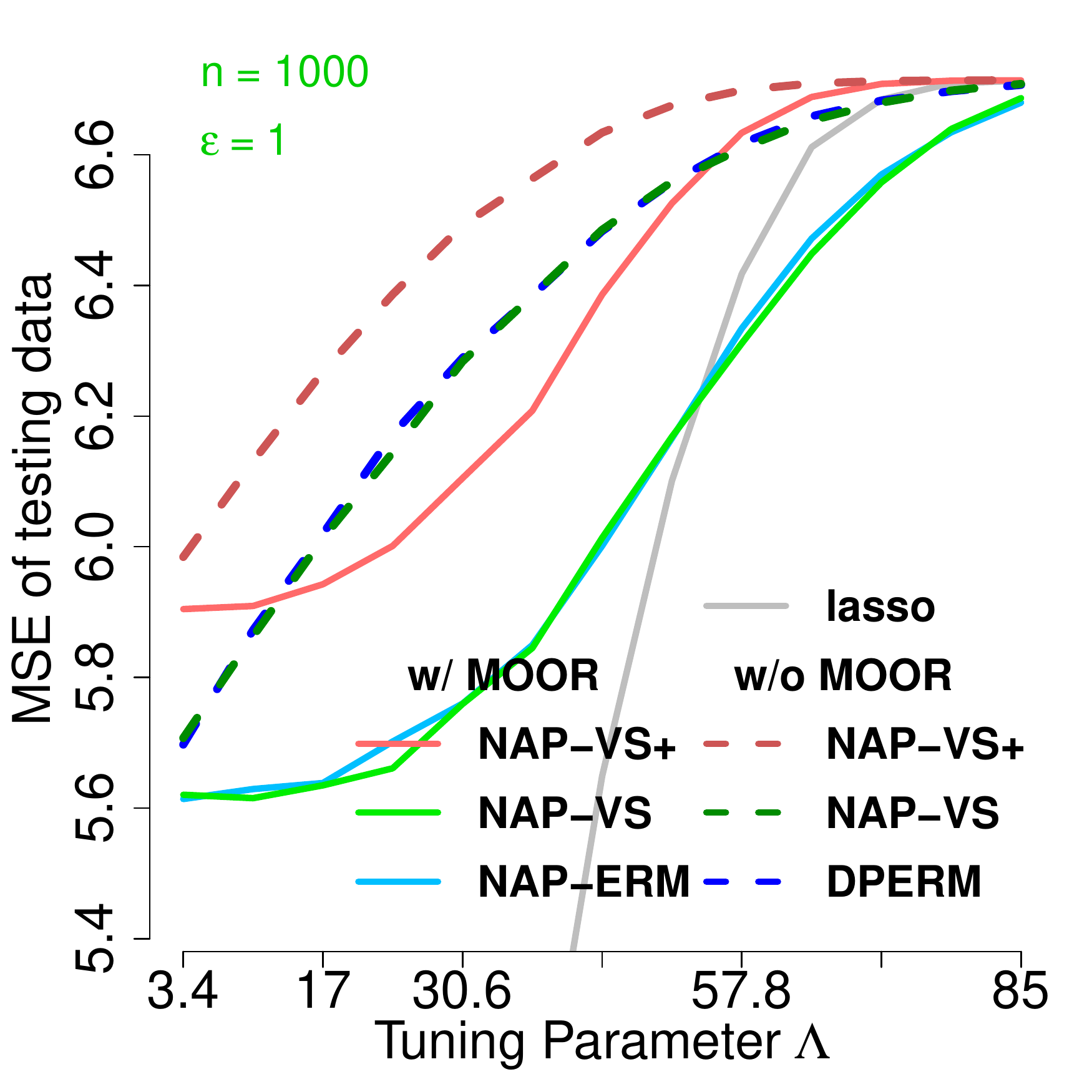}\\\vspace{-3pt}
\footnotesize logistic regression $n=500$ \\\vspace{-1pt}
\includegraphics[width=0.19\linewidth, trim=4pt 9pt 15pt 18pt,clip]{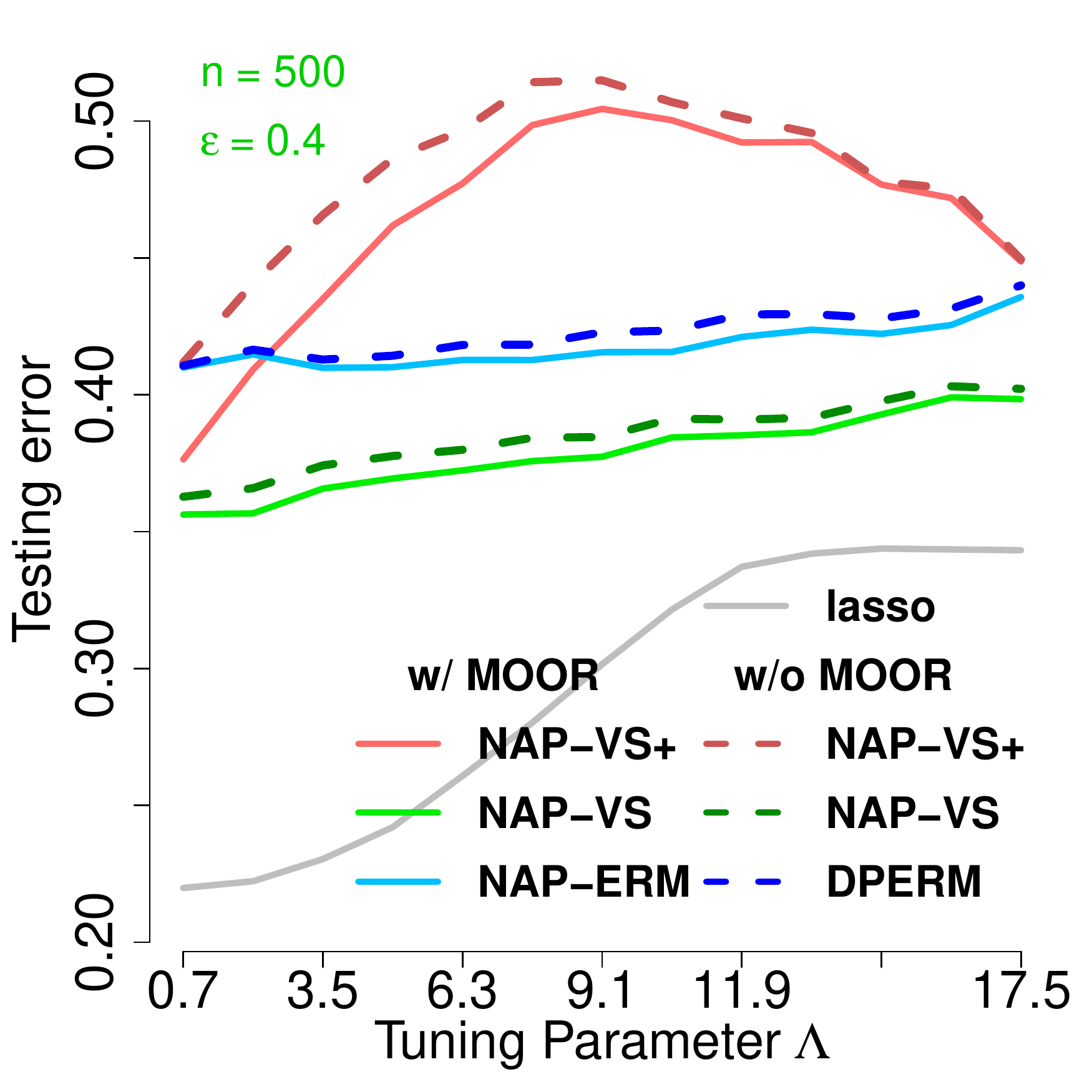}
\includegraphics[width=0.19\linewidth, trim=4pt 9pt 15pt 18pt,clip]{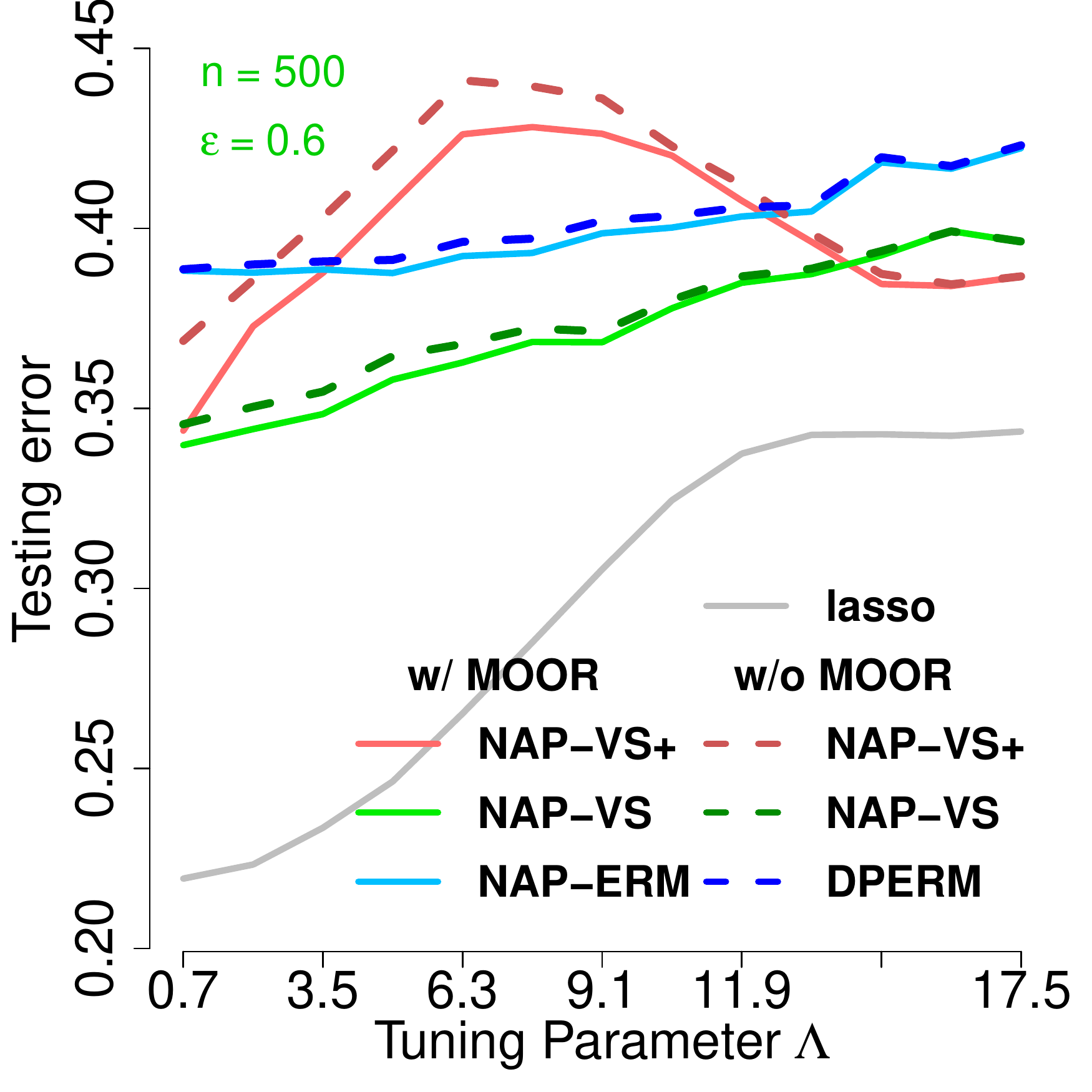}
\includegraphics[width=0.19\linewidth, trim=4pt 9pt 15pt 18pt,clip]{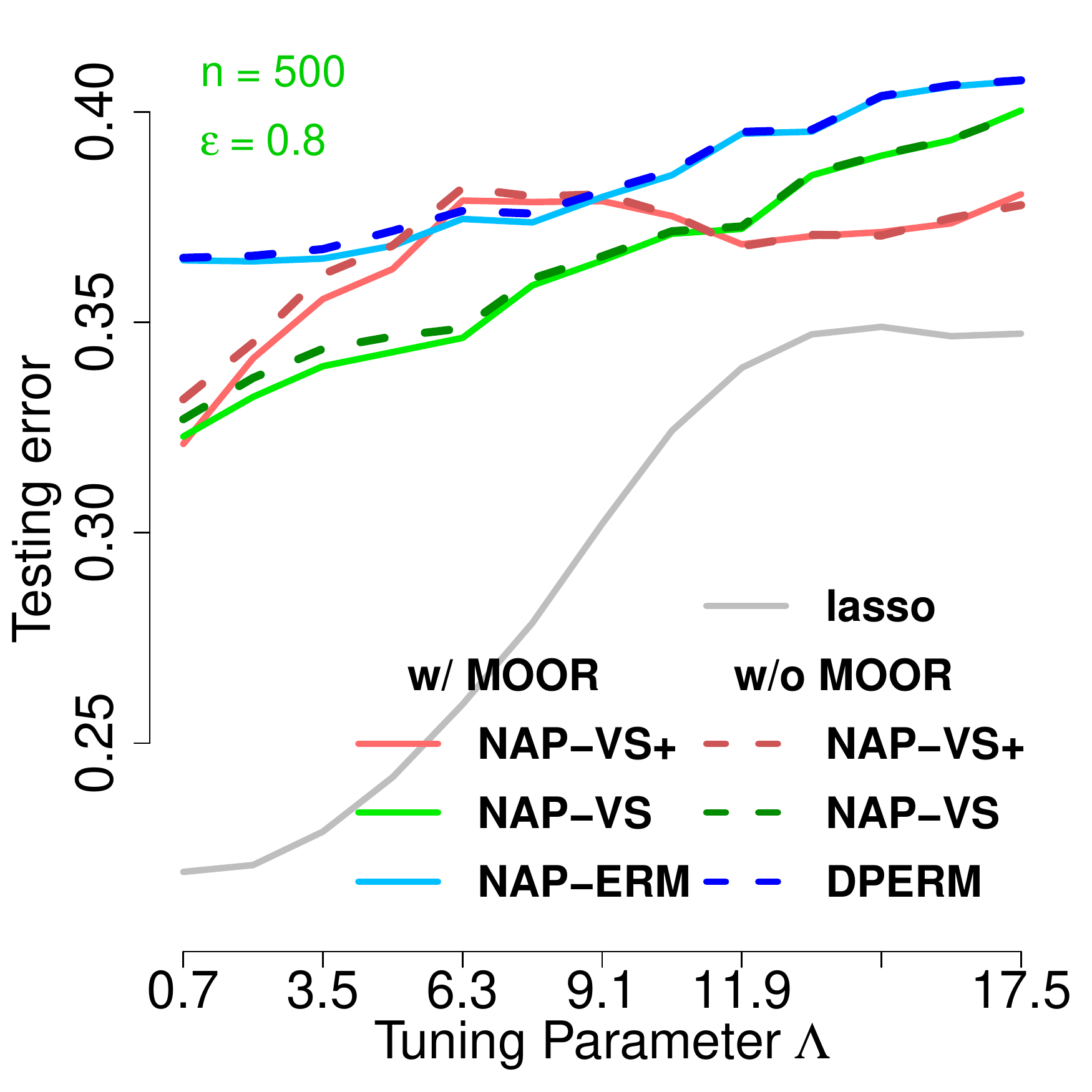}
\includegraphics[width=0.19\linewidth, trim=4pt 9pt 15pt 18pt,clip]{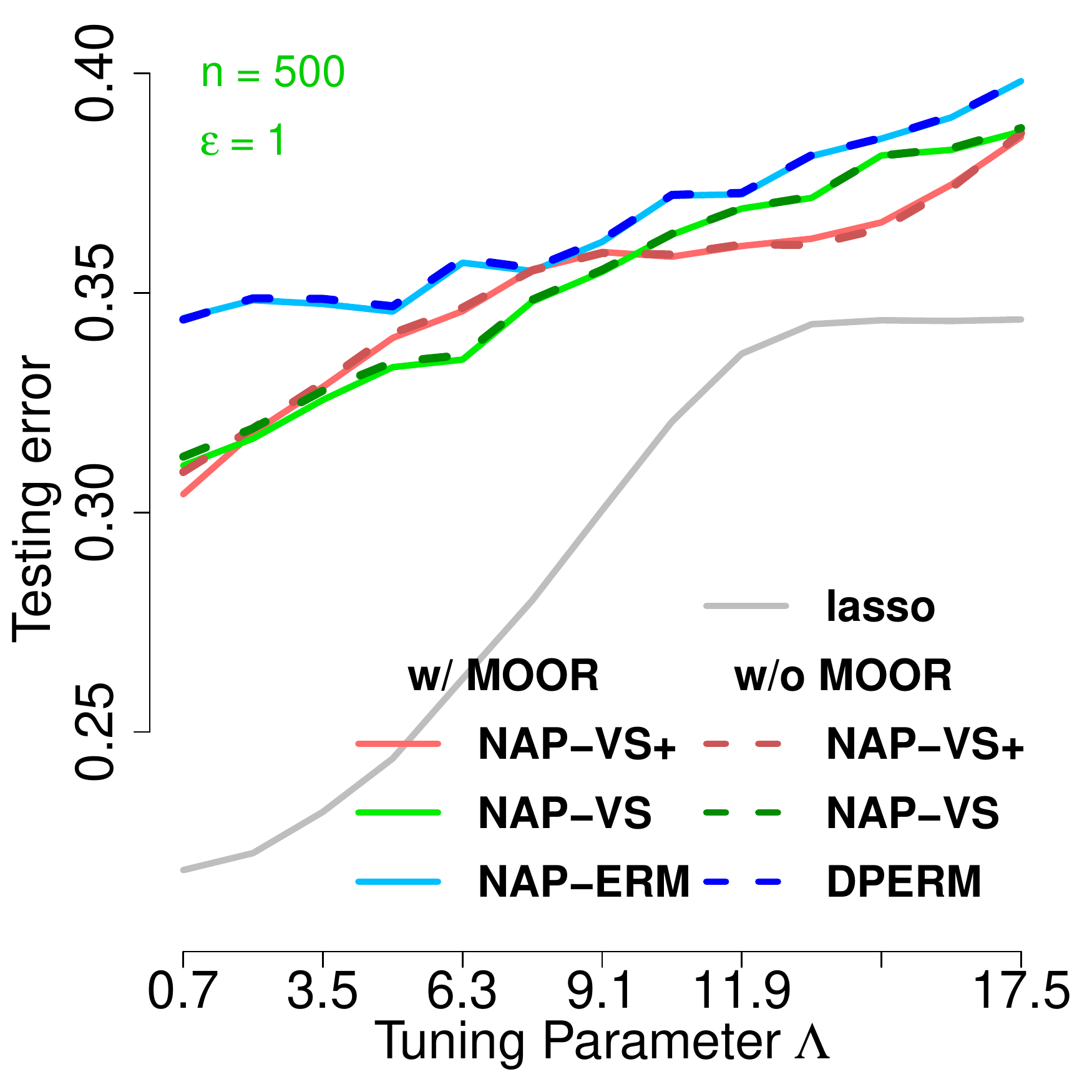}\\\vspace{-3pt}
\footnotesize logistic regression $n=1000$ \\\vspace{-1pt}
\includegraphics[width=0.19\linewidth, trim=4pt 9pt 15pt 18pt,clip]{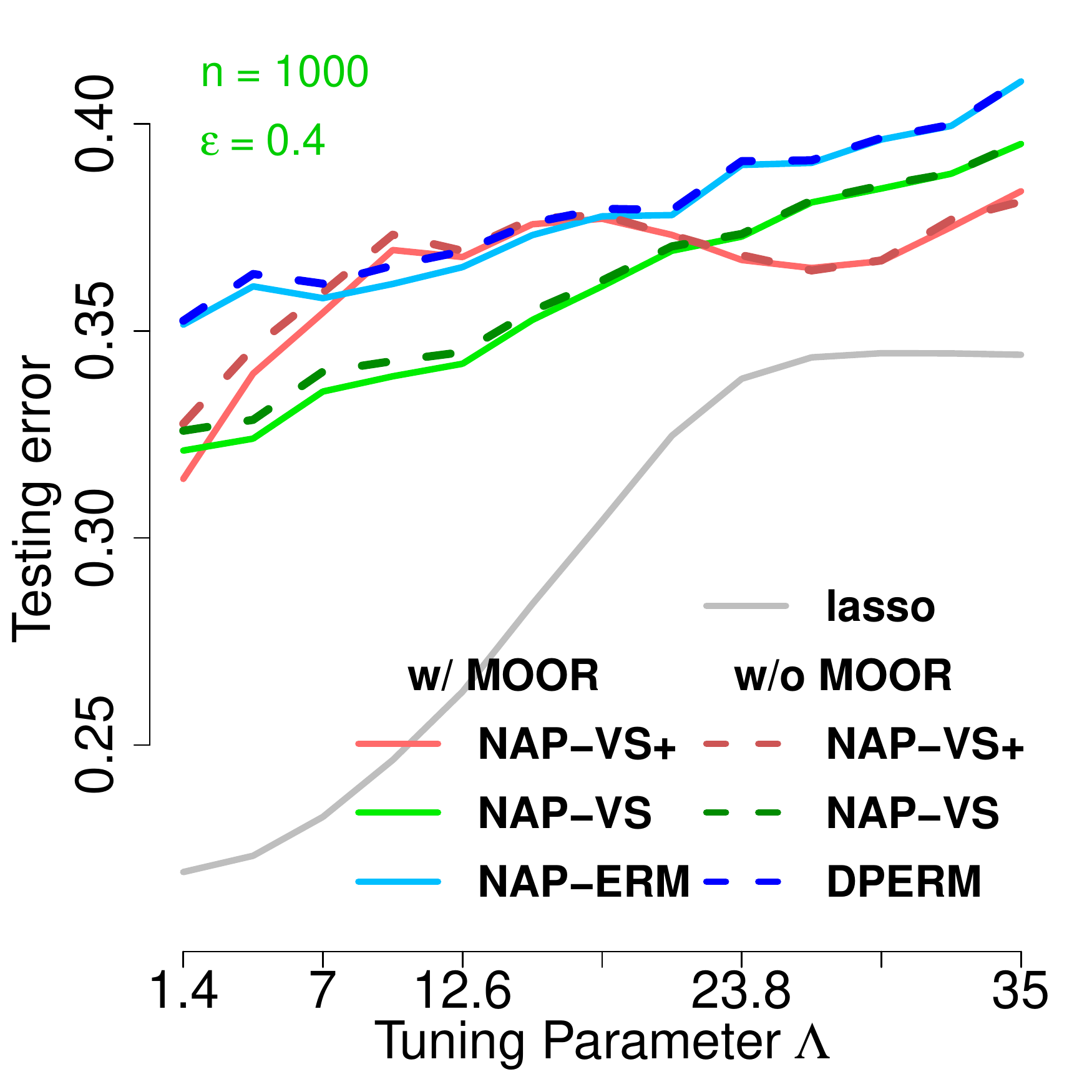}
\includegraphics[width=0.19\linewidth, trim=4pt 9pt 15pt 18pt,clip]{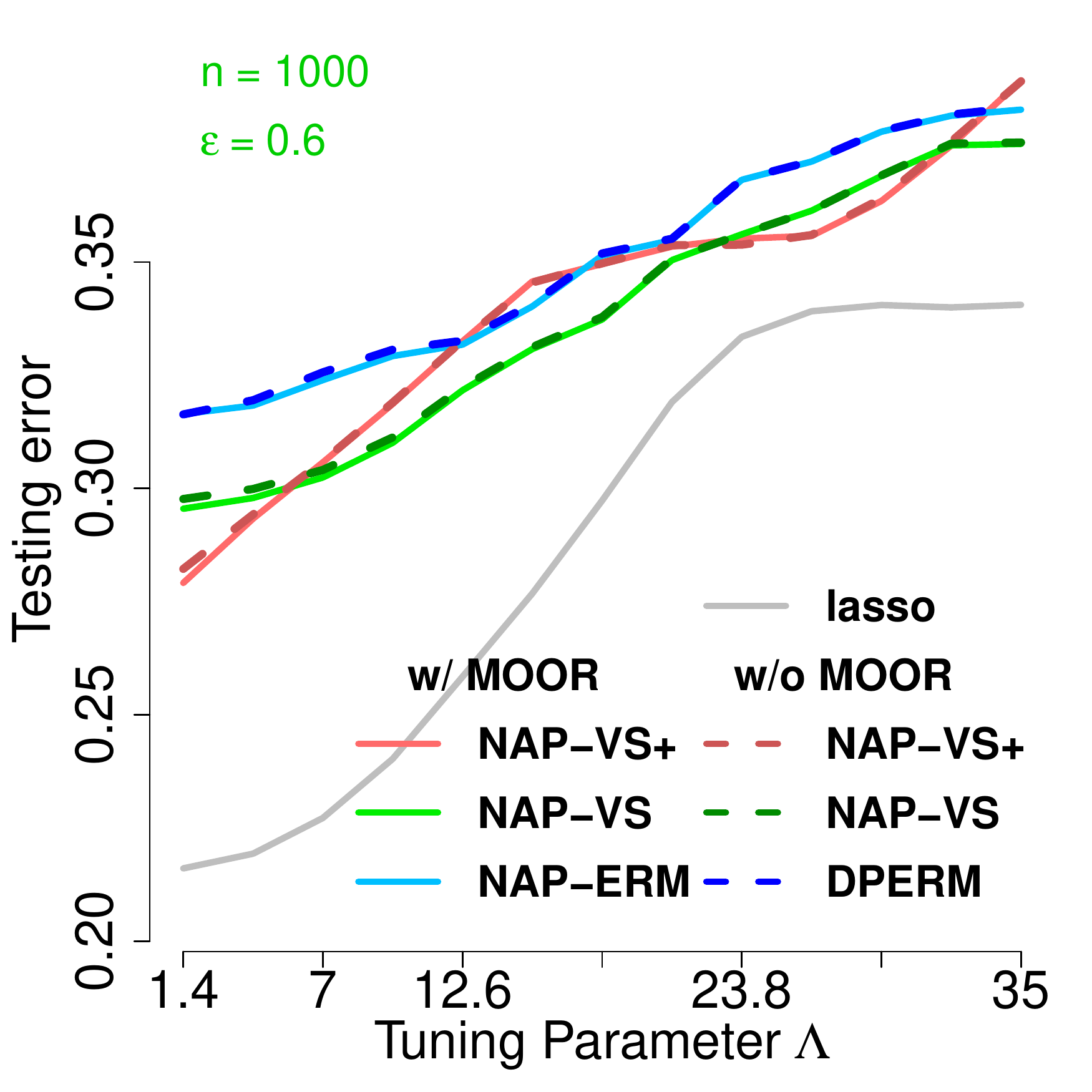}
\includegraphics[width=0.19\linewidth, trim=4pt 9pt 15pt 18pt,clip]{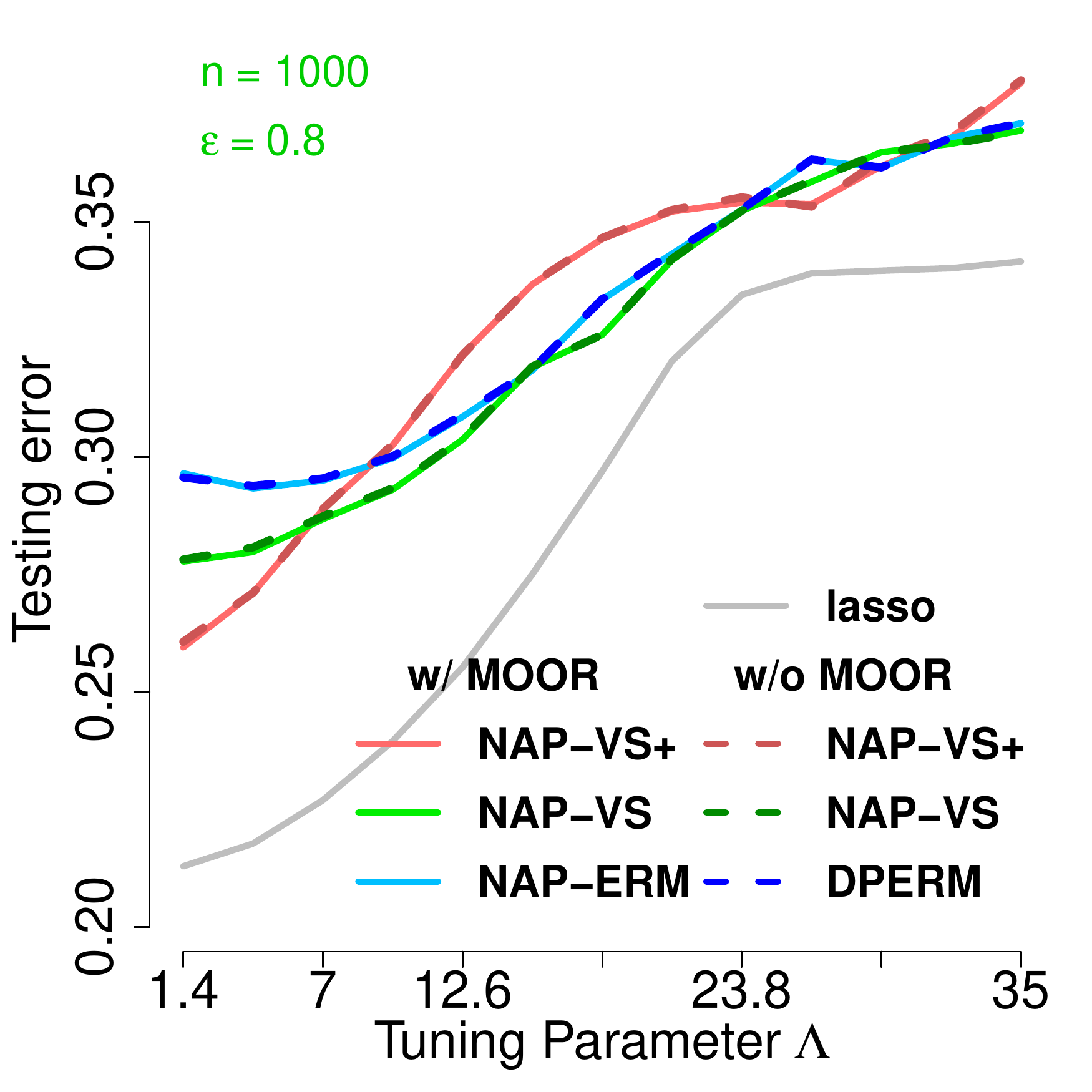}
\includegraphics[width=0.19\linewidth, trim=4pt 9pt 15pt 18pt,clip]{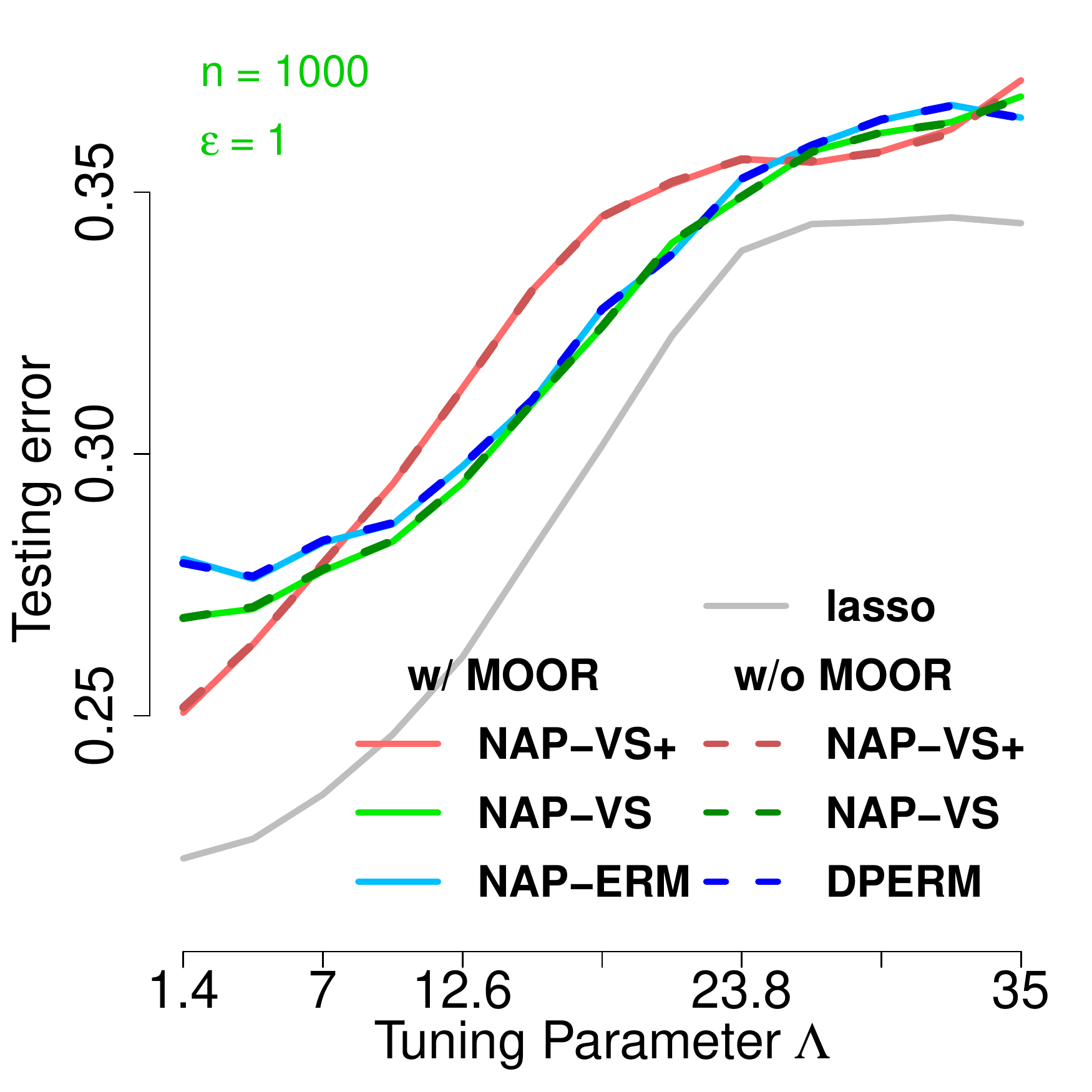}\\
\vspace{-9pt}
\caption{Testing prediction error (MSE in linear and Poisson regression and misclassification rate in logistic regression) at different $\Lambda$, $n$, and $\epsilon$}\label{fig:SIM.prediction}\vspace{-9pt}
\end{figure}

The results on the private prediction in the training data are illustrated in Fig \ref{fig:SIM.prediction} and  summarized as follows. 1) The benefit of MOOR (solid lines) on outcome prediction can be observed from the  smaller MSE or misclassification rates compared to without MOOR  (dashed lines) in general  except in the linear regression  for $n=200$.  2) As  $\Lambda$ increases,  the MSE and the misclassification rate also increase except for NAPP-ERM and the existing DP-ERM  in linear regression at $n=200$ and NAPP-VS+ at $\epsilon<0.8$.  3) NAPP-VS (the green lines) yields the smallest prediction errors compared to NAPP-VS+ (the red lines) and NAPP-ERM (the blue lines) in general.  NAPP-VS+ in some cases generates the largest error (e.g., Poisson regression, and logistic regression at $n=500$).

\begin{figure}[!htb]
\vspace{-6pt}\centering
\includegraphics[width=0.32\linewidth]{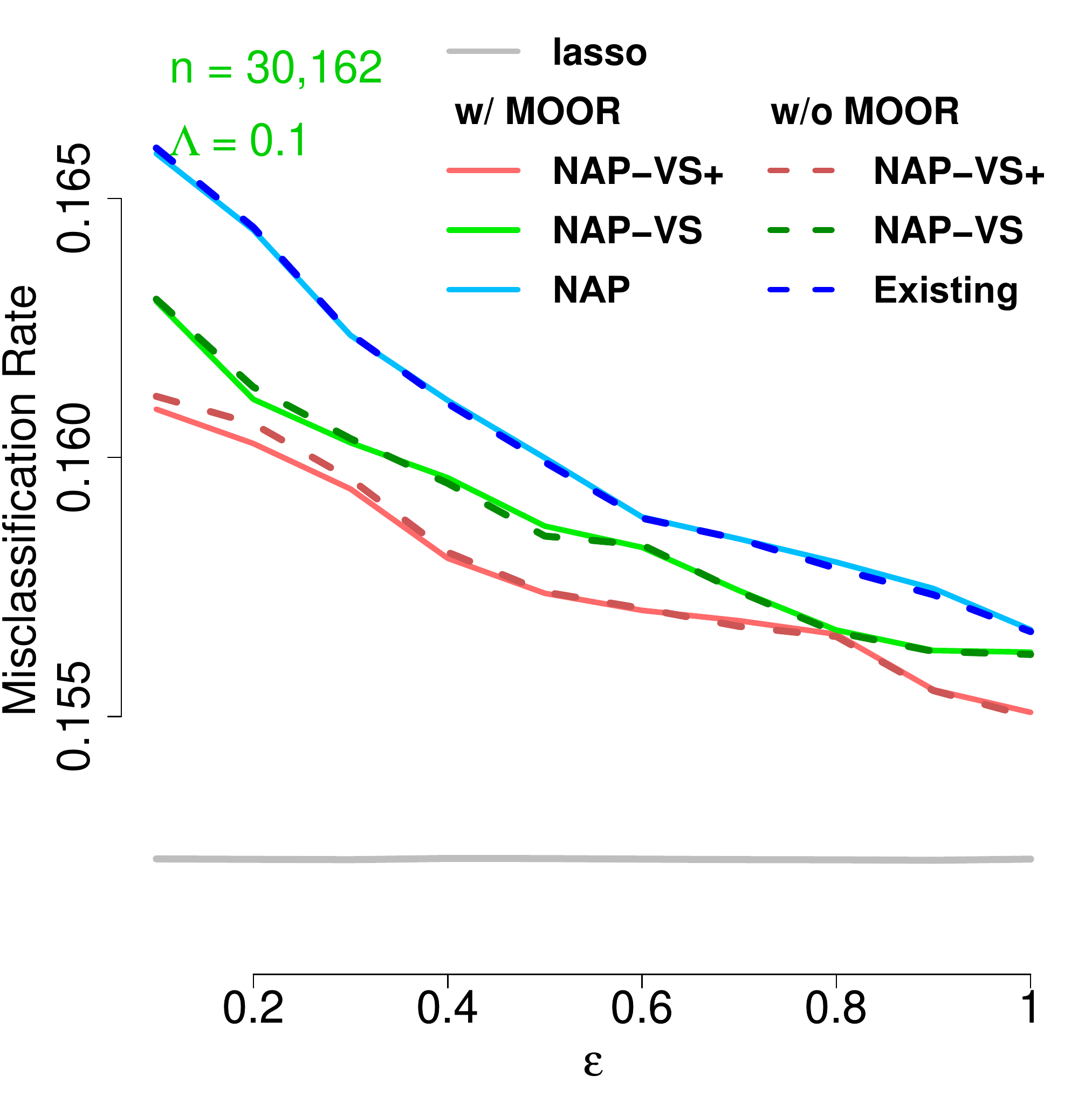}
\includegraphics[width=0.32\linewidth]{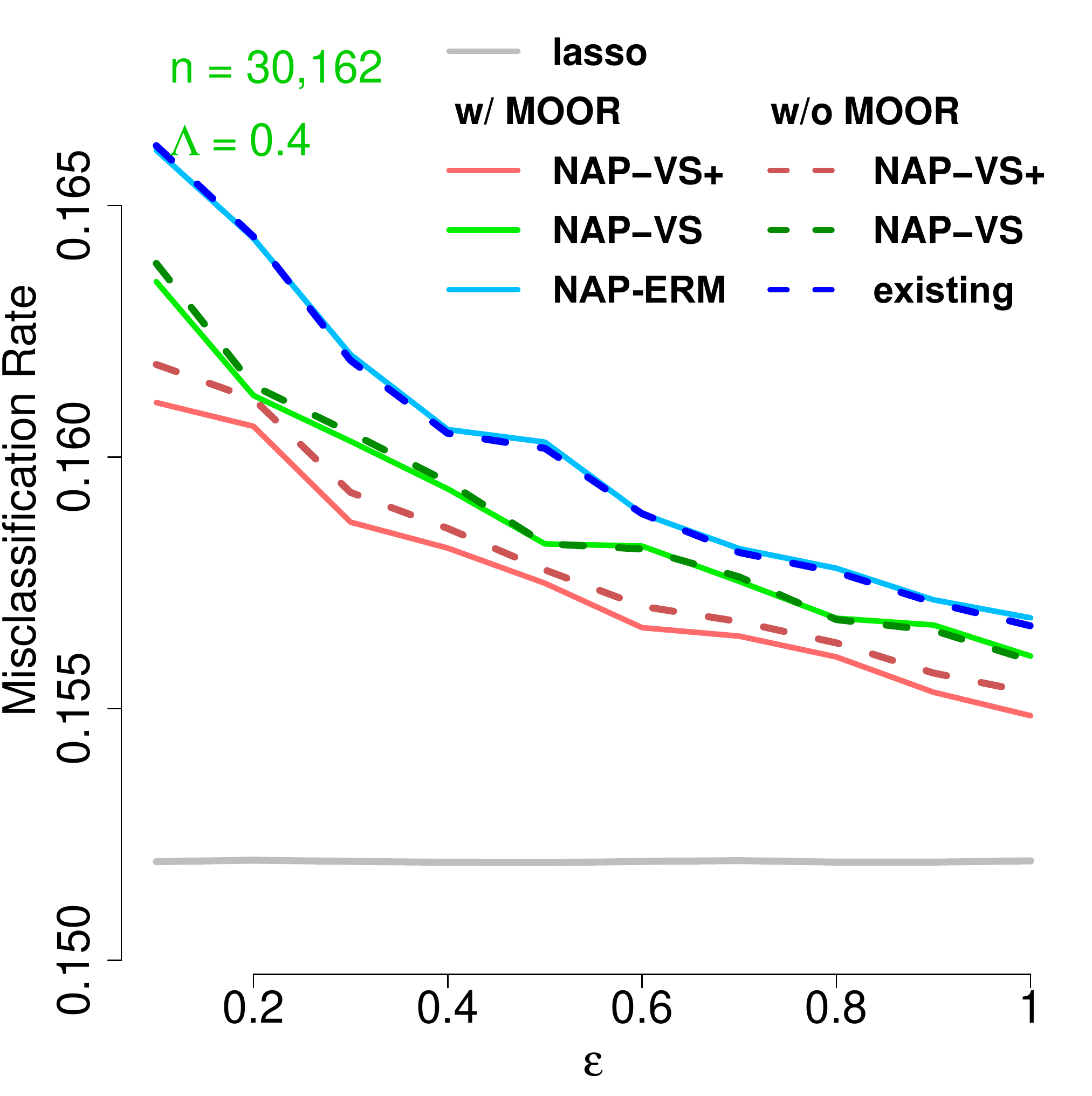}
\includegraphics[width=0.32\linewidth]{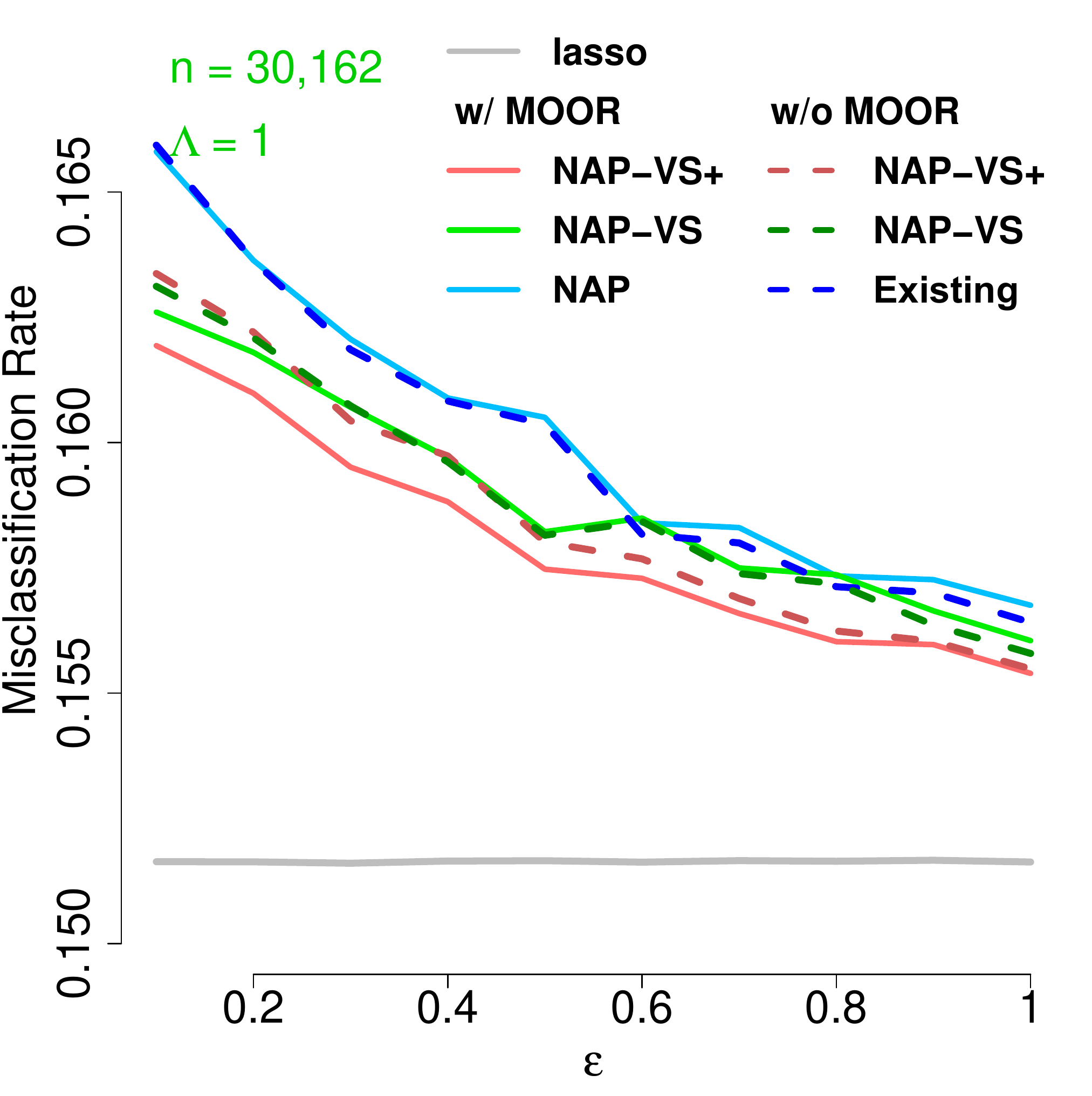}
\vspace{-12pt}
\caption{Misclassification rate in the Adult Data set} \label{fig:case}\vspace{-12pt}
\end{figure}

We also applied the NAPP-ERM and NAPP-VS procedures to the Adult data set (downloaded from \url{https://archive.ics.uci.edu/ml/datasets/Adult}) and compared their prediction performance against the existing DP-ERM approach in a lasso-regularized logistic regression model. The training data set has 22,654 cases of $Y=0$ (income $<50K$) and 7,508 cases of $Y=1$  (income $>50K$). 
the testing data has 11,360 cases of $Y=0$ and 3,700 cases of $Y=1$).   We examined  $\Lambda=\{0.1, 0.4, 1\}$ for the lasso regularizer at privacy cost $\epsilon\in[0.1,1]$. 
For the NAPP-ERM procedure, we set $\Lambda_0=\epsilon^{-1}$ and run $T=80$ iterations. The results are presented in Fig \ref{fig:case}.  NAPP-VS+ (the red lines) performs the best in general with the smallest misclassification rate, followed by NAPP-VS. At  $\epsilon=1$, the misclassification rate via NAPP-VS+ is around 15.5\%,  very close to the non-private misclassification rate 15.2\%. The MOOR effect (the solid lines) is not obvious in most cases except for NAPP-VS+ at $\Lambda=1$.

\vspace{-6pt}\subsection{Privacy Budget Retrieval and Recycling}\label{sec:retrieval}\vspace{-6pt}
In this experiment, we quantify the retrievable privacy budget via NAPP out of the originally allocation to the Jacobian ratio $r_2$ in Eq (\ref{eqn:ratio}), which is set at 1/2 of total $\epsilon$ in all experiment settings.

\begin{figure}[!htb]\centering
\footnotesize{linear regression $n=200$}\\
\includegraphics[width=0.19\linewidth, trim=4pt 9pt 15pt 18pt,clip]{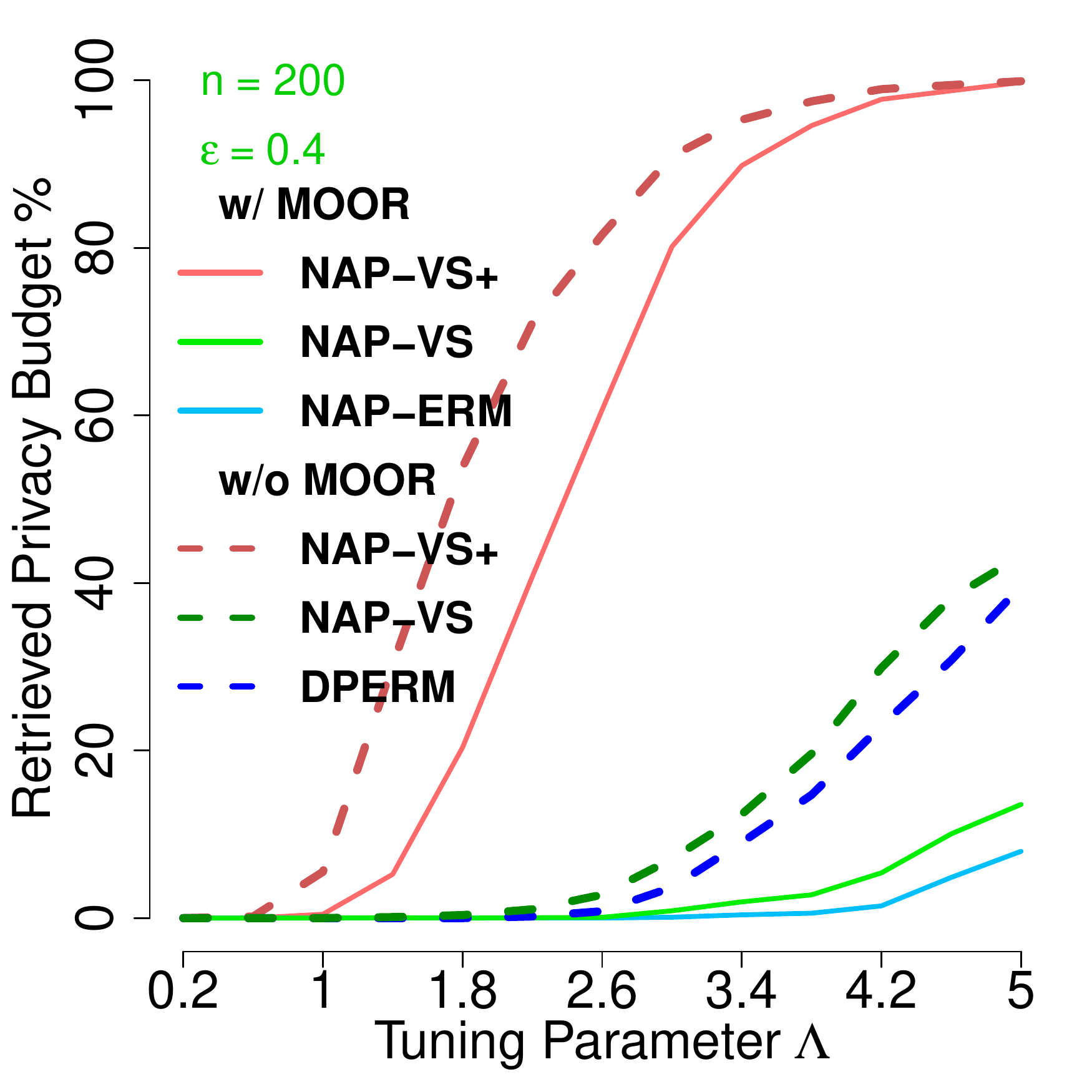}
\includegraphics[width=0.19\linewidth, trim=4pt 9pt 15pt 18pt,clip]{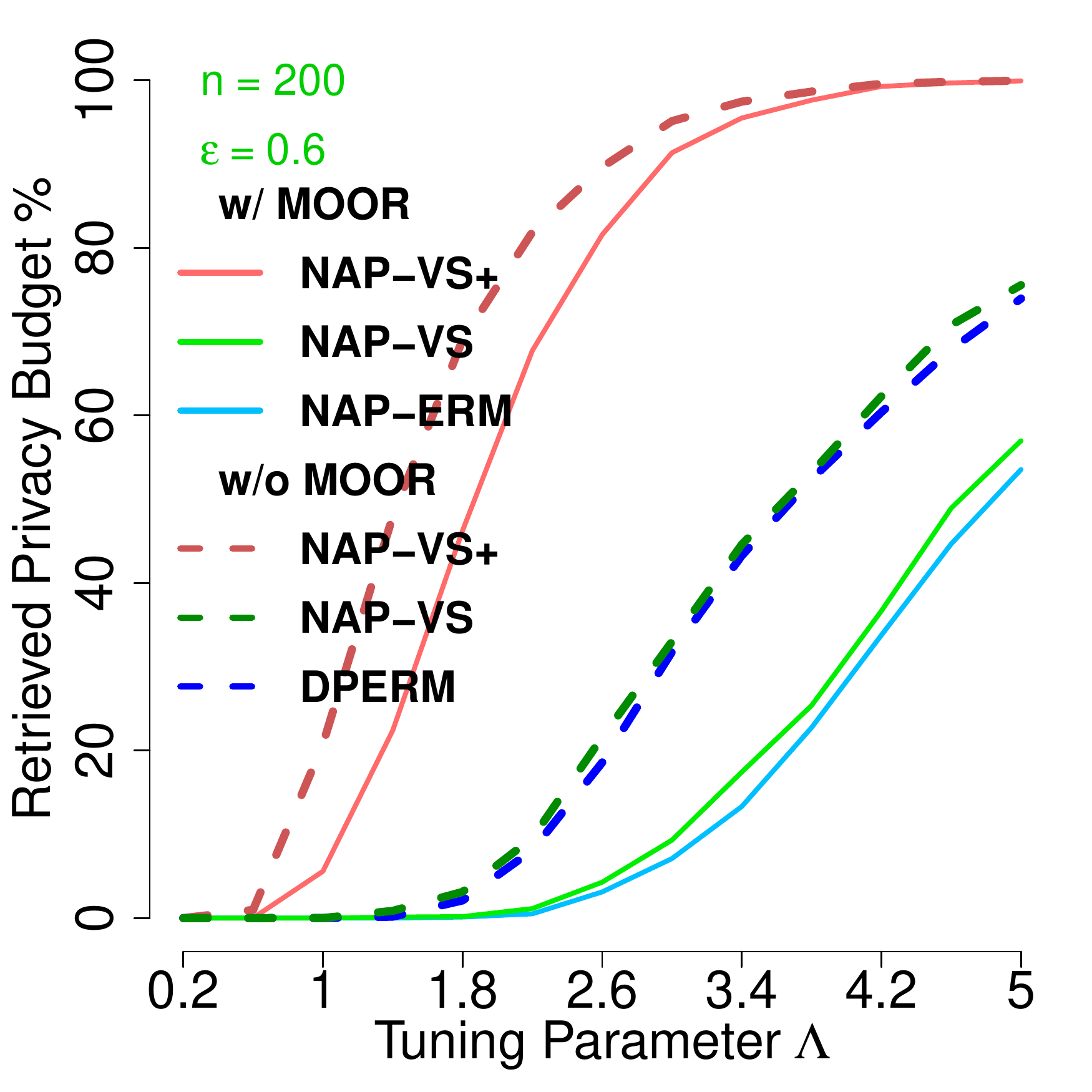}
\includegraphics[width=0.19\linewidth, trim=4pt 9pt 15pt 18pt,clip]{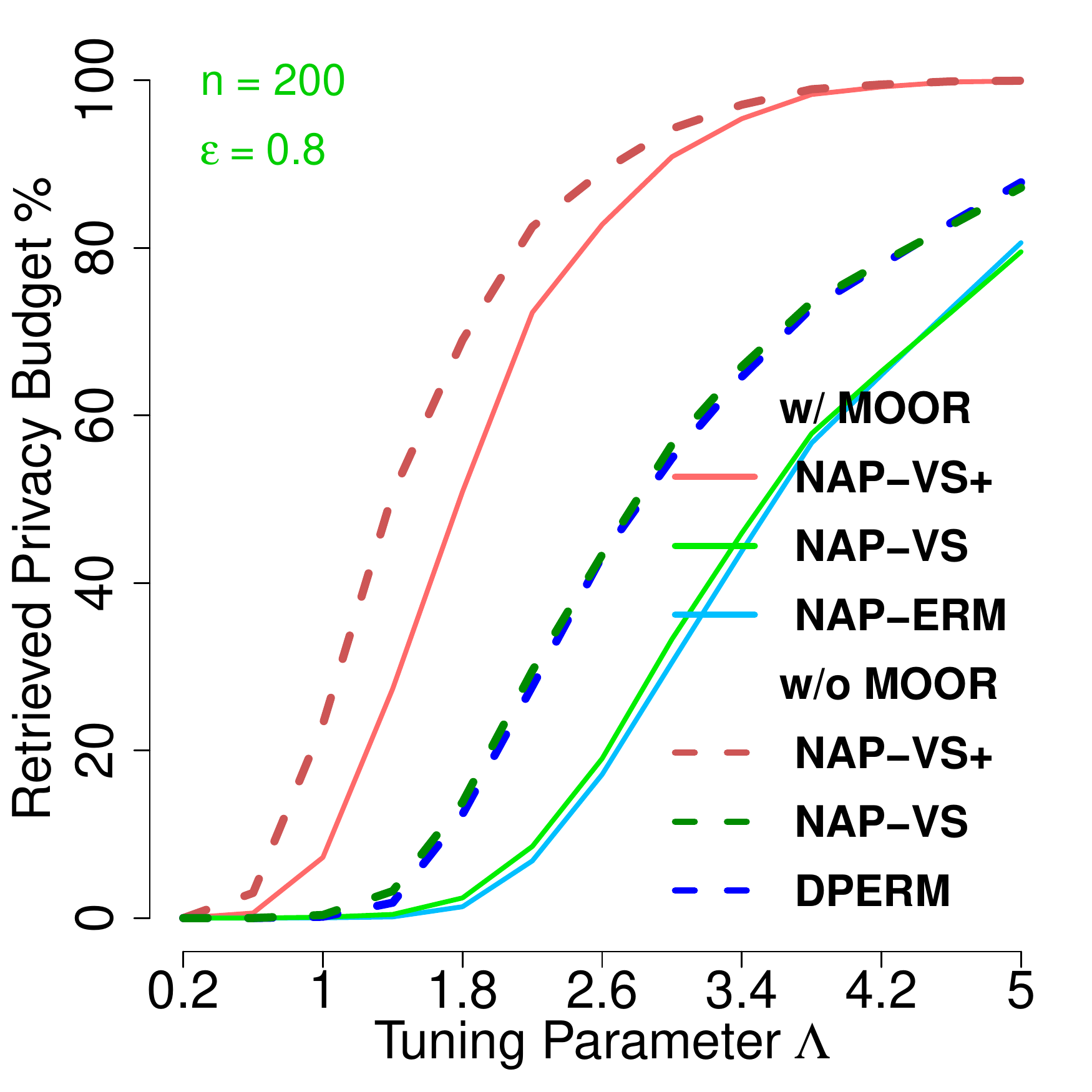}
\includegraphics[width=0.19\linewidth, trim=4pt 9pt 15pt 18pt,clip]{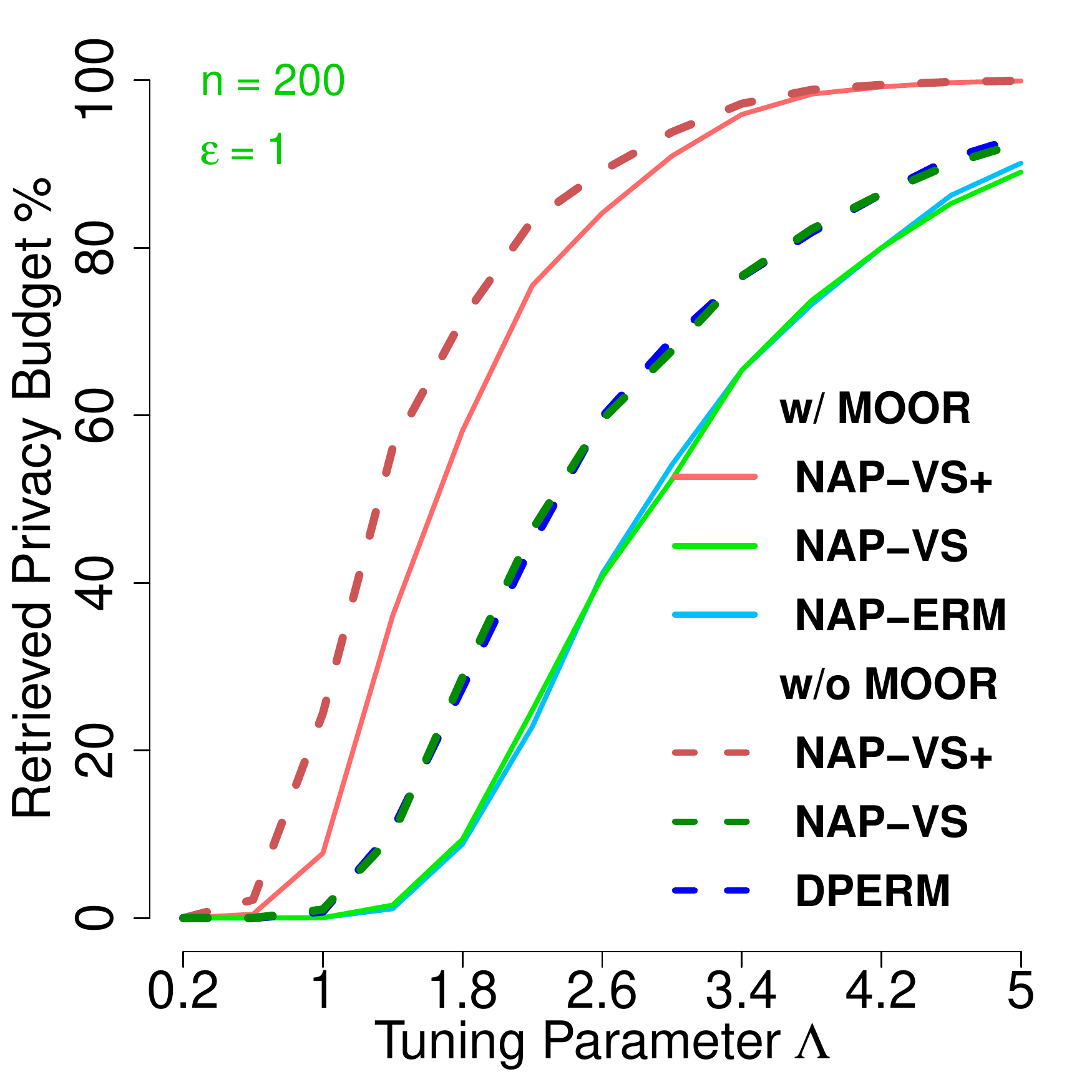}\\ 
\footnotesize linear regression $n=500$\\
\includegraphics[width=0.19\linewidth, trim=4pt 9pt 15pt 18pt,clip]{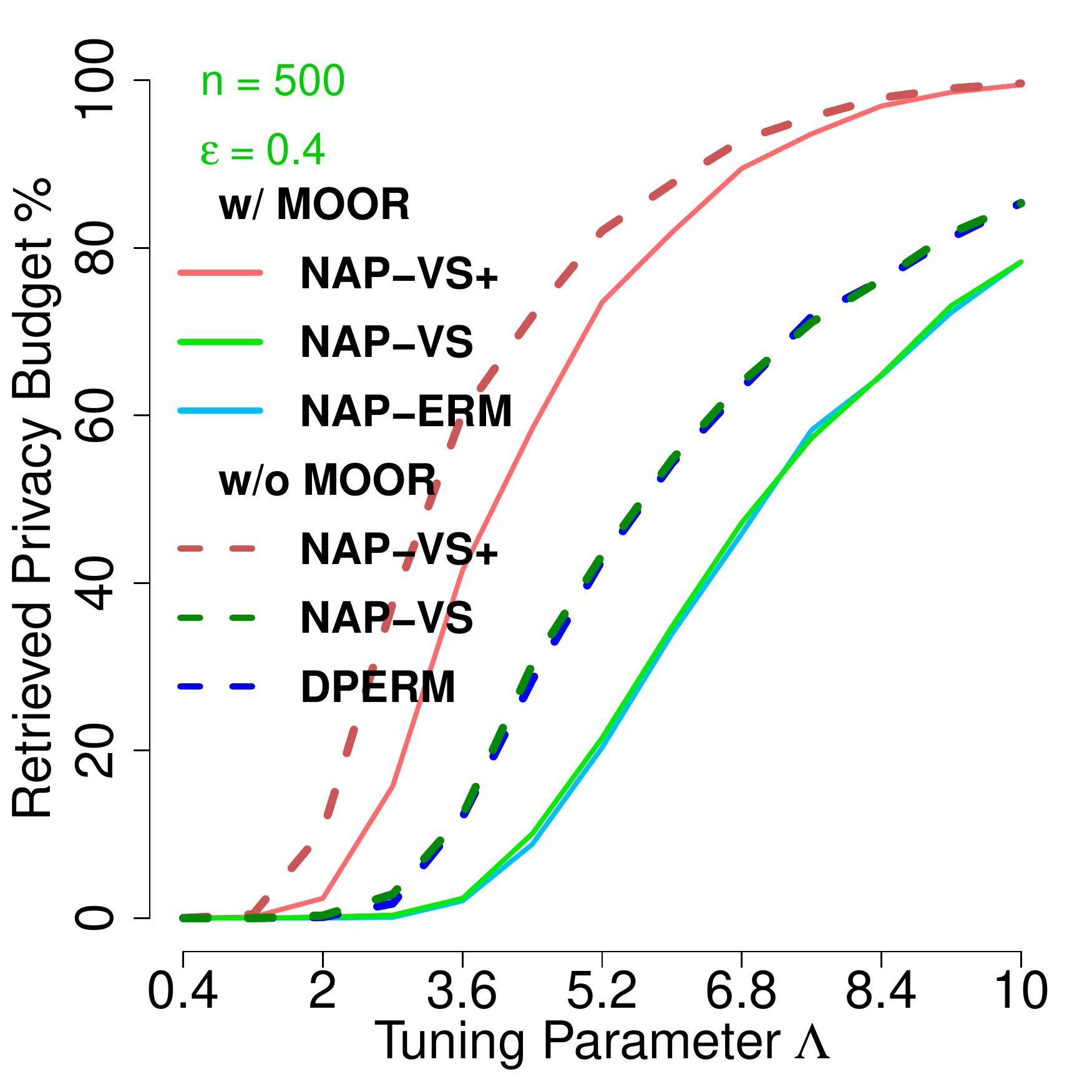}
\includegraphics[width=0.19\linewidth, trim=4pt 9pt 15pt 18pt,clip]{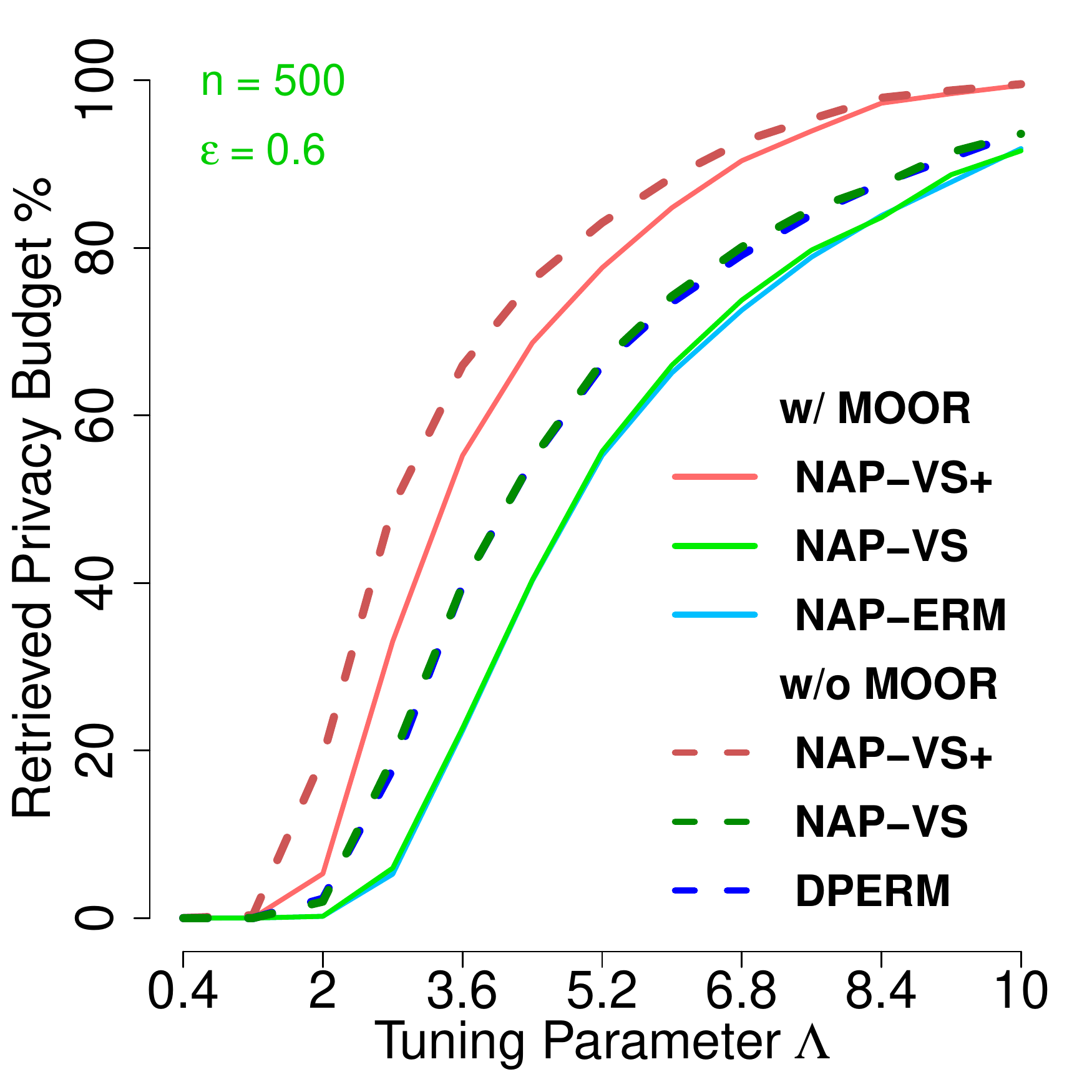}
\includegraphics[width=0.19\linewidth, trim=4pt 9pt 15pt 18pt,clip]{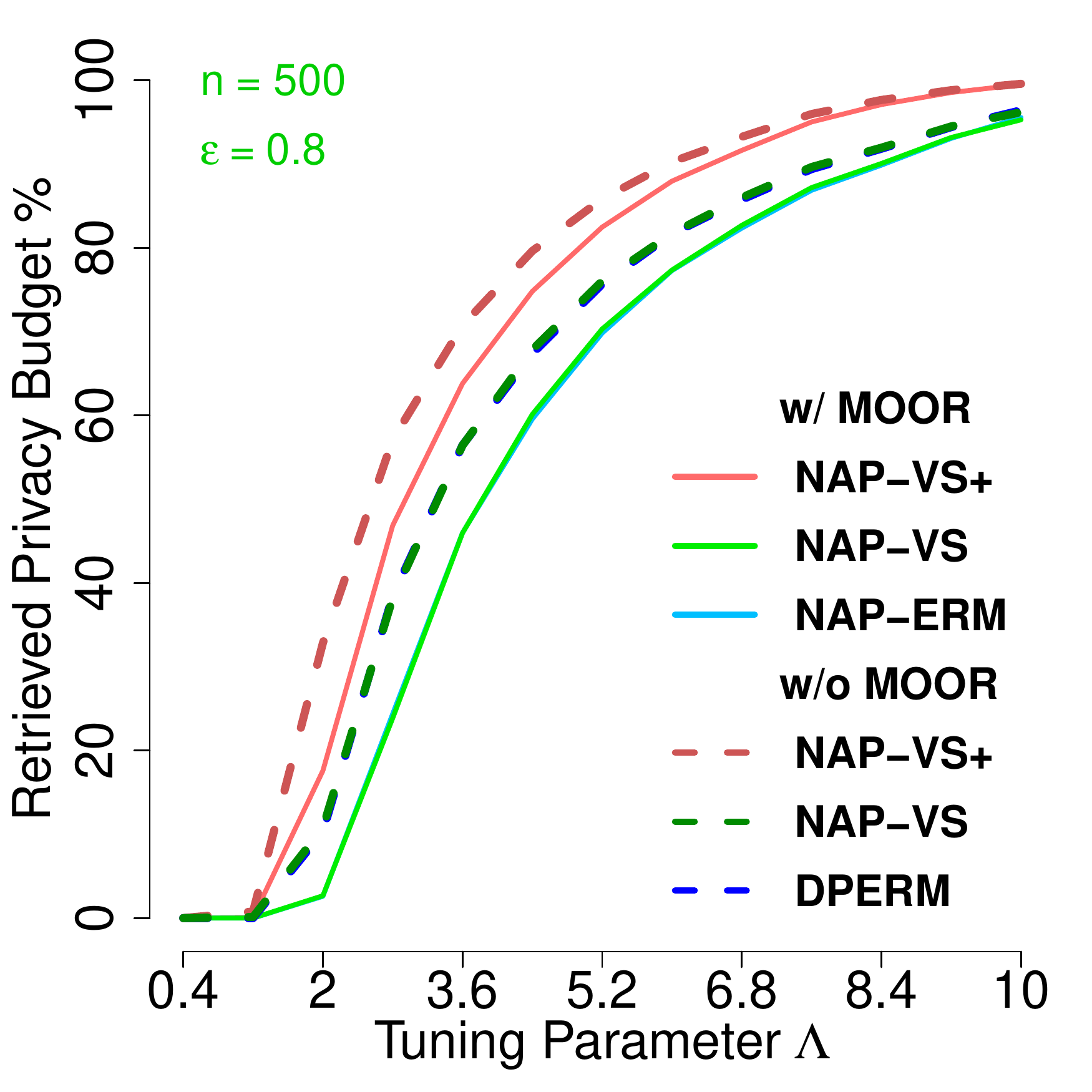}
\includegraphics[width=0.19\linewidth, trim=4pt 9pt 15pt 18pt,clip]{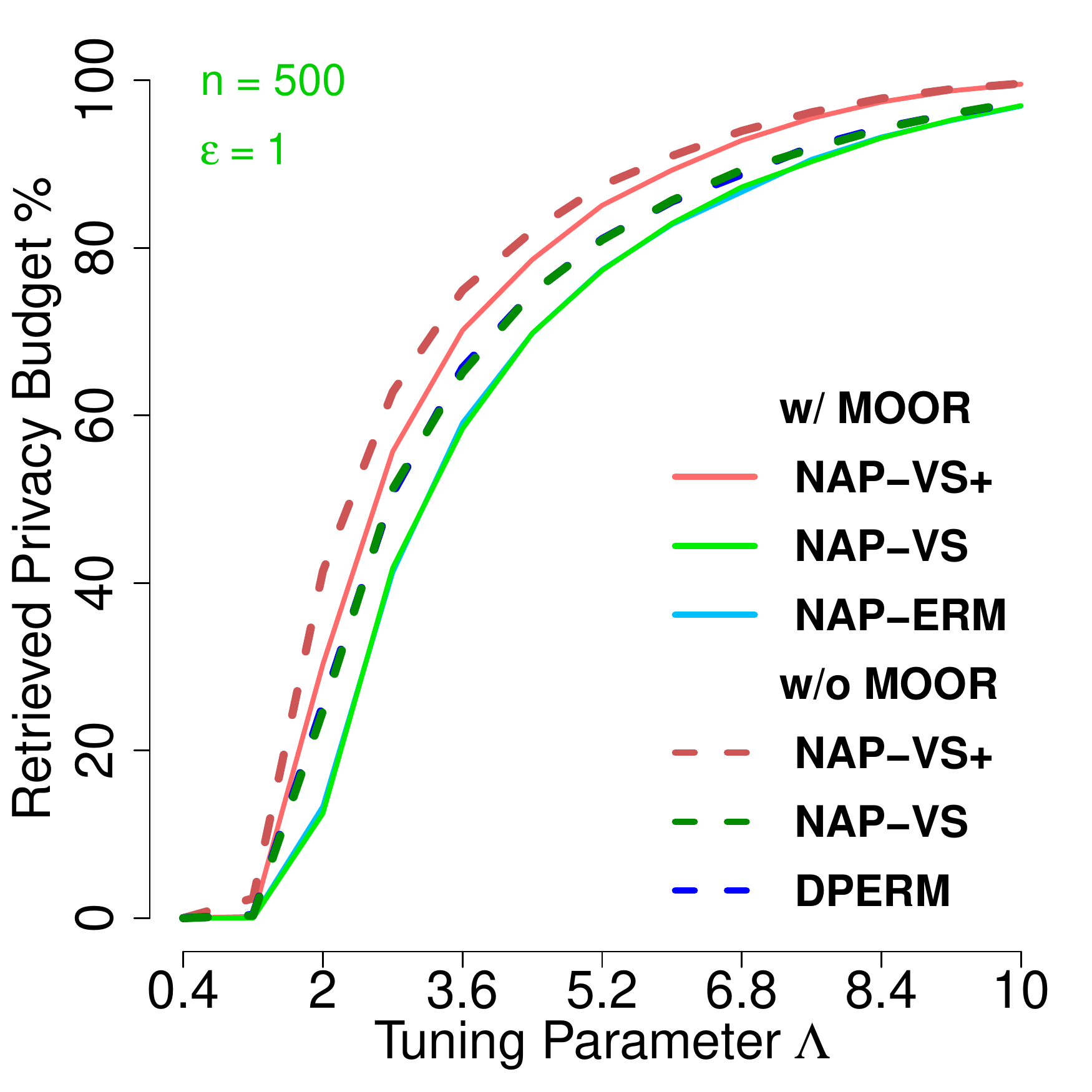}\\\vspace{-3pt}
\footnotesize Poisson regression $n=500$ \\
\includegraphics[width=0.19\linewidth, trim=4pt 9pt 15pt 18pt,clip]{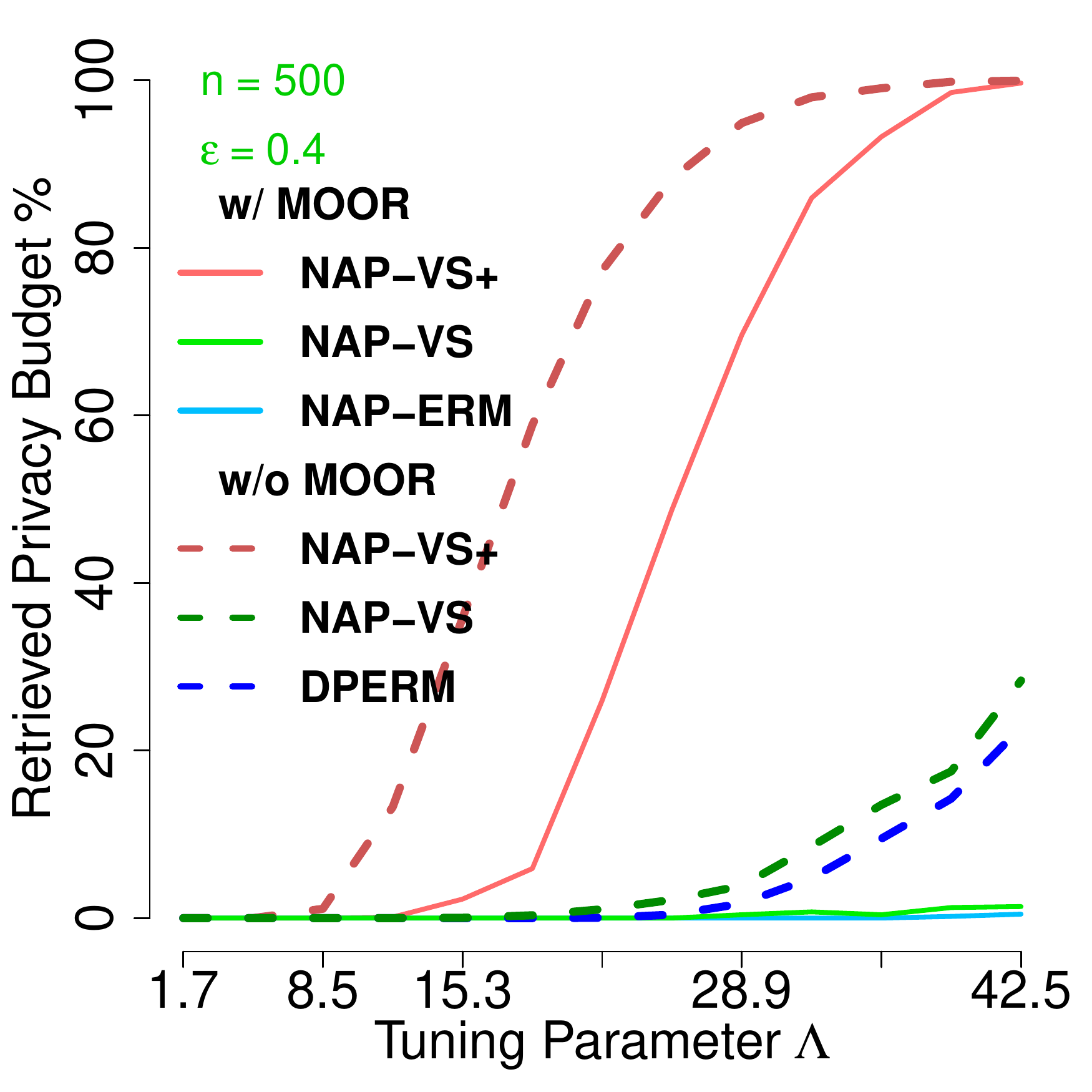}
\includegraphics[width=0.19\linewidth, trim=4pt 9pt 15pt 18pt,clip]{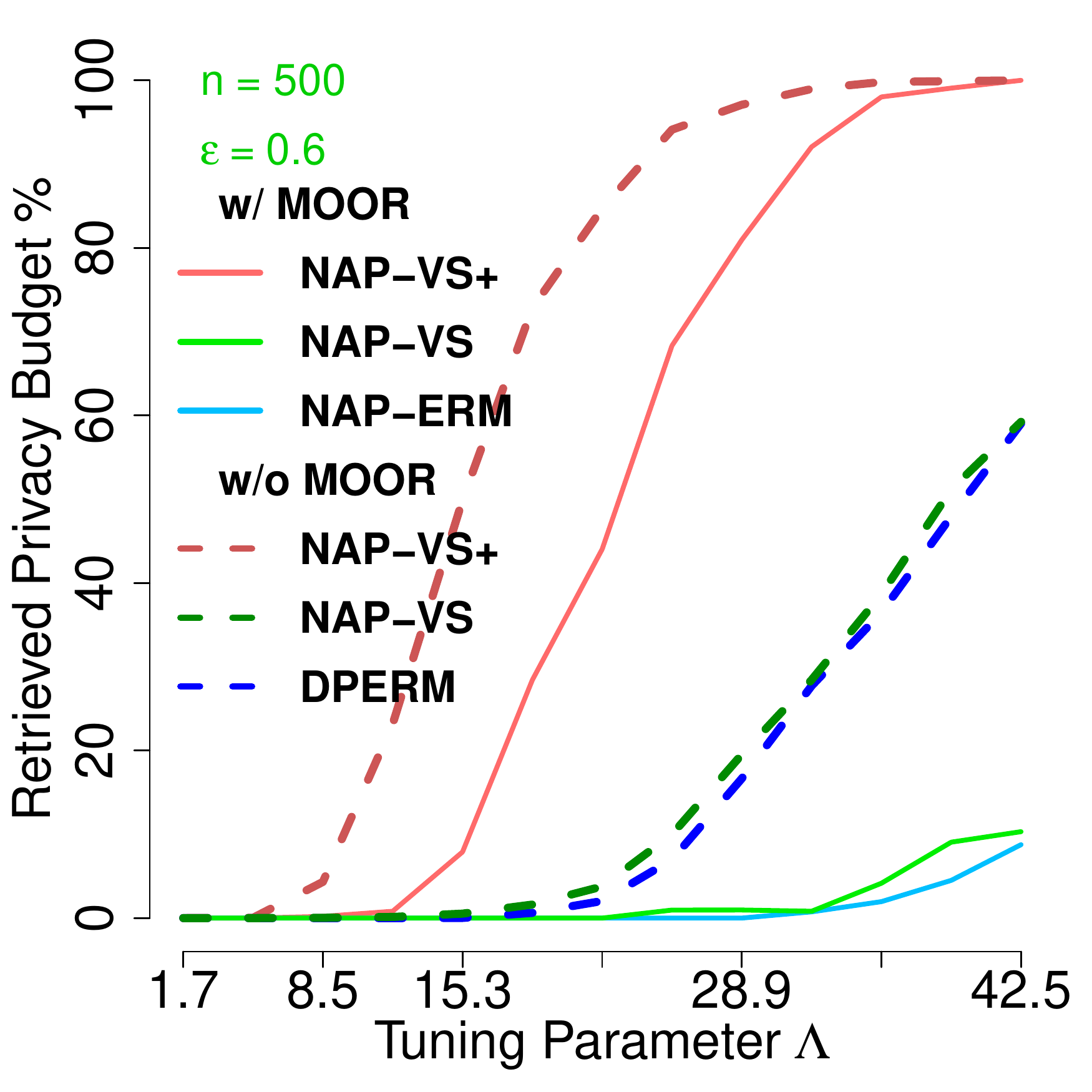}
\includegraphics[width=0.19\linewidth, trim=4pt 9pt 15pt 18pt,clip]{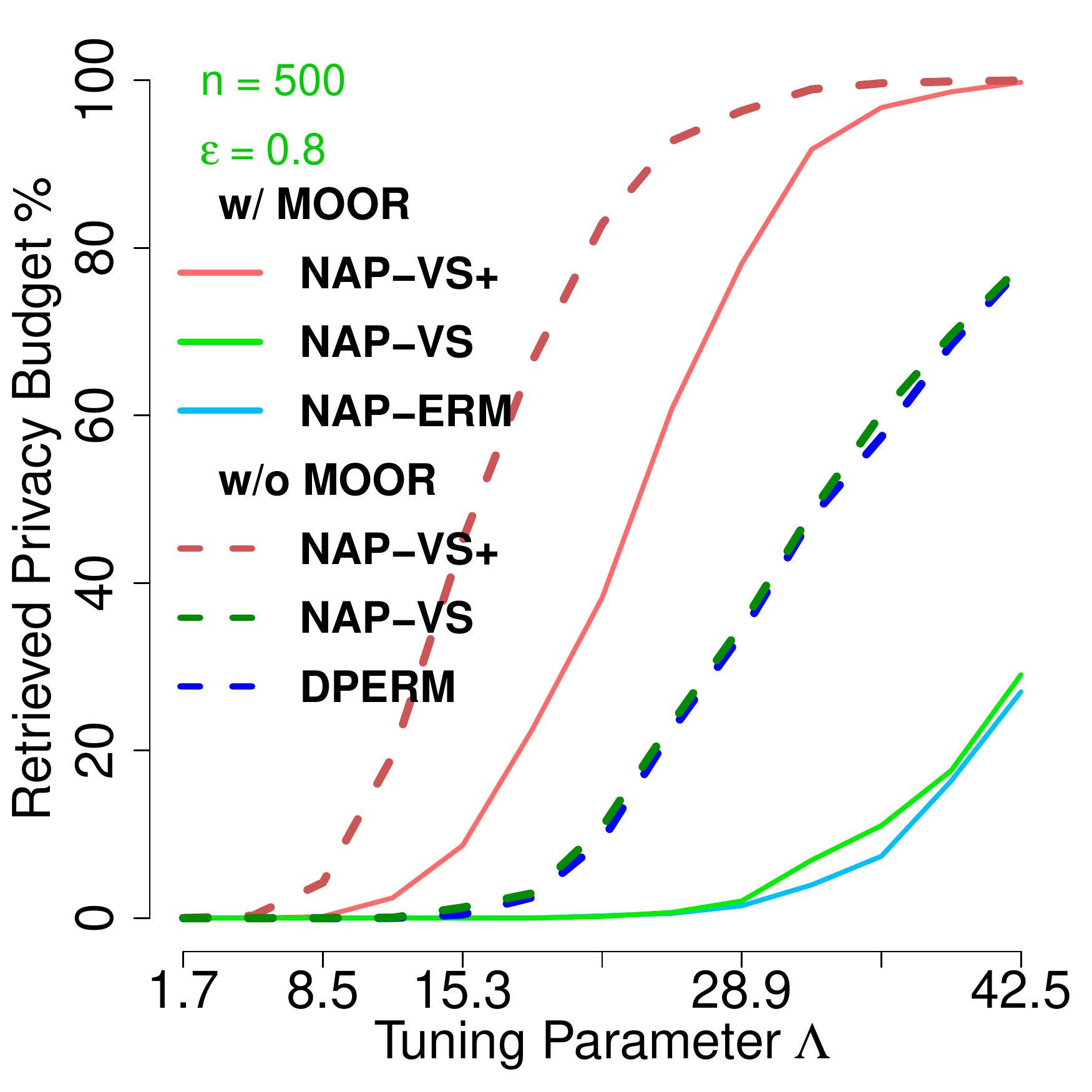}
\includegraphics[width=0.19\linewidth, trim=4pt 9pt 15pt 18pt,clip]{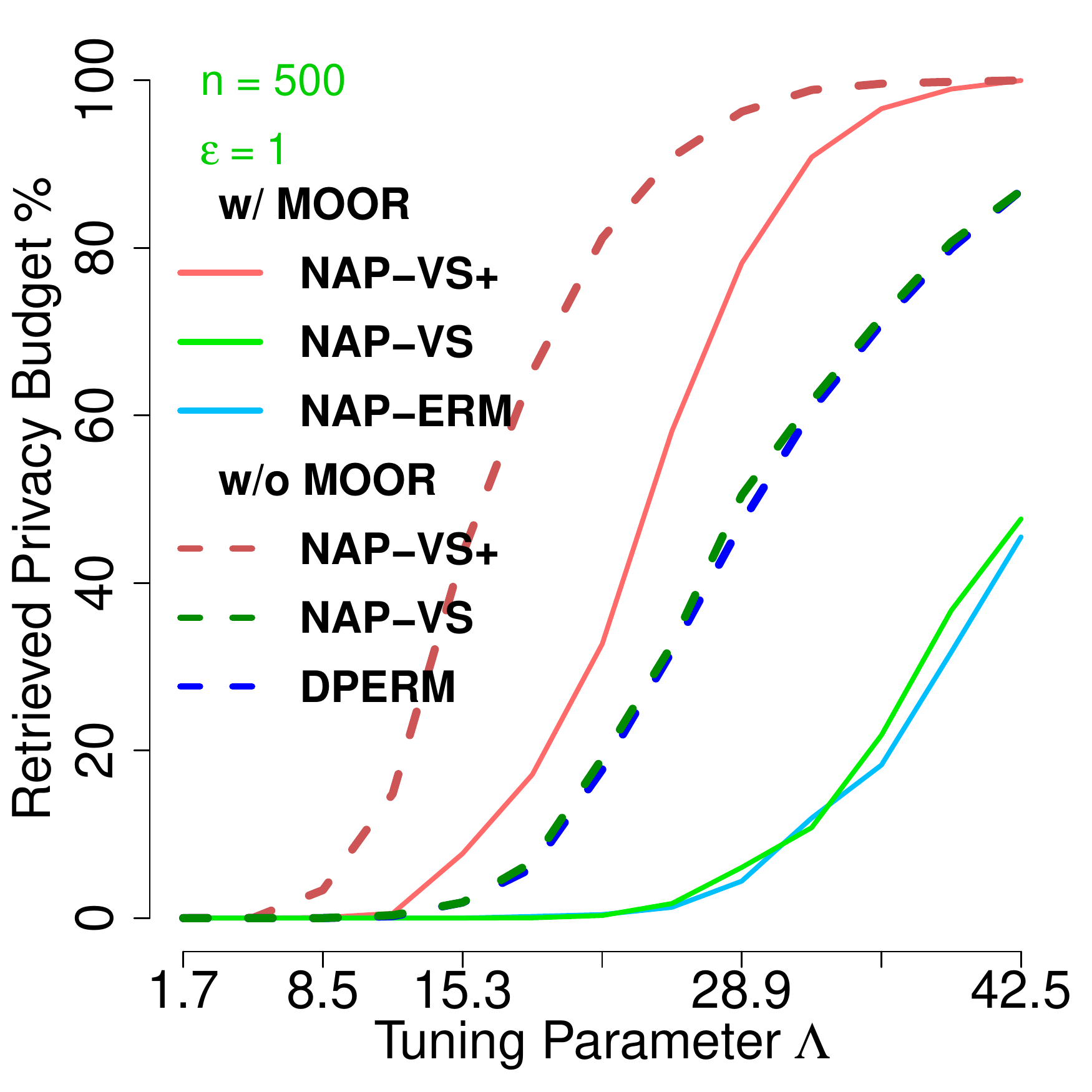}\\ \vspace{-3pt}
\footnotesize Poisson regression $n=1000$\\
\includegraphics[width=0.19\linewidth, trim=4pt 9pt 15pt 18pt,clip]{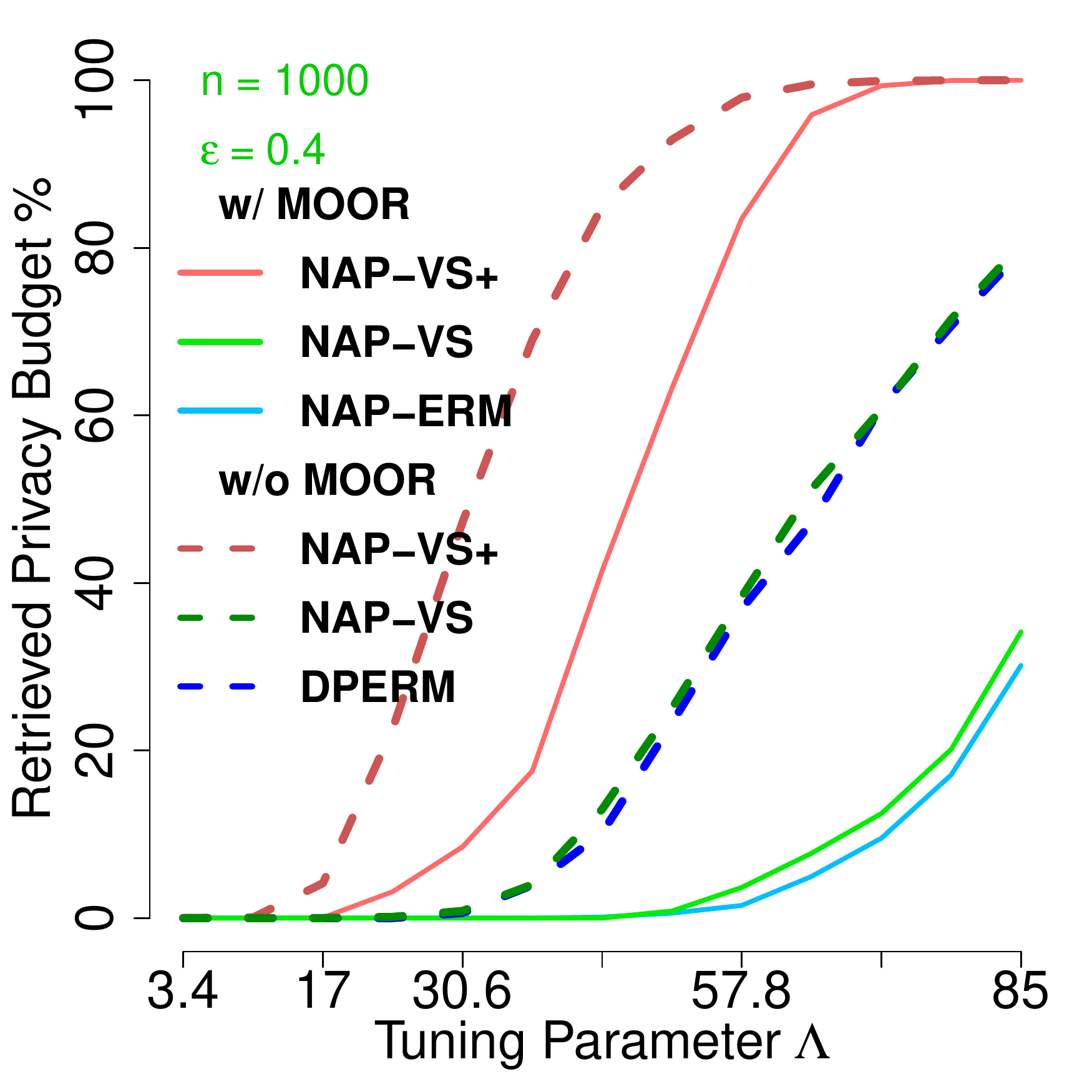}
\includegraphics[width=0.19\linewidth, trim=4pt 9pt 15pt 18pt,clip]{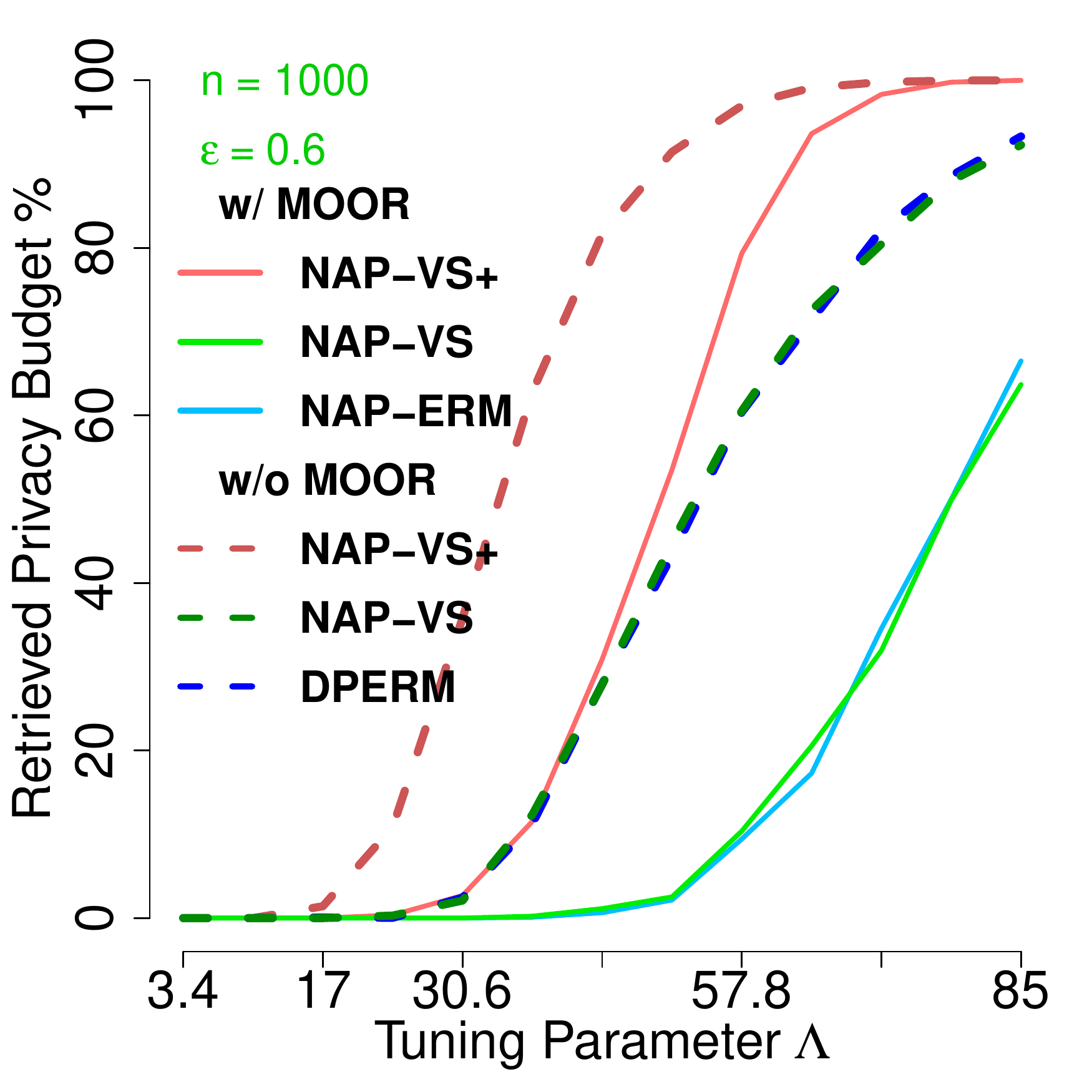}
\includegraphics[width=0.19\linewidth, trim=4pt 9pt 15pt 18pt,clip]{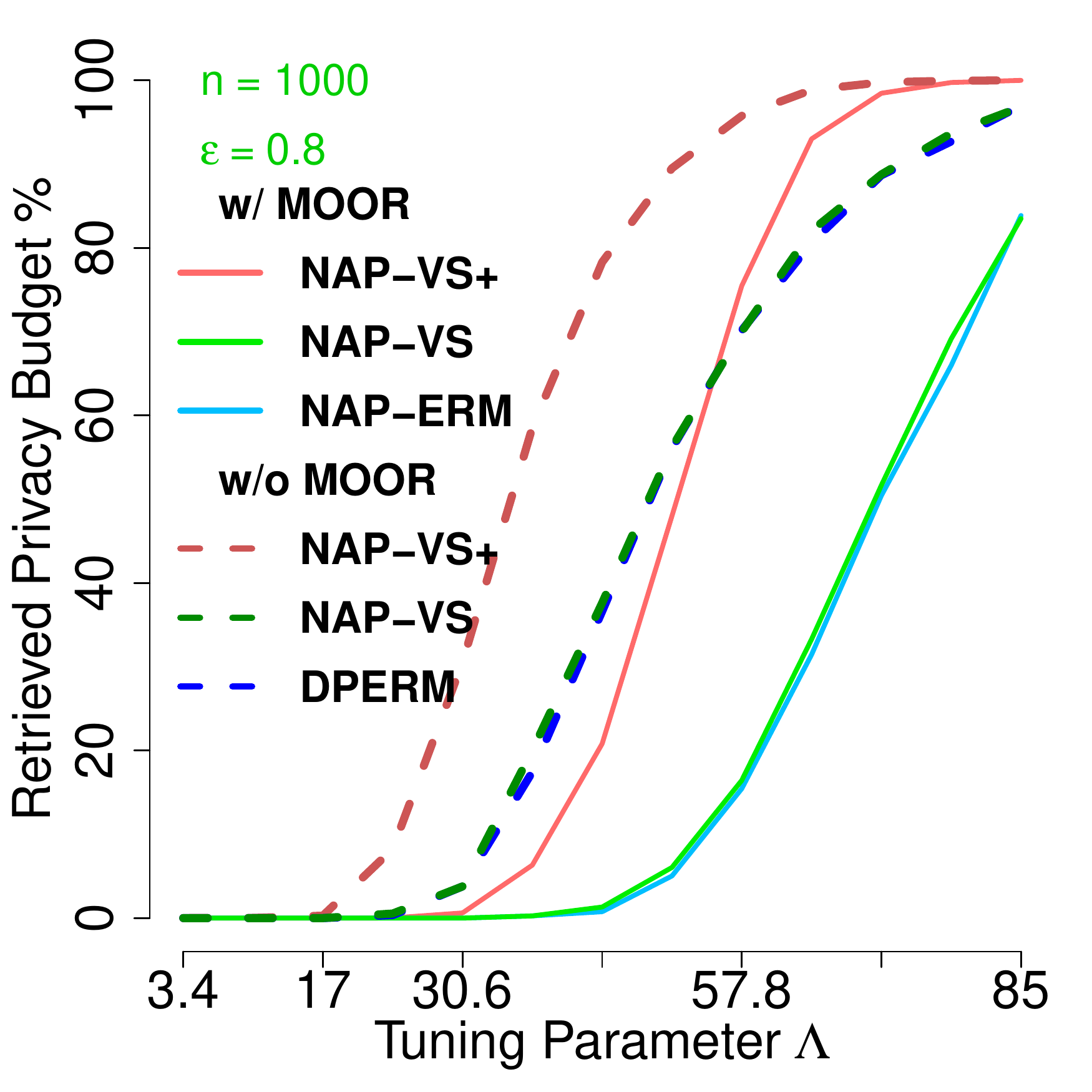}
\includegraphics[width=0.19\linewidth, trim=4pt 9pt 15pt 18pt,clip]{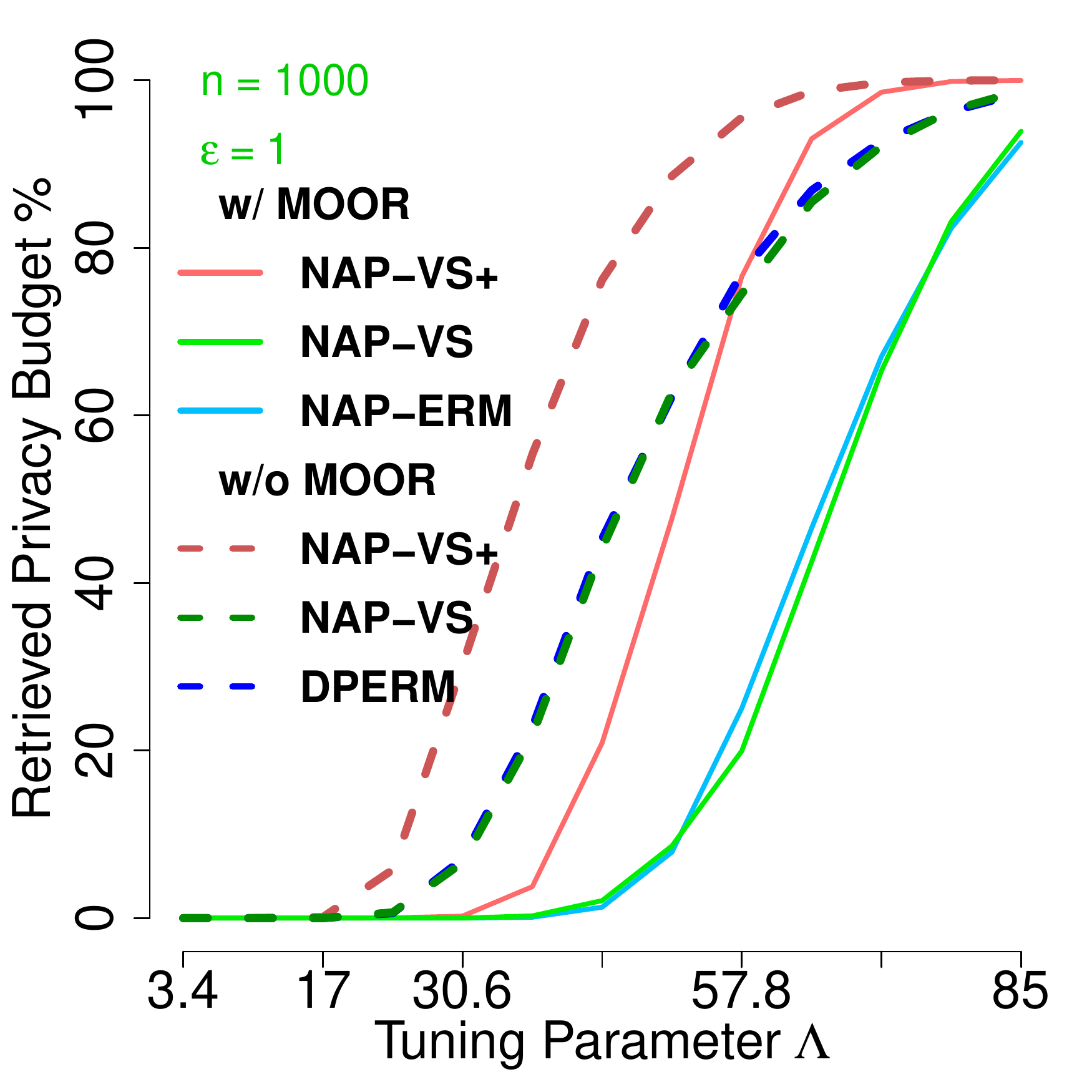}\\\vspace{-3pt}
\footnotesize logistic regression $n=500$ \\
\includegraphics[width=0.19\linewidth, trim=4pt 9pt 15pt 18pt,clip]{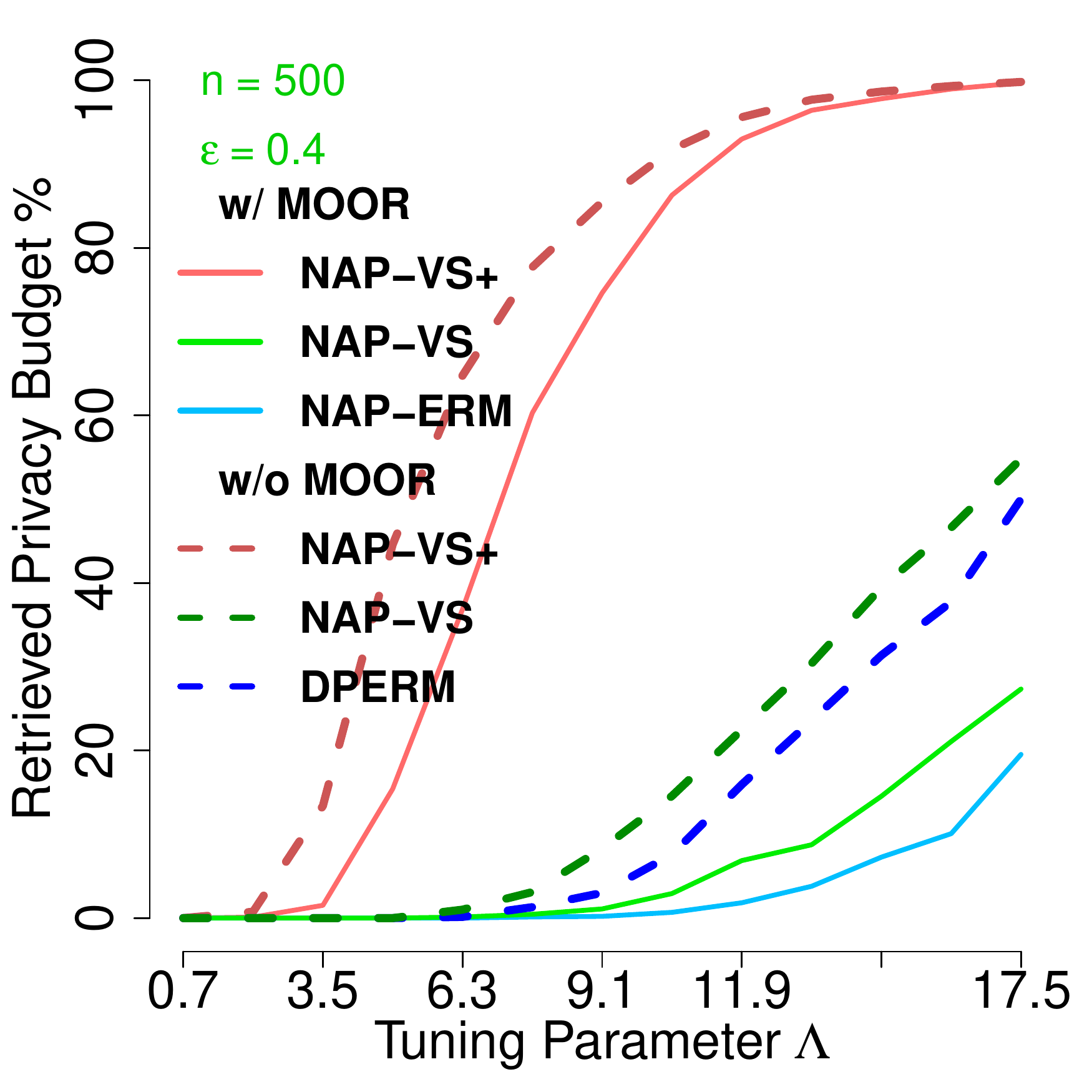}
\includegraphics[width=0.19\linewidth, trim=4pt 9pt 15pt 18pt,clip]{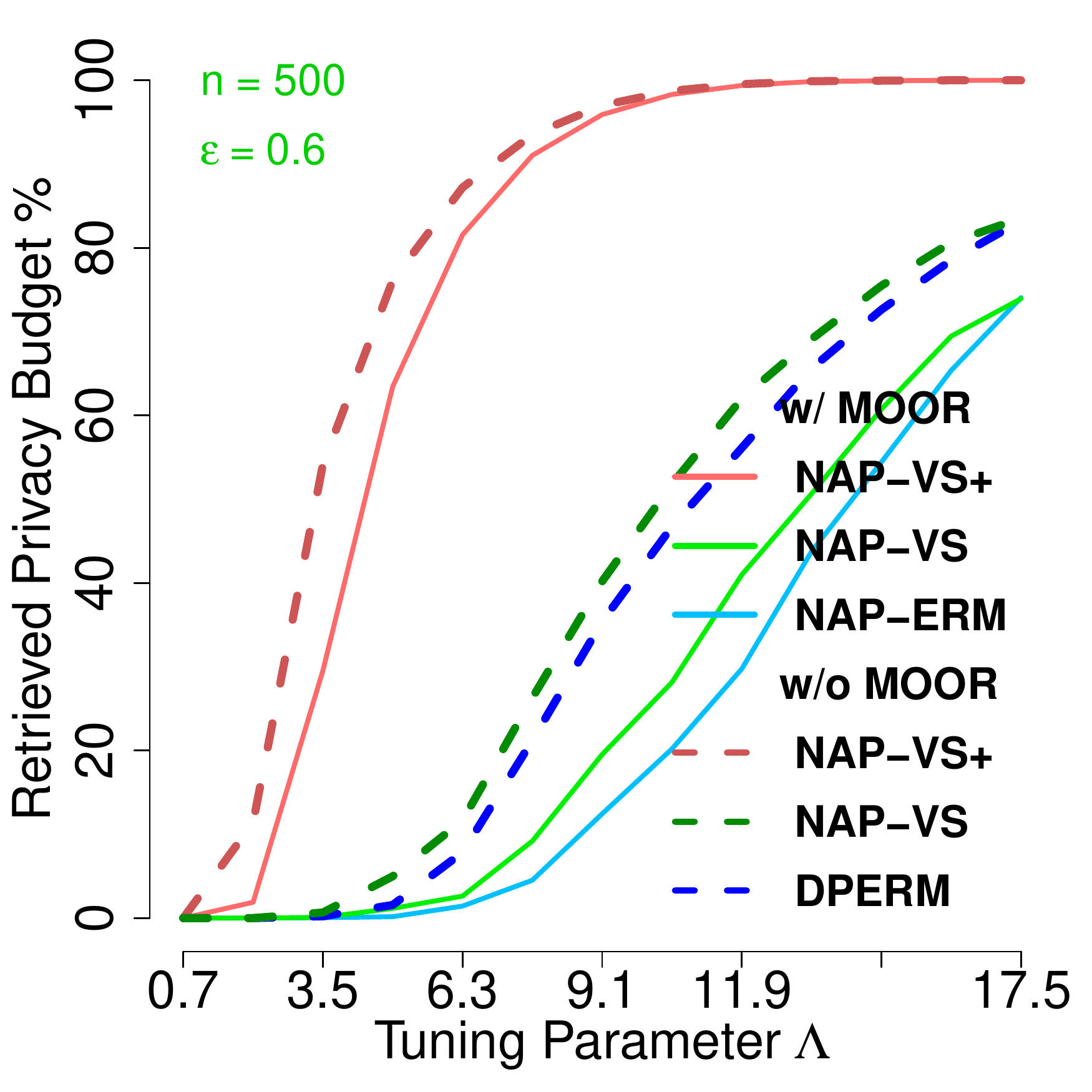}
\includegraphics[width=0.19\linewidth, trim=4pt 9pt 15pt 18pt,clip]{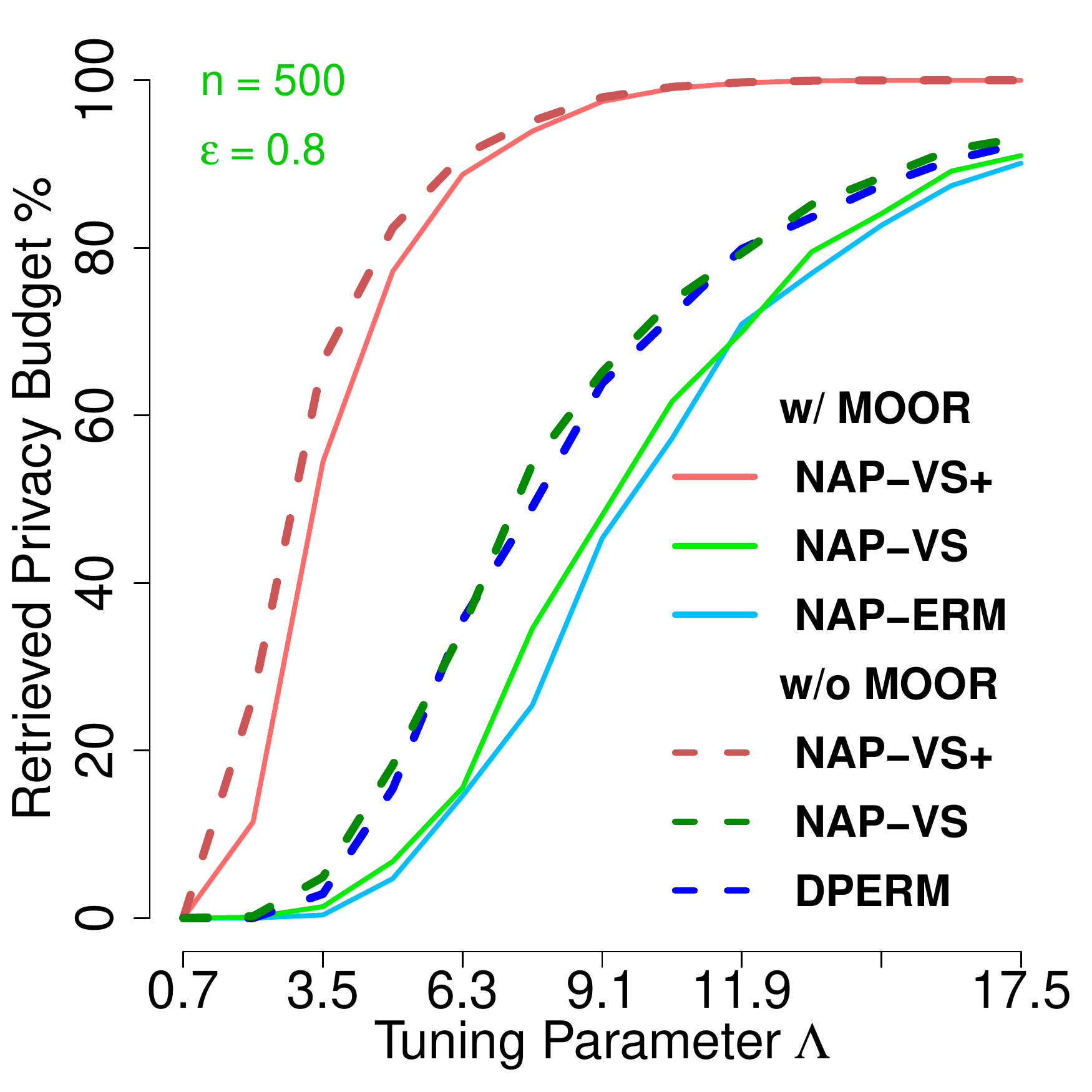}
\includegraphics[width=0.19\linewidth, trim=4pt 9pt 15pt 18pt,clip]{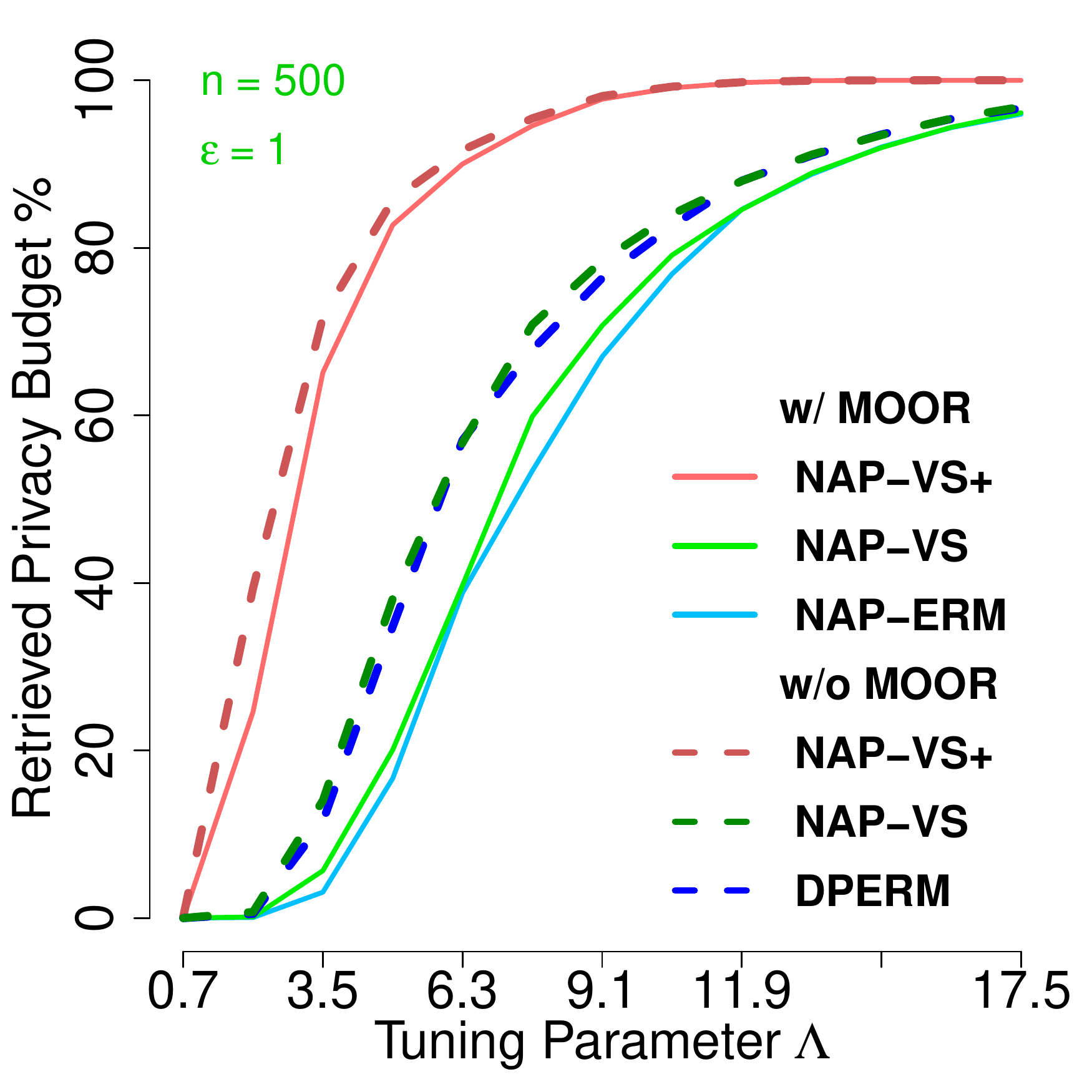}\\\vspace{-3pt}
\footnotesize{logistic regression $n\!=\!1000$} \\
\includegraphics[width=0.19\linewidth, trim=4pt 9pt 15pt 18pt,clip]{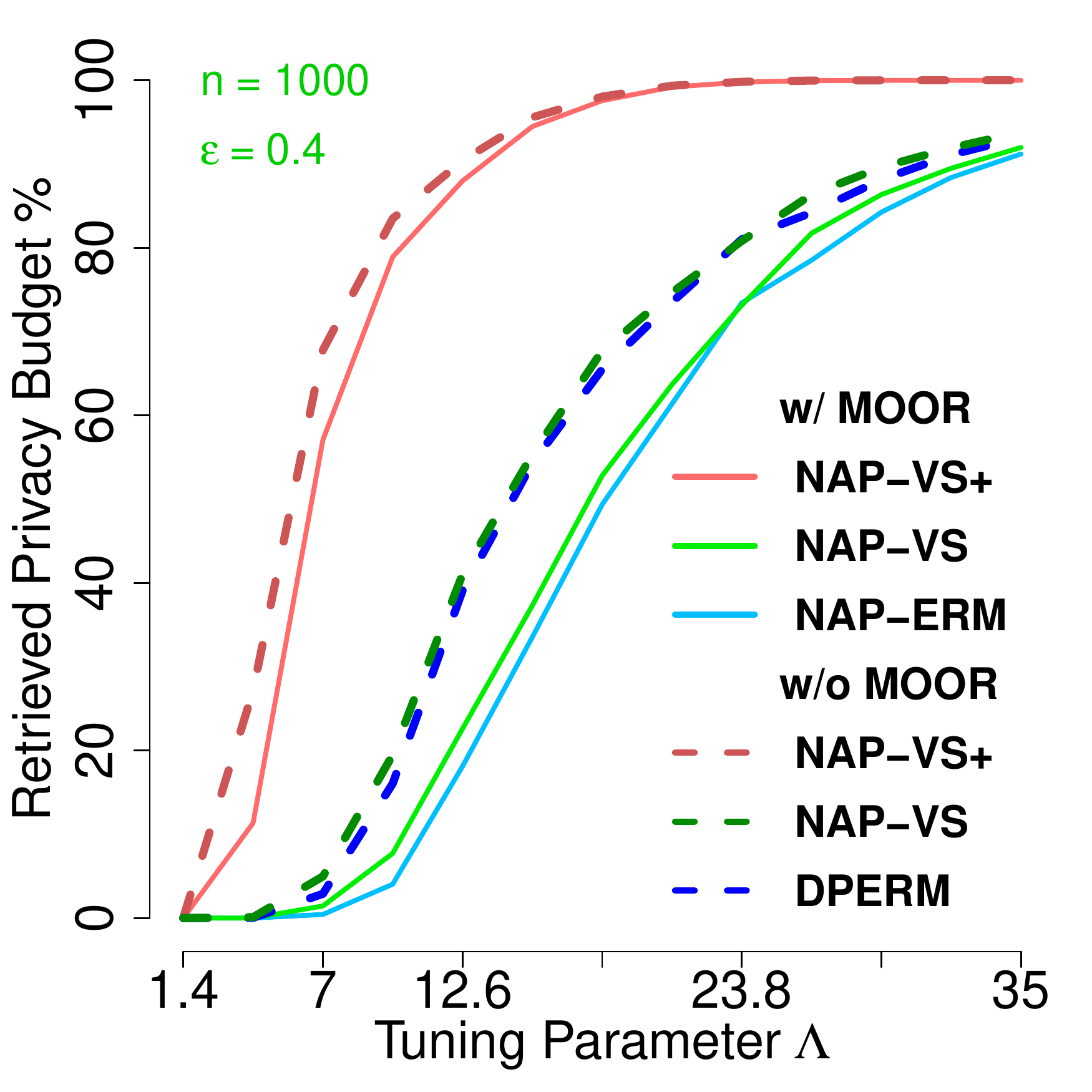}
\includegraphics[width=0.19\linewidth, trim=4pt 9pt 15pt 18pt,clip]{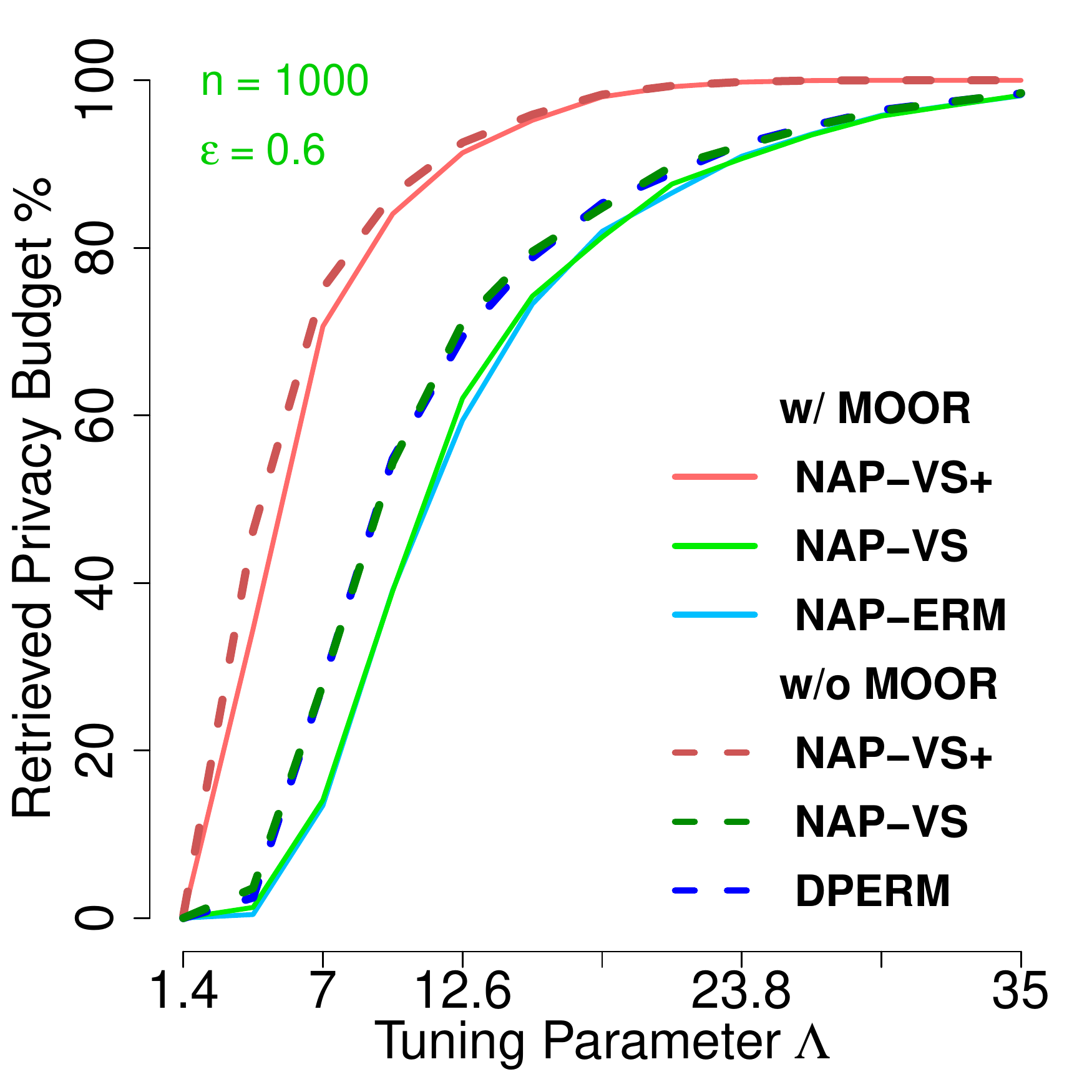}
\includegraphics[width=0.19\linewidth, trim=4pt 9pt 15pt 18pt,clip]{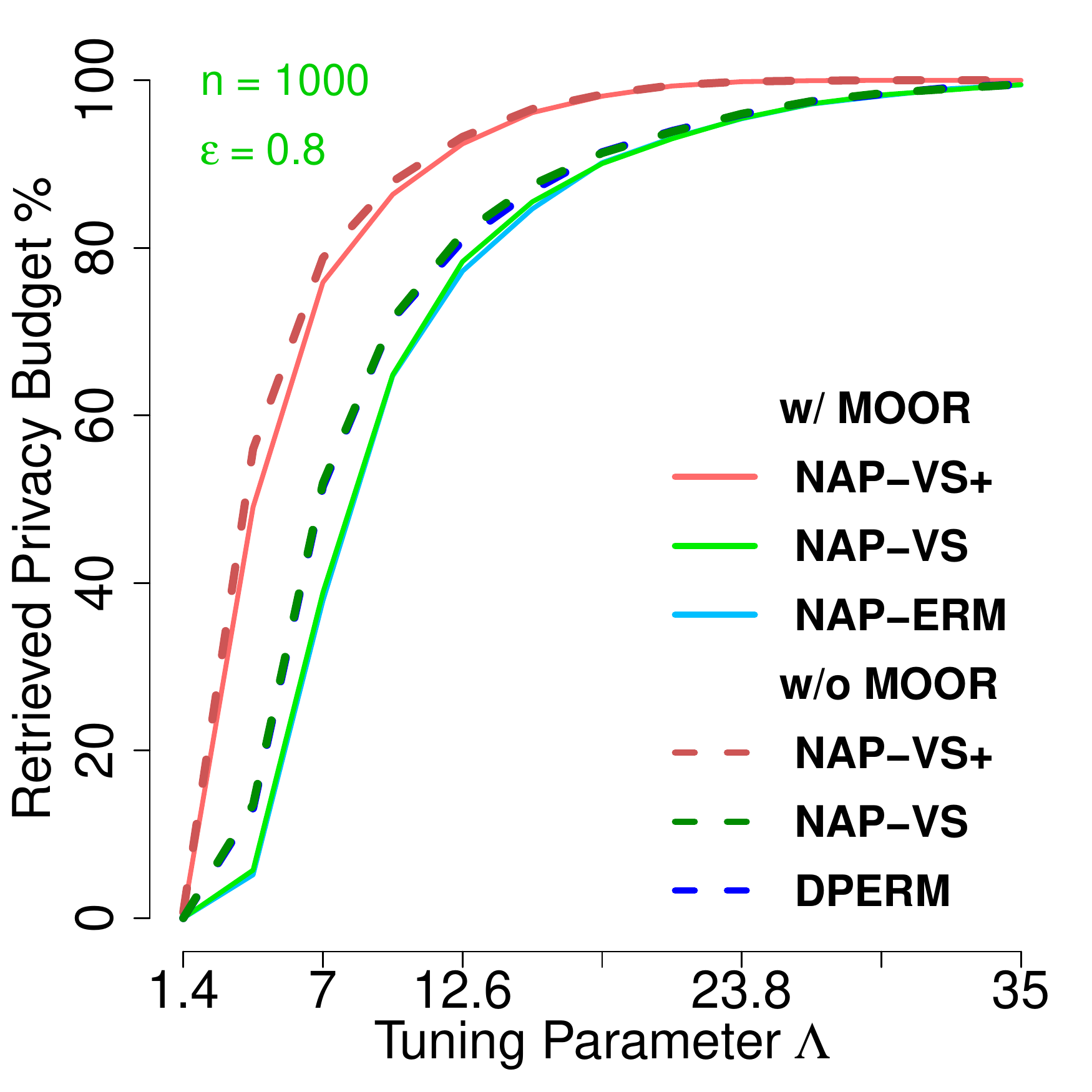}
\includegraphics[width=0.19\linewidth, trim=4pt 9pt 15pt 18pt,clip]{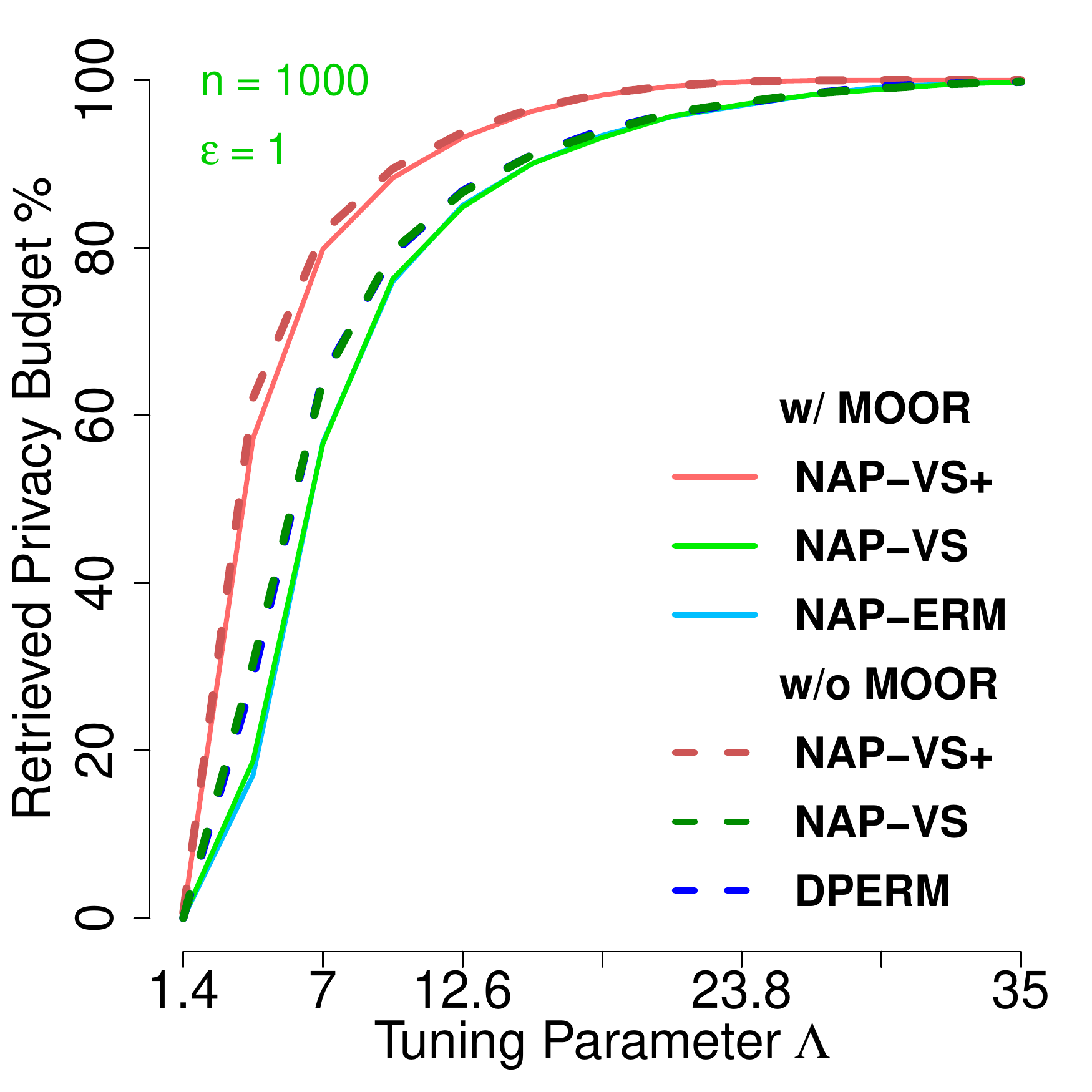}
\vspace{-12pt}
\caption{Portion of retrieved privacy budget originally allocated to  bounding Jacobian ratio}  
\label{fig:retrieval}
\end{figure}

\begin{figure}[!htb]\centering\vspace{-9pt}
\footnotesize variable selection ROC curve via lasso $n=200$ \\
\includegraphics[width=0.20\linewidth, trim=4pt 9pt 15pt 18pt,clip]{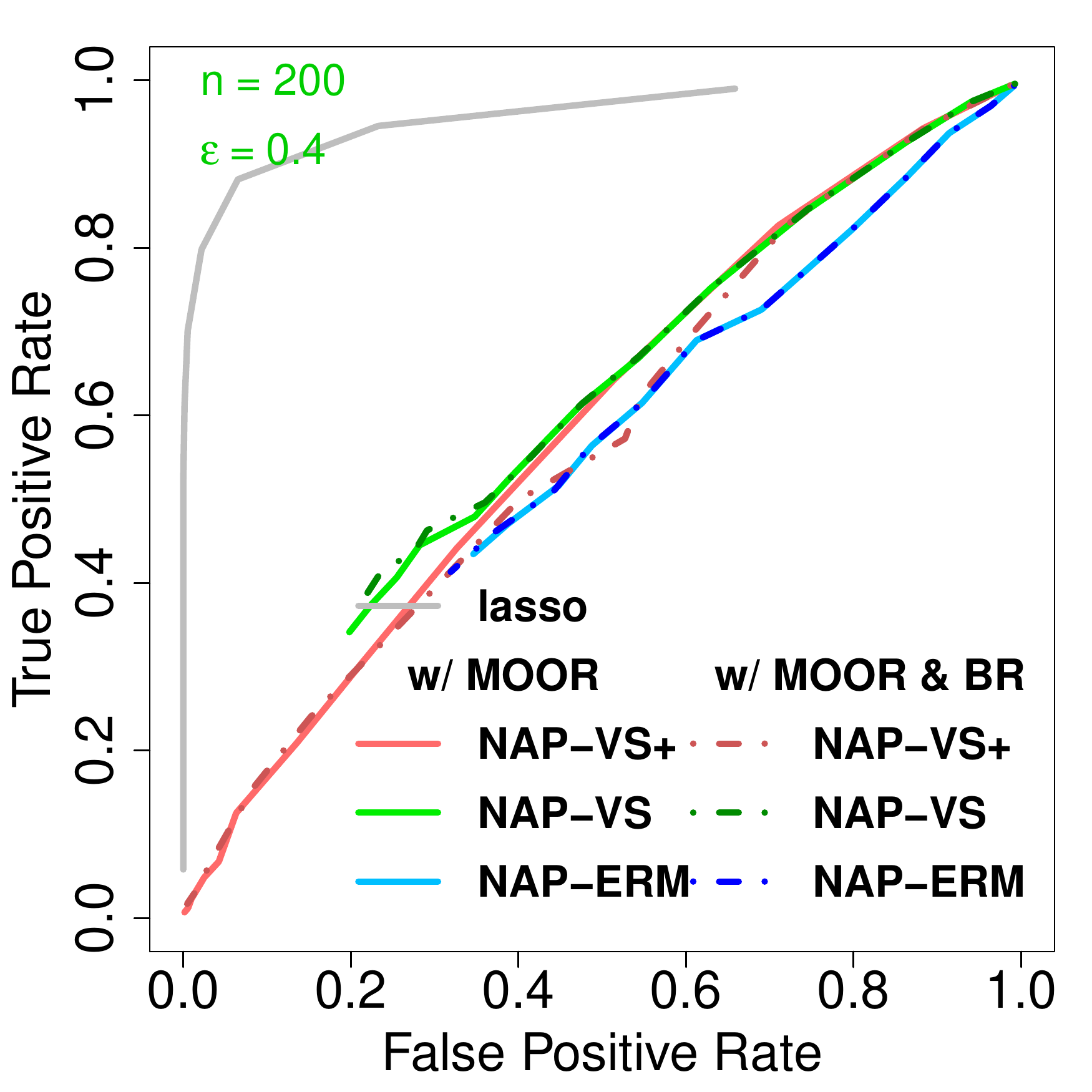}
\includegraphics[width=0.20\linewidth, trim=4pt 9pt 15pt 18pt,clip]{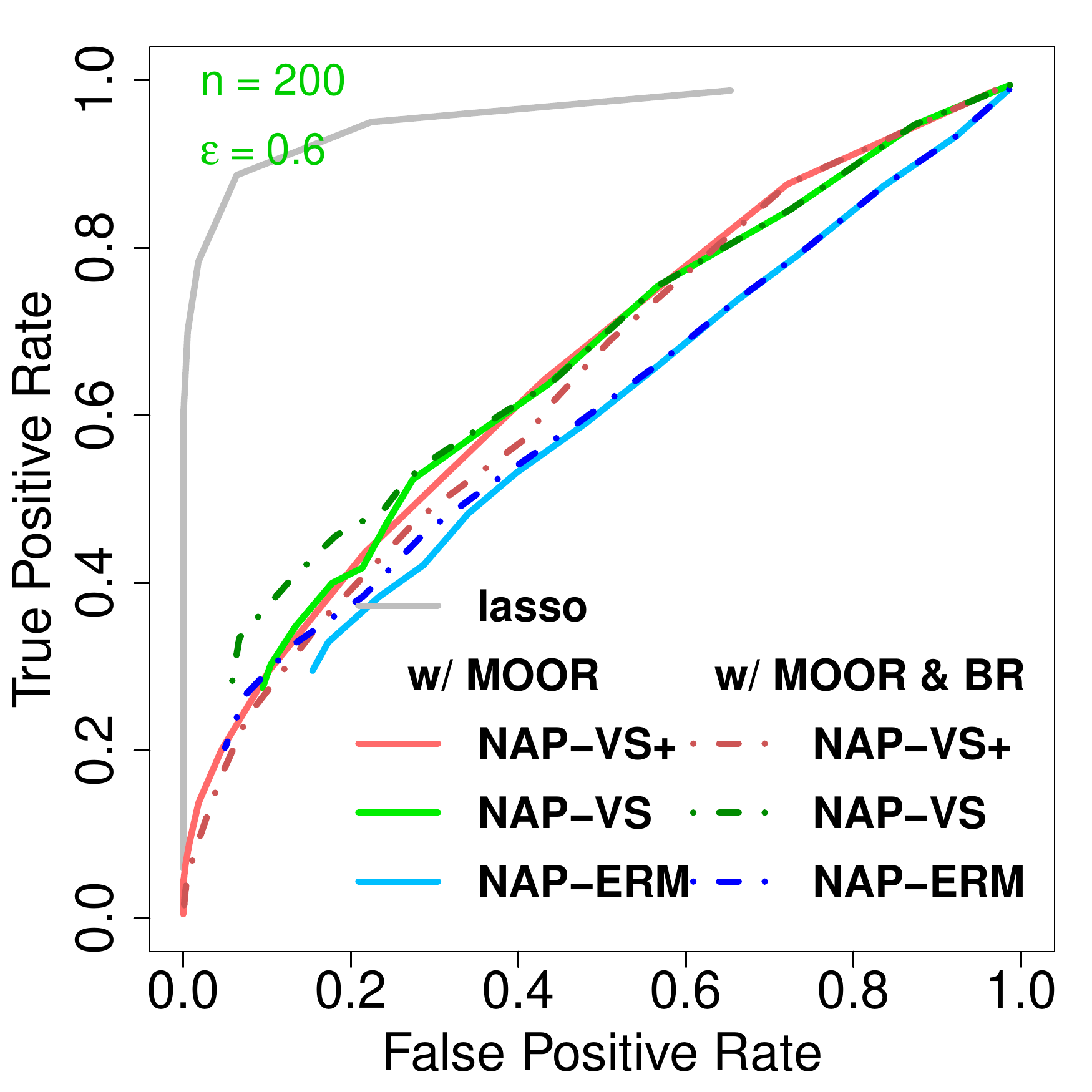}
\includegraphics[width=0.20\linewidth, trim=4pt 9pt 15pt 18pt,clip]{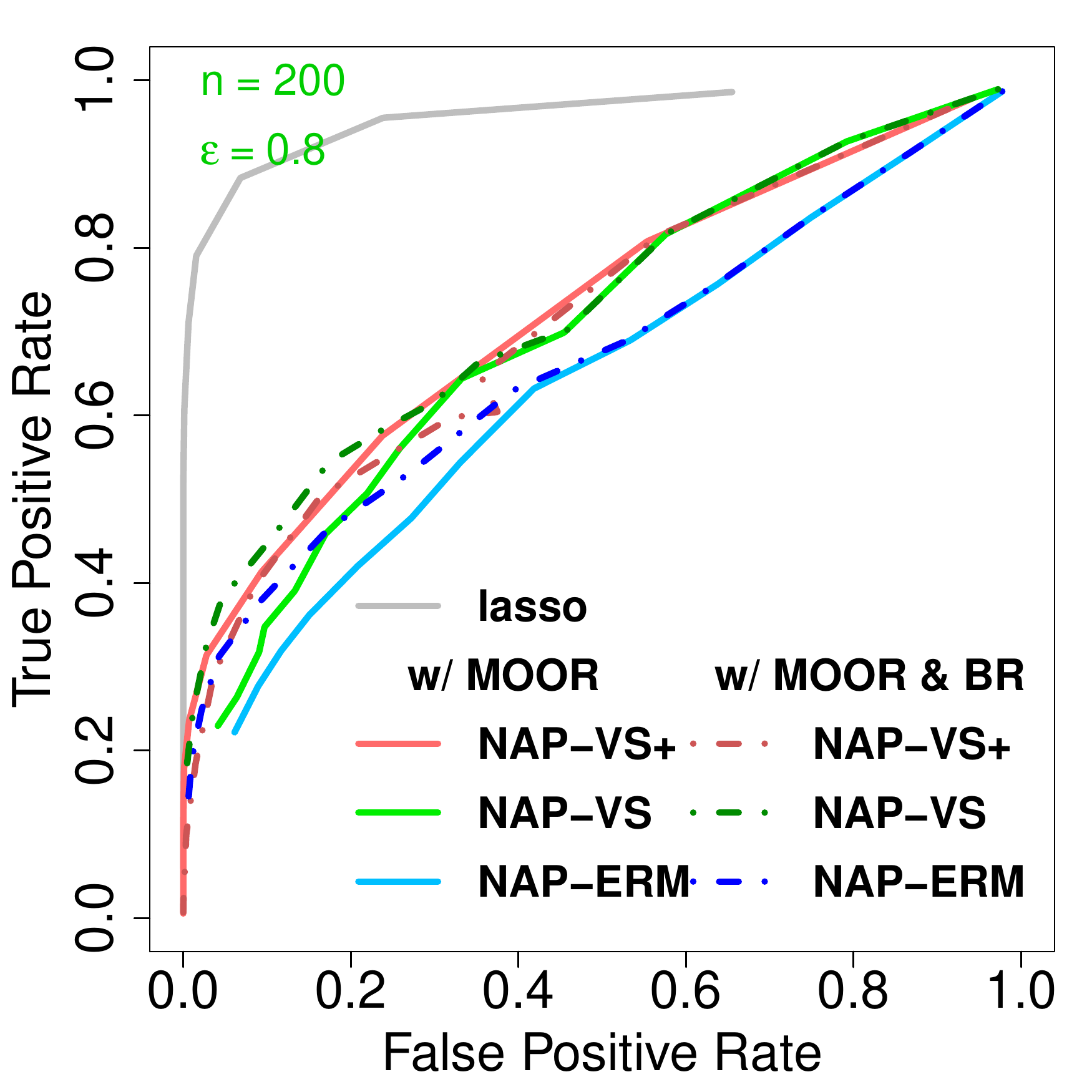}
\includegraphics[width=0.20\linewidth, trim=4pt 9pt 15pt 18pt,clip]{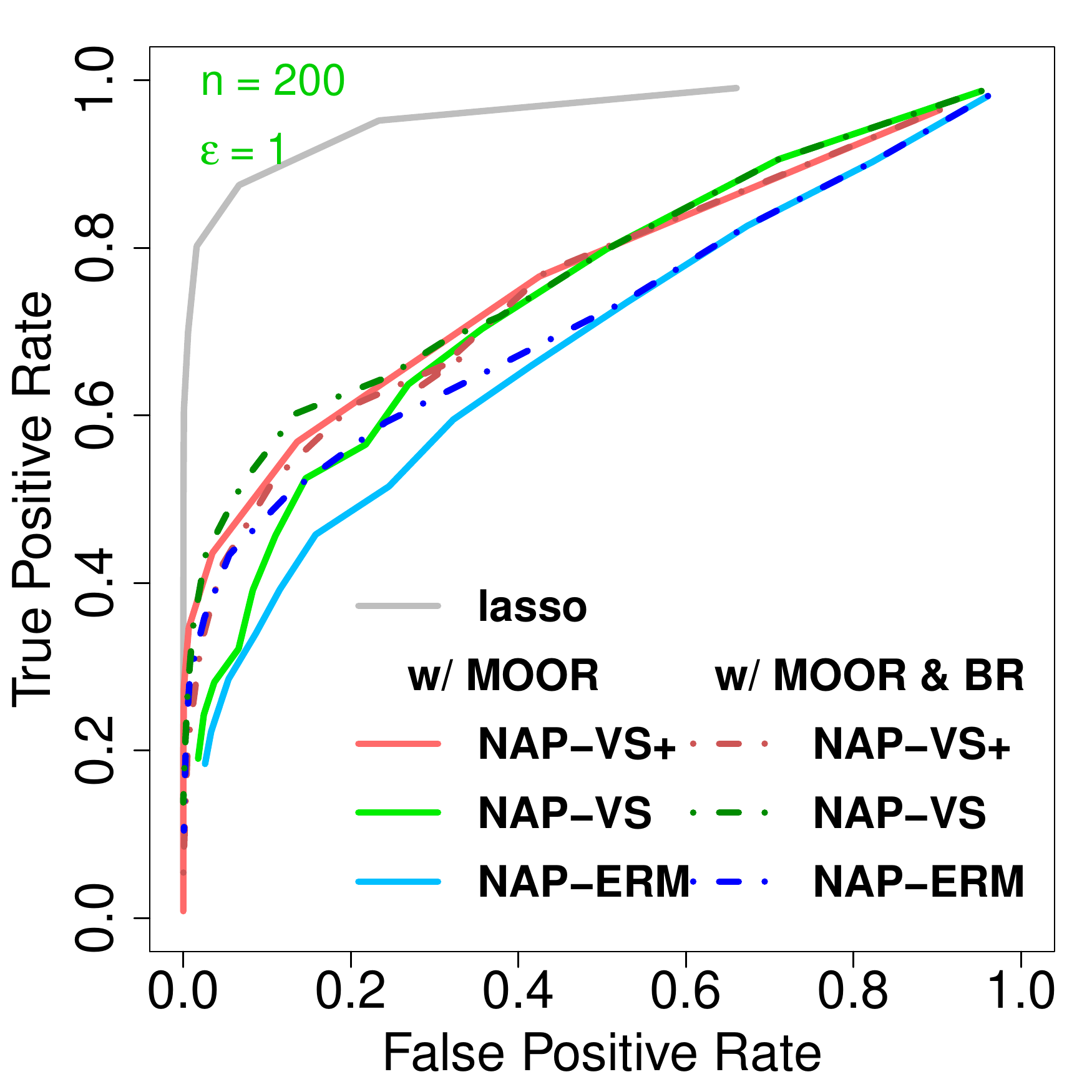}\\
\footnotesize $n=500$\\
\includegraphics[width=0.20\linewidth, trim=4pt 9pt 15pt 18pt,clip]{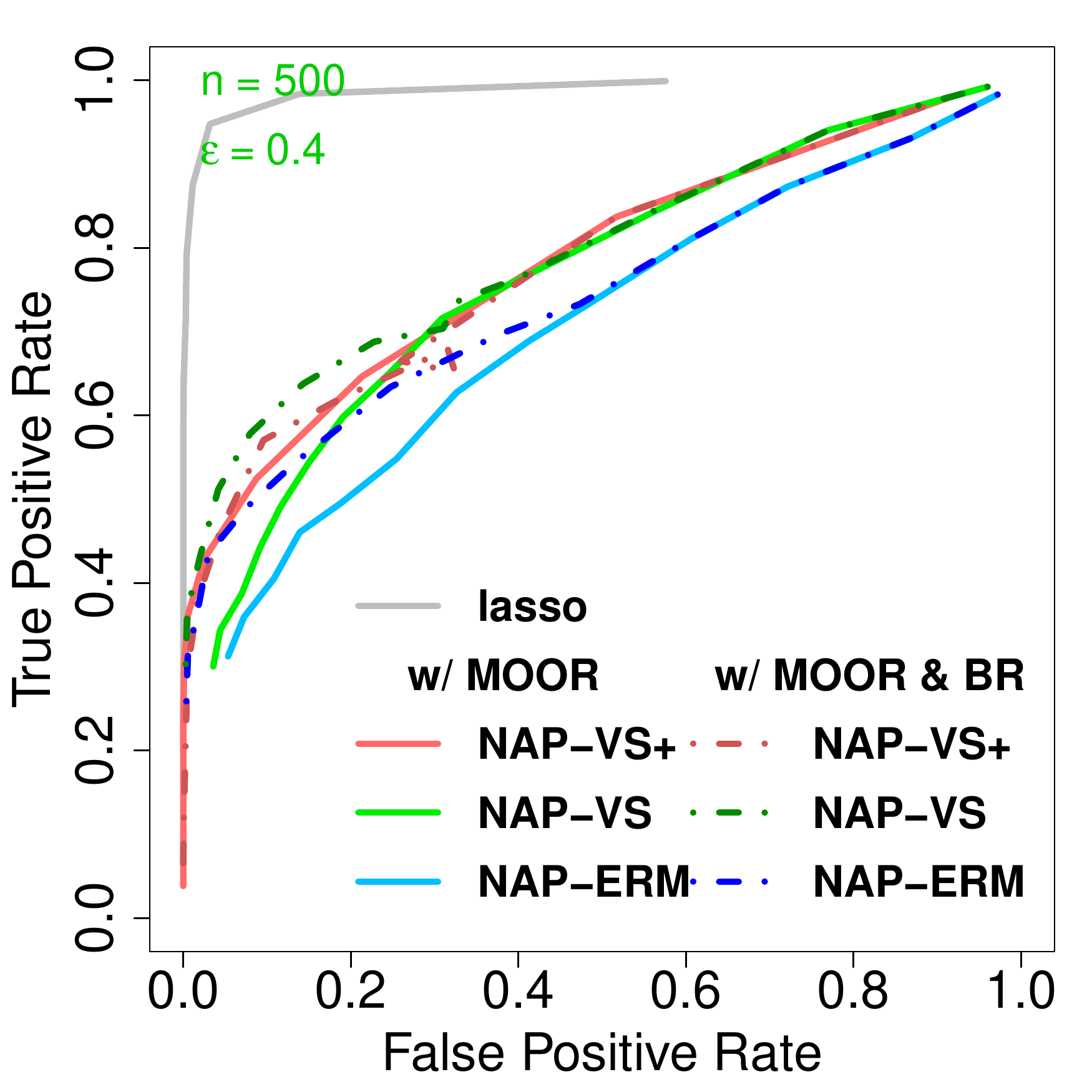}
\includegraphics[width=0.20\linewidth, trim=4pt 9pt 15pt 18pt,clip]{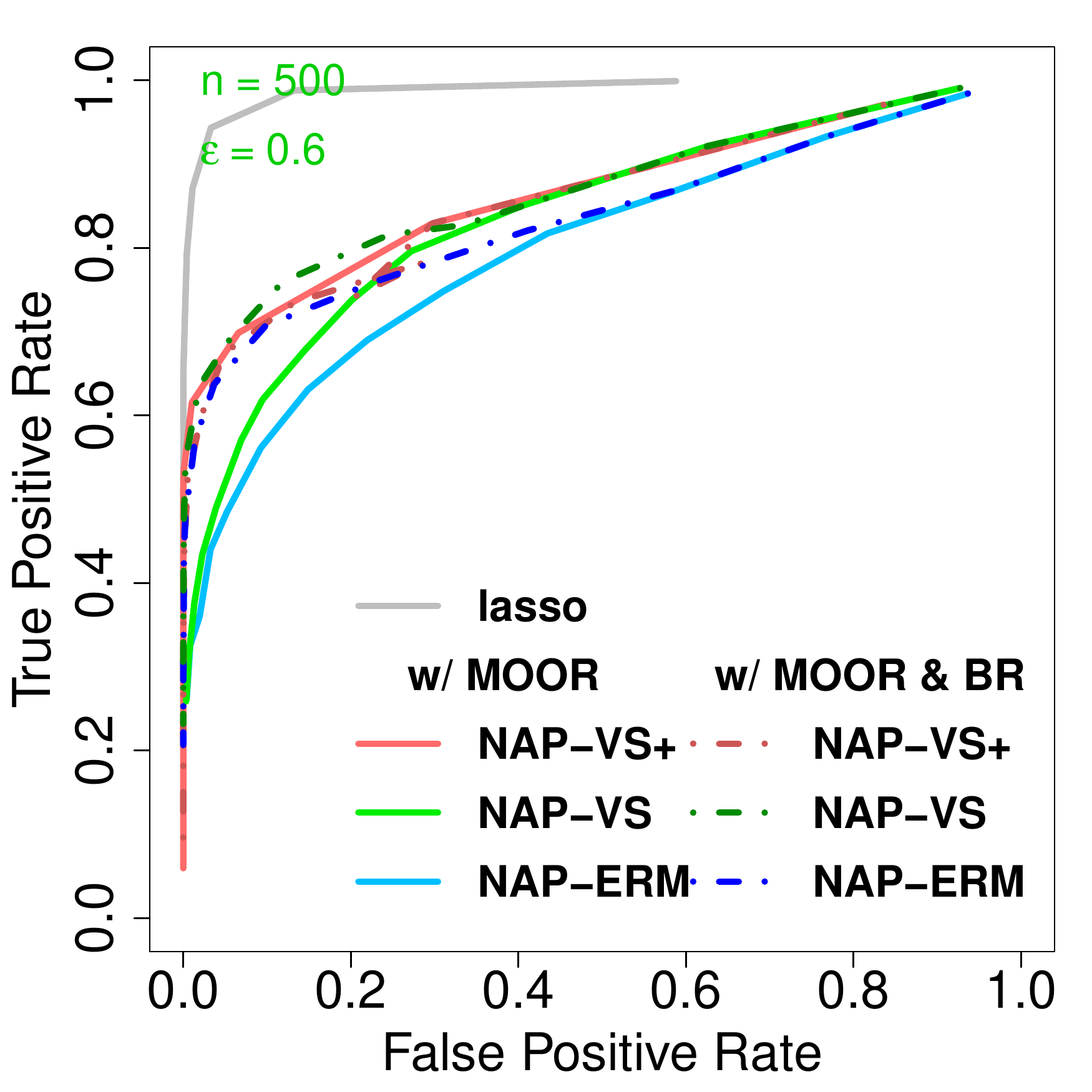}
\includegraphics[width=0.20\linewidth, trim=4pt 9pt 15pt 18pt,clip]{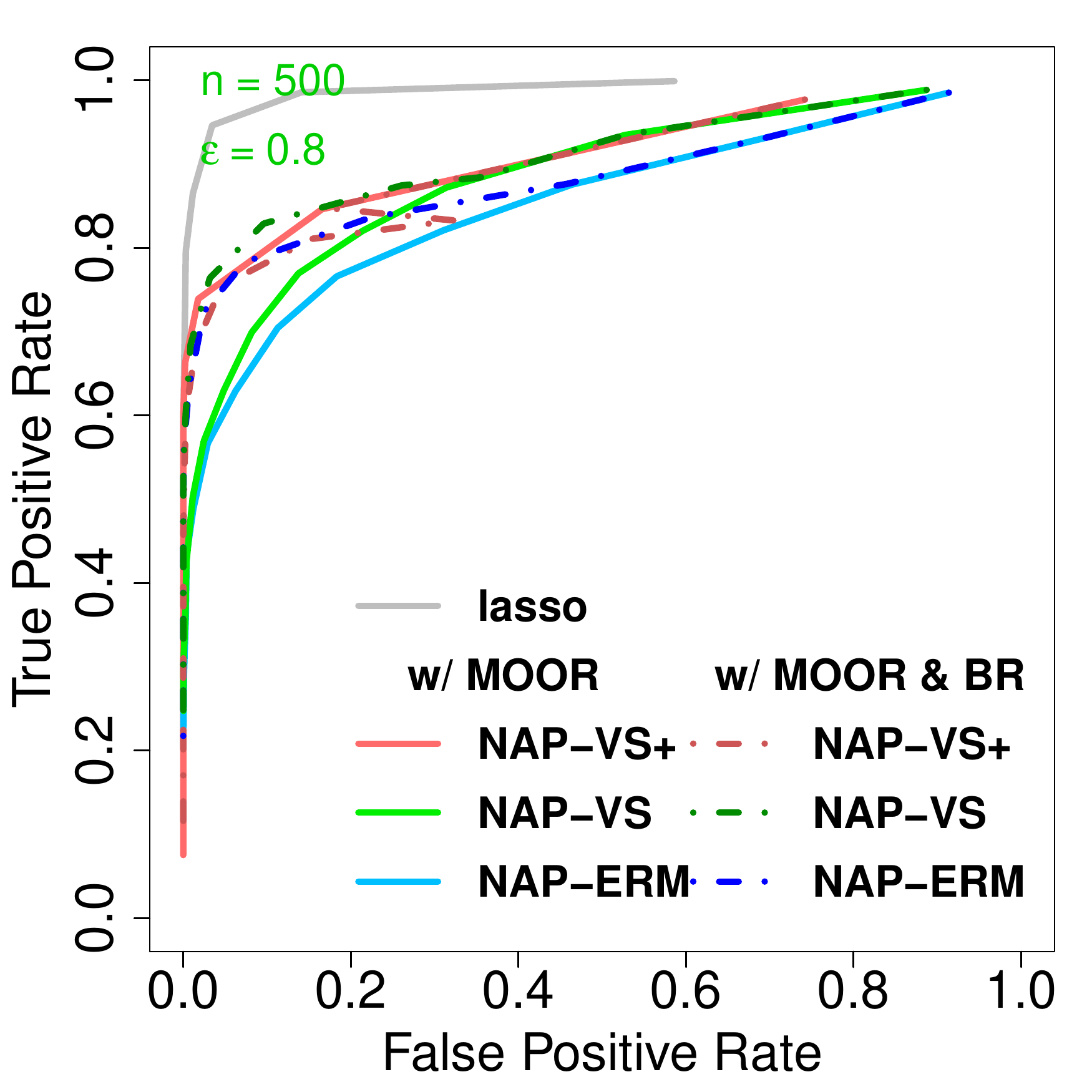}
\includegraphics[width=0.20\linewidth, trim=4pt 9pt 15pt 18pt,clip]{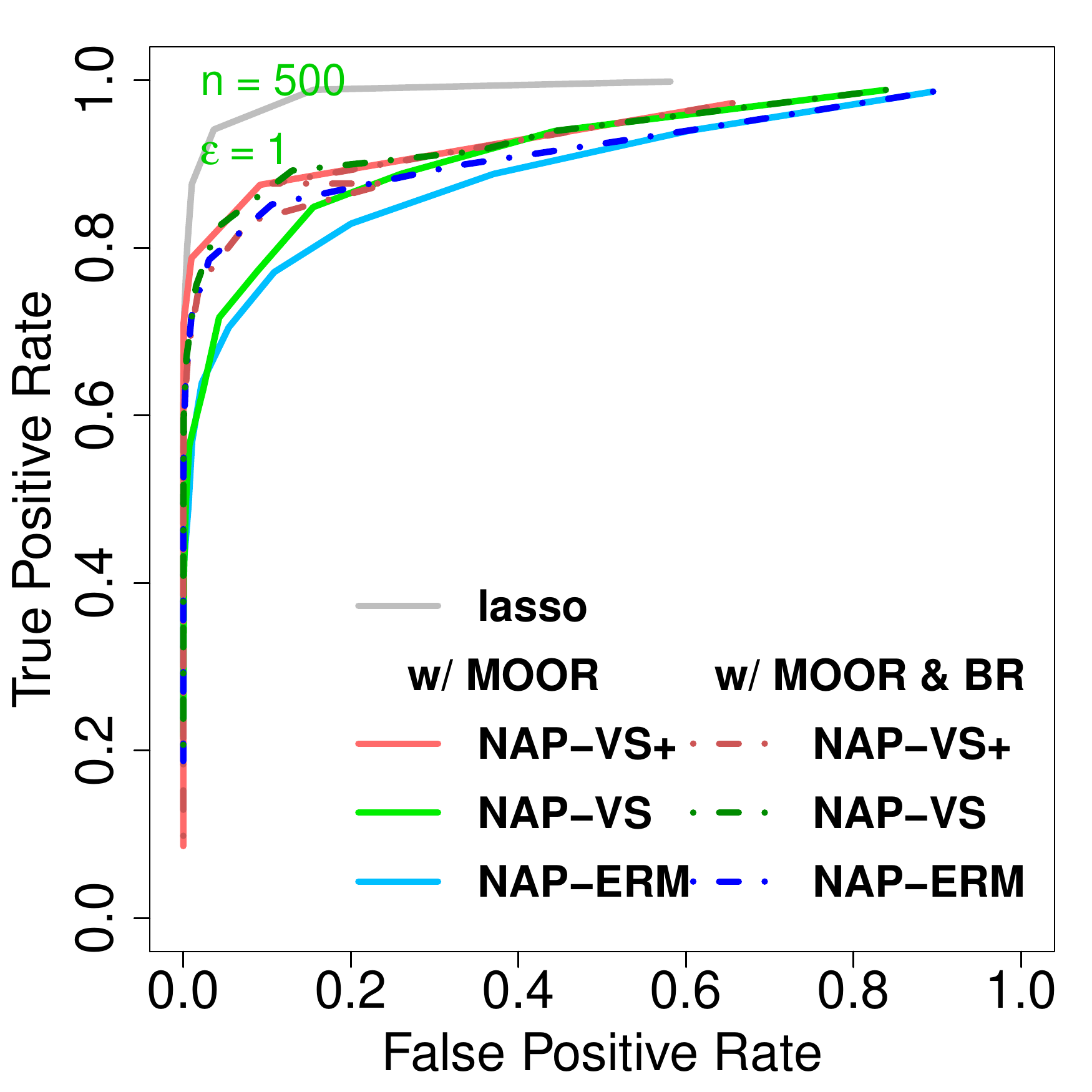} \\
outcome prediction MSE $n=200$ \\
\includegraphics[width=0.20\linewidth, trim=4pt 9pt 15pt 18pt,clip]{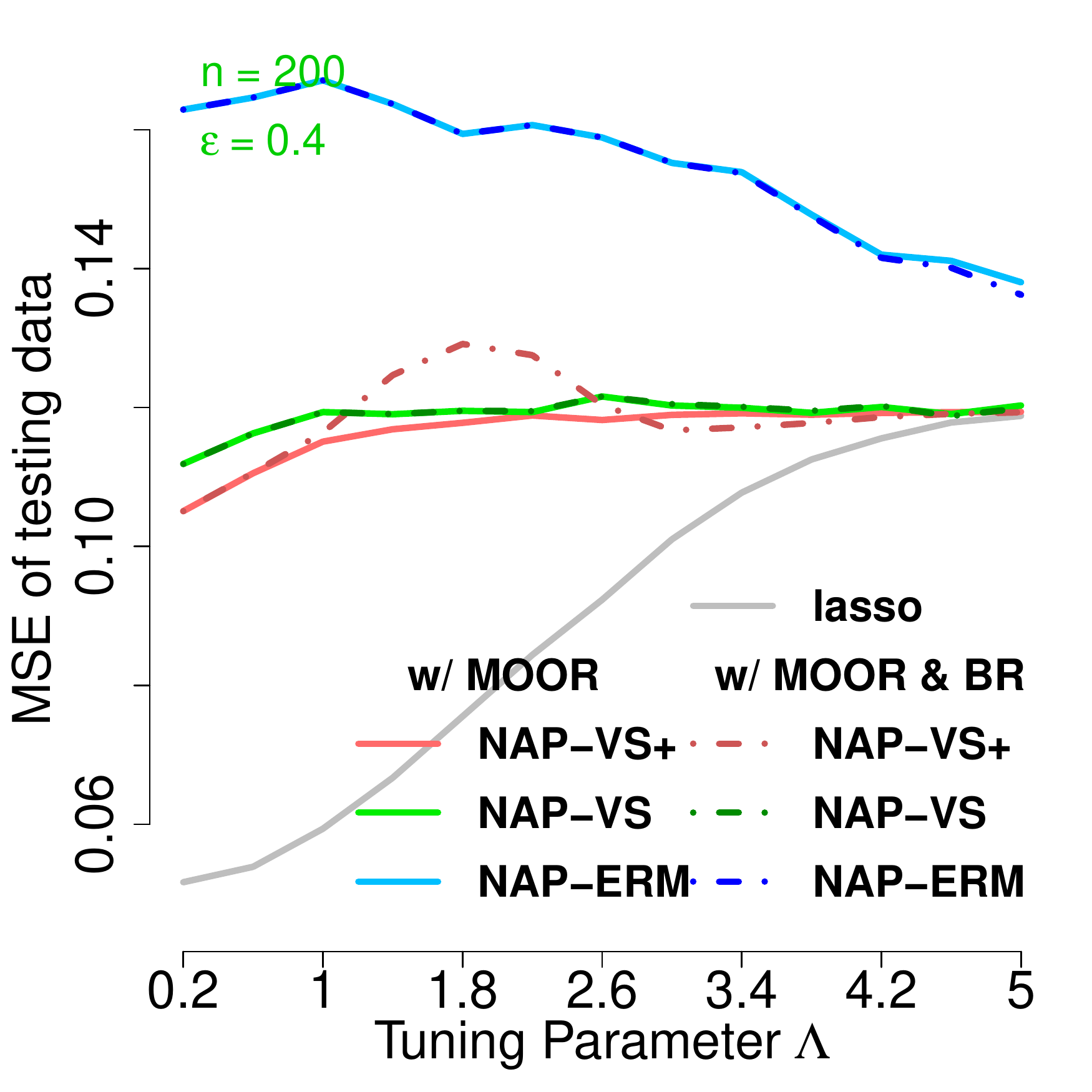}
\includegraphics[width=0.20\linewidth, trim=4pt 9pt 15pt 18pt,clip]{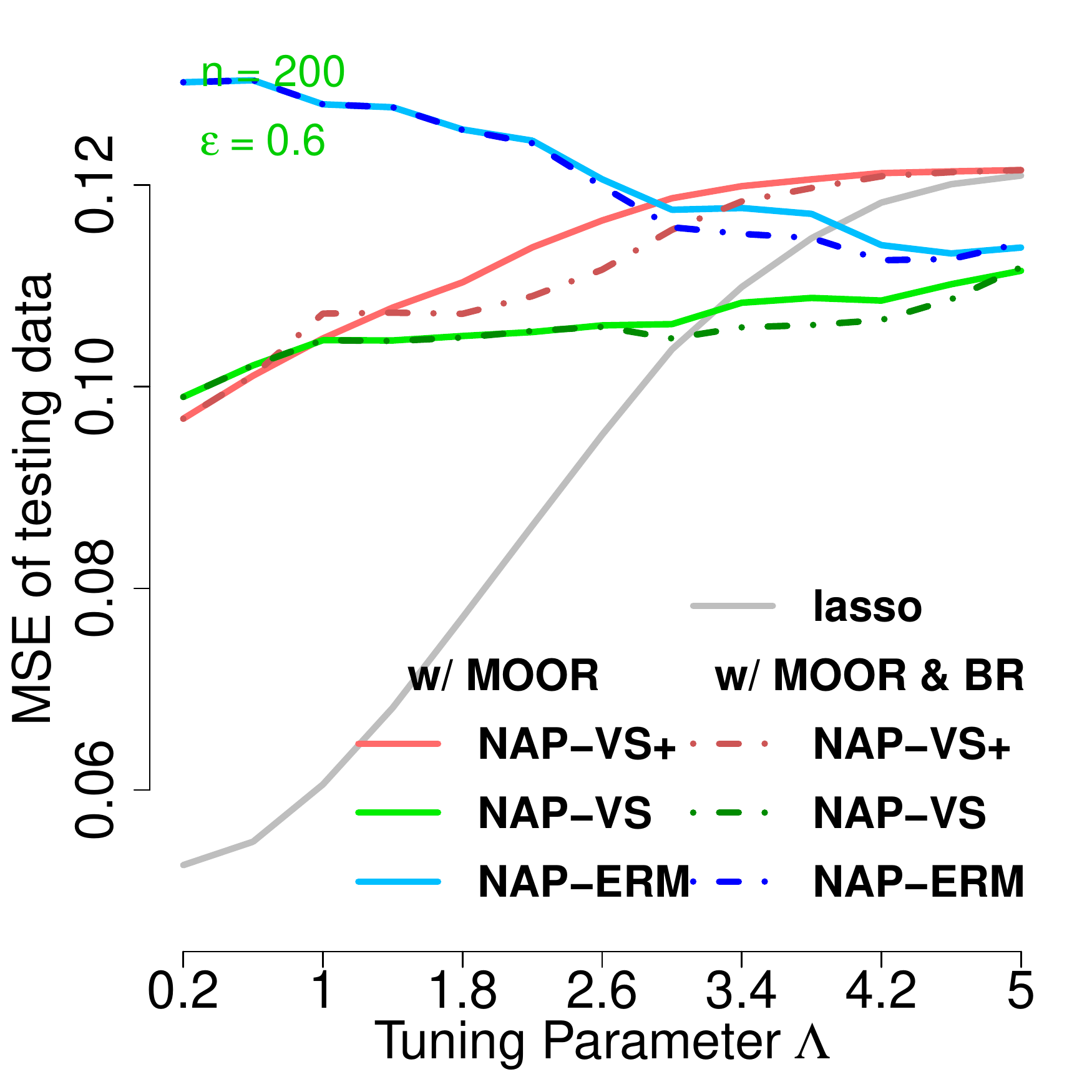}
\includegraphics[width=0.20\linewidth, trim=4pt 9pt 15pt 18pt,clip]{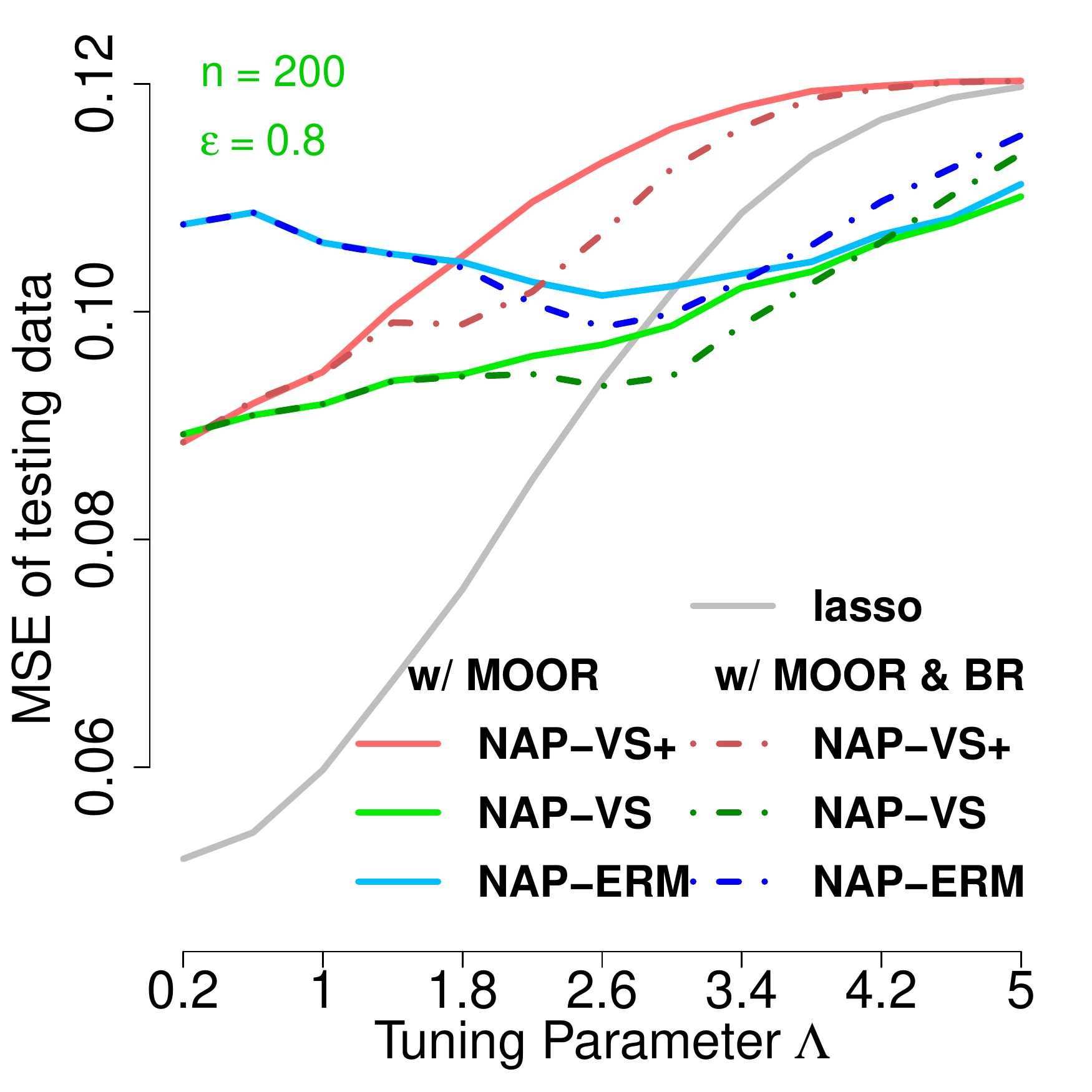}
\includegraphics[width=0.20\linewidth, trim=4pt 9pt 15pt 18pt,clip]{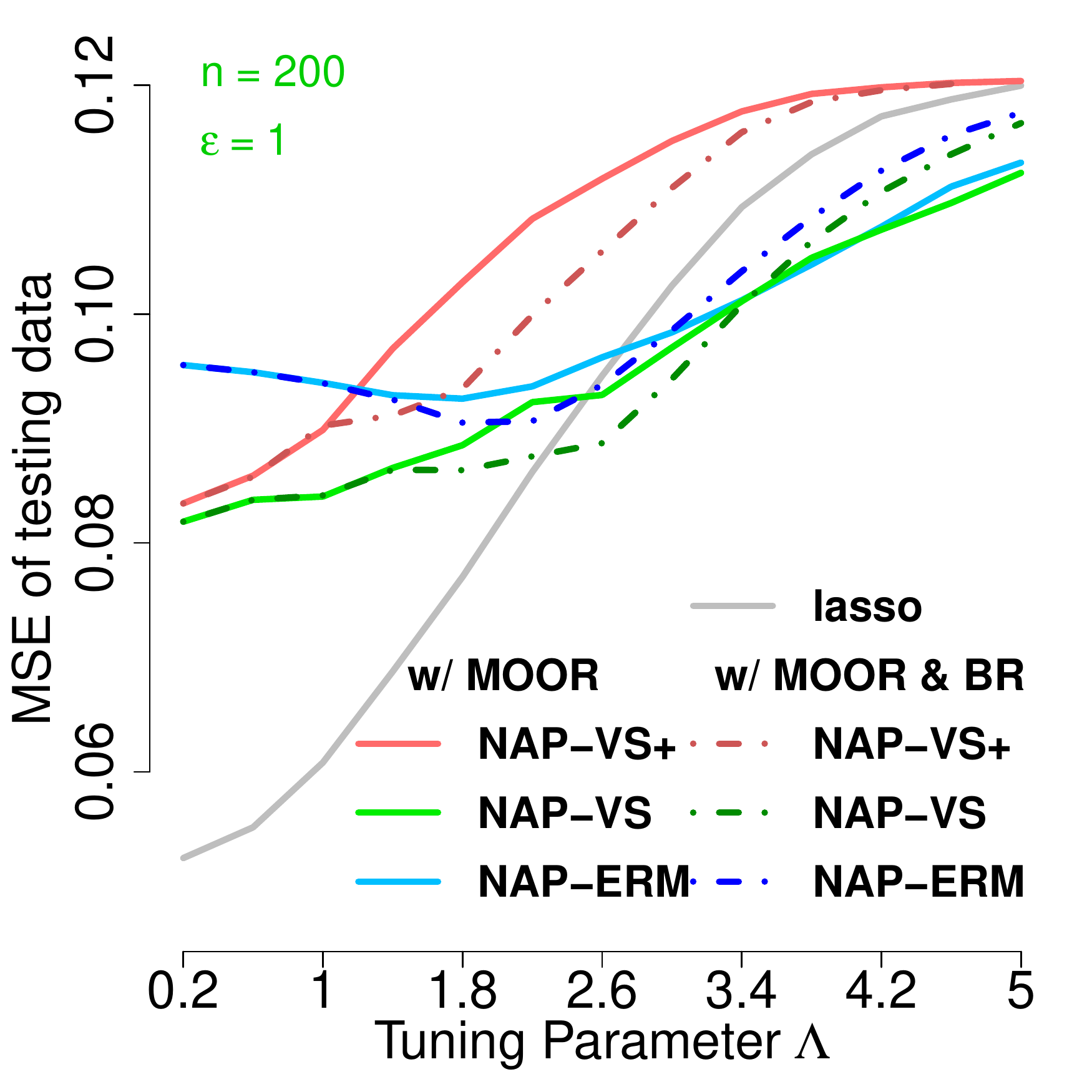}\\
\footnotesize $n=500$\\
\includegraphics[width=0.20\linewidth, trim=4pt 9pt 15pt 18pt,clip]{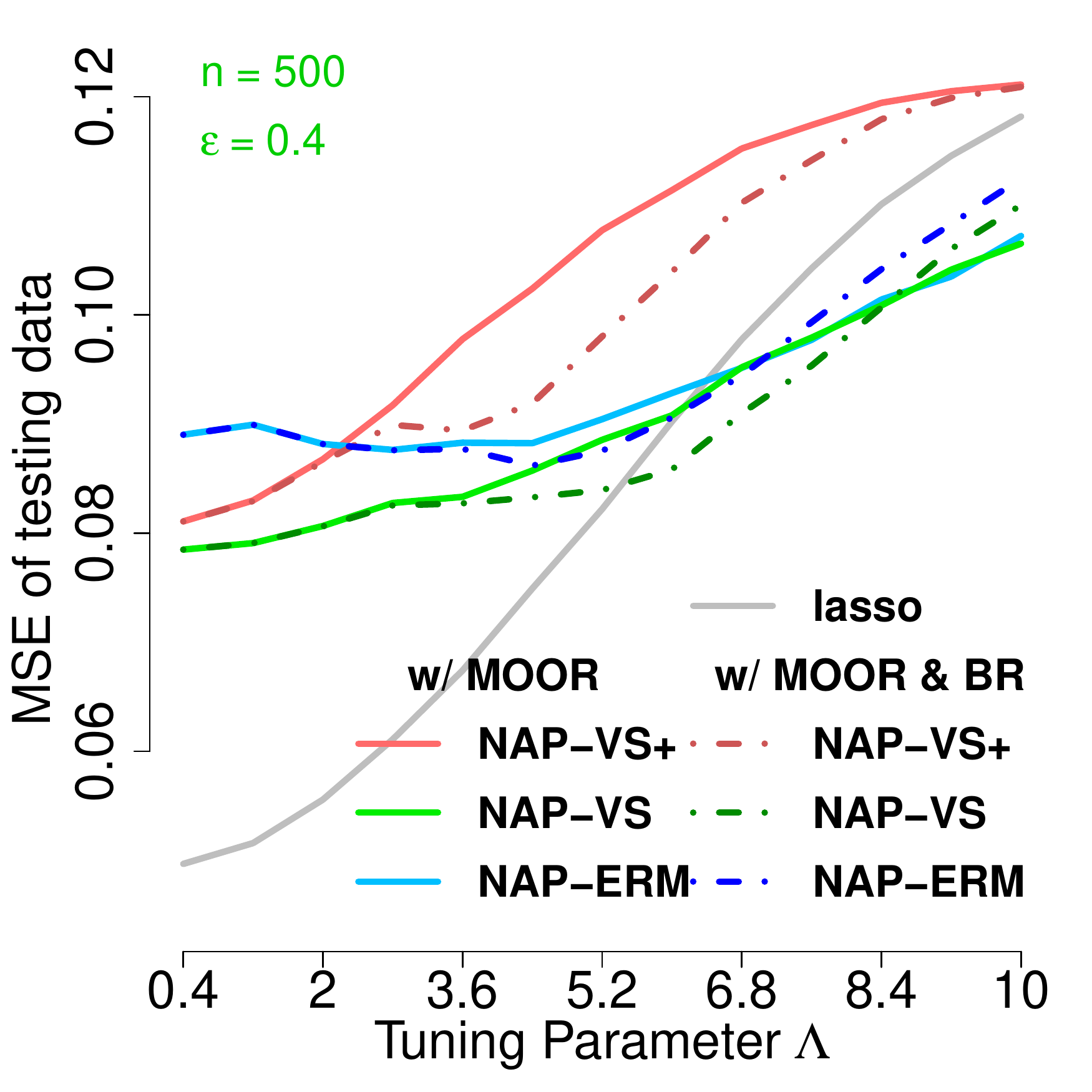}
\includegraphics[width=0.20\linewidth, trim=4pt 9pt 15pt 18pt,clip]{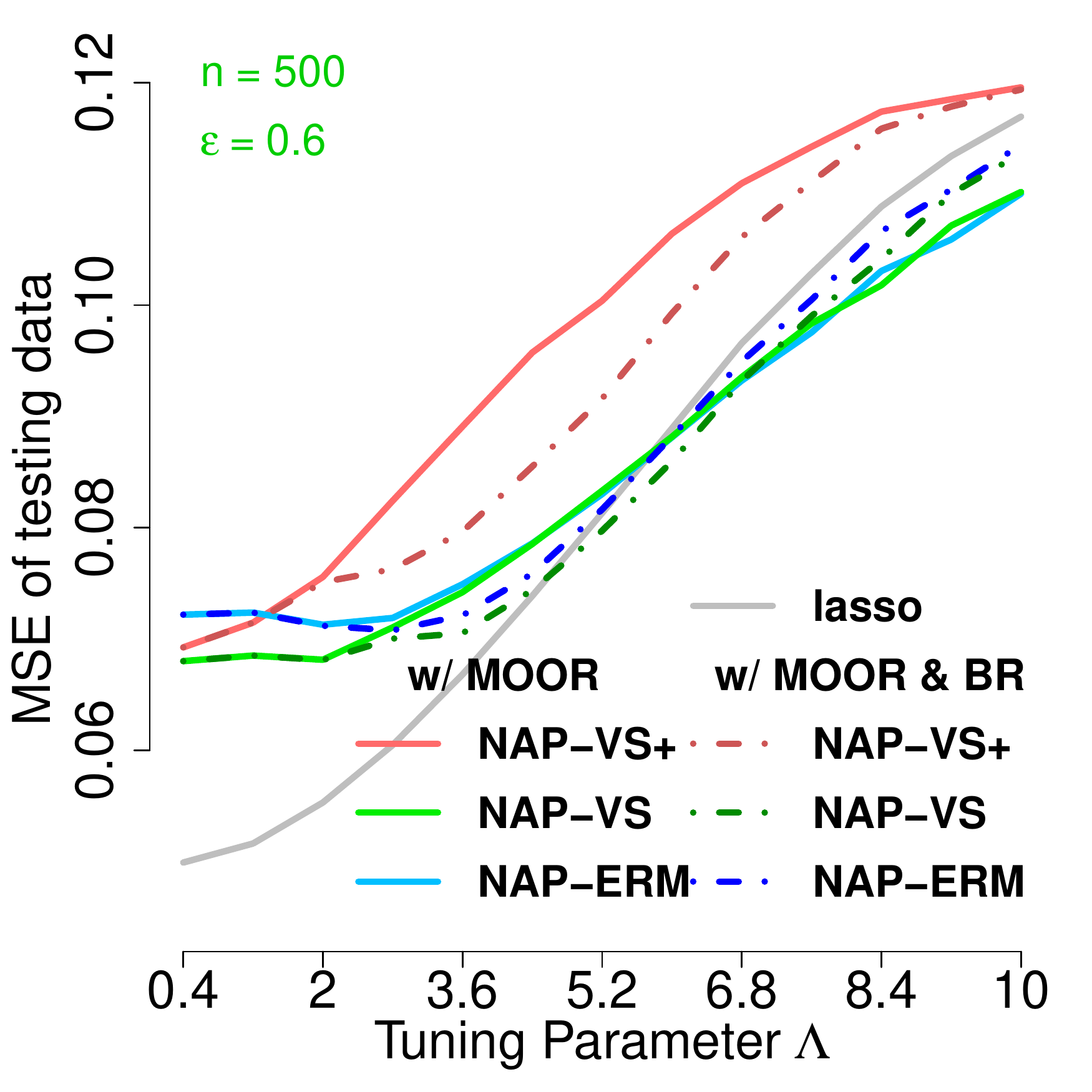}
\includegraphics[width=0.20\linewidth, trim=4pt 9pt 15pt 18pt,clip]{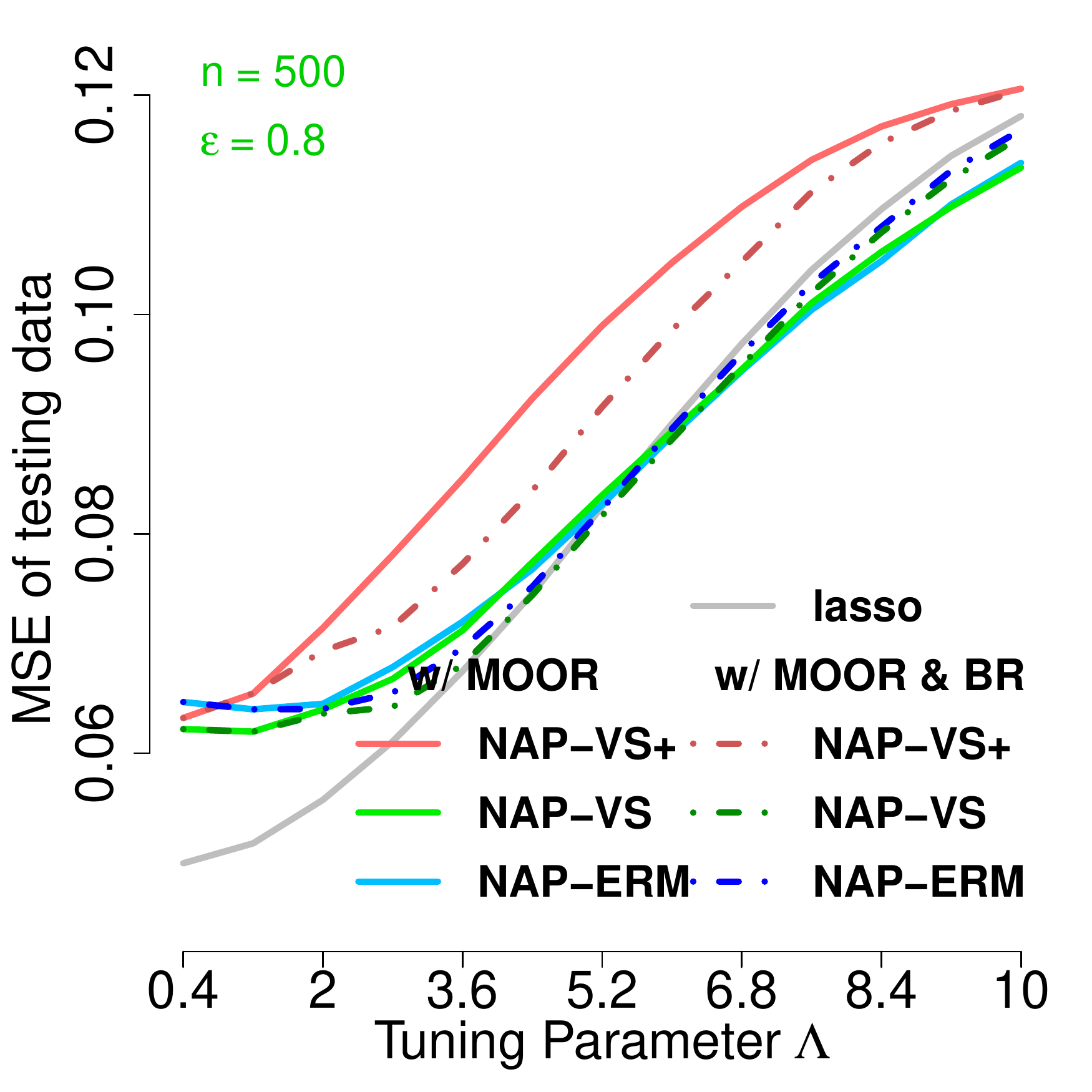}
\includegraphics[width=0.20\linewidth, trim=4pt 9pt 15pt 18pt,clip]{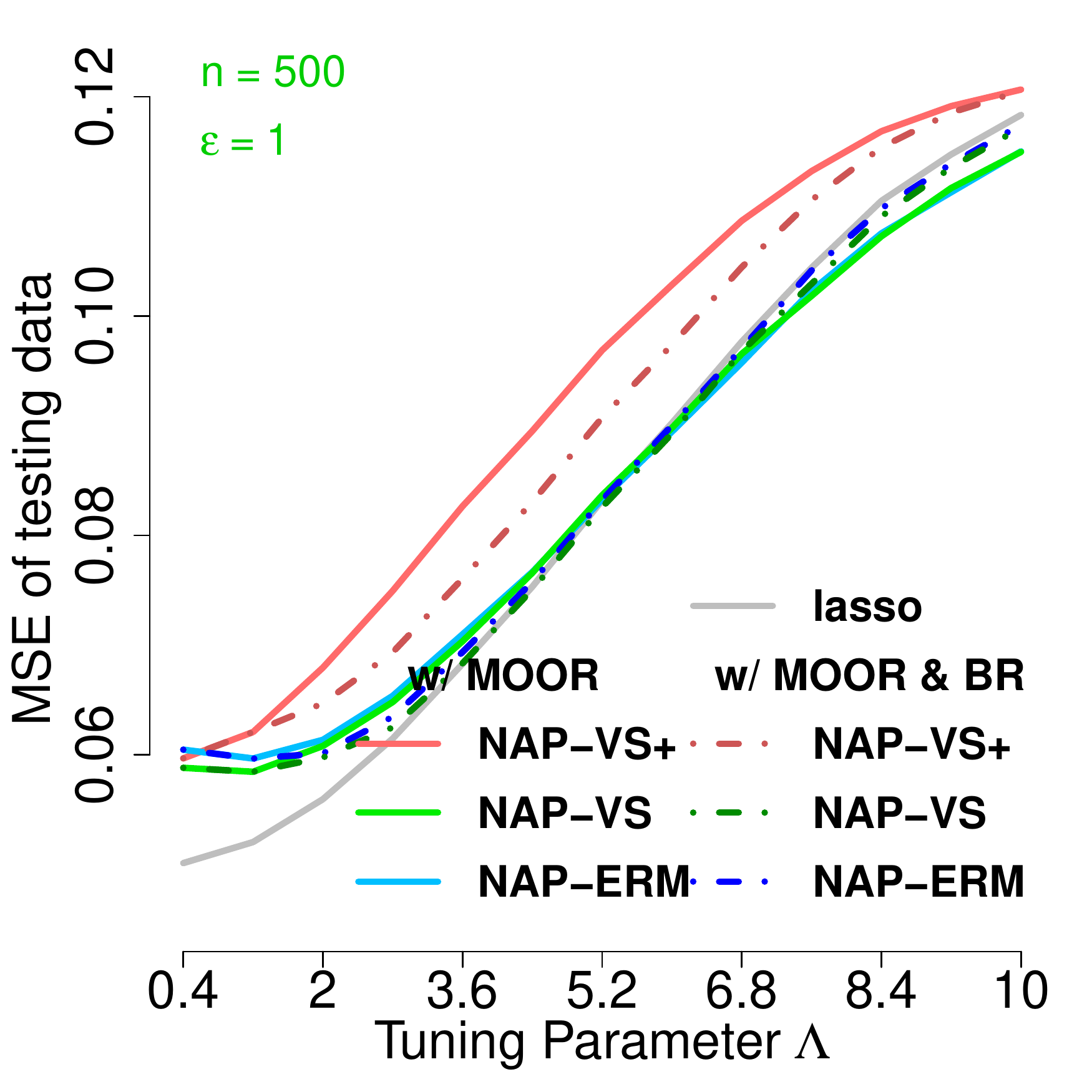}
\vspace{-9pt}
\caption{Variable selection ROC curves and outcome prediction MSE on testing data in linear regression with lasso via NAPP-ERM with vs without recycled privacy budget (BR in the legends stands for Budget Recycling)} \label{fig:recyling} \vspace{-9pt}
\end{figure}

The results are presented in Fig \ref{fig:retrieval} and summarized as follows. 1) The portion of retrievable budget increases with $\Lambda$ and eventually reaches a plateau. In some cases, the retrievable budget is $\sim100\%$, an indication that the target regularization on the loss function is sufficient to guarantee the required strong-convexity needed for bounding $r_2$.  2)  As $\epsilon$ increases, the portion of  retrievable budget  increases in a more significant manner for NAPP-VS and NAPP-ERM than for NAPP-VS+ and eventually stabilizes. 3)  More budget can be retrieved from the regression without MOOR than with MOOR, which is expected, as the over-regularization in the former can be leveraged to achieve  strong convexity of a similar degree as the latter and parameter estimation with similar sparsity as the latter but for a smaller budget allocated to $r_2$. 4) NAPP-VS+ retrieves more budget than NAPP-VS and NAPP-ERM since it targets at variable selection and can generate estimates with similar sparsity for a smaller budget than the latter two. 5) NAPP-VS is similar to NAPP-ERM with vs. without MOOR in most cases and retrieves a slightly higher portion than NAPP-ERM for small $n$ and $\epsilon$. 

The retrieved privacy budget can be recycled back to the NAPP-ERM procedure so to reduce the magnitude of injected DP noise. WLOG, we demonstrate the utility improvement in linear regression  and expect similar findings in other types of regression. We focus on NAPP with MOOR given its non-inferior performance to NAPP without MOOR in both theory and empirical studies in Sec. \ref{sec:moor} and \ref{sec:VS}. The results on the accuracy in variable selection and outcome prediction with one round of budget retrieval  vs. without  are presented in Fig \ref{fig:recyling}. The  improvement in variable selection with budget retrieval is more evident for NAPP-ERM and NAPP-VS where there is room for improvement and when $n$ or $\epsilon$ is large enough to showcase such an improvement. In the case pf NAPP-VS+, the performance is about the same with vs. without the recycled budget. As a result, NAPP-ERM and NAPP-VS with budget retrieval perform at about the same level as NAPP-VS+ without budget retrieval in variable selection. For outcome prediction, the findings are rather the opposite. The decrease in MSE with the recycled privacy budget is more evident for NAPP-VS+ especially when $n$ or $\epsilon$ is large than for NAPP-ERM and NAPP-VS; but the relative magnitudes in the MSE among NAPP-VS+, NAPP-VS and NAPP-ERM follow a similar trend to that in Fig \ref{fig:SIM.prediction}.

\vspace{-6pt}\subsection{Experiment Result Summary}\label{sec:summary}\vspace{-3pt}
The following main conclusions can be made based on the experiment results in this section. First, the MOOR effect brought by the dual-purpose weighted $l_2$ regularization in the proposed NAPP framework can be very effective in maintaining the utility of the privacy-preserving results compared to using two separate terms in the objective function to achieve the target regularization and the strong convexity requirement as employed in the existing DP-ERM framework. Second, our new formulation of the DP-ERM problem for variable selection with a new type of DP noise term that functions as a ``random'' lasso term guarantees sparsity in variable selection compared to the current DP-ERM problem formulation. Third, our proposed privacy budget retrieval scheme works as expected. If recycled back to the privacy-preserving learning procedure, the retrieved privacy budget can help to improve the utility of the private results.

\section{Discussion}\label{sec:discuss}\vspace{-3pt}
Our privacy-preserving ERM framework -- NAPP-ERM -- has two main advantages over the existing framework of privacy-preserving ERM based on the objective function perturbation. First, NAPP-ERM mitigates over-regularization issue experienced by the existing approaches, while still fulfilling the strong convexity requirement. Second, it offers a budget retrieval/recycling scheme to help either reduce the privacy loss or decrease the amount of DP noise at the same privacy loss. In addition, we also propose NAPP for variable selection with a newly designed DP term that guarantees sparsity in variable selection. From an implementation perspective, there is no need to develop ad-hoc optimization algorithms for NAPP-ERM as it can leverage existing non-private ERM solvers on the noise-augmented data.

NAPP-ERM utilizes iterative weighted $l_2$ regularization, which realizes target regularization upon convergence, to cover the strong-convexity requirement as much as possible, regardless of whether the target regularizer is convex or not. Non-convex regularization such as SCAD and $l_0$ can also be realized through noise augmentation as documented in \cite{pandaGLM}. Therefore, NAPP-ERM can accommodate a lot of more regularizers than the current existing DP-ERM framework which requires the target regularizer to be convex. 

Our experiments have focused on GLMs where the loss function  $l(\bs\theta|\D)$ is the negative log-likelihood.  NAPP-ERM also admits other types of loss functions as long as the assumptions for the loss function listed in Assumption \ref{ass} are satisfied, including the $l_2$ loss and the smoothed hinge loss ($l(t)=0$ if $t>1$; $(1-t)^2/2$ if $0<t\leq 1$; $1/2-t$ if $t\leq 0$)
employed by SVMs, where $t=y\x\bs\theta$ and $y=\pm1$ is the observed binary outcome. Since the loss is 0 if $t>1$ and is linear in $t$ if $t\le0$, we can leverage the loss at $t\in(0,1]$ to achieve both the target regularization and privacy guarantees. The noise augmentation scheme for SVMs is  defined in a similar manner as in the logistic regression, but a large $n_e$ should be used to so that $t\in(0,1]$ for a given $\Lambda$.

We currently develop an approach for privacy-preserving confidence intervals (CI) for $\bs\theta$ in the NAPP-ERM framework, aiming to achieve better utility -- tighter intervals with closer to nominal coverage -- compared to the existing methods for privacy-preserving CIs.


\bibliographystyle{plainnat}
\bibliography{NAP.bib}

\newpage

\setcounter{page}{1}
\setcounter{section}{0}
\setcounter{figure}{0}
\setcounter{table}{0}

\renewcommand\thesection{S.\arabic{section}}
\renewcommand\thefigure{S.\arabic{figure}}
\renewcommand\thetable{S.\arabic{table}}

\begin{center}
\large{Supplementary Materials to \emph{\textbf{``Noise-Augmented Privacy-Preserving Empirical Risk Minimization with Dual-purpose Regularizer and Privacy Budget Retrieval and Recycling''}}}

\normalsize{Yinan Li and Fang Liu}\\ 
\normalsize{Applied and Computational Mathematics and Statistics}\\
\normalsize{University of Notre Dame, Notre Dame, Indiana, USA}
\end{center}

\vspace{-18pt}\section{Proof of Proposition 1}\label{app:prop:regularization}\vspace{-9pt}
\begin{proof}
Take the 2nd-order Taylor expansion of $nJ_p^{(t)}(\bs\theta|\D,\e^{(t)})= \sum_{i=1}^{n}l(\bs\theta|\dd_i)+\sum_{i=1}^{n_e}l(\bs\theta|\e^{(t)}_i)$ at $\bs\theta^T\e_i=0$, and $nJ_p^{(t)}(\bs\theta|\D,\e^{(t)})$ can be approximated by 
\begin{align*}
&\! \textstyle \sum_{i=1}^{n}l(\bs\theta|\dd_i)\!+\!\sum_{i=1}^{n_e}l(\bs\eta\!=\!\0)\!+\! l'|_{\bs\eta\!=\!\0}\sum_{i=1}^{n_e}\!\left(\!\bs\theta^T\e_i^{(t)}\!\right)\!+\!\frac{l^{''}|_{\bs\eta\!=\!\0}}{2}\!\sum_{i=1}^{n_e}\!\left(\!\bs\theta^T\e_i^{(t)}\!\right)^2\\
=&\textstyle\sum_{i=1}^{n}l(\bs\theta|\dd_i)\!+C+\!l'|_{\bs\eta\!=\!\0}\!\sum_{i=1}^{n_e}\!\left(\bs\theta^T\!(\e^*+\tilde{\e}_i^{(t)})\!\right)\!+\!\frac{l^{''}|_{\bs\eta\!=\!\0}}{2}\!\sum_{i=1}^{n_e}\!\left(\bs\theta^T\!(\e^*\!+\!\tilde{\e}_i^{(t)})\right)^2\\
=&\!\sum_{i=1}^{n}\!l(\bs\theta|\dd_i)\!+\!C\!+\!\sum_{j=1}^p\!b_j\theta_j\!+\!0\!+\!\frac{n_el^{''}|_{\bs\eta\!=\!\0}}{2}\!\sum_{i=1}^{n_e}\!\sum_{j=1}^p\!\theta^2_j(e_{ij}^*+\tilde{e}_{ij}^{(t)})^2\!+\!l^{''}|_{\bs\eta\!=\!\0}\!\sum_{i=1}^{n_e}\!\sum_{j\neq k}\!\theta_j\theta_k(e_{ij}^*\!+\!\tilde{e}_{ij}^{(t)})(e_{ik}^*\!+\!\tilde{e}_{ik}^{(t)})\\
=&\textstyle\sum_{i=1}^{n}l(\bs\theta|\dd_i)\!+\!\sum_{j=1}^{p}b_j\theta_j\!+\!\frac{n_el^{''}_{\bs\eta\!=\!\0}}{2}\!\sum_{i=1}^{n_e}\sum_{j=1}^p\theta^2_j\tilde{e}_{ij}^{(t)2}\!+\!O(n_e^{-1})\!+\!0\!+\!O(n_e^{-1/2})\\
\rightarrow&\textstyle\sum_{i=1}^{n}l(\bs\theta|\dd_i)+\sum_{j=1}^{p}b_j\theta_j \!+\frac{n_el^{''}_{\bs\eta\!=\!\0}}{2}\sum_{j=1}^{p} \mbox{V}\left(\tilde{e}_{ij}^{(t)}\right)\theta_j^2 \mbox{ as } n_e\rightarrow\infty. 
\end{align*}
\vspace{-18pt}
\end{proof}

\vspace{-18pt}\section{Proof of Theorem 4}\label{app:prop:privacyguarantee}\vspace{-9pt}
\begin{proof} 
To prove DP in iteration $t$, we need to bound the ratio $\frac{\Pr\left(\hat{\bs\theta}^{(t)}\in\mathcal{Q}|\D\right)}{\Pr\left(\hat{\bs\theta}^{(t)}\in\mathcal{Q}|\D'\right)}$ for any subset $\mathcal{Q}$ of possible values of $\hat{\bs\theta}^{(t)}$ and for all $|\D-\D'|=1$ so that
\begin{align}
\epsilon\mbox{-DP: }& \Pr\left(\hat{\bs\theta}^{(t)}\in\mathcal{Q}|\D\right)\leq e^{\epsilon}\Pr\left(\hat{\bs\theta}^{(t)}\in\mathcal{Q}|\D'\right)
\label{eqn:e-privacy}\\
(\epsilon, \delta)\mbox{-DP: }& \Pr\left(\hat{\bs\theta}^{(t)}\in\mathcal{Q}|\D\right)\leq e^{\epsilon}\Pr\left(\hat{\bs\theta}^{(t)}\in\mathcal{Q}|\D'\right)+\delta. \label{eqn:e-d-privacy}
\end{align} 
To solve for $\hat{\bs\theta}^{(t)}$, we set the 1st-order derivative of the objective function with respect to $\bs{\theta}$ at 0, 
\begin{align*}
\textstyle \nabla l(\hat{\bs\theta}^{(t)}|\D)+n_e l''|_{\bs\eta=\0}\sum_{j=1}^{p} \mbox{V}\left(\tilde{e}_{ij}^{(t)}\right)\hat{\theta}^{(t)}_j+\bb\!=\!0 &\mbox{ for general NAP-ERM}\\
\textstyle \nabla l(\hat{\bs\theta}^{(t)}|\D)\!+\!n_e l''|_{\bs\eta=\0}\!\sum_{j=1}^{p}\! \mbox{V}\!\left(\tilde{e}_{ij,1}^{(t)}\right)\!\hat{\theta}^{(t)}_j\!+\!\mbox{sign}\!\left(\hat{\bs\theta}^{(t-1)}\right)\bb\!=\!0 & \mbox{ for NAP-ERM-VS}, 
\end{align*}
which can be re-arranged to obtain
\begin{align}
\bb&=\textstyle -\left( \nabla l(\hat{\bs\theta}^{(t)}|\D)+n_e l''|_{\bs\eta=\0}\sum_{j=1}^{p} \mbox{V}\left(\tilde{e}_{ij}^{(t)}\right)\hat{\theta}^{(t)}_j \right)\mbox{ for general NAP-ERM} \label{eqn:one2one}\\
\bb&=\textstyle -\left( \nabla l(\hat{\bs\theta}^{(t)}|\D)+n_e l''|_{\bs\eta=\0}\sum_{j=1}^{p} \mbox{V}\left(\tilde{e}_{ij}^{(t)}\right)\hat{\theta}^{(t)}_j \right)\mbox{sign}\left(\hat{\bs\theta}^{(t-1)}\right) \mbox{ for NAP-ERM-VS.}\label{eqn:one2onevs}
\end{align}
Since $l(\bs\theta)$ is convex in $\theta$, the relationship between $\bb$ and $\hat{\bs\theta}^{(t)}$ is bijection in Eqs (\ref{eqn:one2one},\ref{eqn:one2onevs}). Given the distribution of $\bb$ (spherical Laplace, Gaussian, or their truncated versions), we can work out the distribution function for  $\hat{\bs\theta}^{(t)}$  using the change of variable technique given the bijection relationship between  $\hat{\bs\theta}^{(t)}$  and $\bb$. The Jacobian of $\bb$ with regard to $\hat{\bs\theta}^{(t)}$ in Eqs (\ref{eqn:one2one}) and (\ref{eqn:one2onevs}) is
\begin{equation}\label{eqn:jacob}
J_{\bb}(\hat{\bs\theta}^{(t)}|\D) =- \nabla^2 l(\hat{\bs\theta}^{(t)}|\D)-n_e l''|_{\bs\eta=\0}\diag  \left(\mbox{V}\left(e_{i1,1}^{(t)}\right),\ldots,\mbox{V}\left(e_{ip,1}^{(t)}\right) \right).
\end{equation}
Denote by $f(\hat{\bs\theta}^{(t)})$ the probability density function of $\hat{\bs{\theta}}^{(t)}$ at iteration $t$. Then  by the change of variable technique, we have
$$\frac{f\left(\hat{\bs\theta}^{(t)}|\D\right)}{f\left(\hat{\bs\theta}^{(t)}|\D'\right)}=\frac{f_{\bb}(\bb^{-1}(\bs\theta^{(t)}|\D))} {f_{\bb}(\bb^{-1}(\bs\theta^{(t)}|\D'))}\times\frac{|\det(J_{\bb}(\hat{\bs\theta}^{(t)}|\D'))|}{|\det(J_{\bb}(\hat{\bs\theta}^{(t)}|\D))|} =r_1r_2,
$$
which is Eqn (12). The rest of the proof is divided into two parts. The first part establishes $\epsilon$-DP, and the second part establishes $(\epsilon,\delta)$-DP.

\underline{Part I}: In the case of $\epsilon$-DP, Eq (\ref{eqn:e-privacy}) $\Pr\left(\hat{\bs\theta}^{(t)}\in\mathcal{Q}|\D\right)\leq e^{\epsilon}\Pr\left(\hat{\bs\theta}^{(t)}\in\mathcal{Q}|\D'\right)$ is the same as $f\left(\hat{\bs\theta}^{(t)}|\D\right)\leq e^{\epsilon}f\left(\hat{\bs\theta}^{(t)}|\D'\right)$.
We now obtain the bounds for the first ratio $r_1$ and the second ratio $r_2$ in Eq (12) separately to show their product is bounded by $e^{\epsilon}$.  Denote by $r$ the portion of $\epsilon$ allocated to ratio $r_1$. Plugging  Eq (4) in the first ratio $r_1$ of Eq (12), we have
\begin{align}
r_1&=\frac{f_\bb(\bb^{-1}(\hat{\bs\theta}^{(t)}|\D))}
{f_\bb(\bb^{-1}(\hat{\bs\theta}^{(t)}|\D')}=
\exp\left(\frac{r\epsilon}{\zeta_1\zeta_2}\left(||\bb^{-1}(\hat{\bs\theta}^{(t)}|\D)||_2-||\bb^{-1}(\hat{\bs\theta}^{(t)}|\D')||_2\right)\right)\notag\\
&\le
\exp\left(\frac{r\epsilon}{\zeta_1\zeta_2}||\bb^{-1}(\hat{\bs\theta}^{(t)}|\D)-\bb^{-1}(\hat{\bs\theta}^{(t)}|\D')||_2\right)\triangleq \exp\left(\frac{r\epsilon}{\zeta_1\zeta_2}||\Gamma||_2\right)\notag\\
&\le e^{r\epsilon}, \mbox{ as 
$||\Gamma||_2=||\nabla l(\bs\theta^{(t)}|\D)- \nabla l(\bs\theta^{(t)}|\D')||_2 
\leq\zeta_1\zeta_2$ per Eq (\ref{eqn:one2one})}.\label{eqn:ratio2}
\end{align}  

Bounding the second ratio $r_2$ leverages the following lemma.
\begin{lem}\label{lem:rank}\citep{dperm}
If $A$ is a full-rank matrix and $E$ is a matrix with rank at most 2, denoting $\lambda_i(*)$ as the $i^{th}$ largest eigenvalue of matrix $*$, then 
\begin{equation}
\frac{|\det(A+E)|-|\det(A)|}{|\det(A)|}= \lambda_1(A^{-1}E)+\lambda_2(A^{-1}E)+\lambda_1(A^{-1}E)\lambda_2(A^{-1}E).
\end{equation}
\end{lem}
Applying the lemma in NAP-ERM, we let $A=\nabla^2 l(\hat{\bs\theta}^{(t)}|\D)+n_e l''|_{\bs\eta=\0}\diag  \left(\mbox{V}\left(\tilde{e}_{i1}^{(t)}\right),\ldots,\mbox{V}\left(\tilde{e}_{ip}^{(t)}\right) \right)$ and $E=\nabla^2l(\hat{\bs\theta}^{(t)}|\dd_{n+1})$. WLOG, assume that $\D'$ has one more observation than $\D$, then $|\mbox{det}(A+E)|\!=\!|\det(J_{\bb}(\hat{\bs\theta}^{(t)}|\D'))|$. The largest eigenvalue of $A^{-1}$ is the inverse of the smallest eigenvalue of $A$, that is, $\lambda_1(A^{-1}) =(\lambda_p(A))^{-1}$. Since $\nabla^2 l(\hat{\bs\theta}^{(t)}|\D)$ is positive semi-definite,  then $\lambda_p(A)>n_e l''|_{\bs\eta=\0}\mbox{V}^{(t)}_{(1)}$, where  $\mbox{V}^{(t)}_{(1)}=\min\limits_{j=1,\ldots, p}\mbox{V}\left(\tilde{e}_{ij}^{(t)}\right)$. In addition, per the design of the variance term of $\tilde{e}$ (Table 1), $\V\left(\tilde{e}_{ij}^{(t)}\right)\ge 2(n_el''|_{\bs\eta=\0})^{-1}\Lambda_0$. All taken together,
\begin{align}\label{eqn:maxeigen}
\lambda_1(A^{-1}) =(\lambda_p(A))^{-1} \leq\left(n_e l''|_{\bs\eta=\0}\mbox{V}^{(t)}_{(1)}\right)^{-1}\leq (2\Lambda_0)^{-1}, \end{align}
By the fact that the largest eigenvalue of a product of two matrices is less than or equal to the product of the largest eigenvalues of the two matrices, we have
$\lambda_1(A^{-1}E) \leq\frac{\lambda_1(E)}{2\Lambda_0}.$
In addition,  $|\lambda_1(E)|\leq|l''(\hat{\bs\theta})|\cdot||\x_{n+1}||_2^2\leq\zeta_3$ per  Assumption 1 
and $\Lambda_0\geq\zeta_3/(2(1-r)\epsilon)$. All taken together with Lemma \ref{lem:rank}, we have
\begin{align}\label{eqn:bound1}
r_2&=\frac{|\mbox{det}(J_{\bb}(\hat{\bs\theta}^{(t)}|\D'))|}{|\mbox{det}(J_{\bb}(\hat{\bs\theta}^{(t)}|\D))|} =\frac{\det(A+E)}{\det(A)}= 1+ \lambda_1(A^{-1}E)+\lambda_2(A^{-1}E)+\lambda_1(A^{-1}E)\lambda_2(A^{-1}E)\notag\\ 
&\leq 1+ \lambda_1(E)/(2\Lambda_0)
\leq 1+2(1-r)\epsilon/2\leq e^{(1-r)\epsilon}.
\end{align}
Eqs  (\ref{eqn:bound1}) and (\ref{eqn:ratio2}) 
taken together, we obtain the $\epsilon-$DP.

\underline{Part II}: In the case of $(\epsilon, \delta)-$DP,
\begin{align}
r_1=&\frac{f_\bb(\bb^{-1}(\hat{\bs\theta}^{(t)}|\D))}
{f_\bb(\bb^{-1}(\hat{\bs\theta}^{(t)}|\D')}=\exp\left( (2\sigma^2)^{-1}\left(||\bb^{-1}(\hat{\bs\theta}^{(t)}|\D)||_2^2-||\bb^{-1}(\hat{\bs\theta}^{(t)}|\D')||_2^2  \right)  \right)\notag\\
=&\exp\left( (2\sigma^2)^{-1}\left(||\bb^{-1}(\hat{\bs\theta}^{(t)}|\D)||_2^2-||\bb^{-1}(\hat{\bs\theta}^{(t)}|\D)-\Gamma||_2^2  \right)  \right)\notag\\
=&\exp\left( (2\sigma^2)^{-1}\left( 2\langle\bb^{-1}(\hat{\bs\theta}^{(t)}|\D),\Gamma \rangle-||\Gamma||_2^2 \right)  \right)
\leq\exp\left( (2\sigma^2)^{-1}\left( 2\left|\langle\bb^{-1}(\hat{\bs\theta}^{(t)}|\D),\Gamma \rangle\right|+||\Gamma||_2^2 \right)  \right)\notag\\
\leq&\exp\left( (2\sigma^2)^{-1}\left(2 \left|\langle\bb^{-1}(\hat{\bs\theta}^{(t)}|\D),\Gamma \rangle\right|+\zeta_1^2\zeta_2^2 \right)  \right).\label{eqn:bound2}
\end{align}
Since $\langle\bb^{-1}(\hat{\bs\theta}^{(t)}|\D),\Gamma \rangle \sim N(0,||\Gamma||_2^2\sigma^2)$, it can be bound  probabilistically using the concentration inequality (i.e., if $Z\sim N(0,1)$, then  $\Pr(|Z|>t)\leq \exp(-t^2/2)$ for $\forall t>0$).  As $||\Gamma||_2\leq\zeta_1\zeta_2$, then $\Pr\left(| \langle\bb^{-1}(\hat{\bs\theta}^{(t)}|\D),\Gamma \rangle|\geq \sigma\zeta_1\zeta_2 t \right)\leq \exp(-t^2/2)$ per the concentration inequality. Define   $\mathcal{S}=\left\{ \bb\in R^p: \left| \langle\bb,\Gamma \rangle \right|\leq\zeta_1\zeta_2\sigma t  \right\}$  as the  set of $\bb$ that leads to $\epsilon$-DP. We require $\Pr(\bb\in\mathcal{S})$ to be at least $1-\delta$. Then
\begin{align*}
&\Pr\left(\hat{\bs\theta}^{(t)}\in\mathcal{Q}|\D \right) \Pr\left(\bb\in\mathcal{S}\right) \Pr\left(\hat{\bs\theta}^{(t)}\in\mathcal{Q}|\bb\in\mathcal{S},\D\right)+
\Pr\left(\bb\in\overline{\mathcal{S}} \right)
\Pr\left(\hat{\bs\theta}^{(t)}\in\mathcal{Q}|\bb\in\overline{\mathcal{S}},\D\right)\\
\leq&\Pr\!\left(\!\hat{\bs\theta}^{(t)}\!\!\in\!\mathcal{Q}|\bb\!\in\!\mathcal{S},\D\!\right)\!+\!\delta, \mbox{ as both} \Pr\!\left(\bb\!\in\!\mathcal{S}\right)\mbox{ and} \Pr\!\left(\!\hat{\bs\theta}^{(t)}\!\!\in\!\mathcal{Q}|\bb\!\in\!\overline{\mathcal{S}},\D\!\right)\!\in\![0,1] \\
\leq& e^{\epsilon}\Pr\left(\hat{\bs\theta}^{(t)}\in\mathcal{Q}|\bb\in\mathcal{S},\D'\right)+\delta \mbox{ per the established $\epsilon$-DP guarantee in Part I if $\bb\in\mathcal{S}$},
\end{align*}
proving $(\epsilon,\delta)$-DP for Gaussian $\bb$.
\end{proof}

\vspace{-15pt}\section{Proof of Corollary 5}\label{app:boundsigma2}\vspace{-6pt}
\begin{proof}
Building upon the proof of Theorem 4, we let $\exp(-t^2/2)=\delta$  (thus $t\!=\!\sqrt{-2\log(\delta)}$)  in the concentration inequality  $\Pr(|\langle\bb(\hat{\bs\theta}^{(t)}|\D),\Gamma \rangle|>t)\leq \exp(-t^2/2)$; therefore \\ $\Pr\!\left(\!\langle\bb(\hat{\bs\theta}^{(t)}|\D),\Gamma \rangle\!\geq\right.\\ \left.\!\zeta_1\zeta_2\sigma\sqrt{-2\log(\delta)} \right)\!\leq\!\delta$. 
Let $|\langle\bb(\hat{\bs\theta}^{(t)}|\D),\Gamma\rangle|= \zeta_1\zeta_2\sigma\sqrt{-2\log(\delta)}$ and plug it in Eq (\ref{eqn:bound2}), we have 
\begin{equation}\label{eqn:bound21}
\frac{f_{\bb}(\bb^{-1}(\hat{\bs\theta}^{(t)}|\D))}{f_{\bb}(\bb^{-1}(\hat{\bs\theta}^{(t)}|\D'))}\leq \exp\left( (2\sigma^2)^{-1}\left(2\zeta_1\zeta_2\sigma \sqrt{-2\log(\delta)}+\zeta_1^2\zeta_2^2 \right)  \right) \leq \exp(\epsilon/2). 
\end{equation}
Solving for $\sigma^2$ from the quadratic inequality above leads to Eqn (13).
\end{proof}

\vspace{-15pt}\section{Proof of Proposition 6}\label{app:retrieve}\vspace{-9pt}
\begin{proof}
Per  Eq (\ref{eqn:bound1}), 
$r_2\leq 1+ \lambda_1(E)/(2\Lambda_0)$. Per Eq (\ref{eqn:maxeigen}), $(2\Lambda_0)^{-1}\ge\big(l''|_{\bs\eta=\0}\mbox{V}^{(t)}_{(1)} \big)^{-1}$ and $\lambda_1(E)\le\zeta_3$; therefore, $r_2\leq 1+\zeta_3\big(n_e l''|_{\bs\eta=\0}\mbox{V}^{(t)}_{(1)} \big)^{-1}\le 1+2(1-r)\epsilon\Lambda_0\big( 2n_e l''|_{\bs\eta=\0}\mbox{V}^{(t)}_{(1)} \big)^{-1}$, since $\Lambda_0\geq\frac{\zeta_3}{2(1-r)\epsilon}$, $\leq \exp\big)((1-r)\epsilon\Lambda_0\big)( n_e l''|_{\bs\eta=\0}\mbox{V}_{(1)}^{(t)} \big))^{-1}\big))$. 
In other words, the actual budget used on $r_2$ upon convergence is $(1-r)\epsilon\Lambda_0\big(n_e l''|_{\bs\eta=\0}\mbox{V}_{(1)}^{(T)} \big)^{-1}$; When it is less than $(1-r)\epsilon$, there is unspent budget from bounding $r_1$, the amount of which is
$\Delta_\epsilon^{(t)}=(1-r)\epsilon\big(1-\Lambda_0\big)( n_e l''|_{\bs\eta=\0}\mbox{V}_{(1)}^{(t)}\big)^{-1} \big)$.
\end{proof}

\vspace{-15pt}\section{Proof of Lemma 7}\label{app:lossdiff}\vspace{-3pt}
\begin{proof}
$J(\!\hat{\bs\theta}^{(t)*}|\D)\!-\!J(\hat{\bs\theta}|\D)=
J(\!\hat{\bs\theta}^{(t)*}|\D)\!-J(\!\hat{\bs\theta}^{(t)}|\D)+ J(\!\hat{\bs\theta}^{(t)}|\D)\!-J(\hat{\bs\theta}|\D)=$
\begin{align}
\textstyle n^{-1}\!\left[\left(\sum_{i=1}^nl(\hat{\bs\theta}^{(t)*}|\dd_i)\!+\!R(\hat{\bs\theta}^{(t)*})\!\right)\!-\!
\left(\sum_{i=1}^nl(\hat{\bs\theta}^{(t)}|\dd_i)\!+\!R(\hat{\bs\theta}^{(t)})\!\right)\!+\notag\right.\\
\textstyle\left.\left(\sum_{i=1}^nl(\hat{\bs\theta}^{(t)}|\dd_i)\!+\!R(\hat{\bs\theta}^{(t)})\!\right)\!-\!\left(\sum_{i=1}^nl(\hat{\bs\theta}|\dd_i)\!+\!R(\hat{\bs\theta})\!\right)\right].\label{eqn:lossD}
\end{align}
Since $n^{-1}\sum_{i=1}^{n}\!l(\hat{\bs\theta}^{(t)*}|\dd_i)=J^{(t)}_p(\hat{\bs\theta}^{(t)*}|\D)-n^{-1}\!R^{(t)}(\hat{\bs\theta}^{(t)*})$, similarly for $n^{-1}\sum_{i=1}^{n}\!l(\hat{\bs\theta}^{(t)}|\dd_i)$ and \\ $n^{-1}\sum_{i=1}^{n}\!l(\hat{\bs\theta}|\dd_i)$, then Eq (\ref{eqn:lossD}) becomes
\begin{align*}
&\textstyle\small(J_p^{(t)}(\hat{\bs\theta}^{(t)*}|\D)
\!-\!J_p^{(t)}(\hat{\bs\theta}^{(t)}|\D)\!+\!(J_p^{(t)}(\hat{\bs\theta}^{(t)}|\D)\!-\!J_p^{(t)}(\hat{\bs\theta}|\D))\!+\! n^{-1}\left(\!R^{(t)}(\hat{\bs\theta})\!-\!R(\hat{\bs\theta})\! \right)\!-\!n^{-1}\left(\! R^{(t)}(\hat{\bs\theta}^{(t)*})\!-\!R(\hat{\bs\theta}^{(t)*}) \!\right)\\
& \leq\textstyle ||\bb||_2^2 \left(nn_e l''|_{\bs\eta=\0}\mbox{V}^{(t)}_{(1)} \right)^{-1}\!\!+n^{-1}\left( R^{(t)}(\hat{\bs\theta})-R(\hat{\bs\theta}) \right), 
\end{align*}
where the inequality is obtained by applying  $J_p^{(t)}(\hat{\bs\theta}^{(t)*}|\D)-J_p^{(t)}(\hat{\bs\theta}^{(t)}|\D)\leq ||\bb||_2^2 \left(nn_e l''|_{\bs\eta=\0}\mbox{V}^{(t)}_{(1)} \right)^{-1}$ (Lemma 24 and Corollary 25 in \cite{dpermhd}) and noticing that $J_p^{(t)}(\hat{\bs\theta}^{(t)}|\D)-J_p^{(t)}(\hat{\bs\theta}|\D)<0$ and $R^{(t)}(\hat{\bs\theta}_j^{(t)*})-R(\hat{\bs\theta}^{(t)*}) >0$.
\end{proof}

\vspace{-12pt}\section{Proof of Theorem 8}\label{app:excessriskbound}\vspace{-6pt}
\begin{proof}
Lemma 17 in \cite{dperm}  states that if $X\sim \mbox{gamma}(\alpha,\beta)$ with the shape $\alpha$ and rate $\beta$, then $\Pr\left(X< (\alpha\beta\log(\alpha\pi^{-1})\right)\ge 1-\pi$ for $\pi\in[0,1]$. In the case of $\epsilon$-DP,  $||\bb||_2\sim \Gamma(1,\zeta_1\zeta_2(r\epsilon)^{-1})$ in Eq (19), thus with probability at least $1-\pi$, $||\bb||_2\leq p\zeta_1\zeta_2(r\epsilon)^{-1}\log(p\pi^{-1})$. Lemma 2 in \cite{boundb2007} states that if $X\sim N(\0,I_p)$, then for any $\xi>1$, $\Pr\left(||X||_2\geq\sqrt{\xi p}\right)\leq\exp\left(-p(\xi-1-\log(\xi))/2\right)$.  In the case of $(\epsilon,\delta)$-DP, following the same logic as  Lemma 28 in \cite{dpermhd},   $||\bb||_2\leq \\ 2(r\epsilon)^{-1}\zeta_1\zeta_2\left(p \left(r\epsilon+\log(2/\delta) \right)\log(\pi^{-1}) \right)^{1/2}$ with probability $\ge1-\pi$. 
Replacing $||\bb||_2$ in Eq (19) with the bounds in the case of $\epsilon$-DP and $(\epsilon,\delta)$-DP, respectively, we obtain
\begin{align}
J(\hat{\bs\theta}^{(t)*}|\D)-J(\hat{\bs\theta}|\D)&\leq 
B_1(\hat{\bs\theta},n,p,\Lambda_0,\zeta_1,\zeta_2,\epsilon,\pi) \mbox{ for $\epsilon$-DP}\label{eqn:lossdiff1}\\
J(\hat{\bs\theta}^{(t)*}|\D)-J(\hat{\bs\theta}|\D)&\leq B_2(\hat{\bs\theta},n,p,\Lambda_0,\zeta_1,\zeta_2,\epsilon,\delta,\pi)\mbox{ for $(\epsilon,\delta)$-DP} \label{eqn:lossdiff2}
\end{align}
with probability $\ge\!1\!-\!\pi$. Taking the expectations of Eqs (\ref{eqn:lossdiff1}) and (\ref{eqn:lossdiff2})  does not alter the bounds since the latter holds with probability $1\!-\!\pi$ for any $\bb$, leading to Eqs (20), (21). 
\end{proof}

\vspace{-18pt}\section{Proofs of Lemma 9 and Theorem 10}\label{app:sc}\vspace{-3pt}
\begin{proof}
Let $\dd_i$ be an i.i.d. sample for $i=1,\ldots,n$. \vspace{-6pt}
\begin{align}
\textstyle L(\hat{\bs\theta}^{\!(t)*})=& L( \bs\theta^{0})+ \bar{J}_p^{(t)}(\hat{\bs\theta}^{\!(t)*})-\bar{J}_p^{(t)}(\bar{\bs\theta}^{(t)})+ \bar{J}_p^{(t)}(\bar{\bs\theta}^{(t)} )-\bar{J}_p^{(t)}({\bs\theta}^0) +n^{-1}(R^{(t)}(\bs\theta^0)-R^{(t)}(\hat{\bs\theta}^{(t)*}))\notag\\
\leq &\textstyle L(\bs\theta^{0})+ \bar{J}_p^{(t)}(\hat{\bs\theta}^{(t)*})-\bar{J}_p^{(t)}(\bar{\bs\theta}^{(t)})  +n^{-1} R^{(t)}(\bs\theta^0), \mbox{ dropping the negative terms}.\label{eqn:SSineq}
\end{align}
By Eq (25) of Lemma 23 in \cite{dperm}, with probability at least $1-\pi'$,\vspace{-6pt}
\begin{align}
&\bar{J}_p^{(t)}\!\left(\hat{\bs\theta}^{(t)*} \right)\!-\!\bar{J}_p^{(t)}\!\left(\hat{\bs\theta}^{(t)} \right)\!
\leq 2 \left( {J}_p^{(t)}\!\left(\hat{\bs\theta}^{(t)*}|\D \right)\!-\!{J}_p^{(t)}\!\left(\hat{\bs\theta}^{(t)}|\D \right) \! \right) \!+\!O\left( -\log(\pi') \right)\!\left(2n_e l''|_{\bs\eta=\0}\mbox{V}^{(t)}_{(1)} \right)^{-1}\notag\\
&=2 \left( {J}_p^{(t)}(\hat{\bs\theta}^{(t)*}|\D)-{J}_p^{(t)}(\hat{\bs\theta}^{(t)}|\D) \right)-C' \log(\pi')\left(2n_e l''|_{\bs\eta=\0}\mbox{V}^{(t)}_{(1)} \right)^{-1}\!\!.\label{eqn:boundJ}
\end{align}
${J}_p^{(t)}(\hat{\bs\theta}^{(t)*}|\D)-{J}_p^{(t)}(\hat{\bs\theta}^{(t)}|\D)$ can be regarded as a special case of ${J}(\hat{\bs\theta}^{(t)*}|\D)-J(\hat{\bs\theta}|\D )$ when strong convexity with modulus $2\Lambda_0$ is automatically fulfilled by $\Lambda R(\bs{\theta})$. Therefore, we may apply the bounds in Eqs (\ref{eqn:lossdiff1}, \ref{eqn:lossdiff2}) to  ${J}_p^{(t)}(\hat{\bs\theta}^{(t)*}|\D)-{J}_p^{(t)}(\hat{\bs\theta}^{(t)}|\D)$ in Eq (24). Taken together with Eqs (\ref{eqn:boundJ}) and (\ref{eqn:SSineq}), then with probability at least $1-\pi-\pi'$,
\begin{align}\label{eqn:boundL}
L\!\left(\hat{\bs\theta}^{(t)*}\right)\!\leq\! L\!\left( \bs\theta^{0}\right)\!+\!n^{-1}\! R^{(t)}\!\left(\bs\theta^0\right)\!+\!\left(n^{-1}C\!-\!2^{-1}C' \log(\pi')\right)\!\left(\!n_e l''|_{\bs\eta=\0}\mbox{V}^{(t)}_{(1)} \right)^{-1}.
\end{align}
Letting $\varrho= n^{-1}R^{(t)}\!\left(\bs\theta_j^0\right)\!+\!\left(n^{-1}C\!-\!2^{-1}C' \log(\pi')\right)\!\left(\!n_e l''|_{\bs\eta=\0}\mbox{V}^{(t)}_{(1)} \right)^{-1}\!\!$, we can solve for $n$ to get the sample complexity in Eq (25).
\end{proof}

\vspace{-6pt}\section{Additional Experiment Results}\vspace{-6pt}
The results in the experiments with $\epsilon-$DP are presented in this section. 
In summary, the results are similar to those with $(\epsilon,\delta=0.001)$-DP in Sec 5 of the main manuscript. 


\begin{figure}[H] 
\begin{minipage}{0.08\textwidth}
\footnotesize linear\\regression $n=200$ \end{minipage}
\begin{minipage}{0.92\textwidth}
\includegraphics[width=0.24\linewidth, trim=4pt 12pt 15pt 18pt,clip]{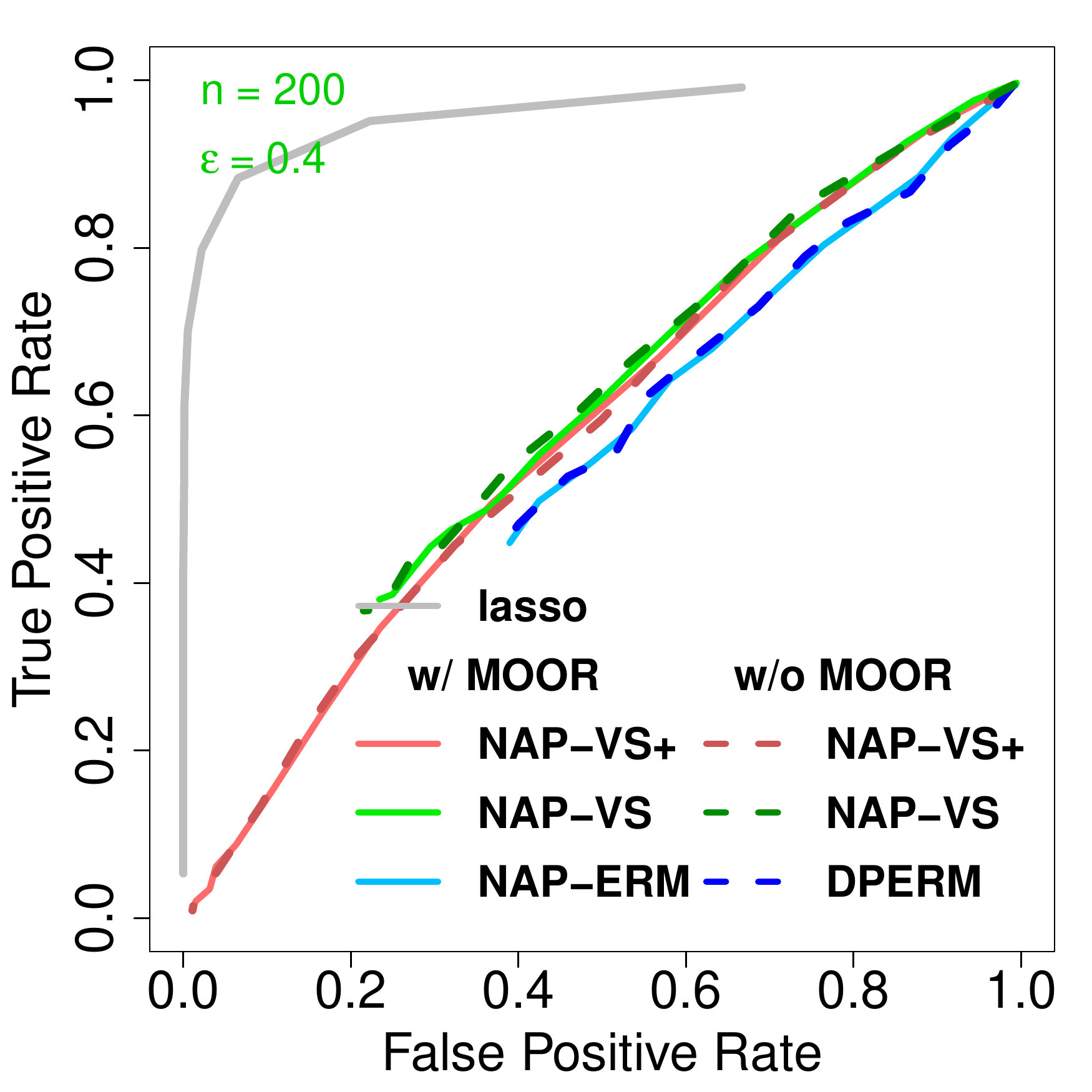}
\includegraphics[width=0.24\linewidth, trim=4pt 12pt 15pt 18pt,clip]{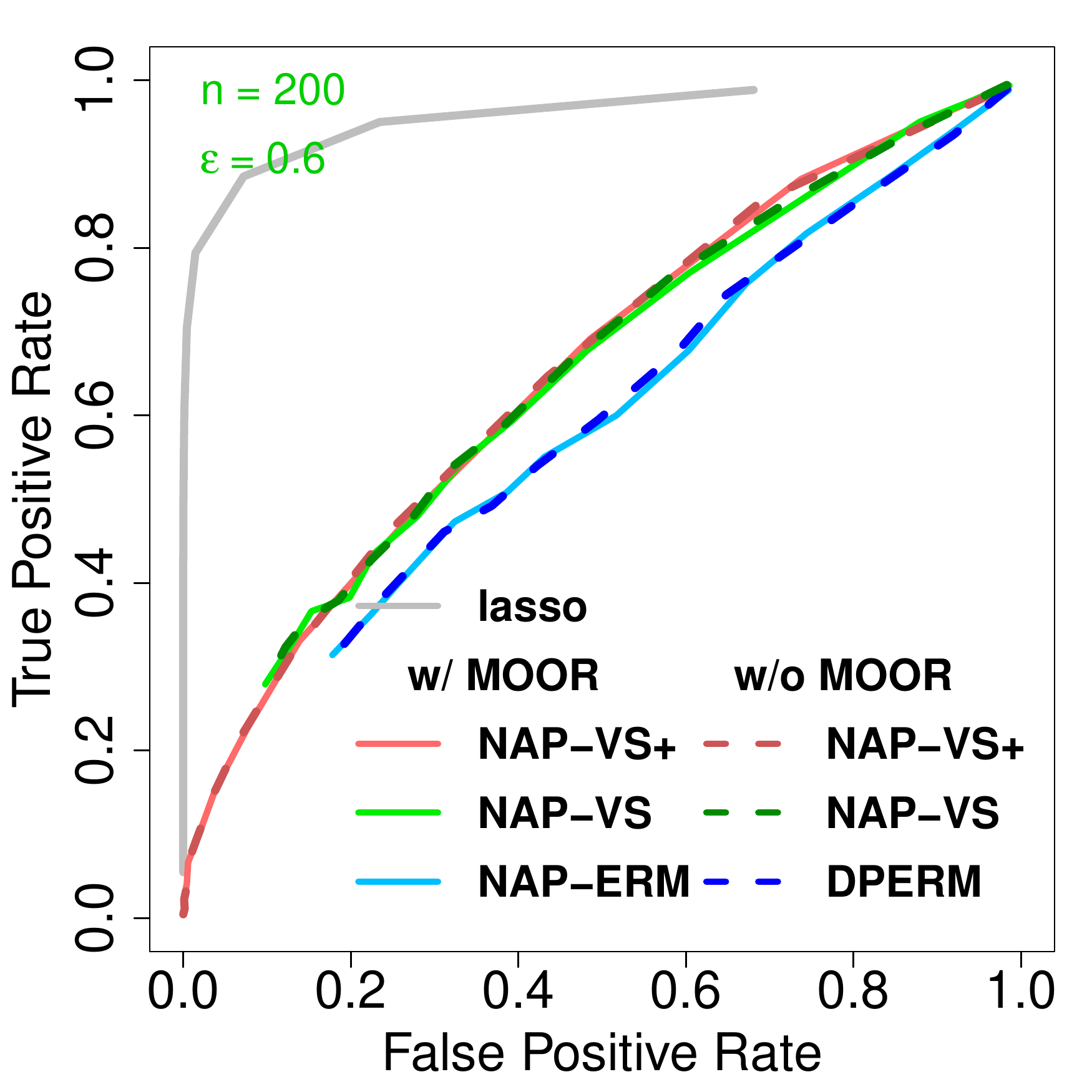}
\includegraphics[width=0.24\linewidth, trim=4pt 12pt 15pt 18pt,clip]{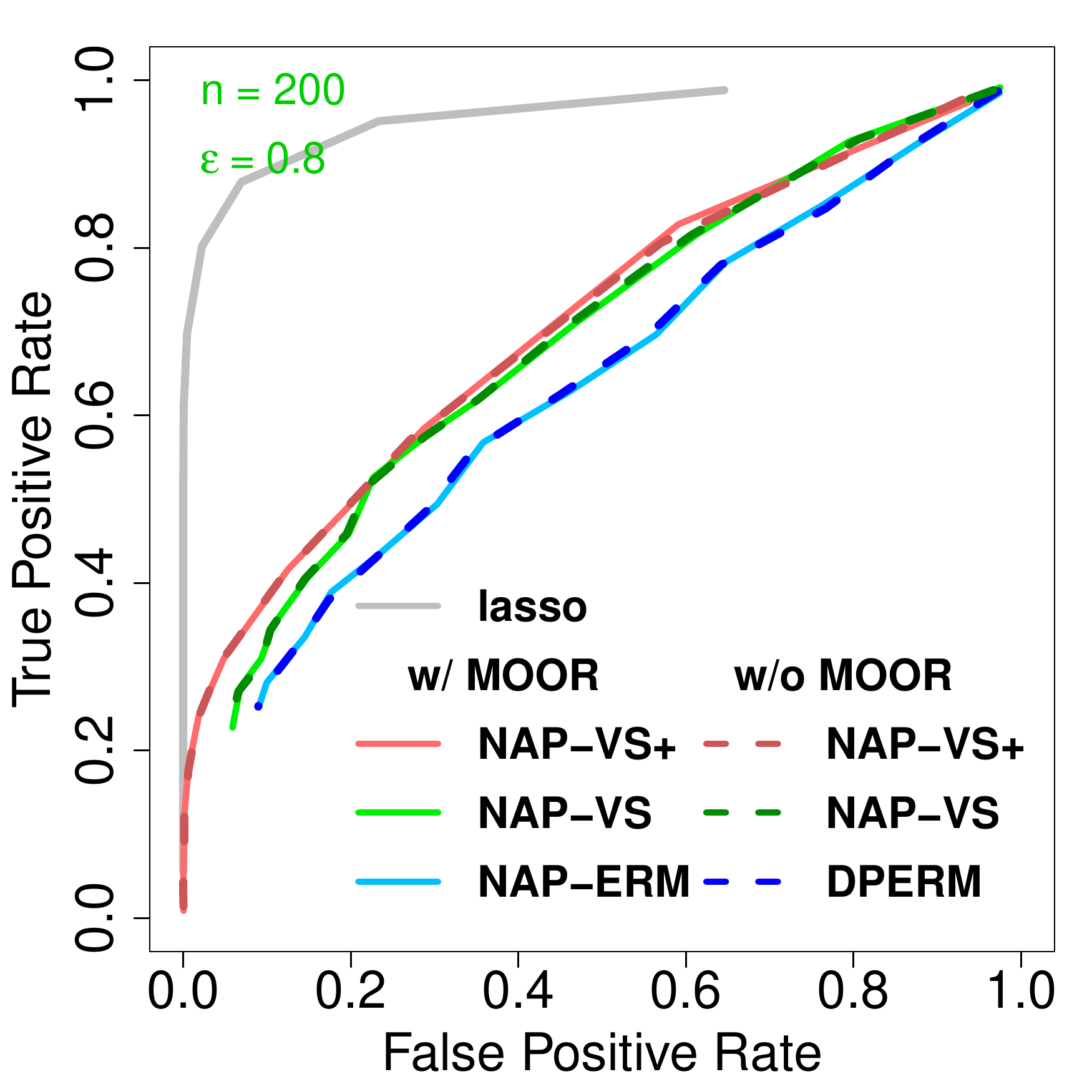}
\includegraphics[width=0.24\linewidth, trim=4pt 12pt 15pt 18pt,clip]{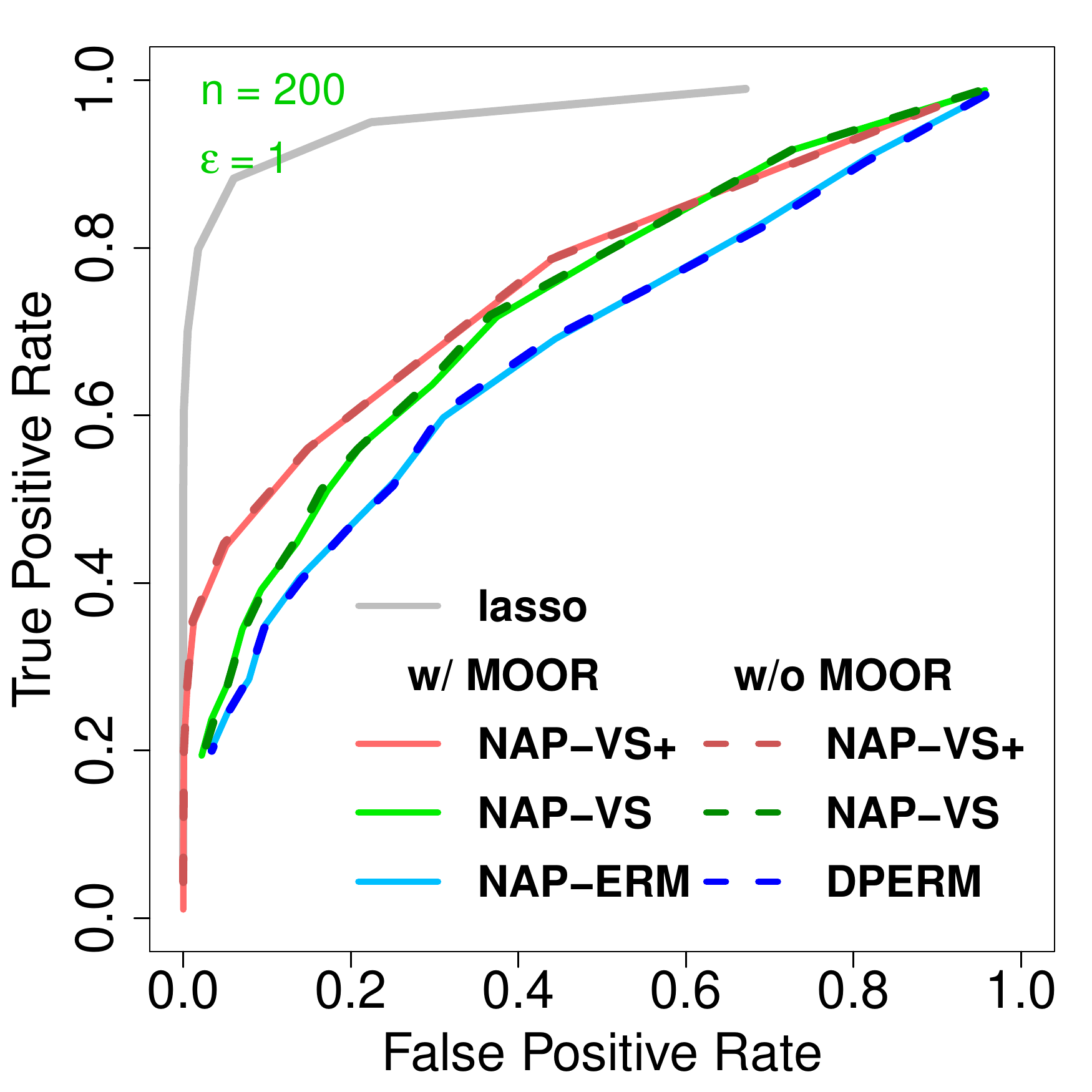}
\end{minipage}
\begin{minipage}{0.08\textwidth}\footnotesize $n=500$ \end{minipage}
\begin{minipage}{0.92\textwidth}
\includegraphics[width=0.24\linewidth, trim=4pt 12pt 15pt 18pt,clip]{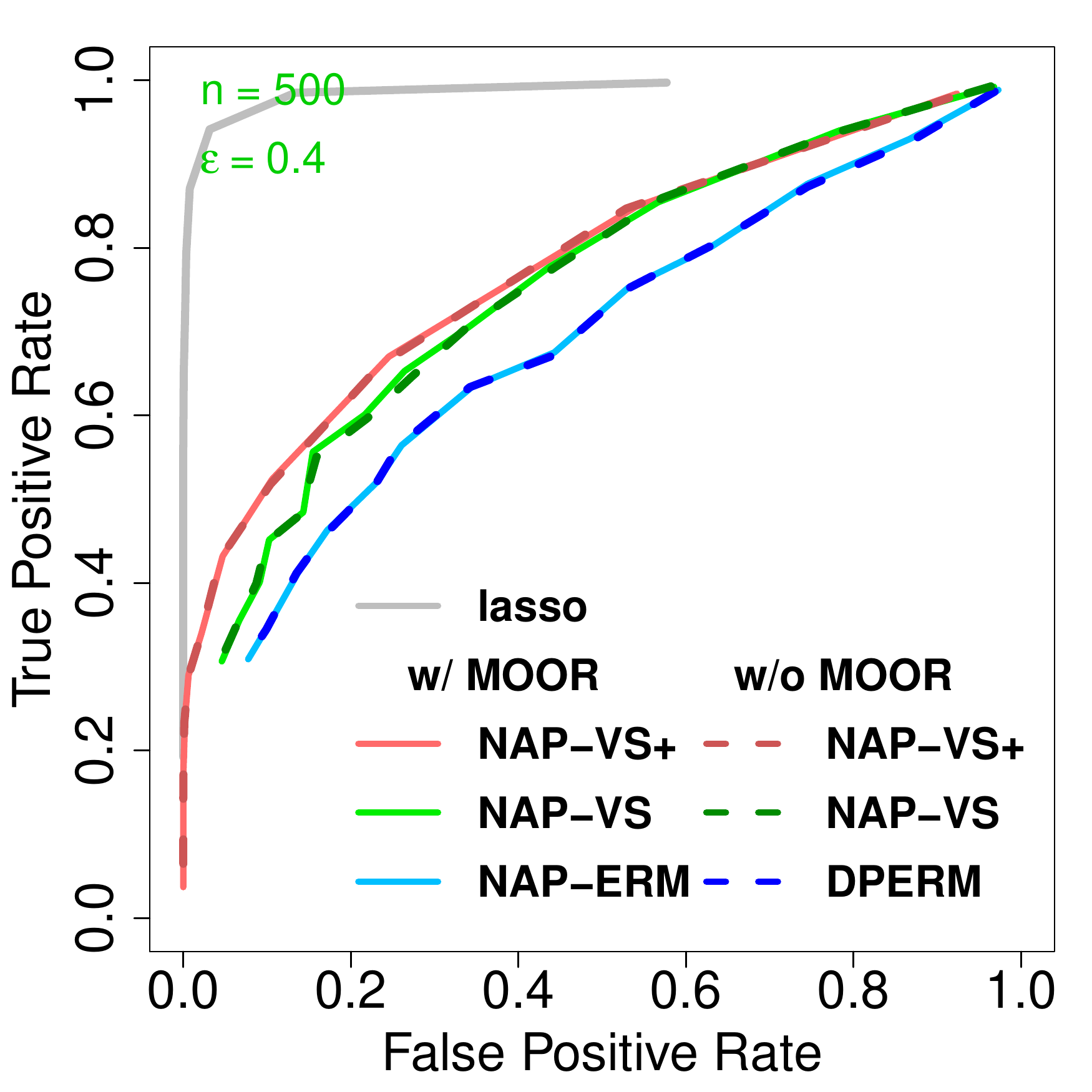}
\includegraphics[width=0.24\linewidth, trim=4pt 12pt 15pt 18pt,clip]{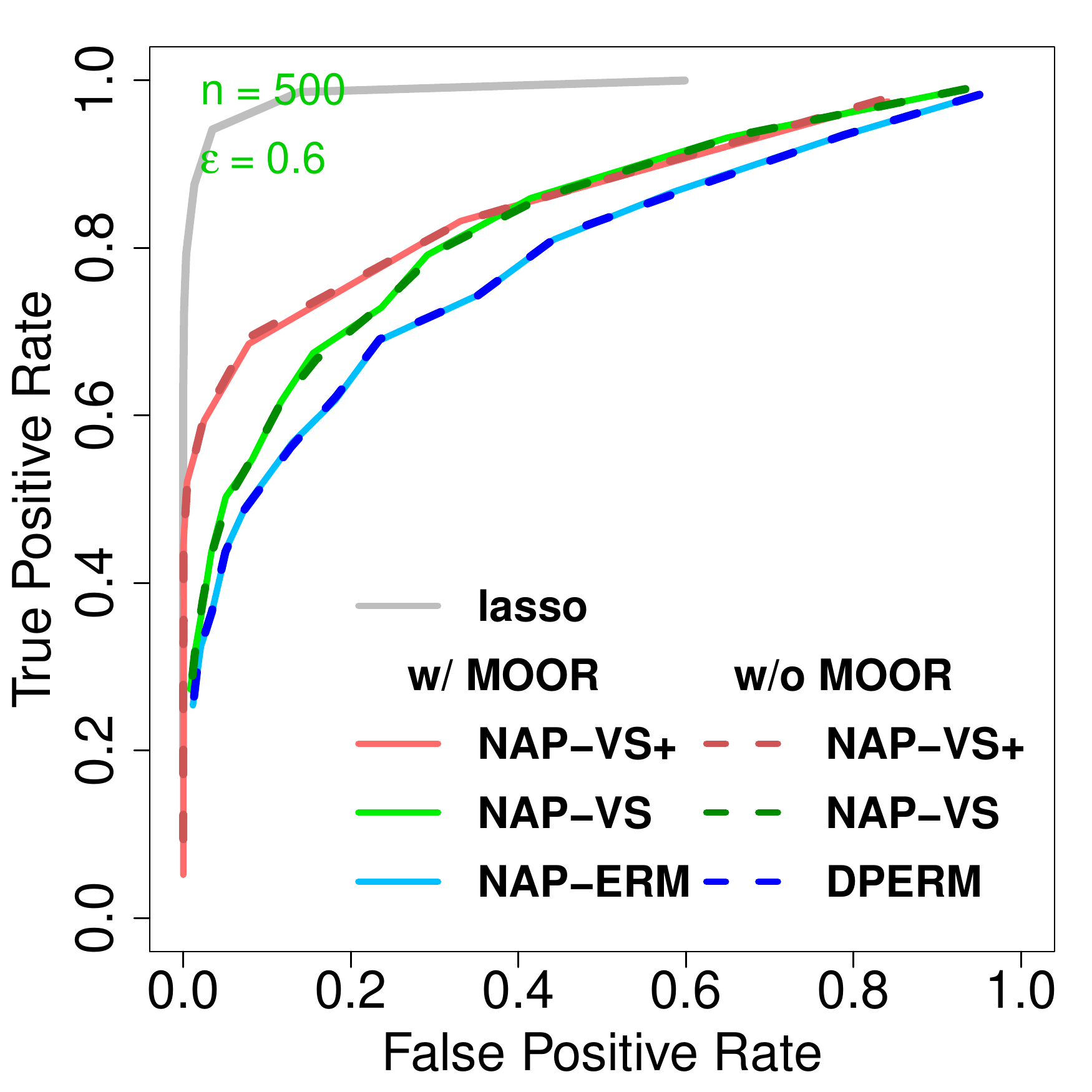}
\includegraphics[width=0.24\linewidth, trim=4pt 12pt 15pt 18pt,clip]{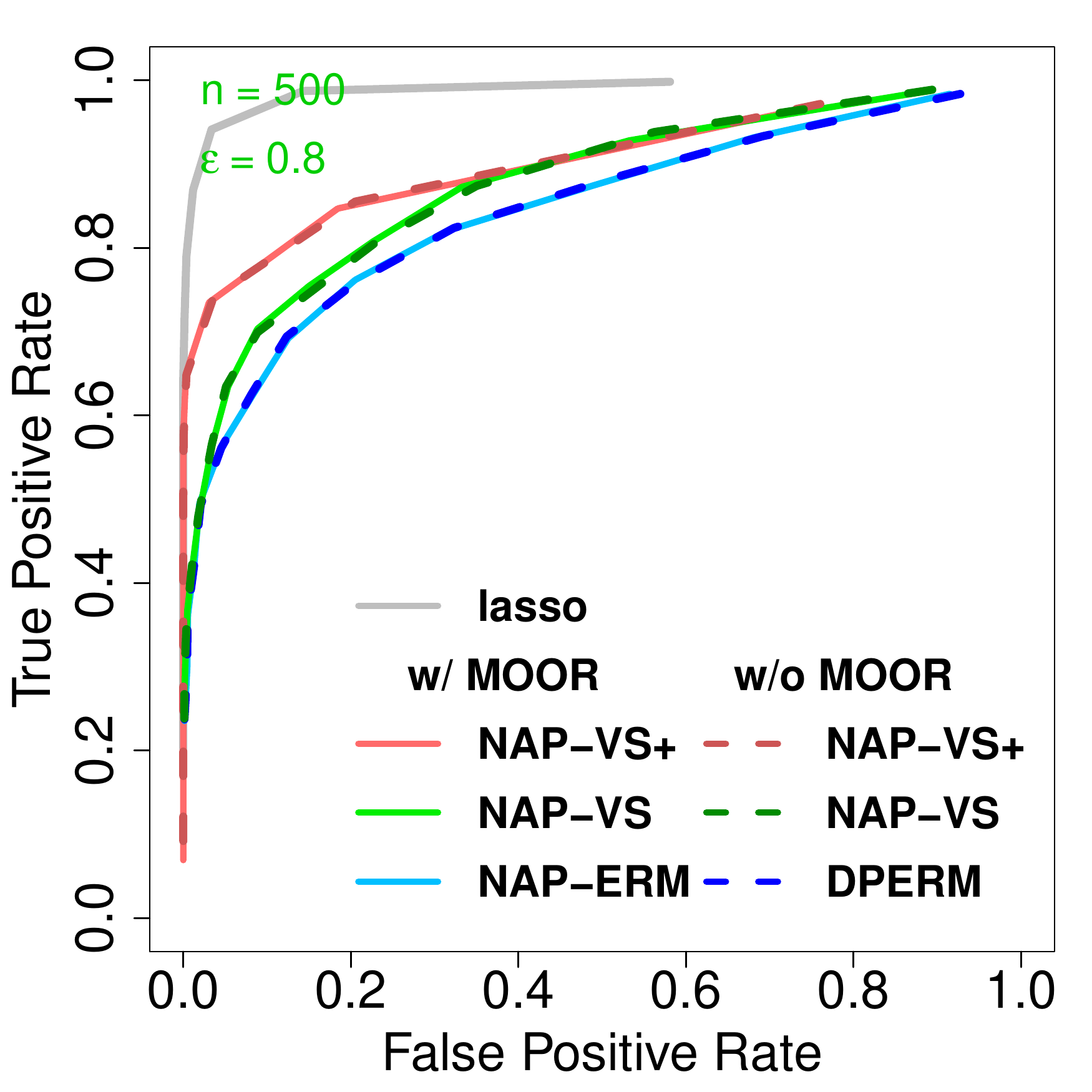}
\includegraphics[width=0.24\linewidth, trim=4pt 12pt 15pt 18pt,clip]{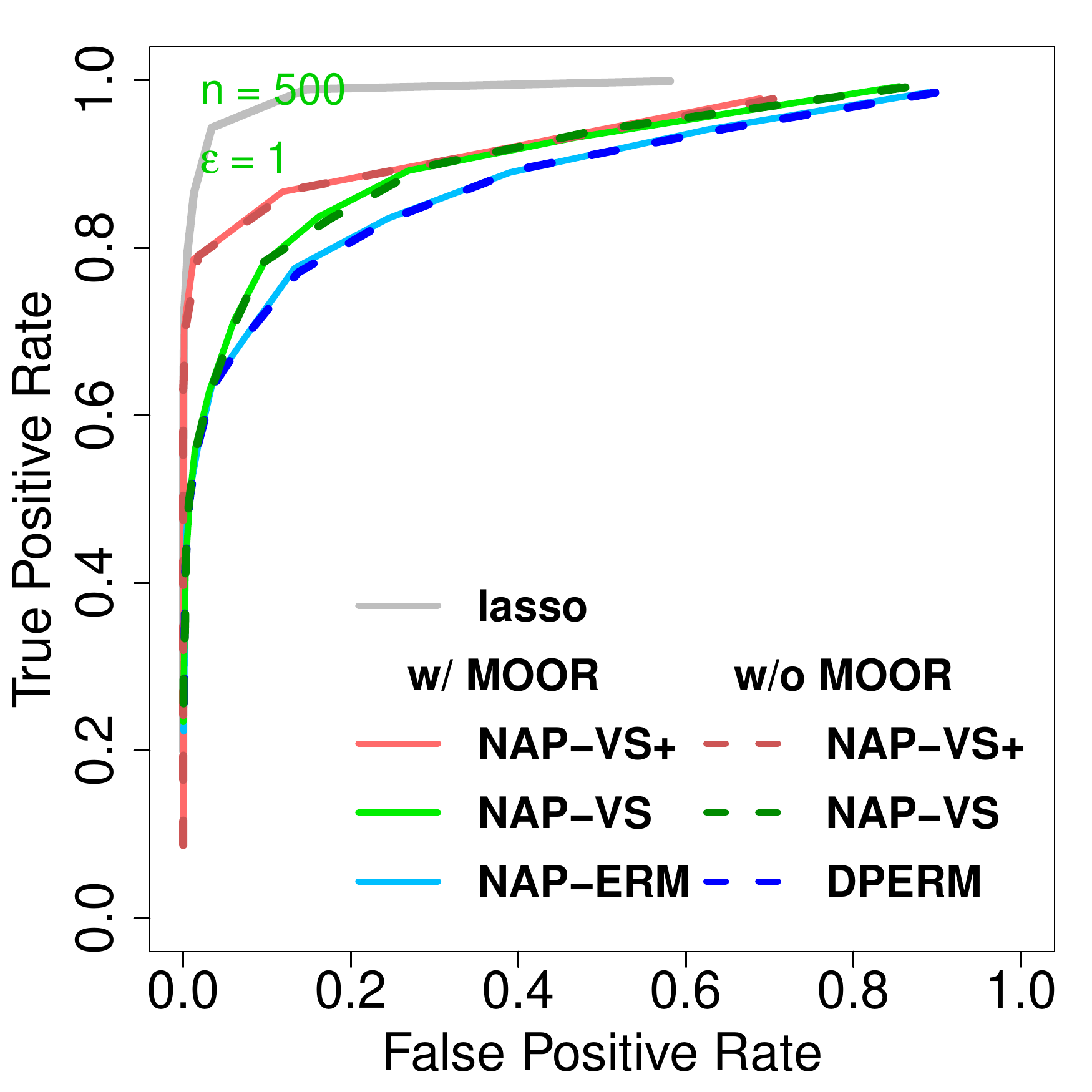}
\end{minipage}

\begin{minipage}{0.08\textwidth}
\footnotesize logistic regression $n=500$ \end{minipage}
\begin{minipage}{0.92\textwidth}
\includegraphics[width=0.24\linewidth, trim=4pt 9pt 15pt 18pt,clip]{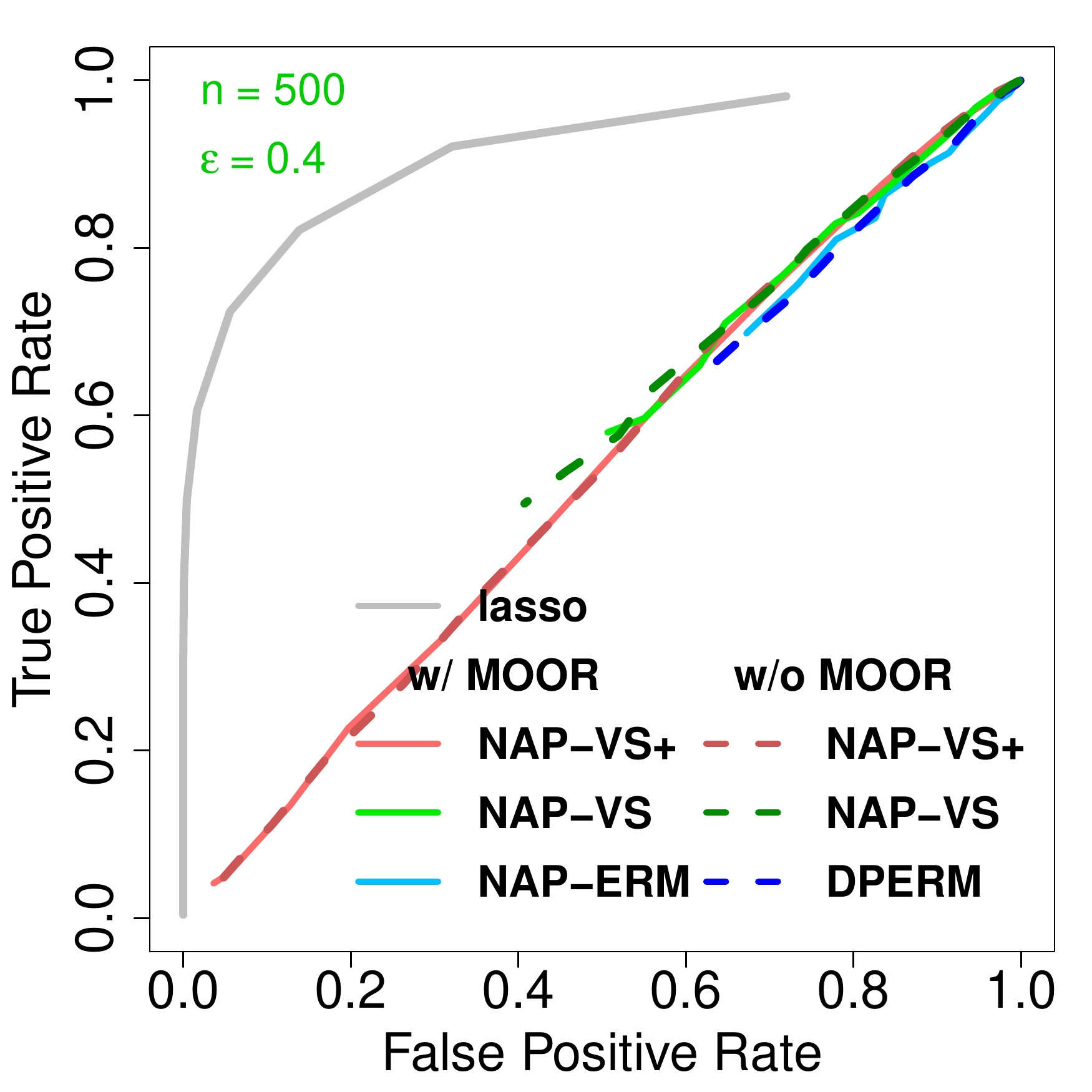}
\includegraphics[width=0.24\linewidth, trim=4pt 9pt 15pt 18pt,clip]{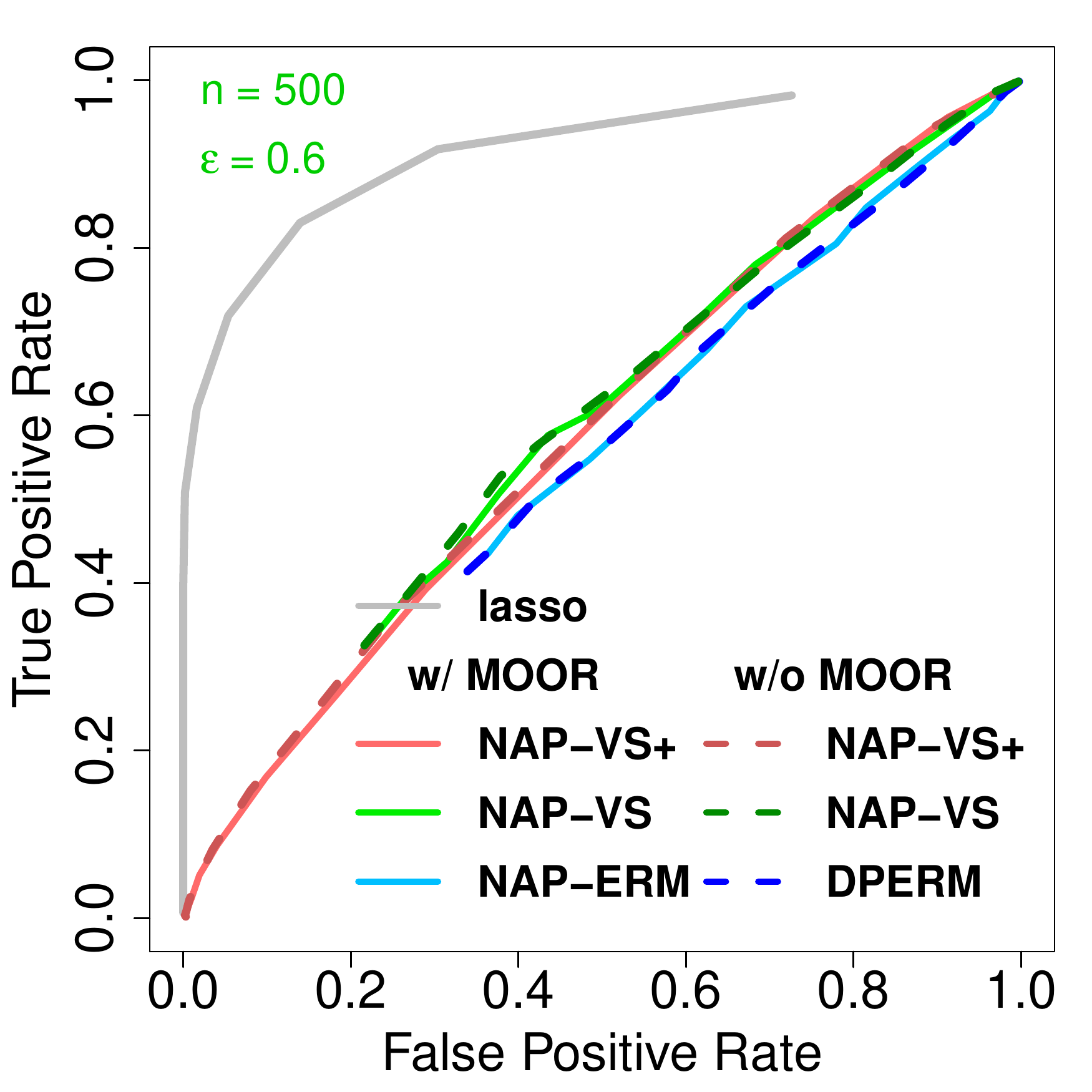}
\includegraphics[width=0.24\linewidth, trim=4pt 9pt 15pt 18pt,clip]{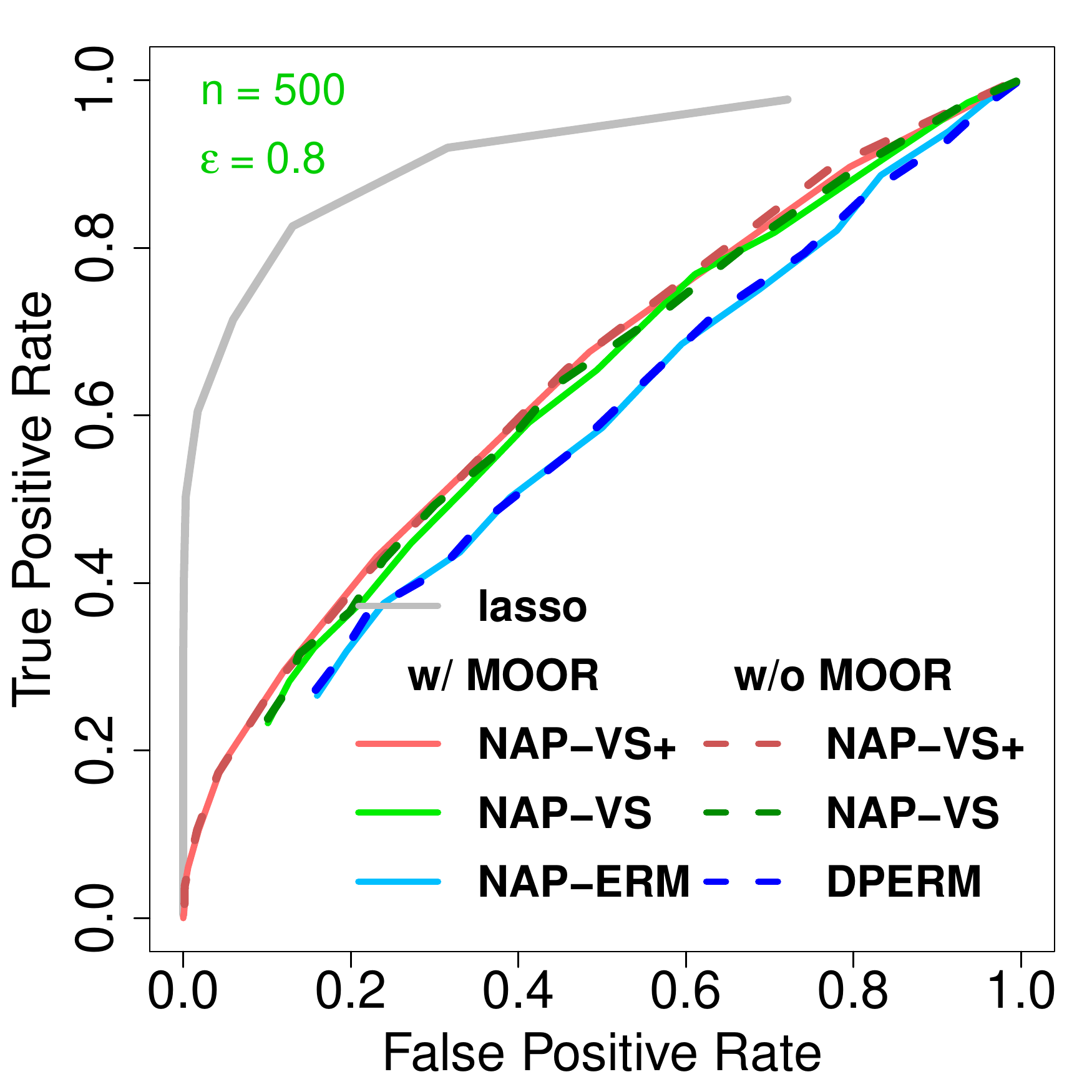}
\includegraphics[width=0.24\linewidth, trim=4pt 9pt 15pt 18pt,clip]{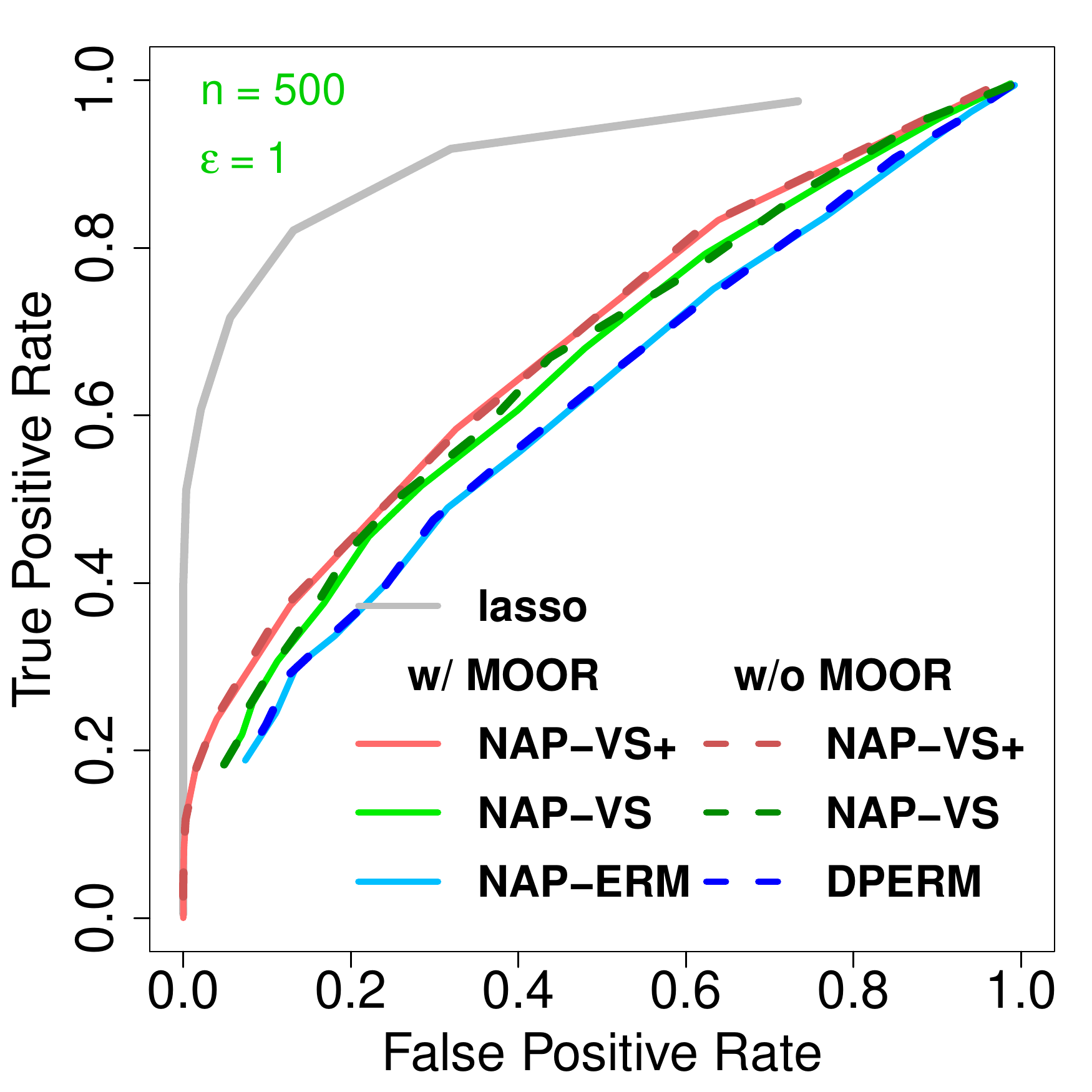}
\end{minipage}
\begin{minipage}{0.08\textwidth}
\footnotesize $n=1000$ \end{minipage}
\begin{minipage}{0.92\textwidth}
\includegraphics[width=0.24\linewidth, trim=4pt 9pt 15pt 18pt,clip]{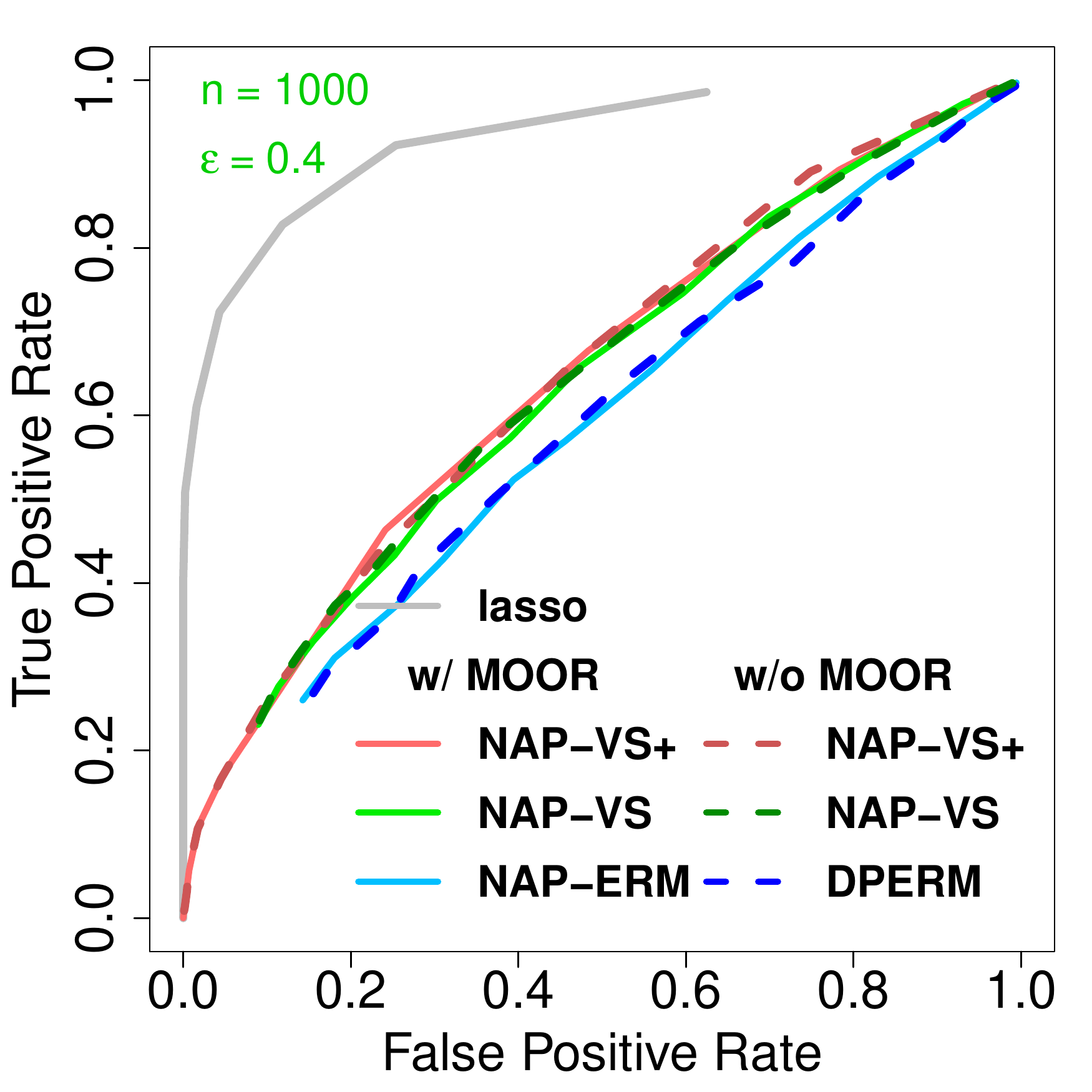}
\includegraphics[width=0.24\linewidth, trim=4pt 9pt 15pt 18pt,clip]{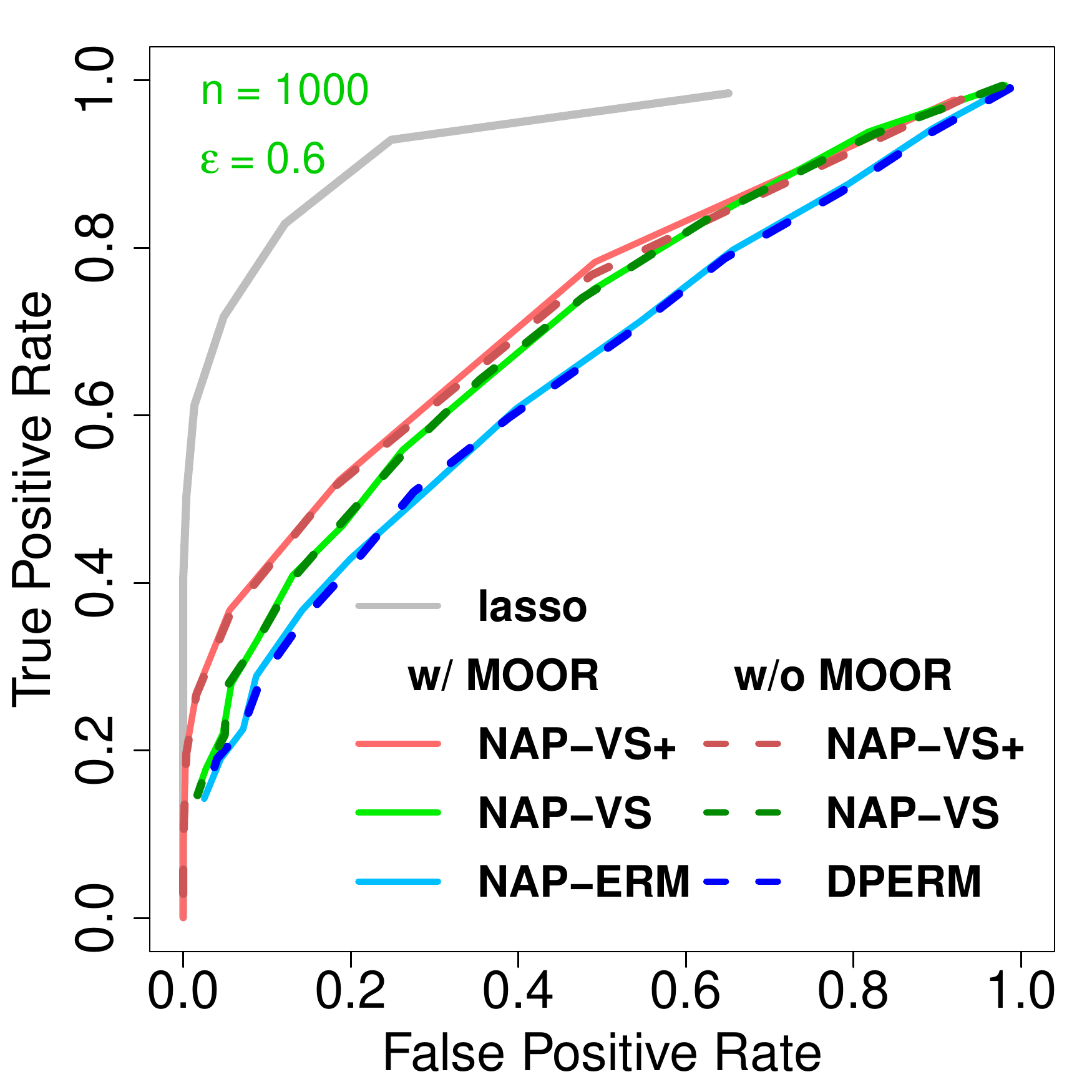}
\includegraphics[width=0.24\linewidth, trim=4pt 9pt 15pt 18pt,clip]{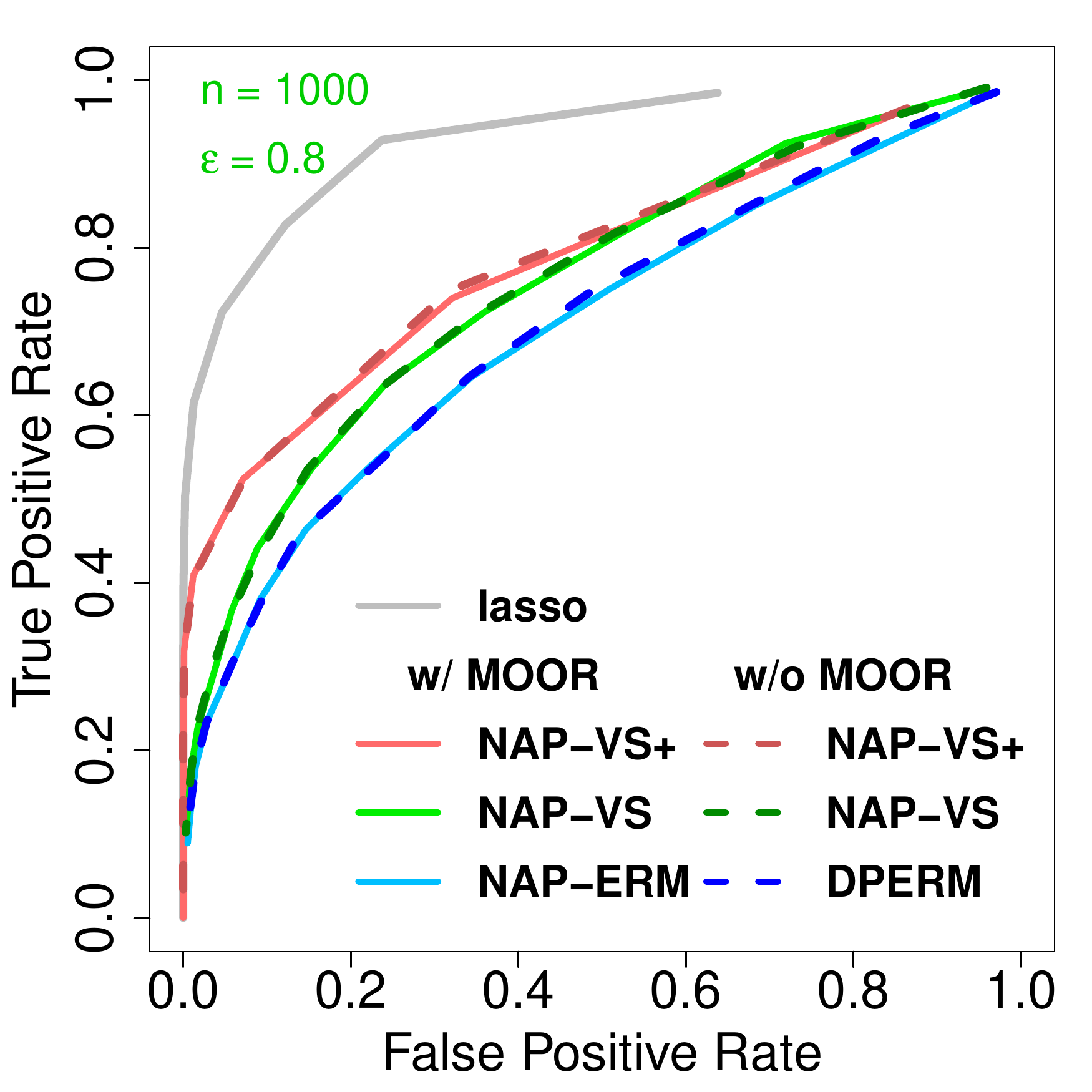}
\includegraphics[width=0.24\linewidth, trim=4pt 9pt 15pt 18pt,clip]{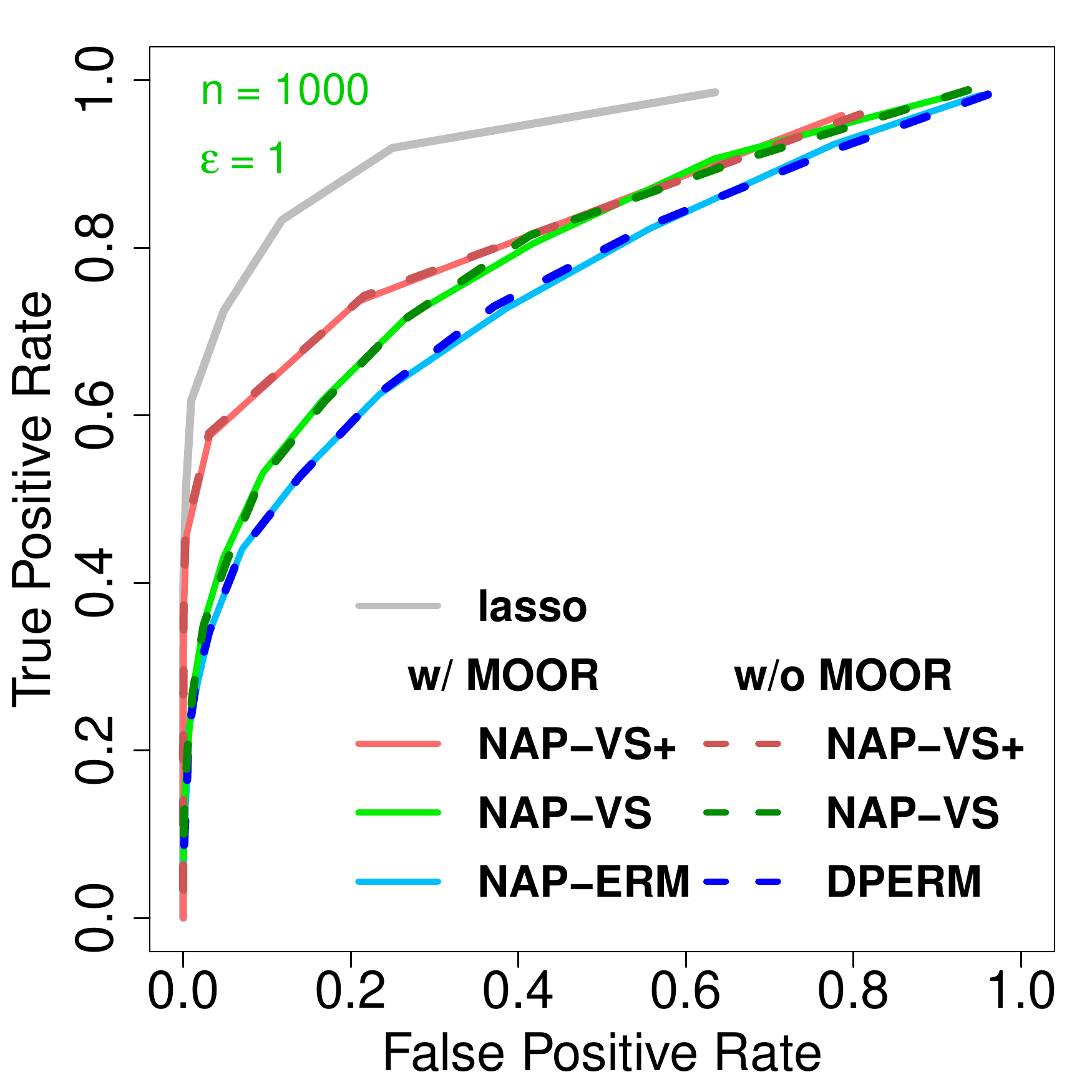}
\end{minipage}

\begin{minipage}{0.08\textwidth}
\footnotesize Poisson regression $n=500$ \end{minipage}
\begin{minipage}{0.92\textwidth}
\includegraphics[width=0.24\linewidth, trim=4pt 9pt 15pt 18pt,clip]{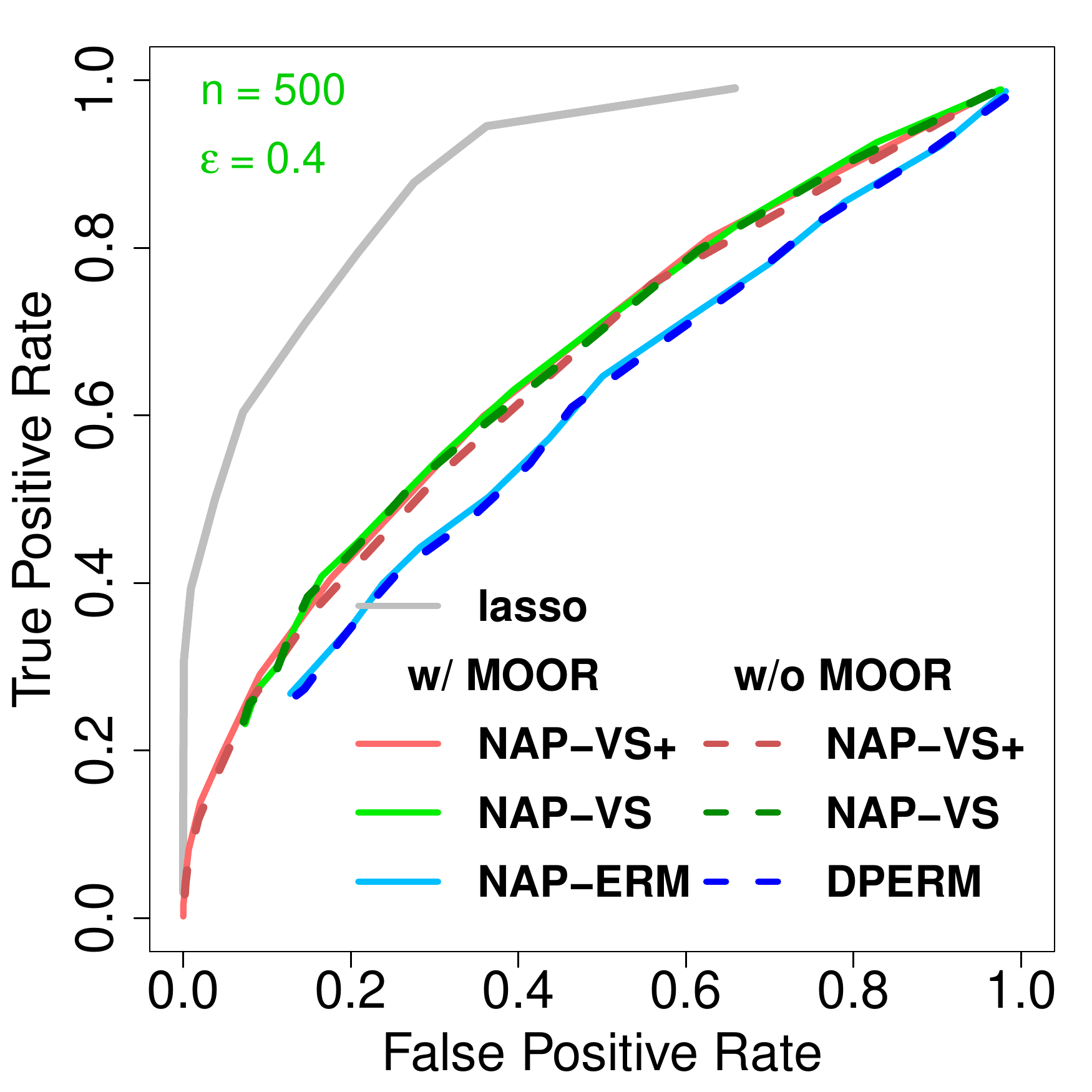}
\includegraphics[width=0.24\linewidth, trim=4pt 9pt 15pt 18pt,clip]{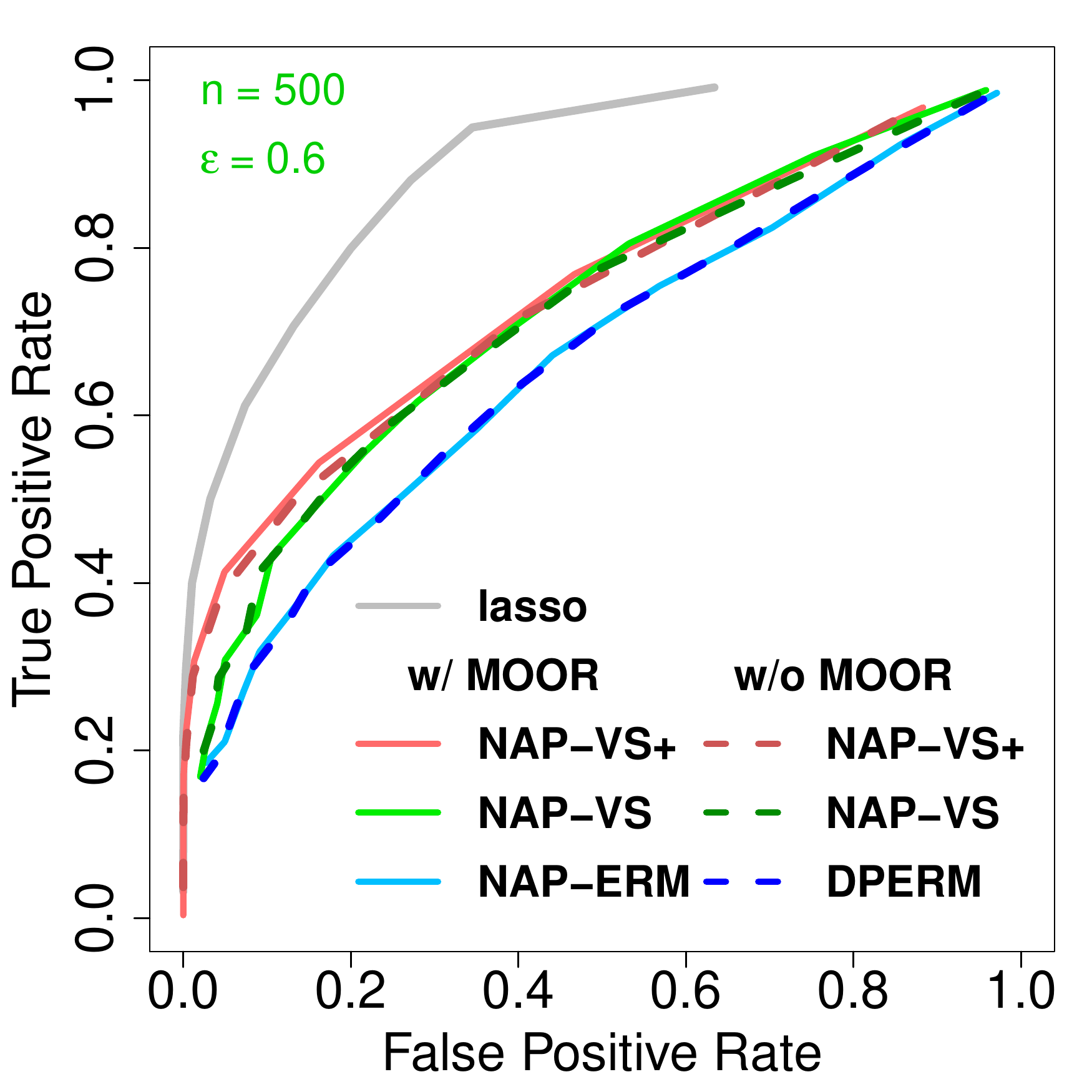}
\includegraphics[width=0.24\linewidth, trim=4pt 9pt 15pt 18pt,clip]{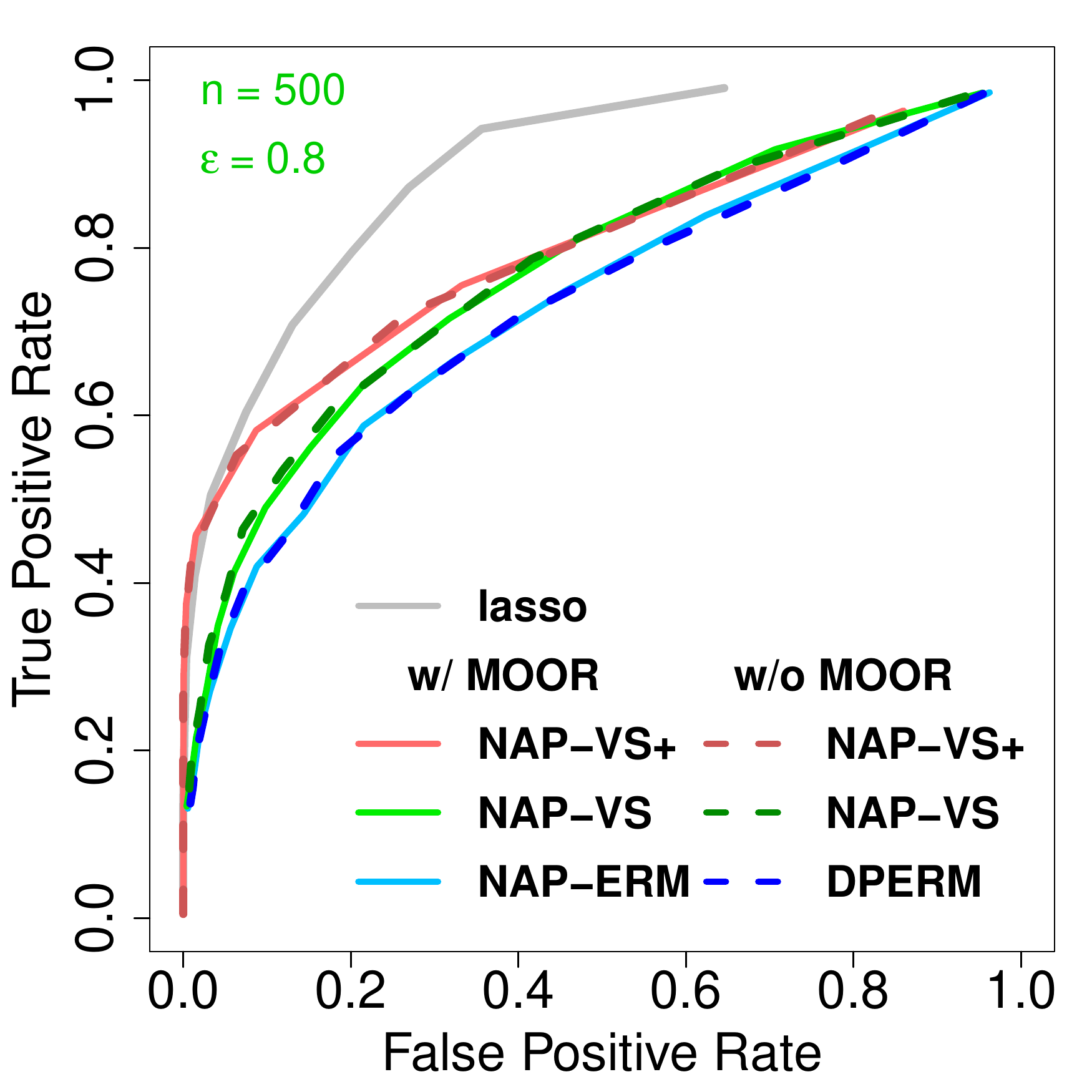}
\includegraphics[width=0.24\linewidth, trim=4pt 9pt 15pt 18pt,clip]{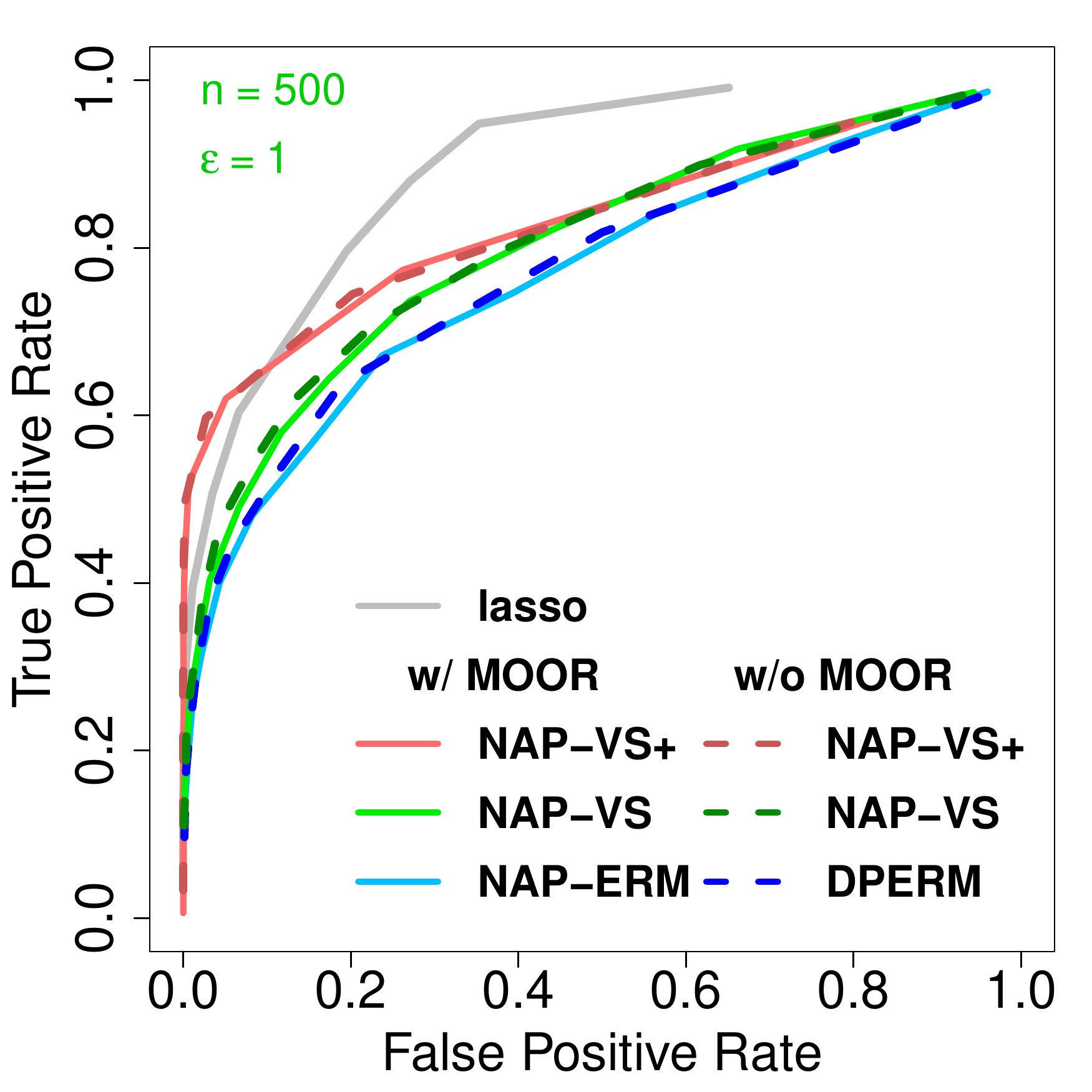}
\end{minipage}
\begin{minipage}{0.08\textwidth}\footnotesize $n=1000$ \end{minipage}
\begin{minipage}{0.92\textwidth}
\includegraphics[width=0.24\linewidth, trim=4pt 9pt 15pt 18pt,clip]{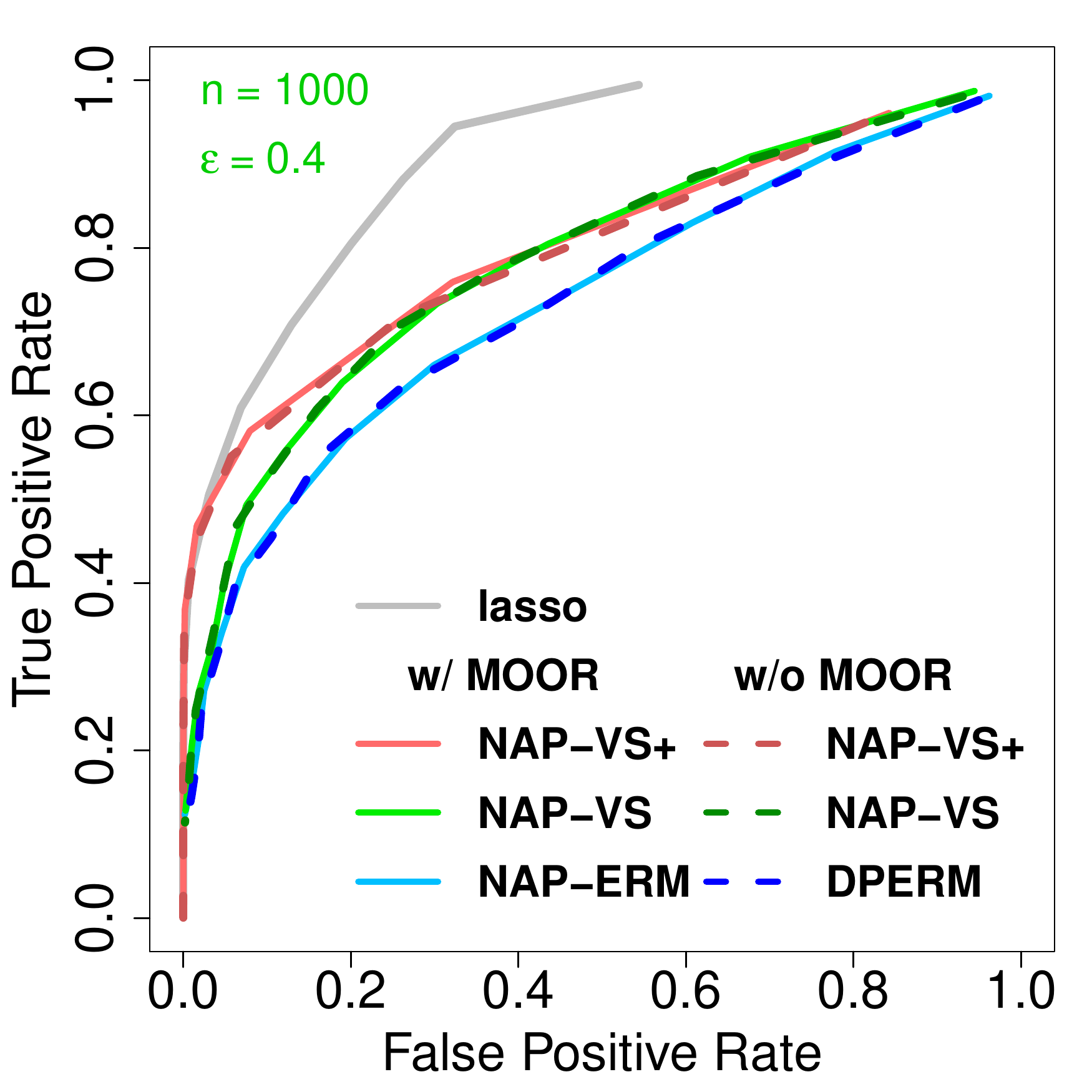}
\includegraphics[width=0.24\linewidth, trim=4pt 9pt 15pt 18pt,clip]{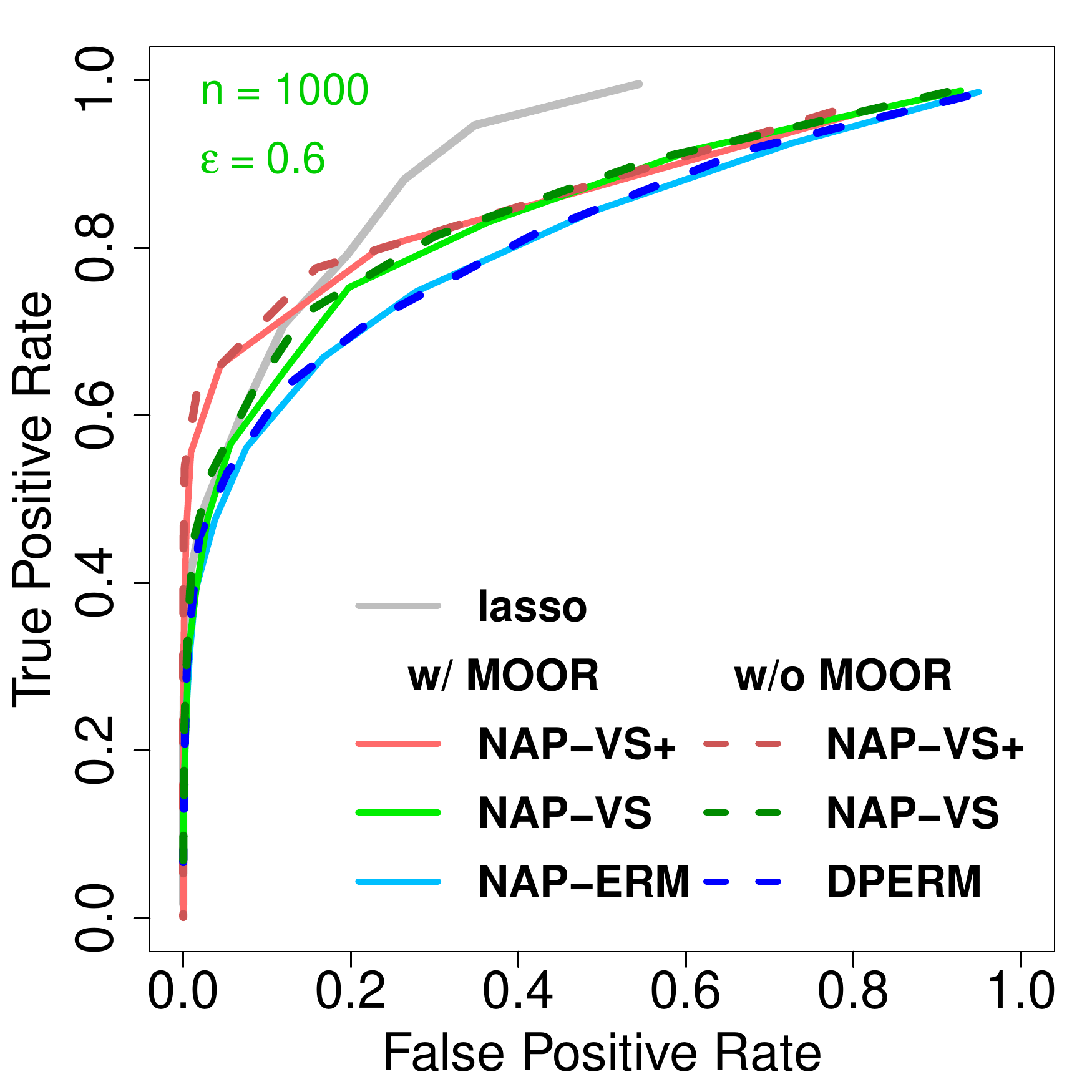}
\includegraphics[width=0.24\linewidth, trim=4pt 9pt 15pt 18pt,clip]{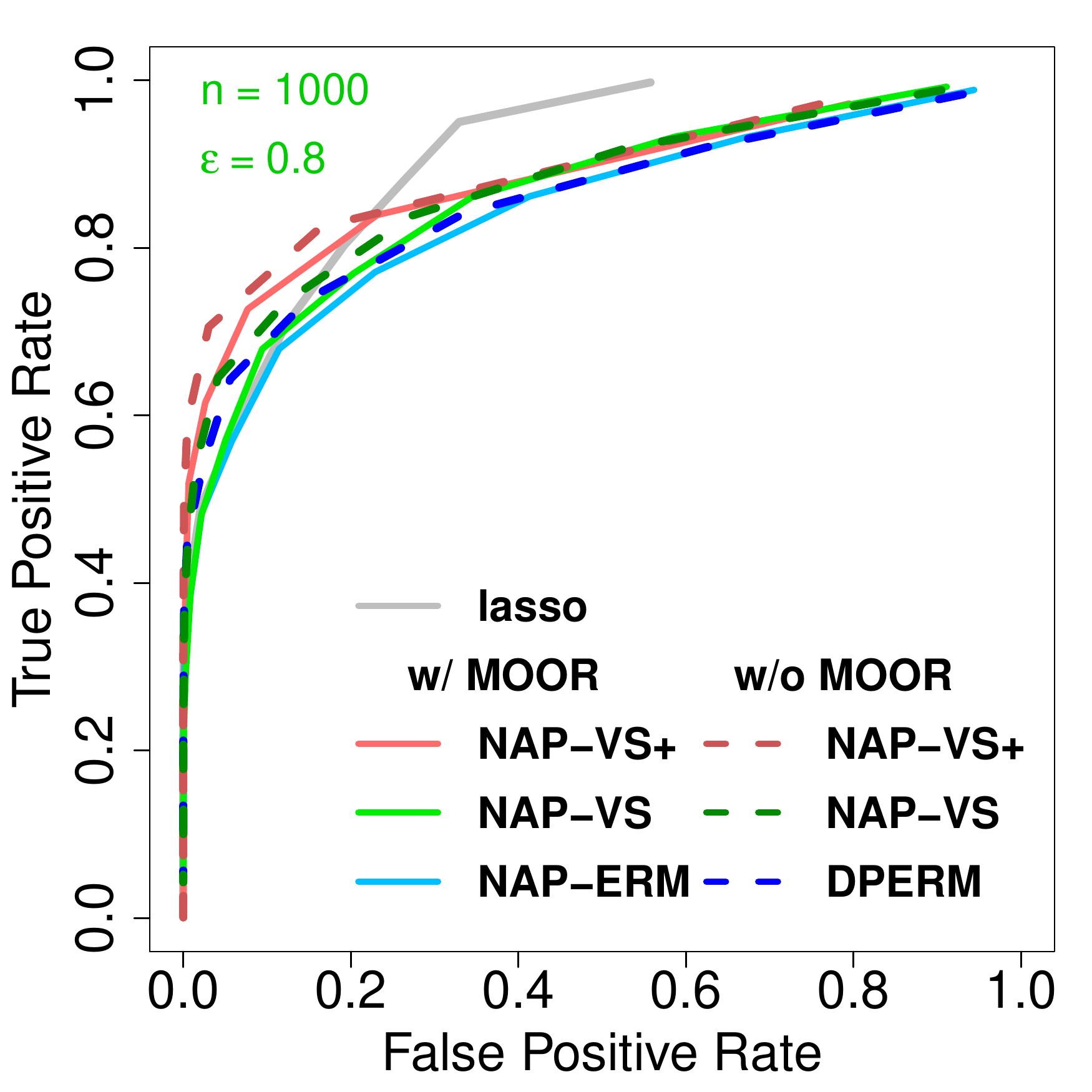}
\includegraphics[width=0.24\linewidth, trim=4pt 9pt 15pt 18pt,clip]{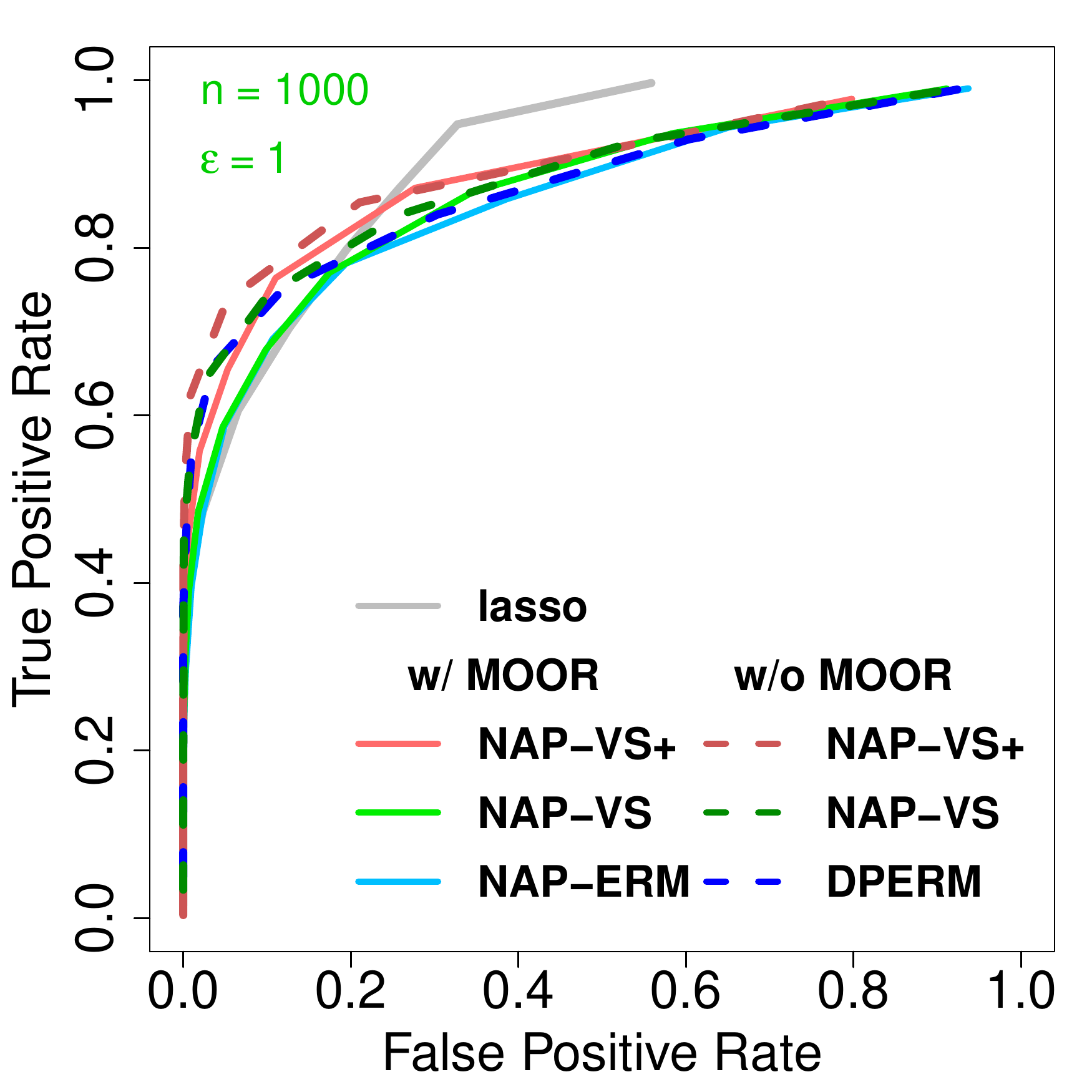}
\end{minipage} \vspace{-12pt}
\caption{Variable Selection ROC curves by varying  tuning parameter $\Lambda$, sample size $n$, and privacy budget $\epsilon$ ($\delta=0.0001$)} \label{fig:SIM21}
\end{figure}

\begin{figure}[H]
\begin{minipage}{0.08\textwidth}
\footnotesize linear regression $n=200$ \end{minipage}
\begin{minipage}{0.92\textwidth}
\includegraphics[width=0.24\linewidth, trim=4pt 9pt 15pt 18pt,clip]{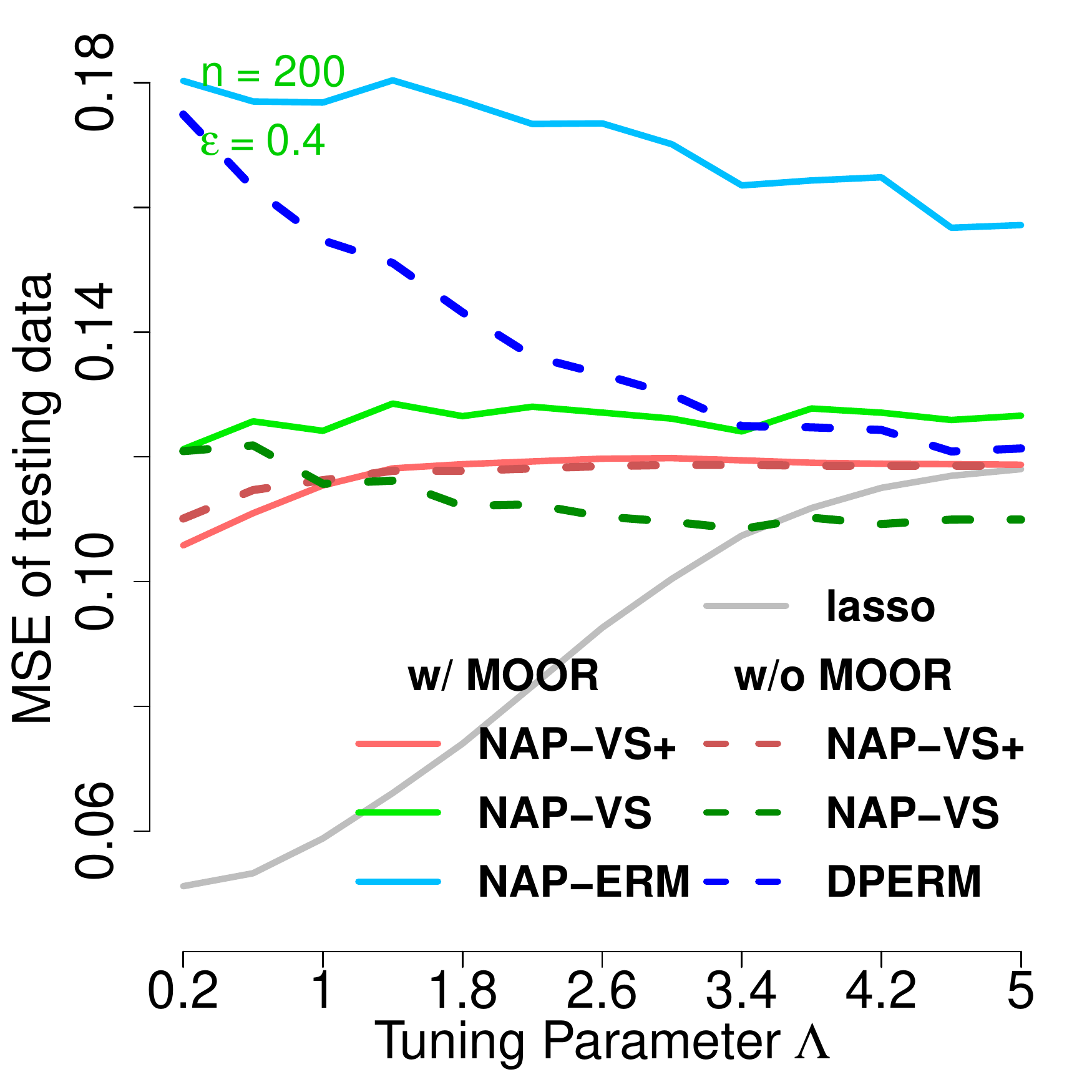}
\includegraphics[width=0.24\linewidth, trim=4pt 9pt 15pt 18pt,clip]{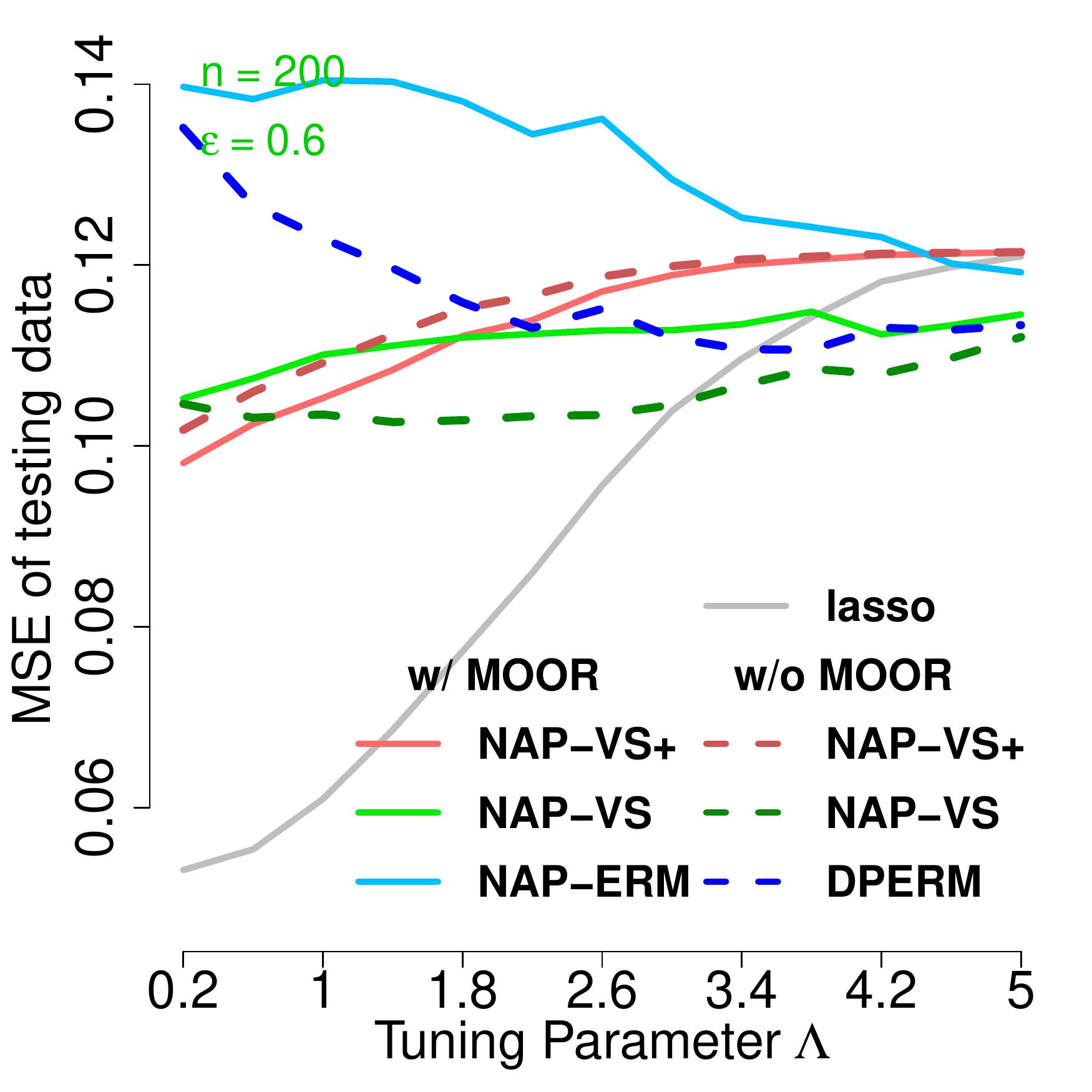}
\includegraphics[width=0.24\linewidth, trim=4pt 9pt 15pt 18pt,clip]{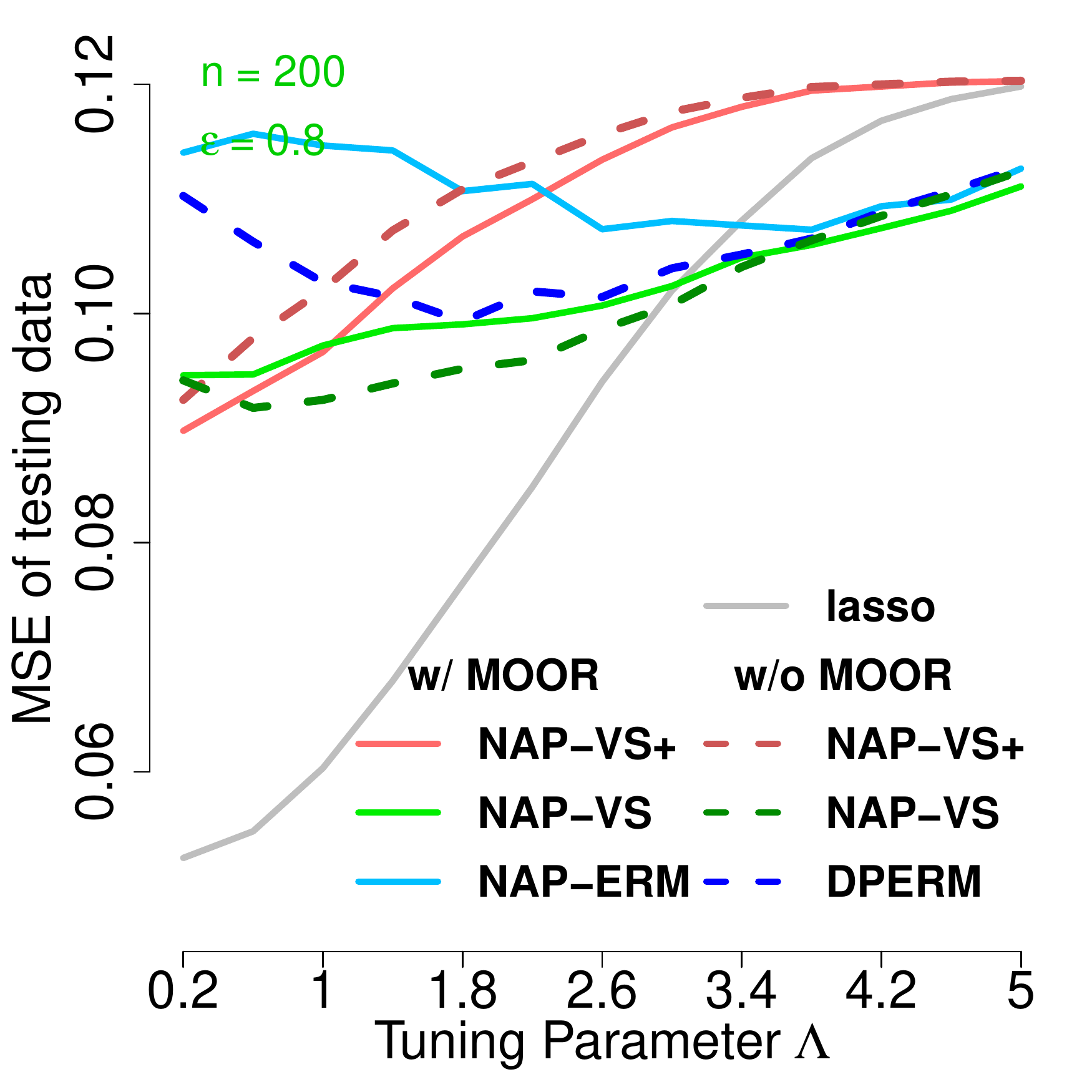}
\includegraphics[width=0.24\linewidth, trim=4pt 9pt 15pt 18pt,clip]{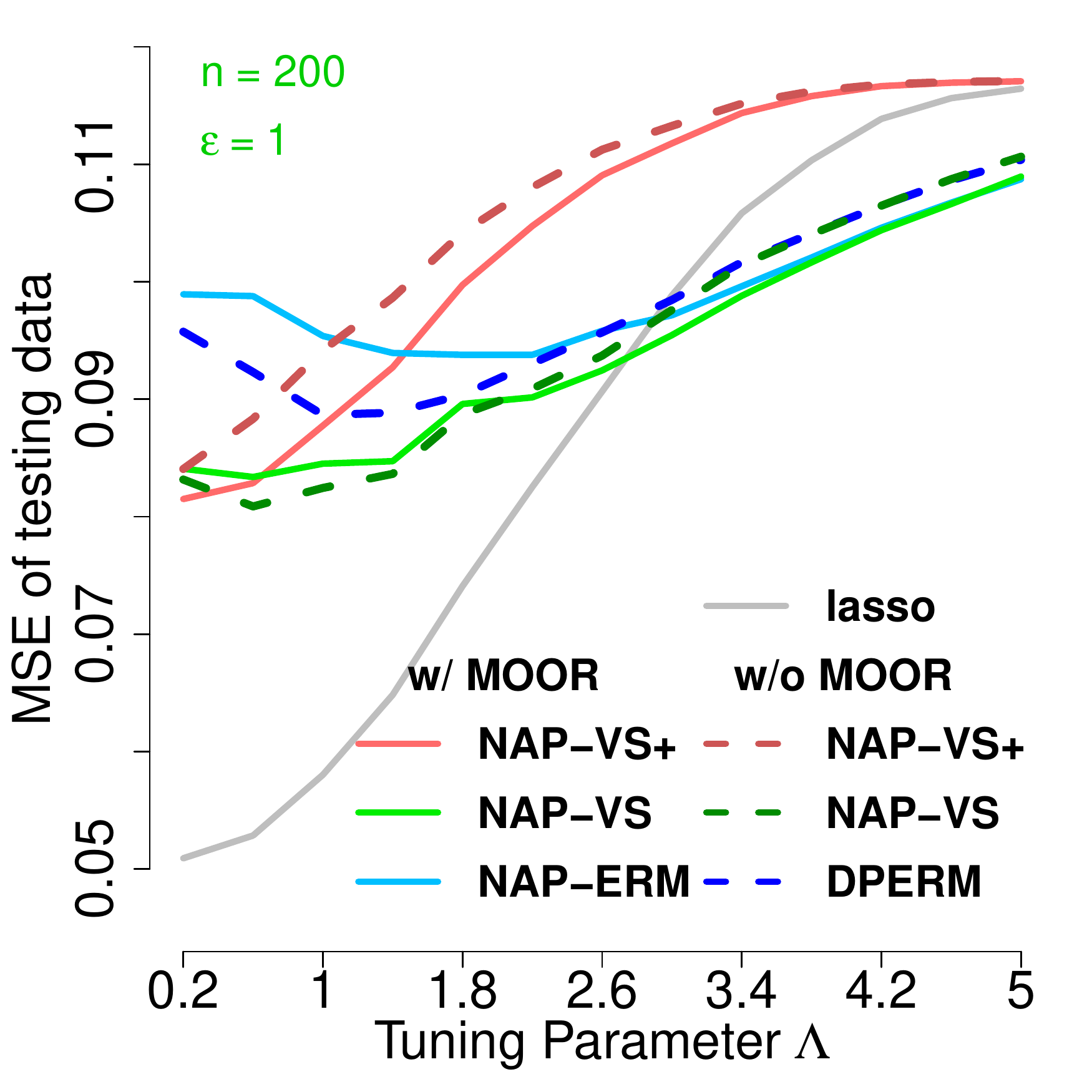}
\end{minipage}
\begin{minipage}{0.08\textwidth}\footnotesize $n=500$ \end{minipage}
\begin{minipage}{0.92\textwidth}
\includegraphics[width=0.24\linewidth, trim=4pt 9pt 15pt 18pt,clip]{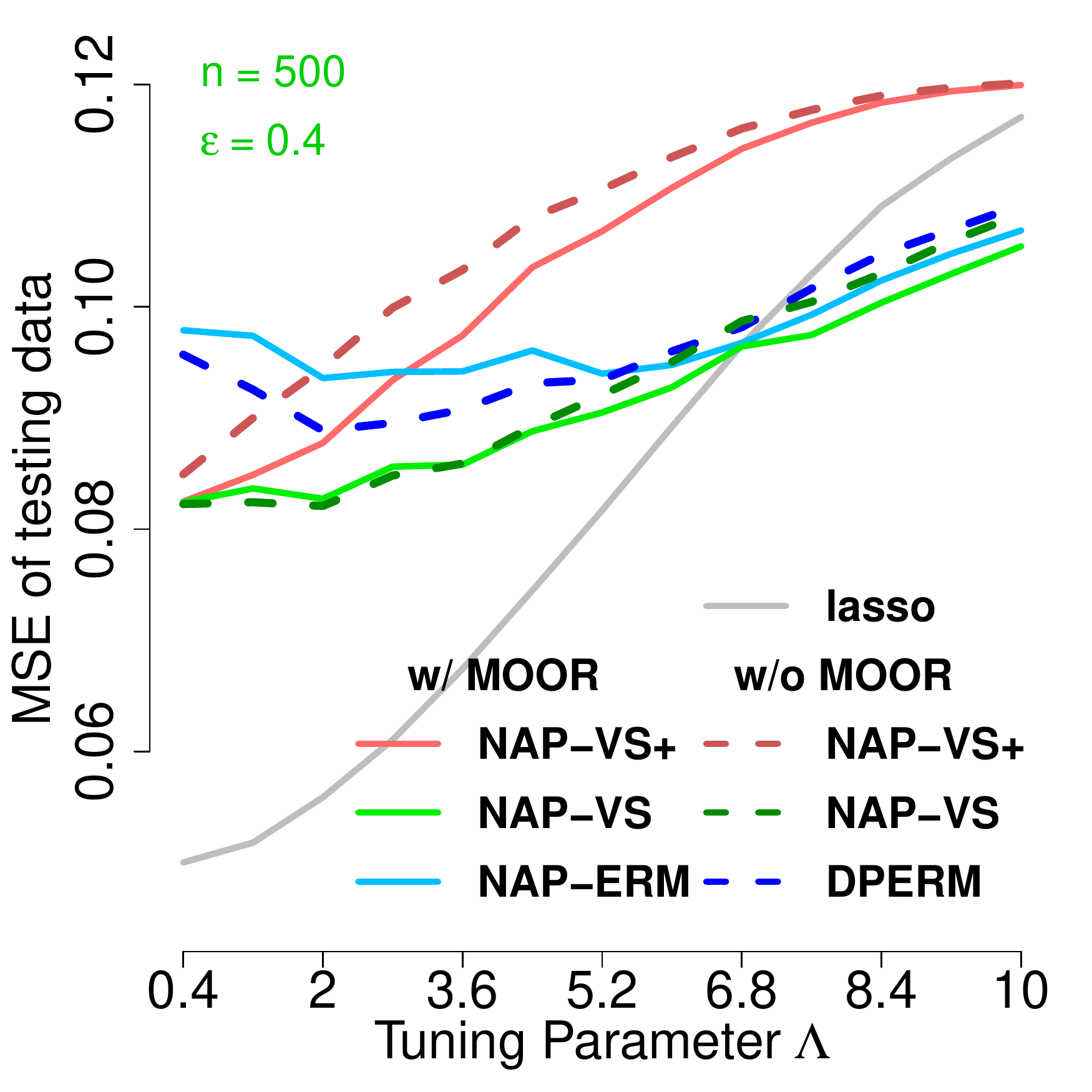}
\includegraphics[width=0.24\linewidth, trim=4pt 9pt 15pt 18pt,clip]{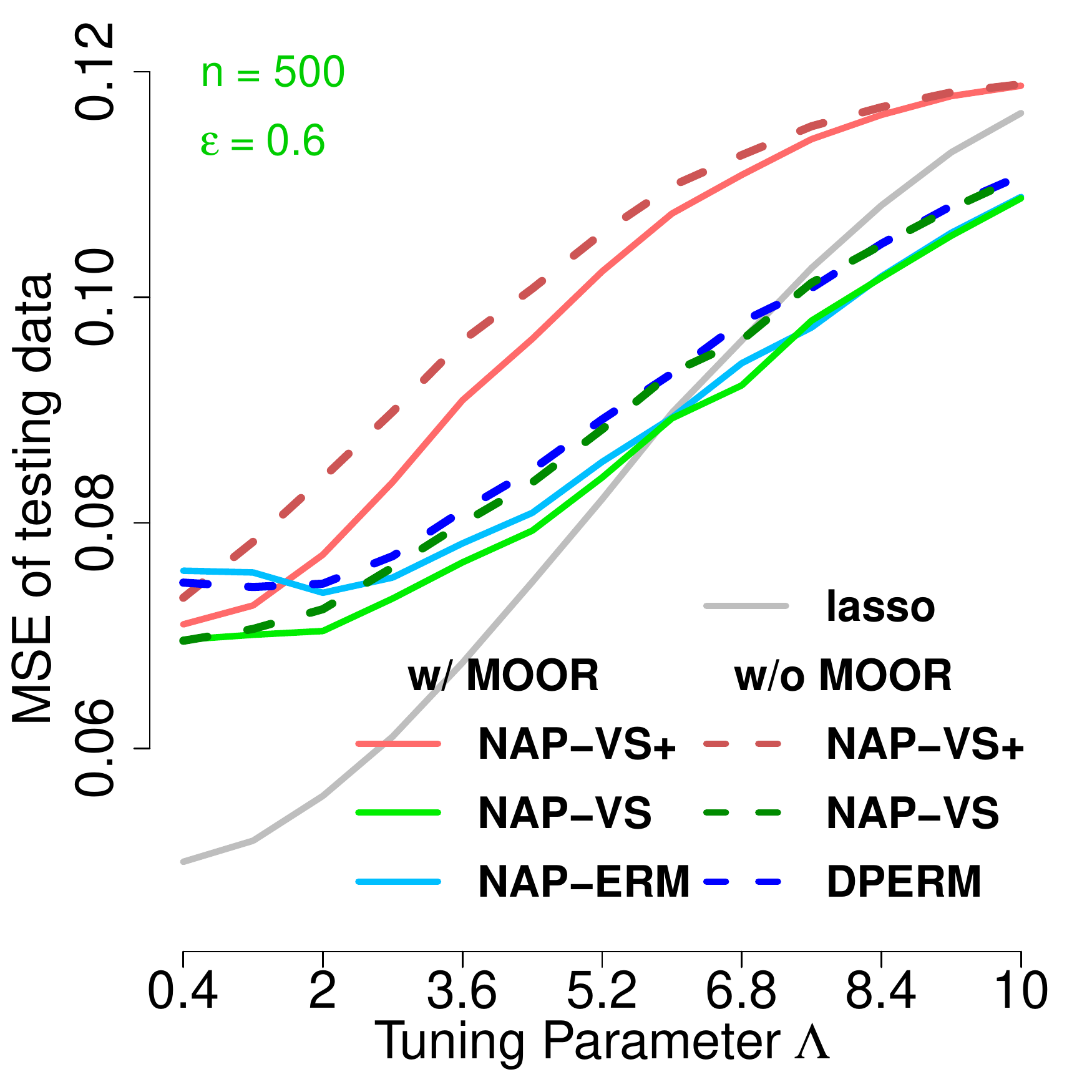}
\includegraphics[width=0.24\linewidth, trim=4pt 9pt 15pt 18pt,clip]{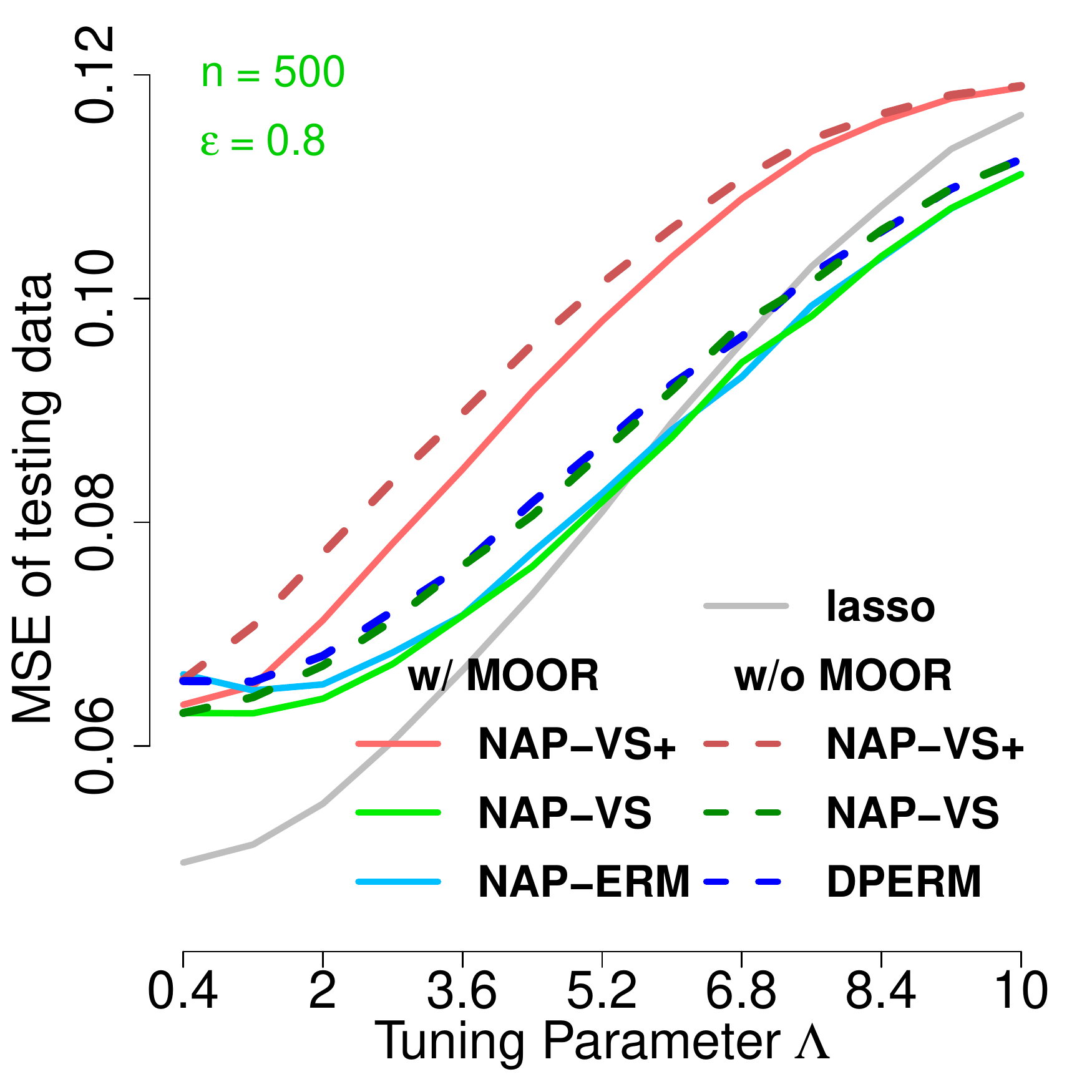}
\includegraphics[width=0.24\linewidth, trim=4pt 9pt 15pt 18pt,clip]{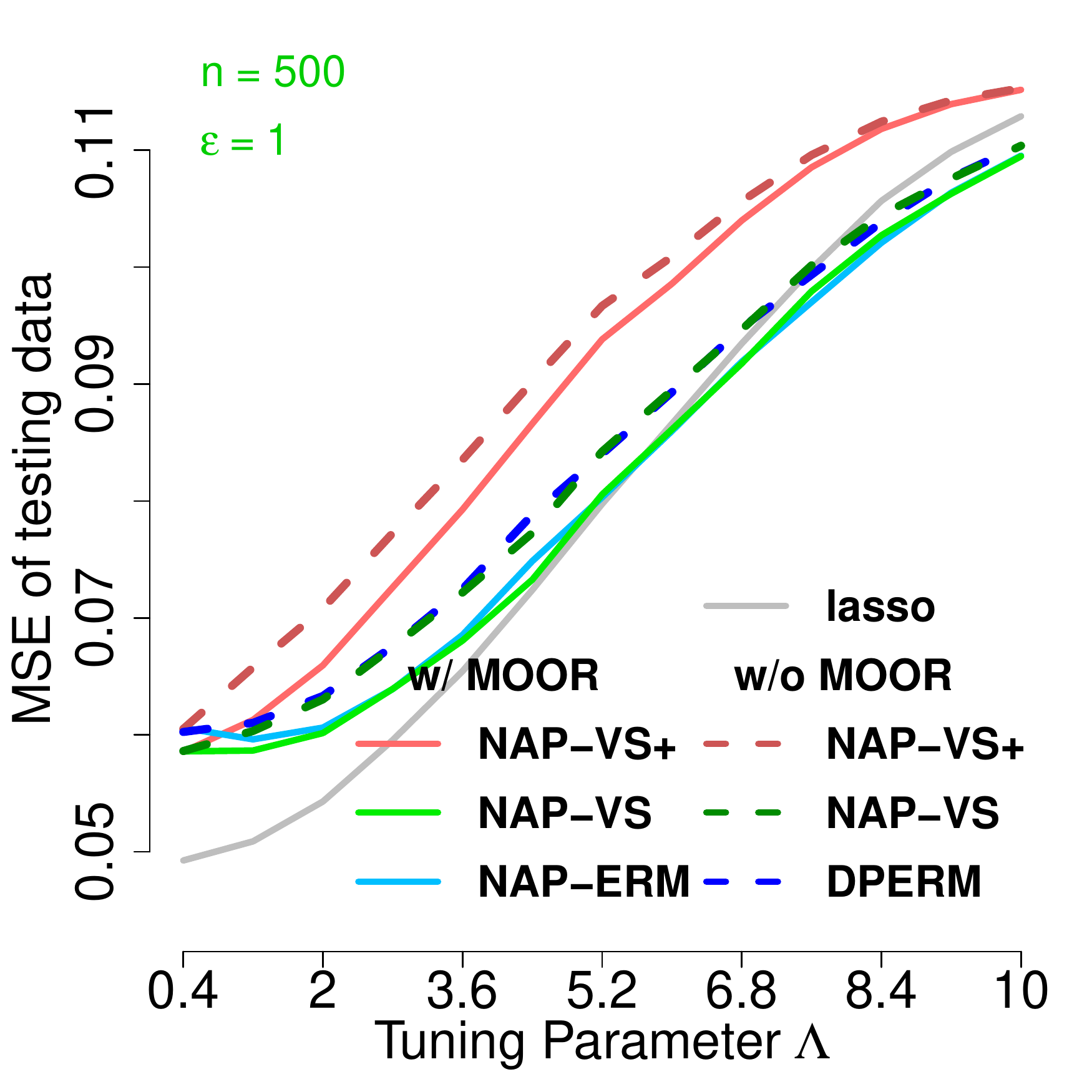}
\end{minipage}
\begin{minipage}{0.08\textwidth}
\footnotesize Poisson regression $n=500$ \end{minipage}
\begin{minipage}{0.92\textwidth}
\includegraphics[width=0.24\linewidth, trim=4pt 9pt 15pt 18pt,clip]{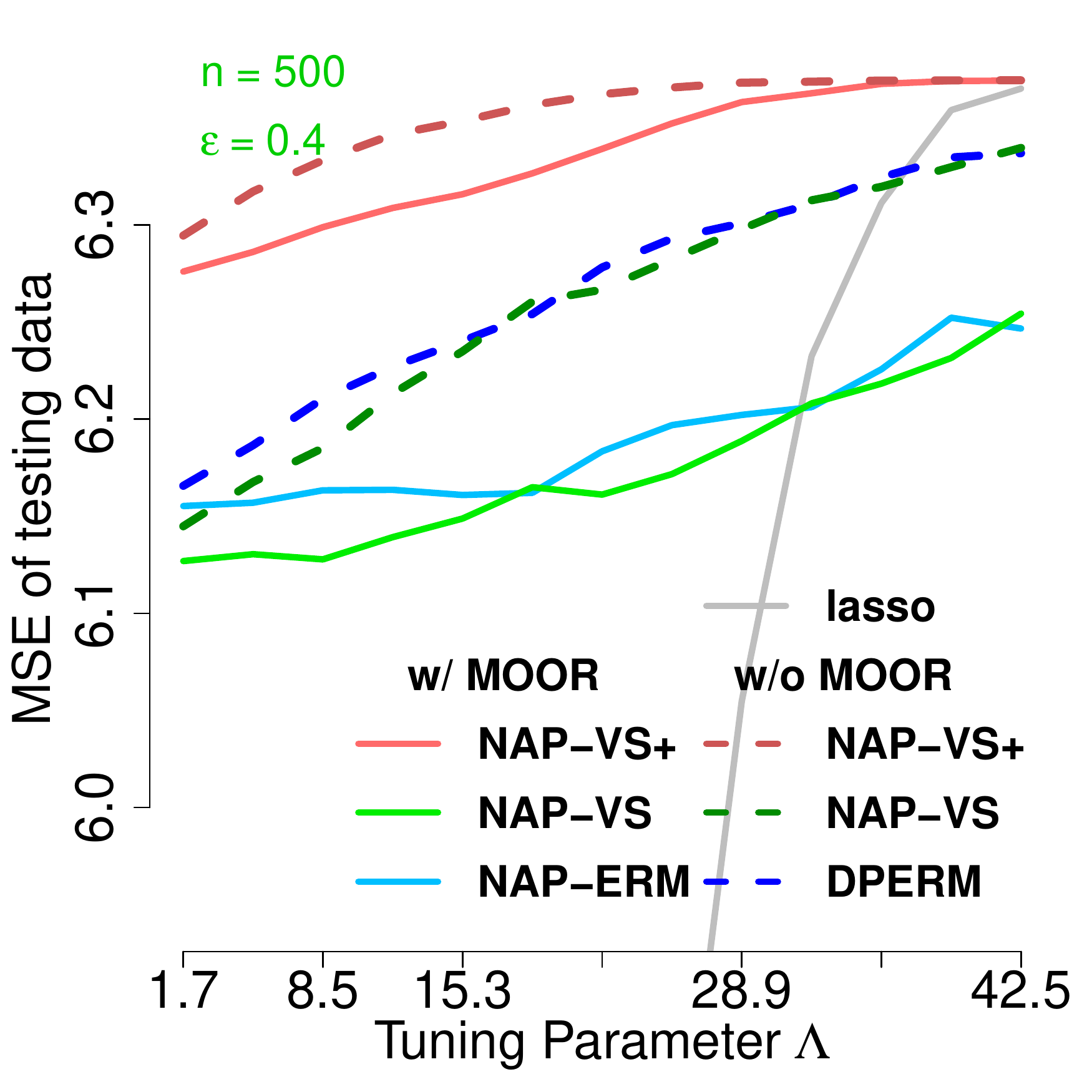}
\includegraphics[width=0.24\linewidth, trim=4pt 9pt 15pt 18pt,clip]{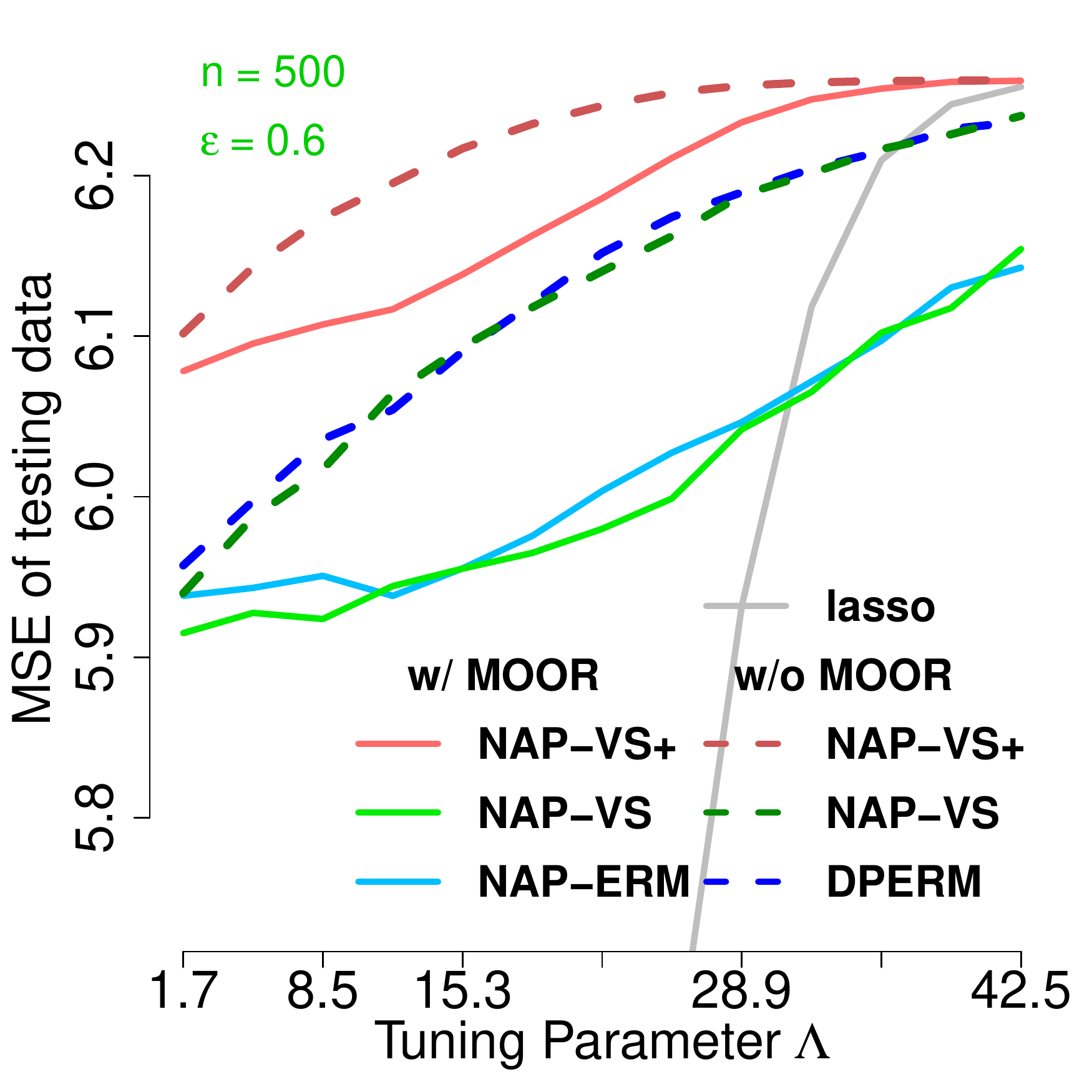}
\includegraphics[width=0.24\linewidth, trim=4pt 9pt 15pt 18pt,clip]{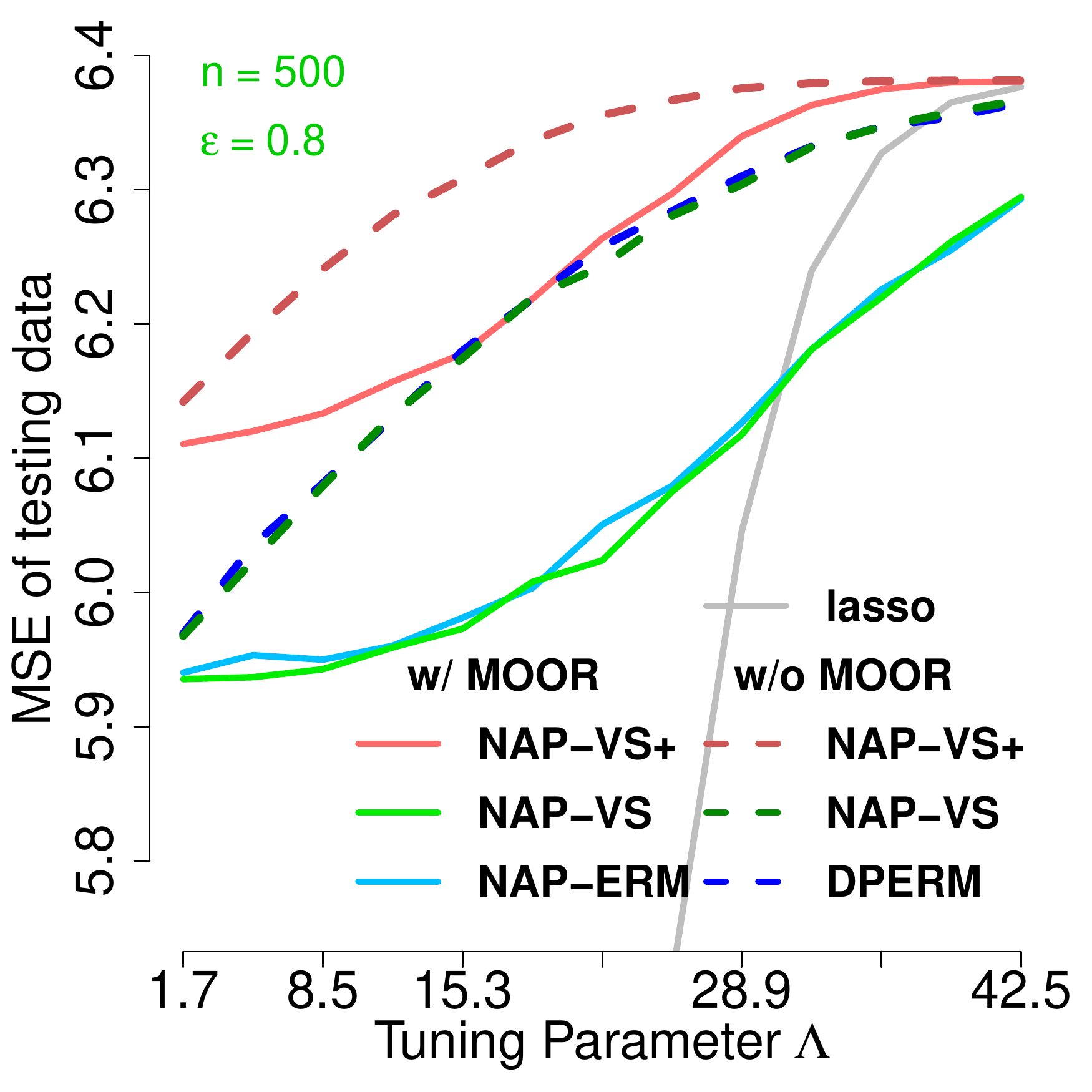}
\includegraphics[width=0.24\linewidth, trim=4pt 9pt 15pt 18pt,clip]{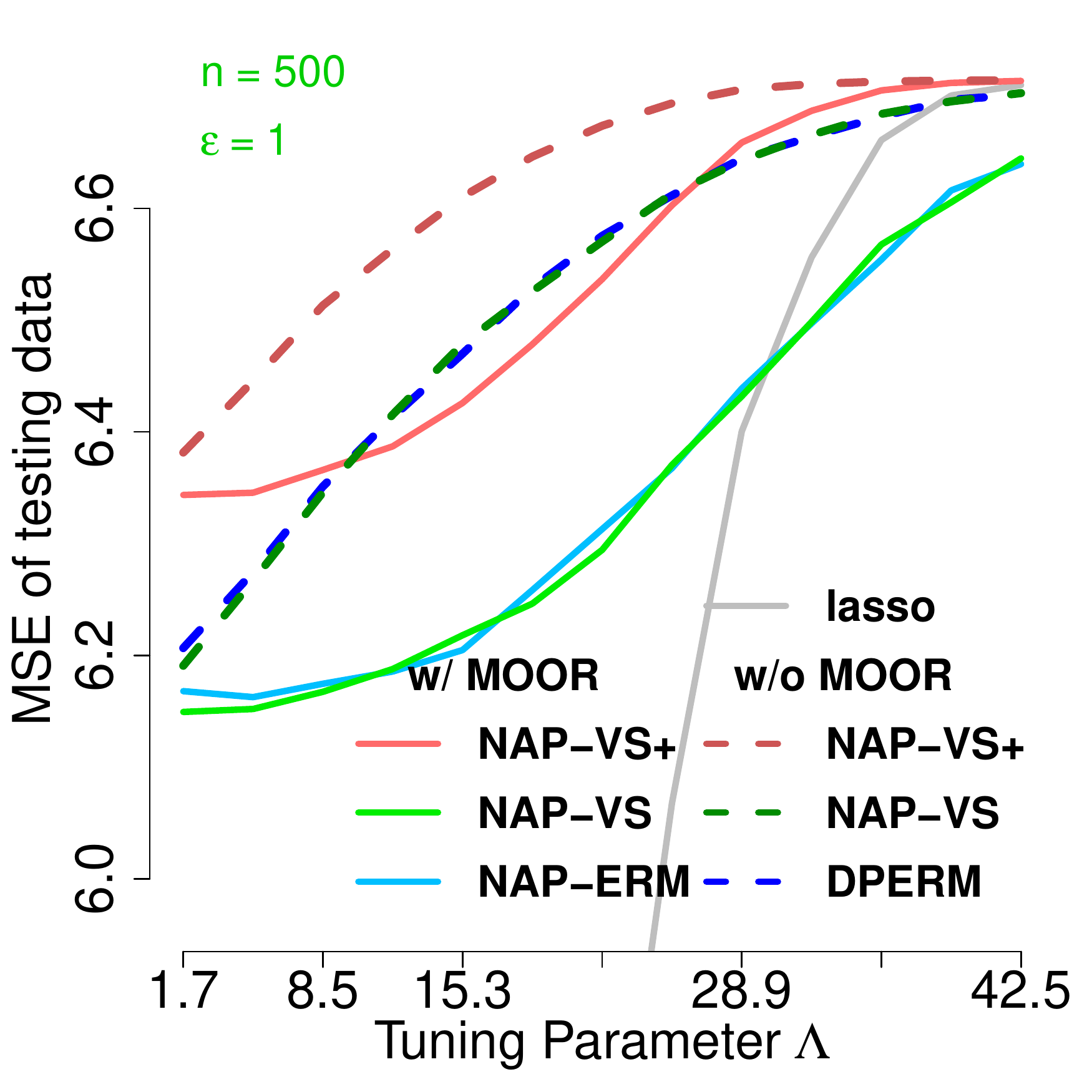}
\end{minipage}
\begin{minipage}{0.08\textwidth}\footnotesize $n=1000$ \end{minipage}
\begin{minipage}{0.92\textwidth}
\includegraphics[width=0.24\linewidth, trim=4pt 9pt 15pt 18pt,clip]{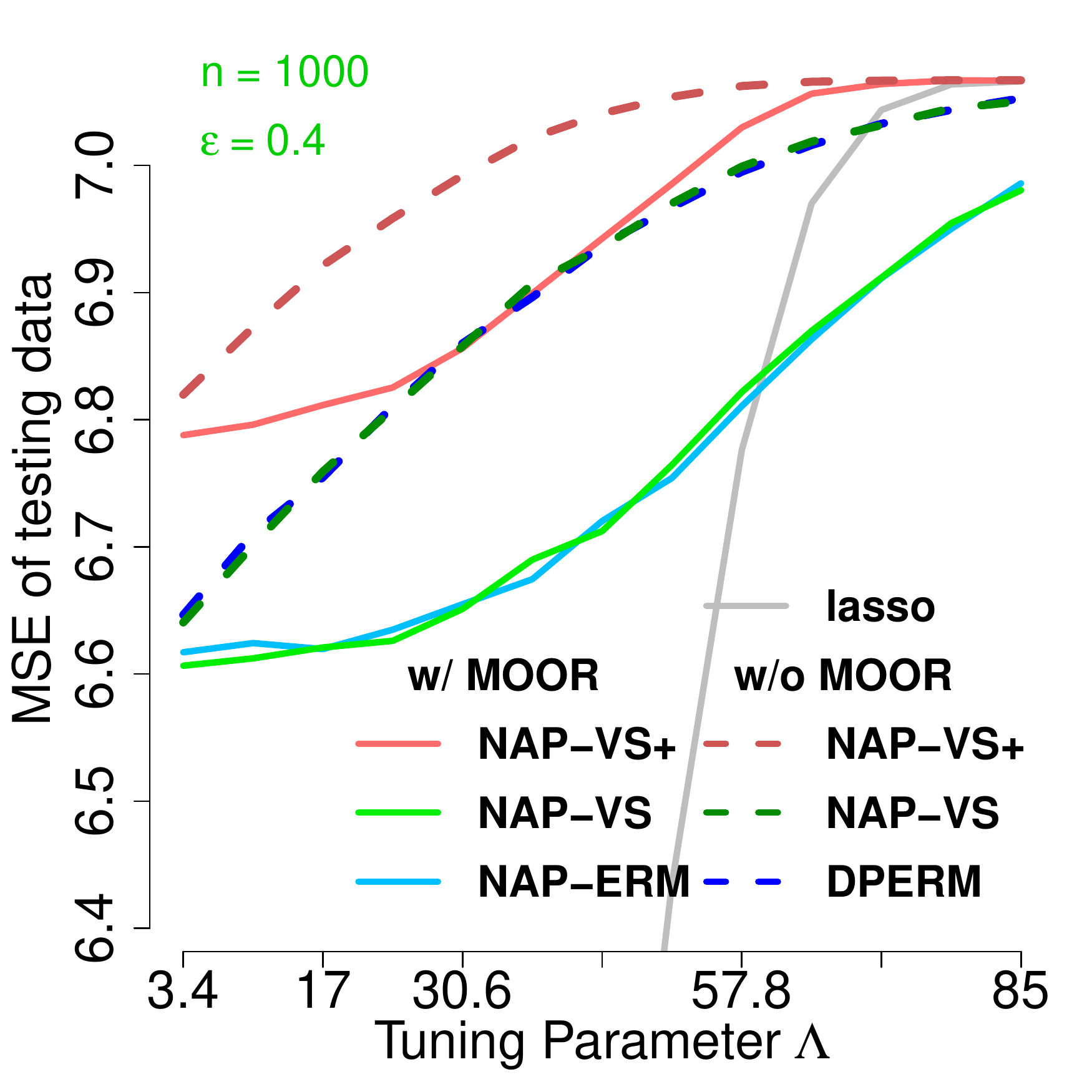}
\includegraphics[width=0.24\linewidth, trim=4pt 9pt 15pt 18pt,clip]{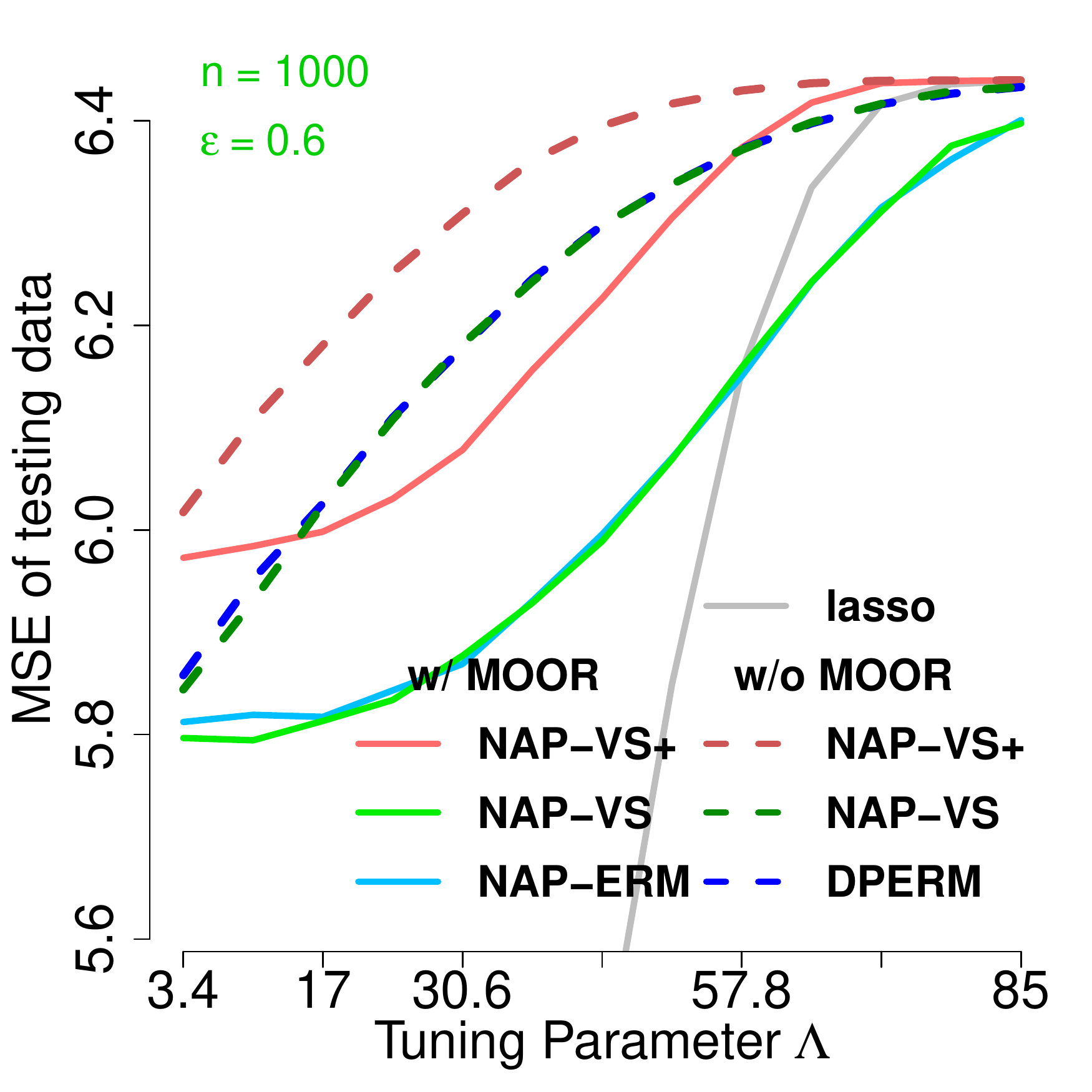}
\includegraphics[width=0.24\linewidth, trim=4pt 9pt 15pt 18pt,clip]{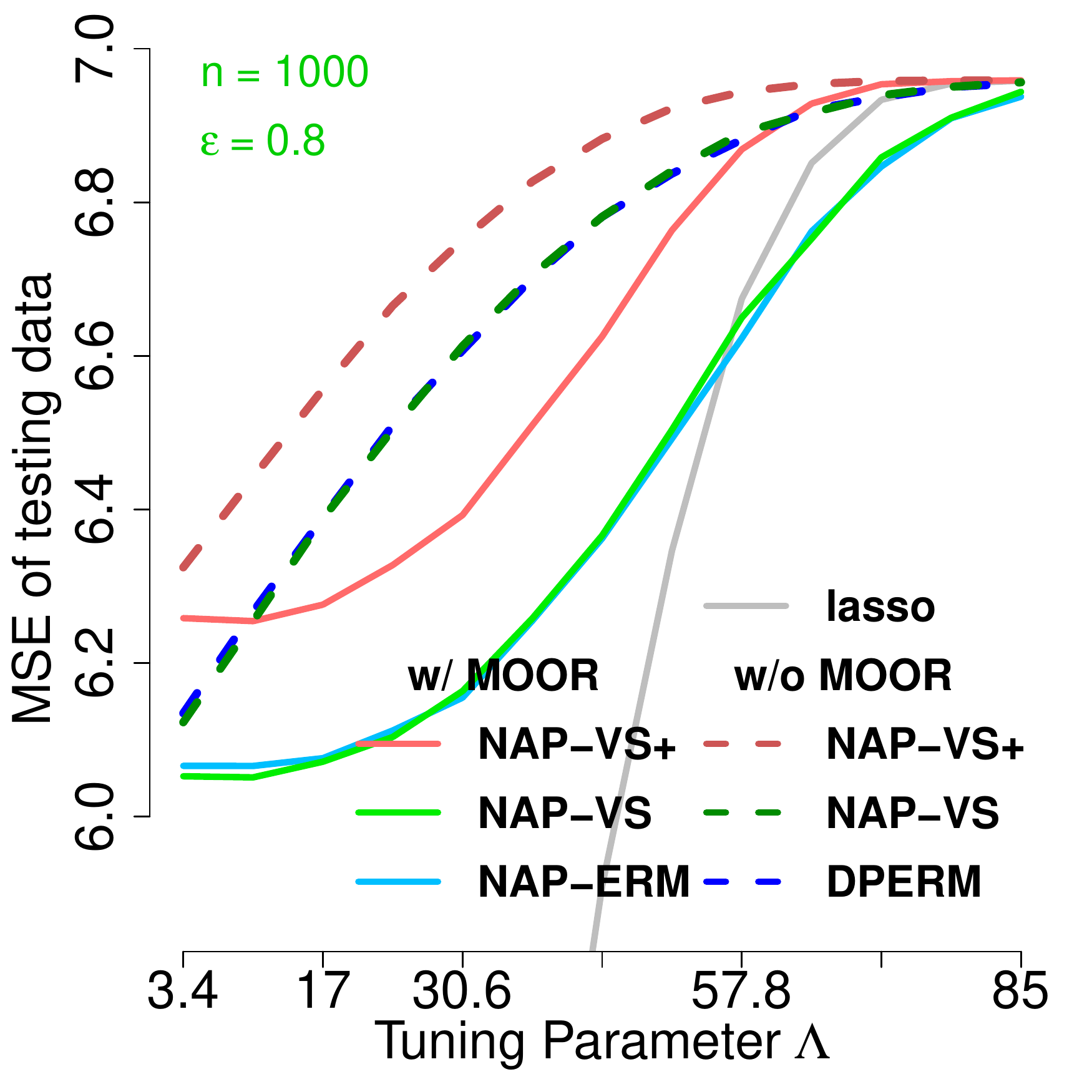}
\includegraphics[width=0.24\linewidth, trim=4pt 9pt 15pt 18pt,clip]{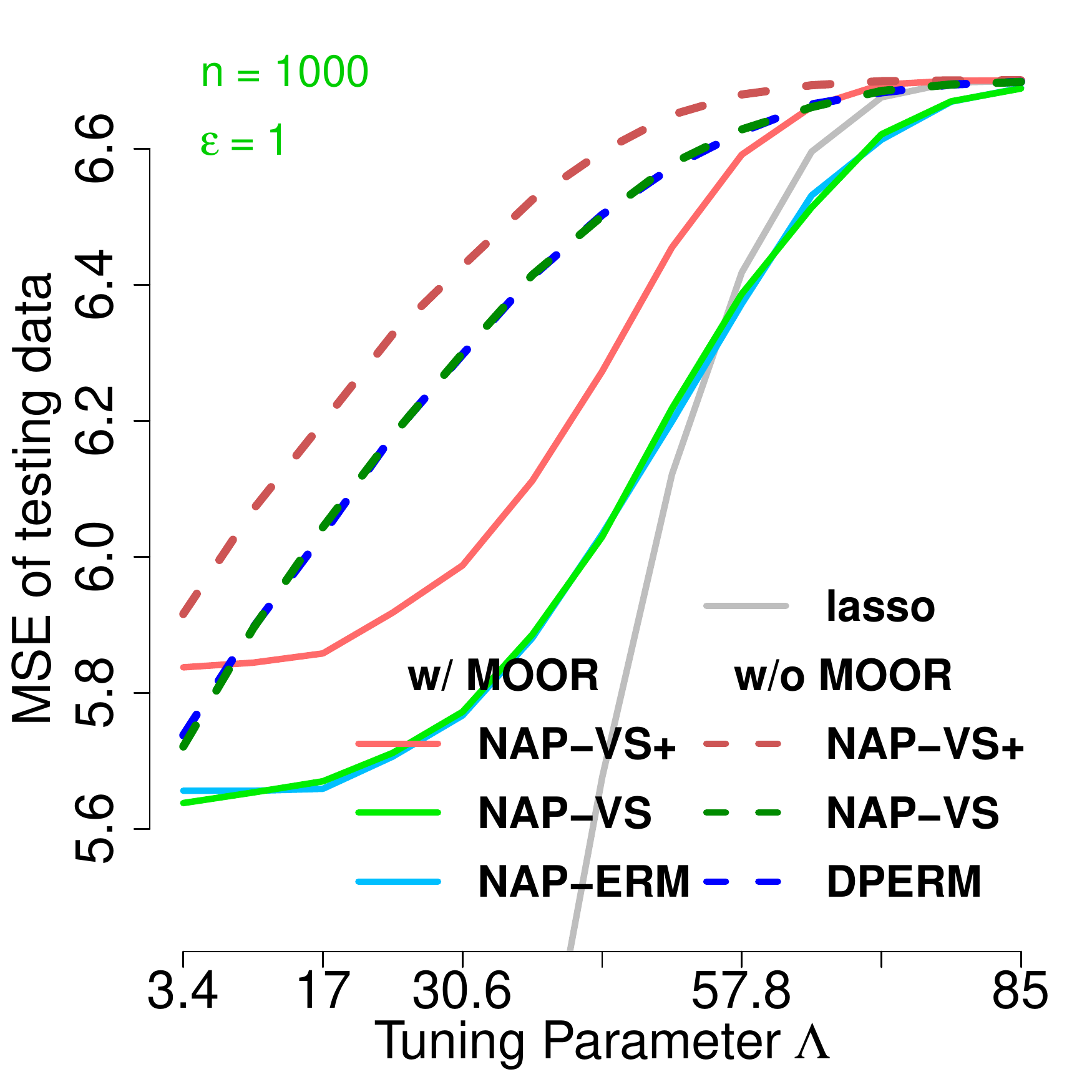}
\end{minipage}
\begin{minipage}{0.08\textwidth}
\footnotesize logistic regression $n=500$ \end{minipage}
\begin{minipage}{0.92\textwidth}
\includegraphics[width=0.24\linewidth, trim=4pt 9pt 15pt 18pt,clip]{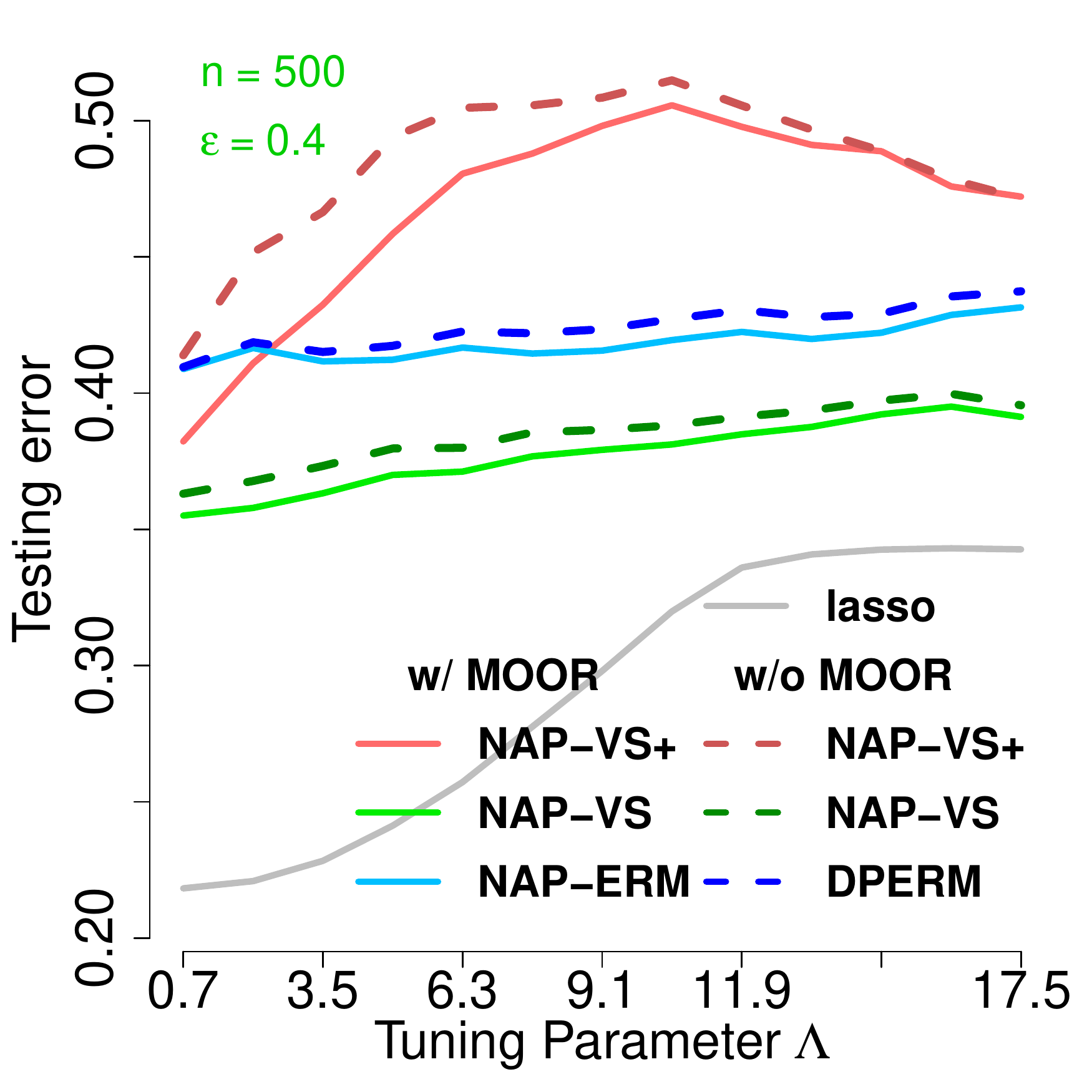}
\includegraphics[width=0.24\linewidth, trim=4pt 9pt 15pt 18pt,clip]{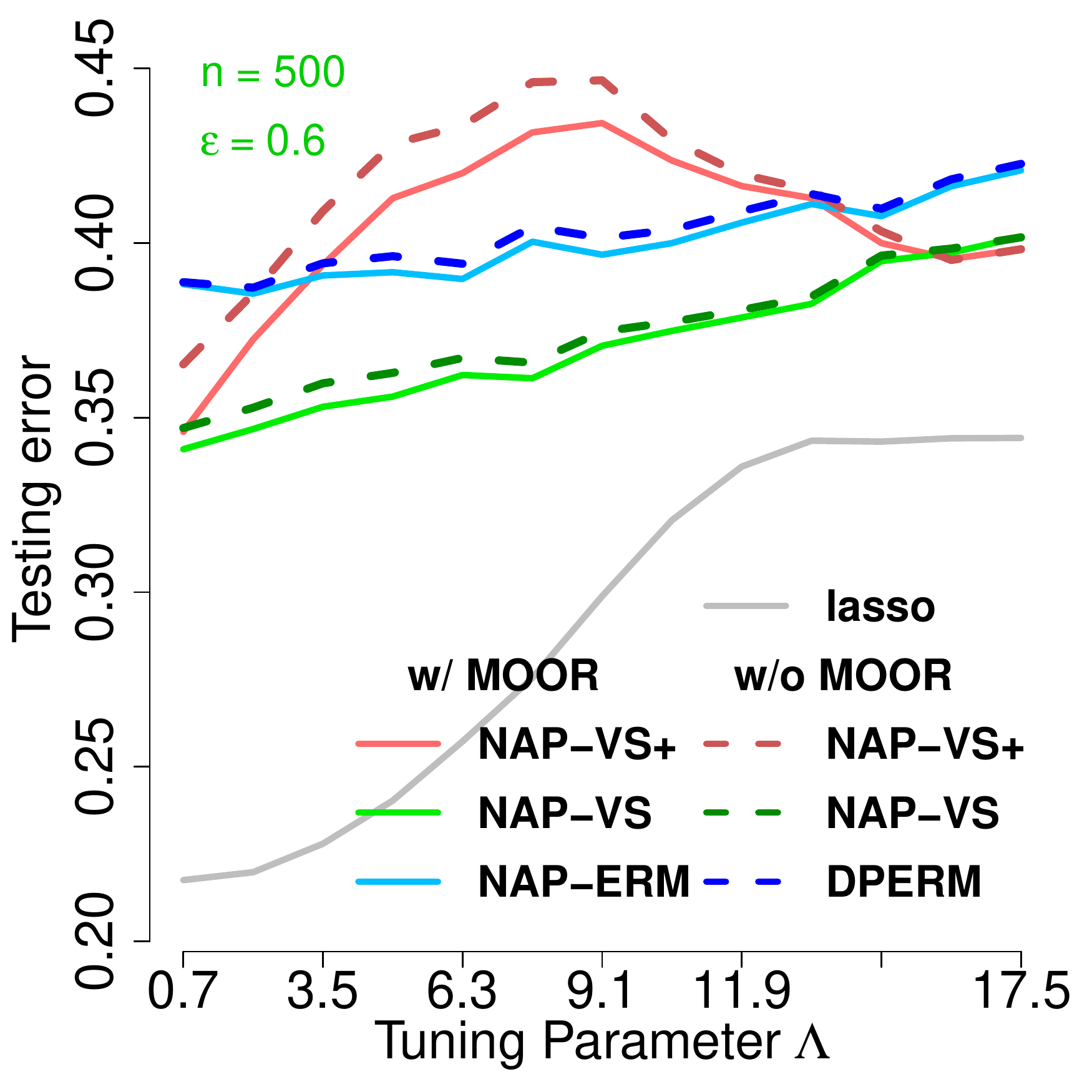}
\includegraphics[width=0.24\linewidth, trim=4pt 9pt 15pt 18pt,clip]{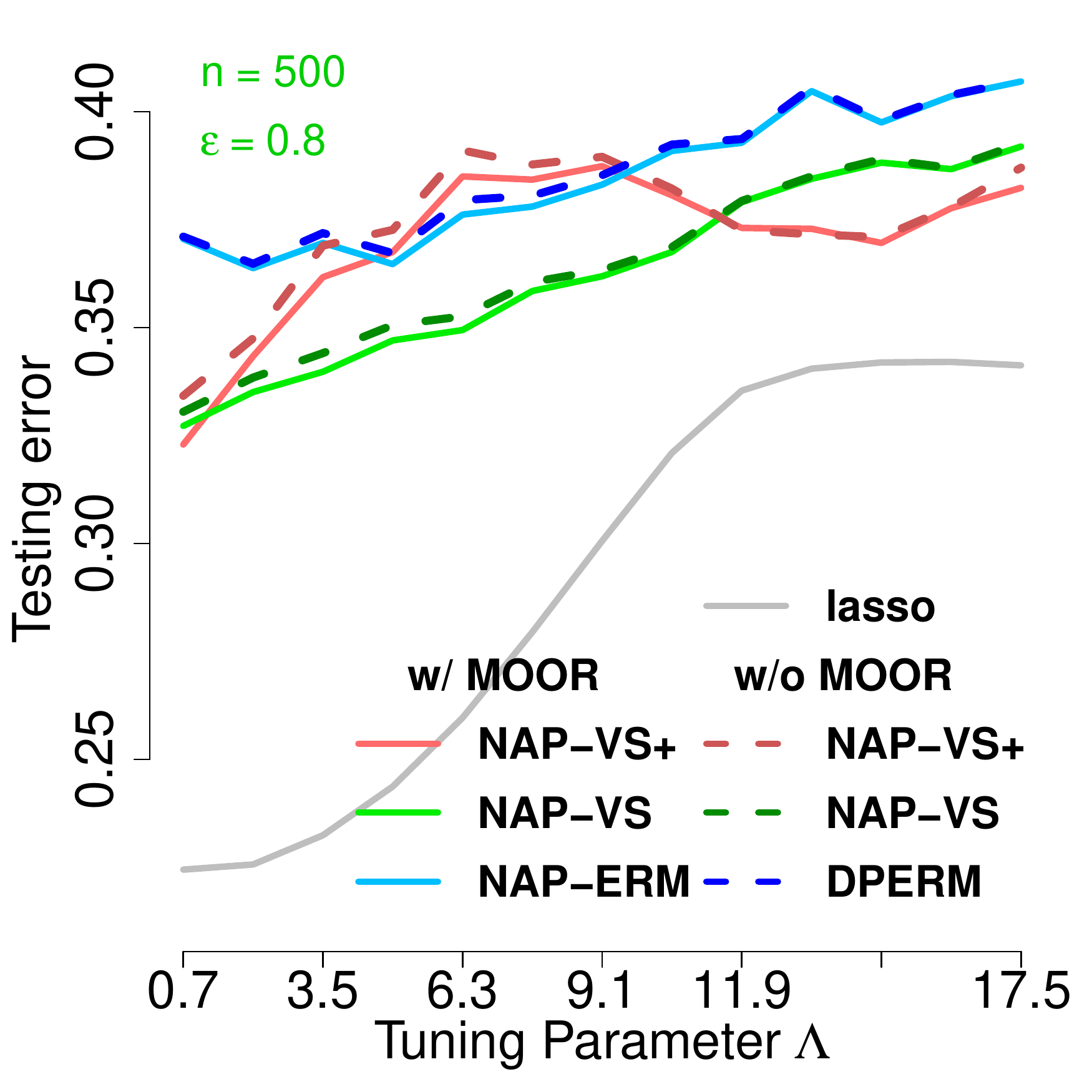}
\includegraphics[width=0.24\linewidth, trim=4pt 9pt 15pt 18pt,clip]{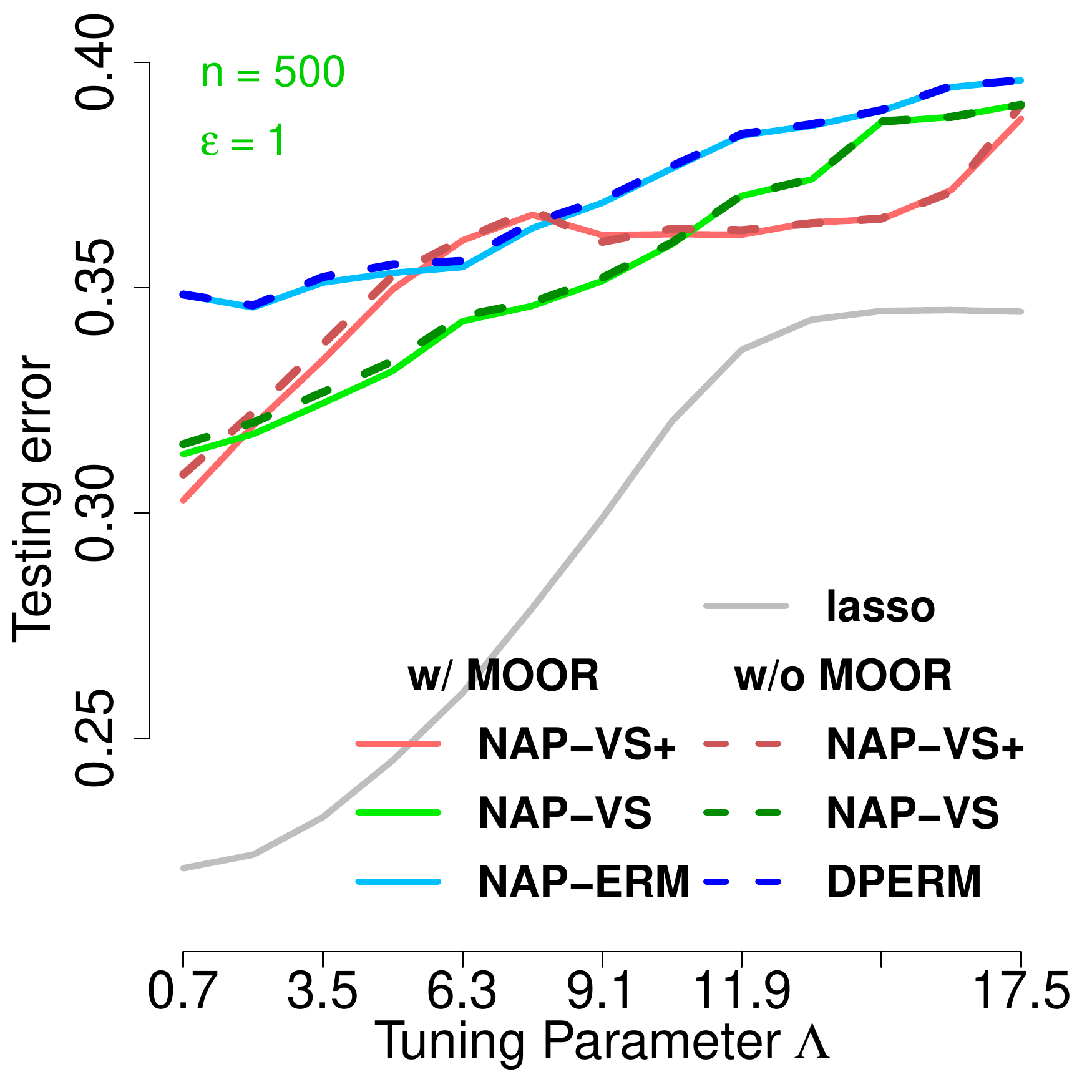}
\end{minipage}
\begin{minipage}{0.08\textwidth}\footnotesize $n=1000$ \end{minipage}
\begin{minipage}{0.92\textwidth}
\includegraphics[width=0.24\linewidth, trim=4pt 9pt 15pt 18pt,clip]{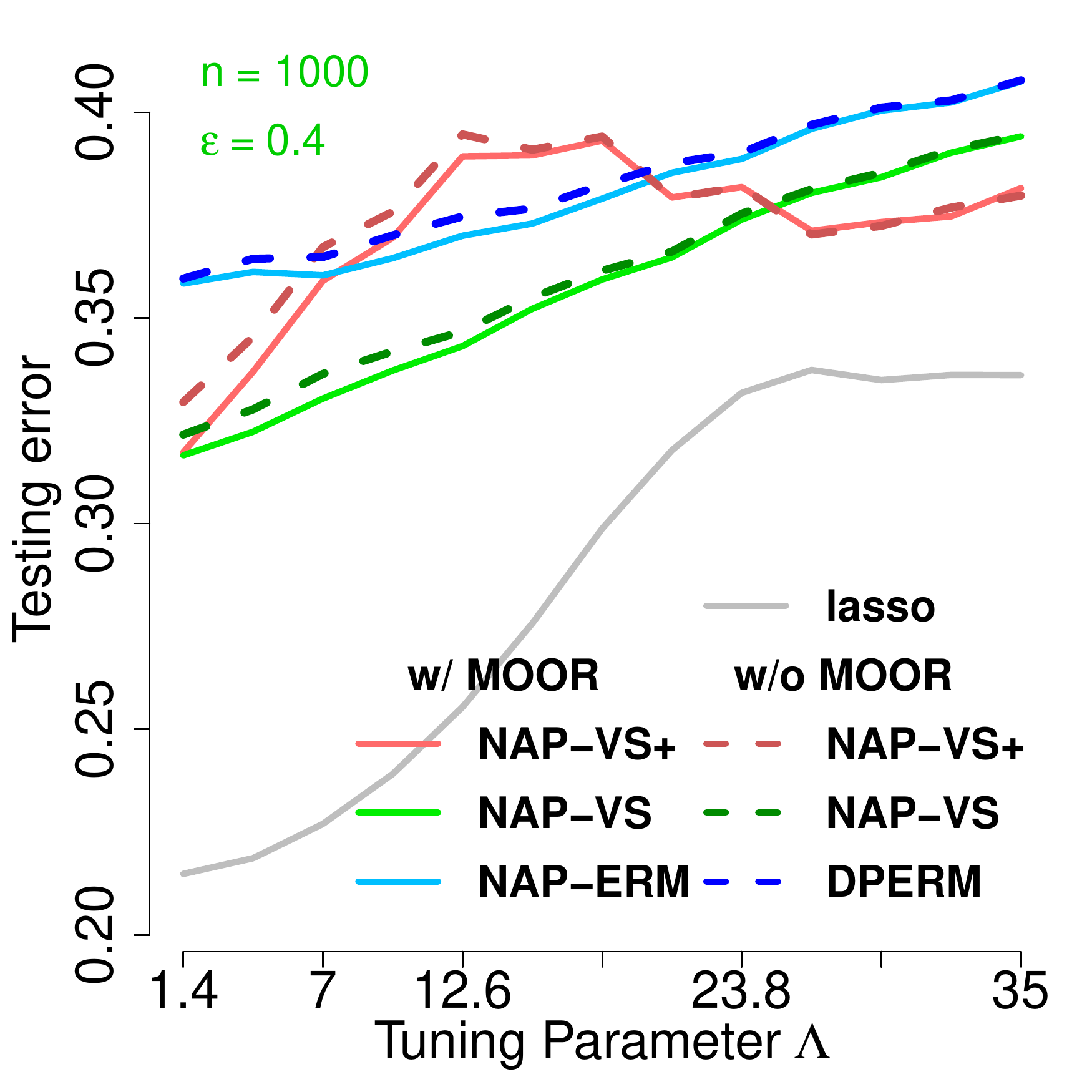}
\includegraphics[width=0.24\linewidth, trim=4pt 9pt 15pt 18pt,clip]{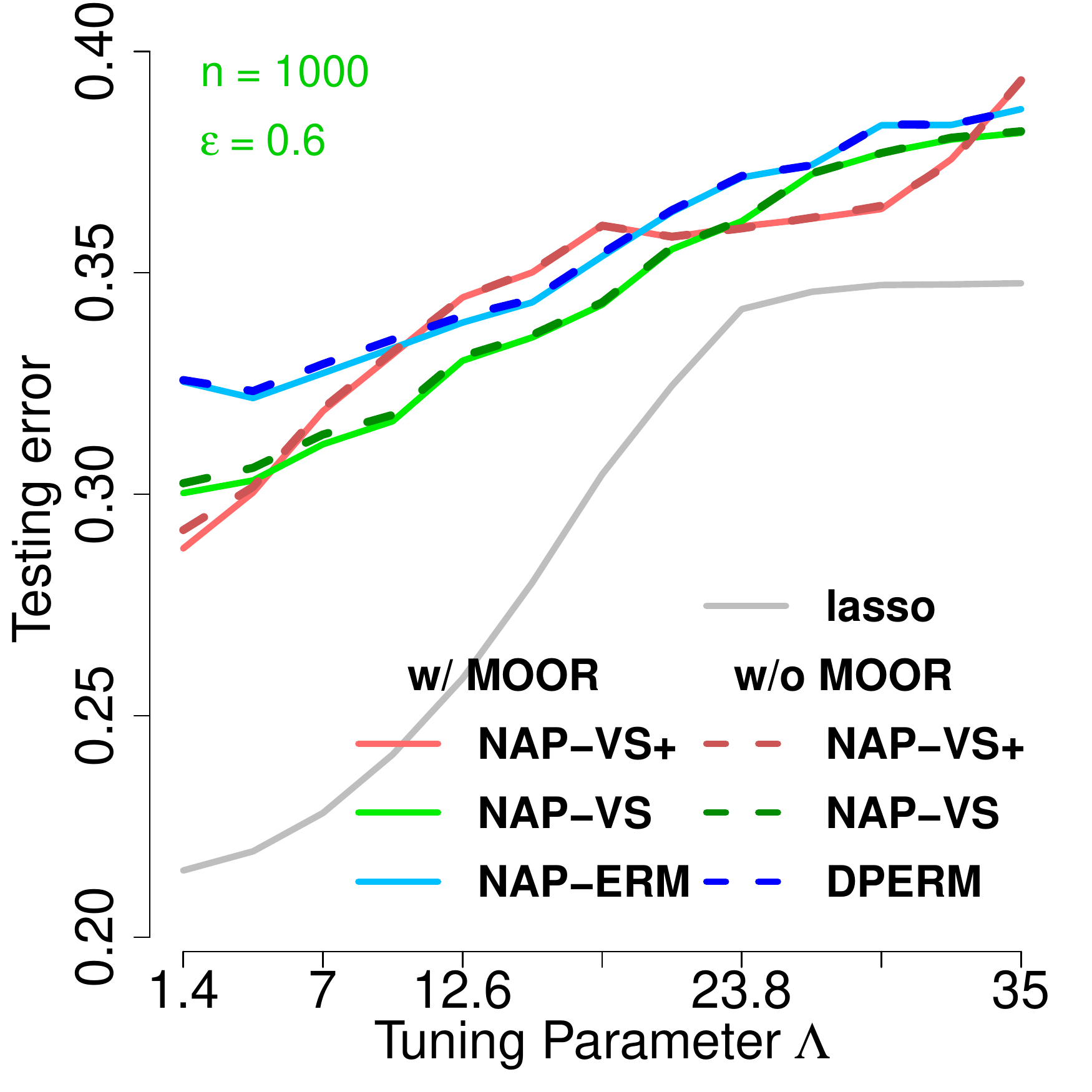}
\includegraphics[width=0.24\linewidth, trim=4pt 9pt 15pt 18pt,clip]{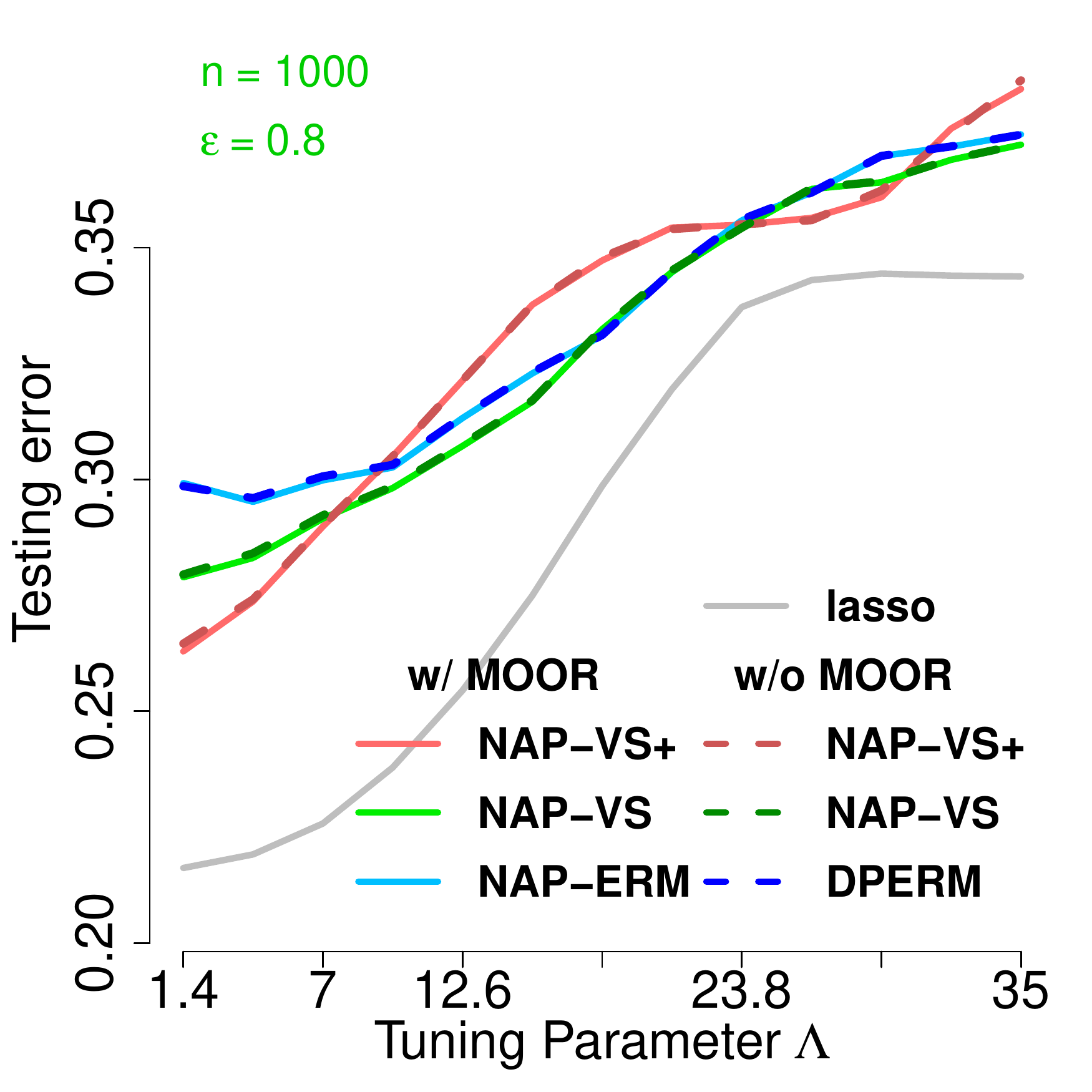}
\includegraphics[width=0.24\linewidth, trim=4pt 9pt 15pt 18pt,clip]{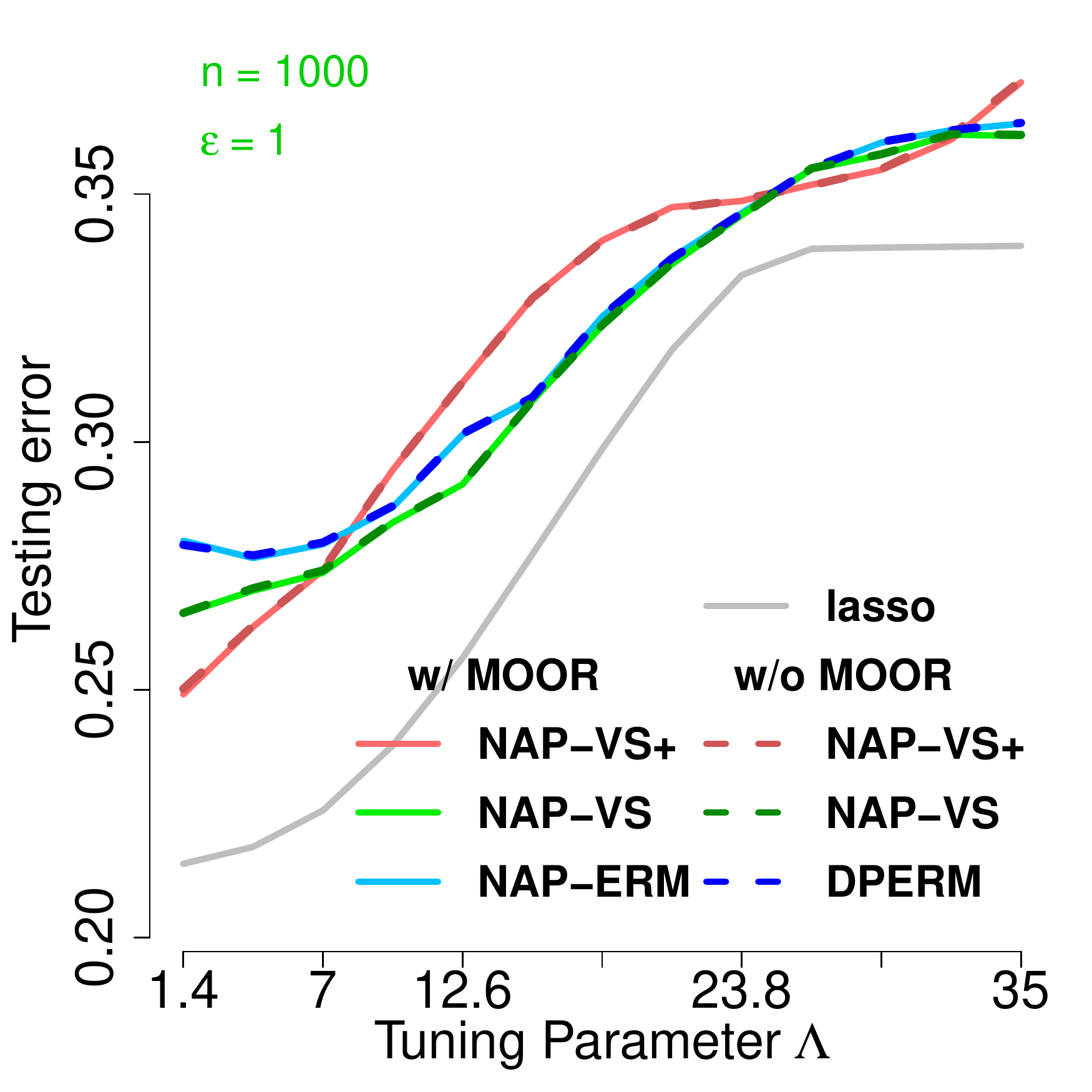}
\end{minipage}\vspace{-9pt}
\caption{Testing data prediction error (MSE in linear and Poisson regression and misclassification rate in logistic regression) at different tuning parameter $\Lambda$, sample size $n$, and privacy budget $\epsilon$ ($\delta=0.0001$)}.
\label{fig:SIM.prediction1}
\end{figure}

\begin{figure}[H]
\begin{minipage}{0.08\textwidth}
\footnotesize linear regression $n=200$ \end{minipage}
\begin{minipage}{0.9\textwidth}
\includegraphics[width=0.24\linewidth, trim=4pt 9pt 15pt 18pt,clip]{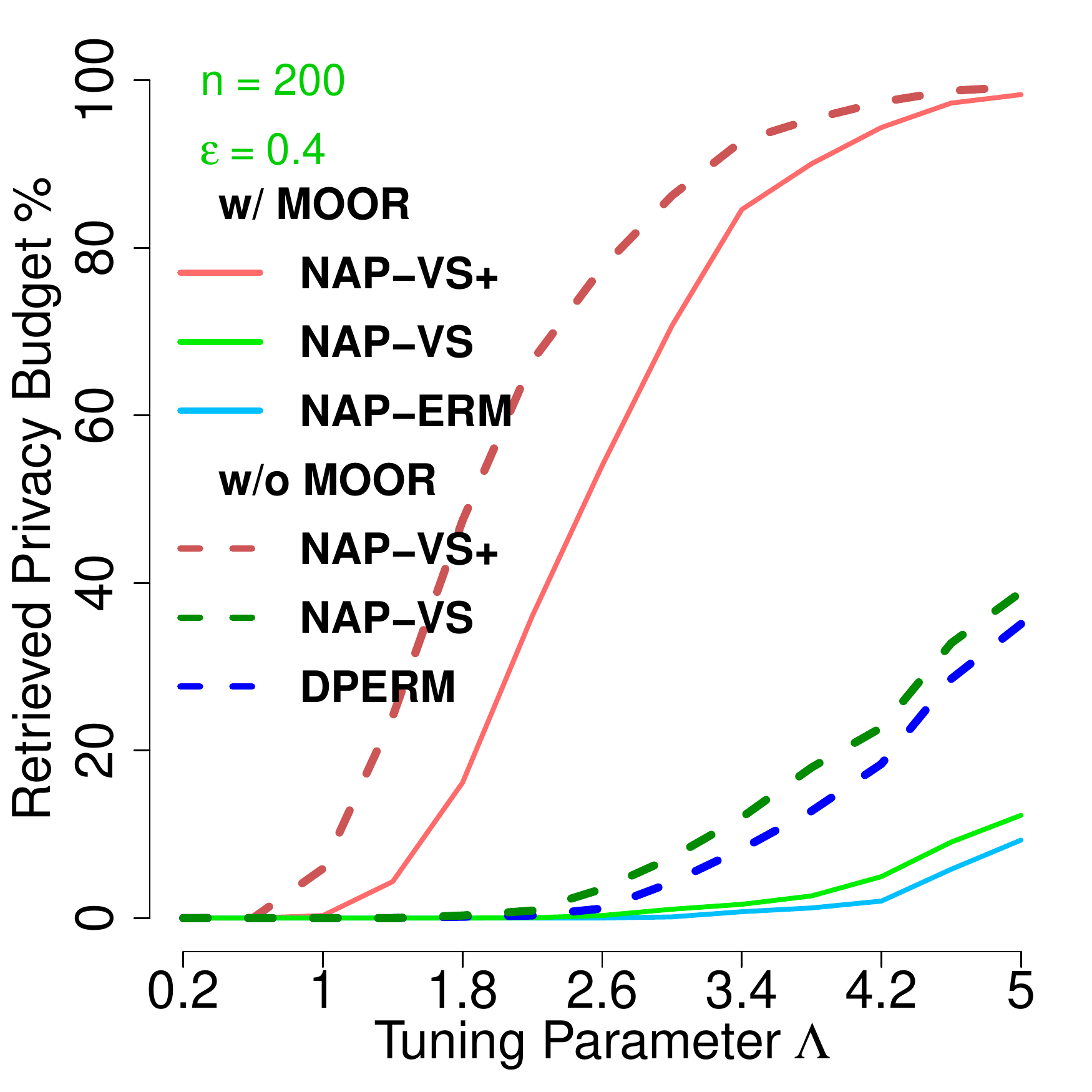}
\includegraphics[width=0.24\linewidth, trim=4pt 9pt 15pt 18pt,clip]{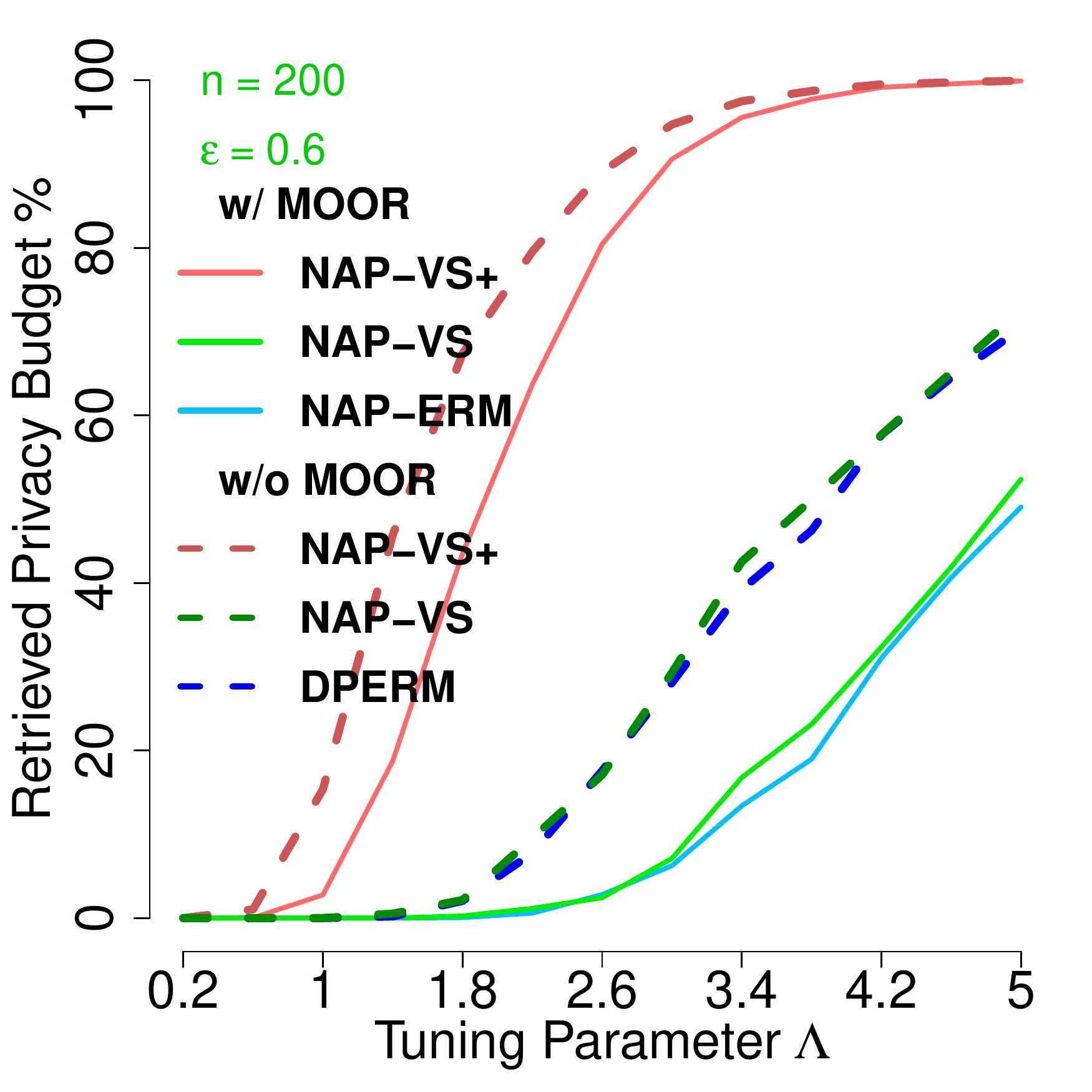}
\includegraphics[width=0.24\linewidth, trim=4pt 9pt 15pt 18pt,clip]{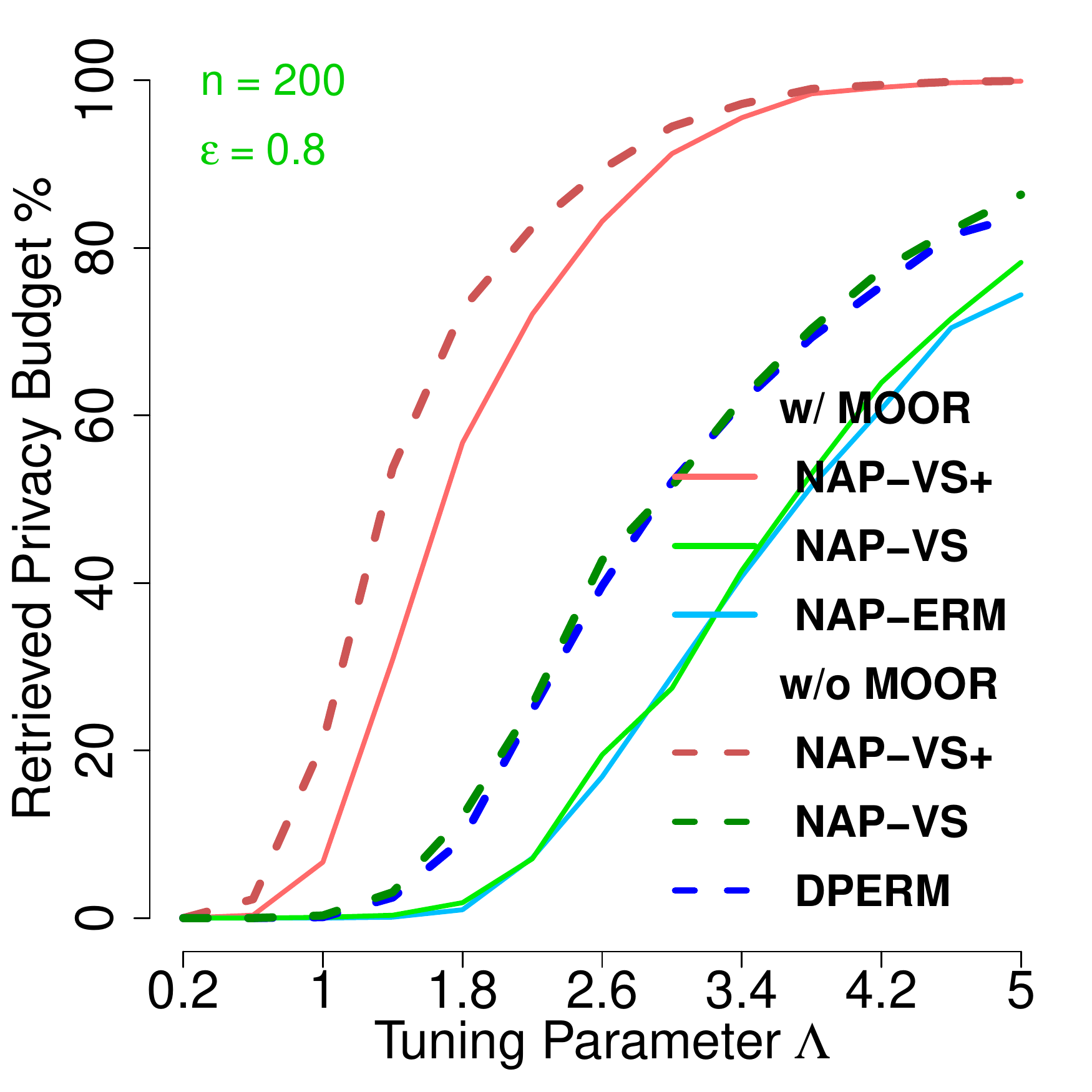}
\includegraphics[width=0.24\linewidth, trim=4pt 9pt 15pt 18pt,clip]{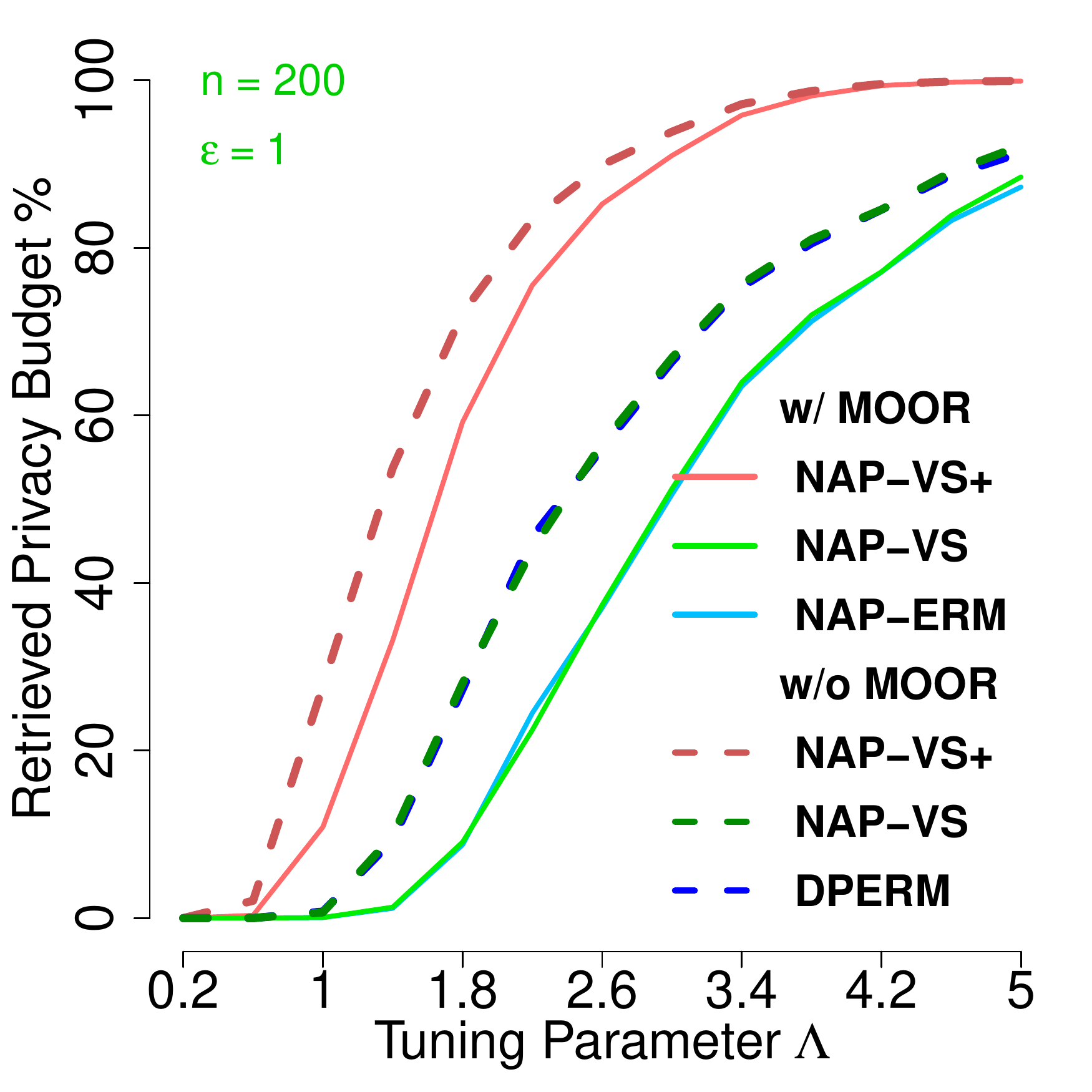}
\end{minipage}

\begin{minipage}{0.08\textwidth}\footnotesize $n=500$ \end{minipage}
\begin{minipage}{0.91\textwidth}
\includegraphics[width=0.24\linewidth, trim=4pt 9pt 15pt 18pt,clip]{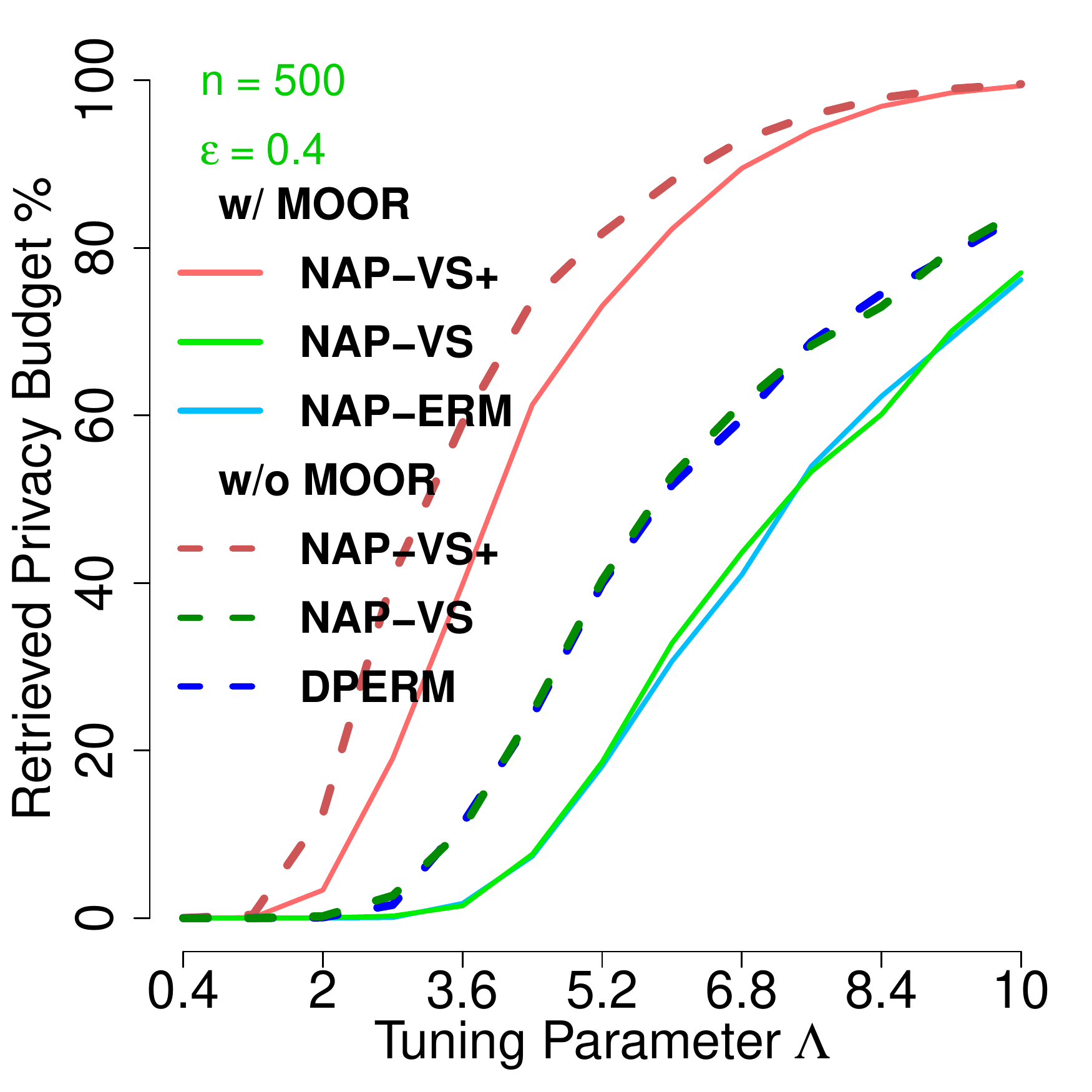}
\includegraphics[width=0.24\linewidth, trim=4pt 9pt 15pt 18pt,clip]{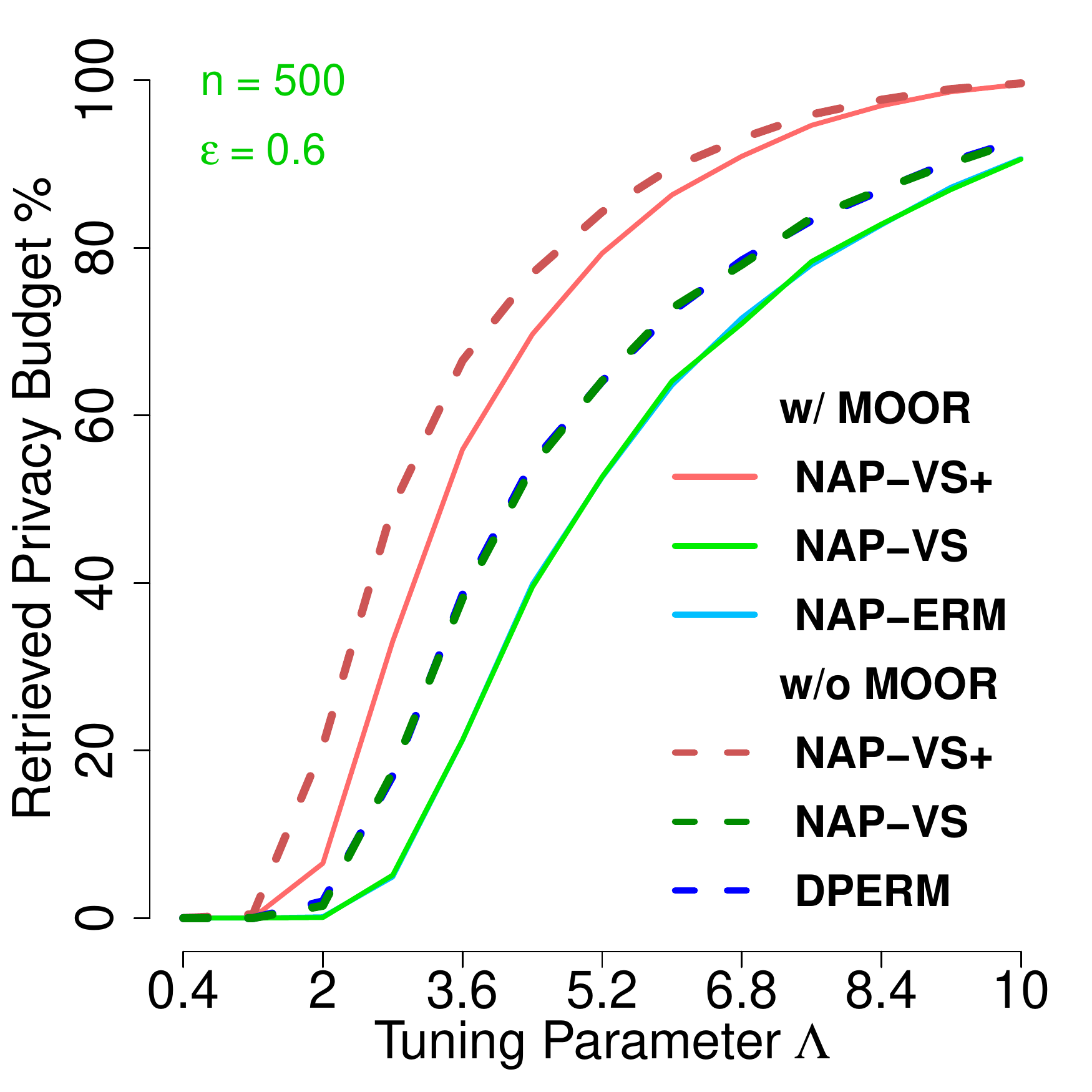}
\includegraphics[width=0.24\linewidth, trim=4pt 9pt 15pt 18pt,clip]{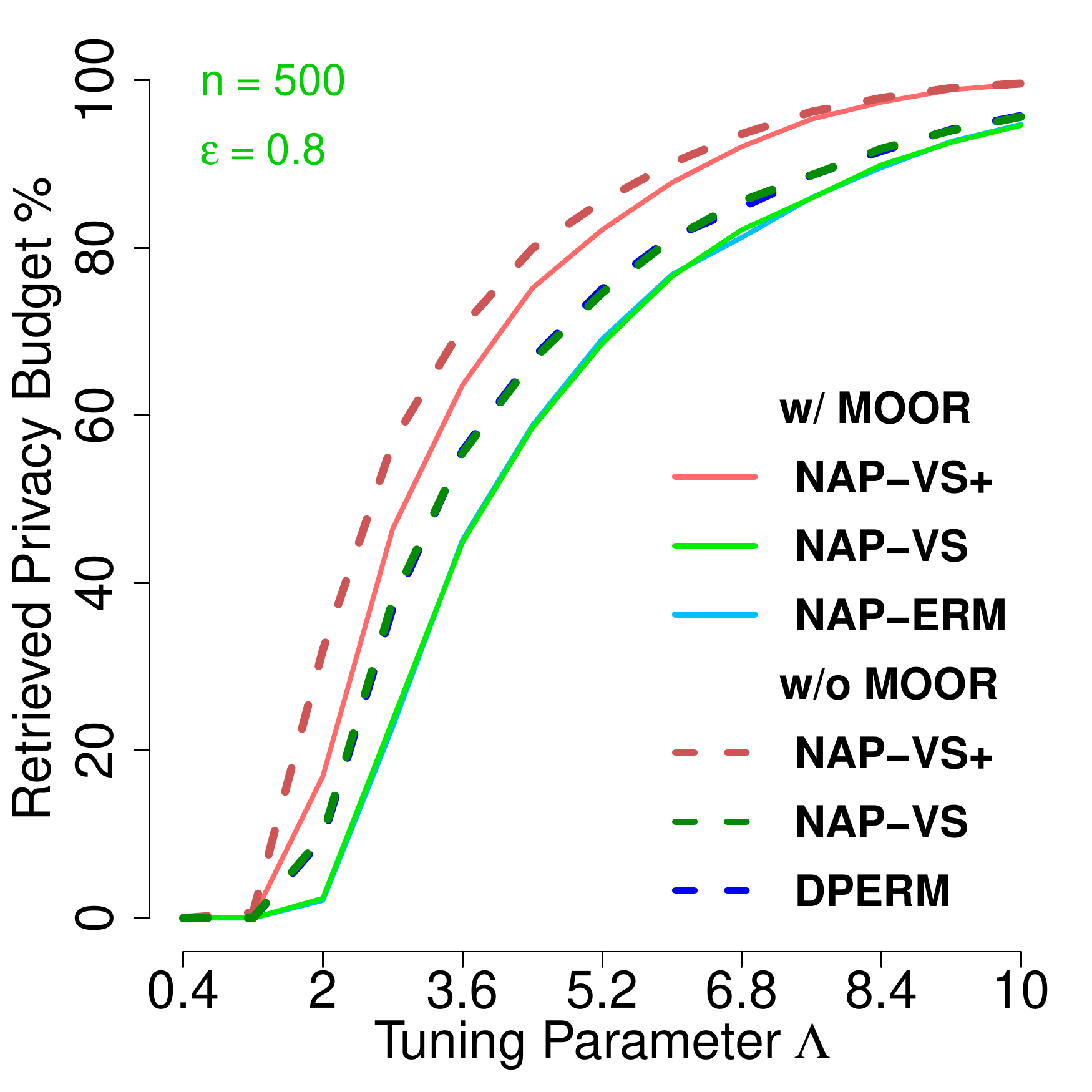}
\includegraphics[width=0.24\linewidth, trim=4pt 9pt 15pt 18pt,clip]{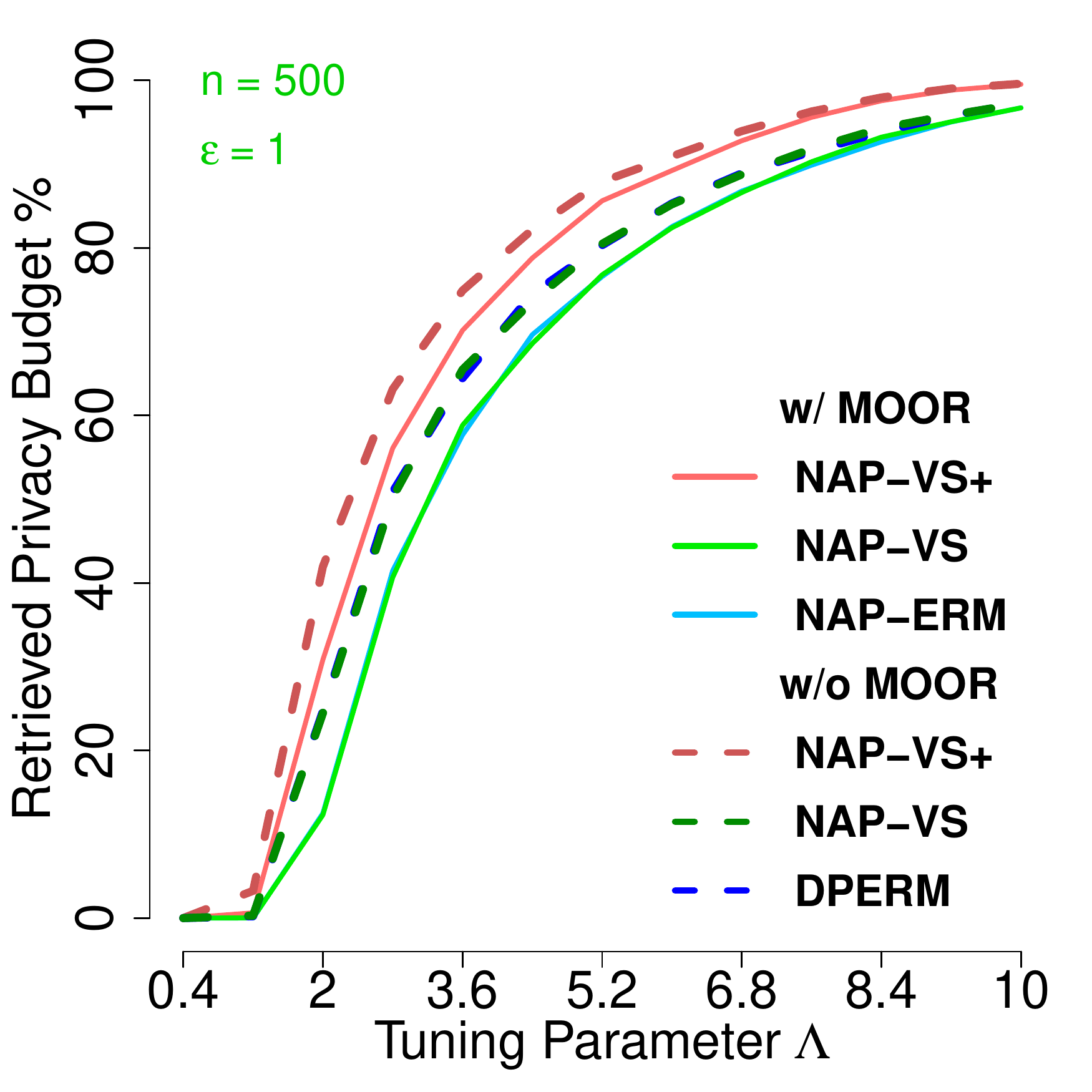}
\end{minipage}

\begin{minipage}{0.08\textwidth}
\footnotesize Poisson regression $n=500$ \end{minipage}
\begin{minipage}{0.91\textwidth}
\includegraphics[width=0.24\linewidth, trim=4pt 9pt 15pt 18pt,clip]{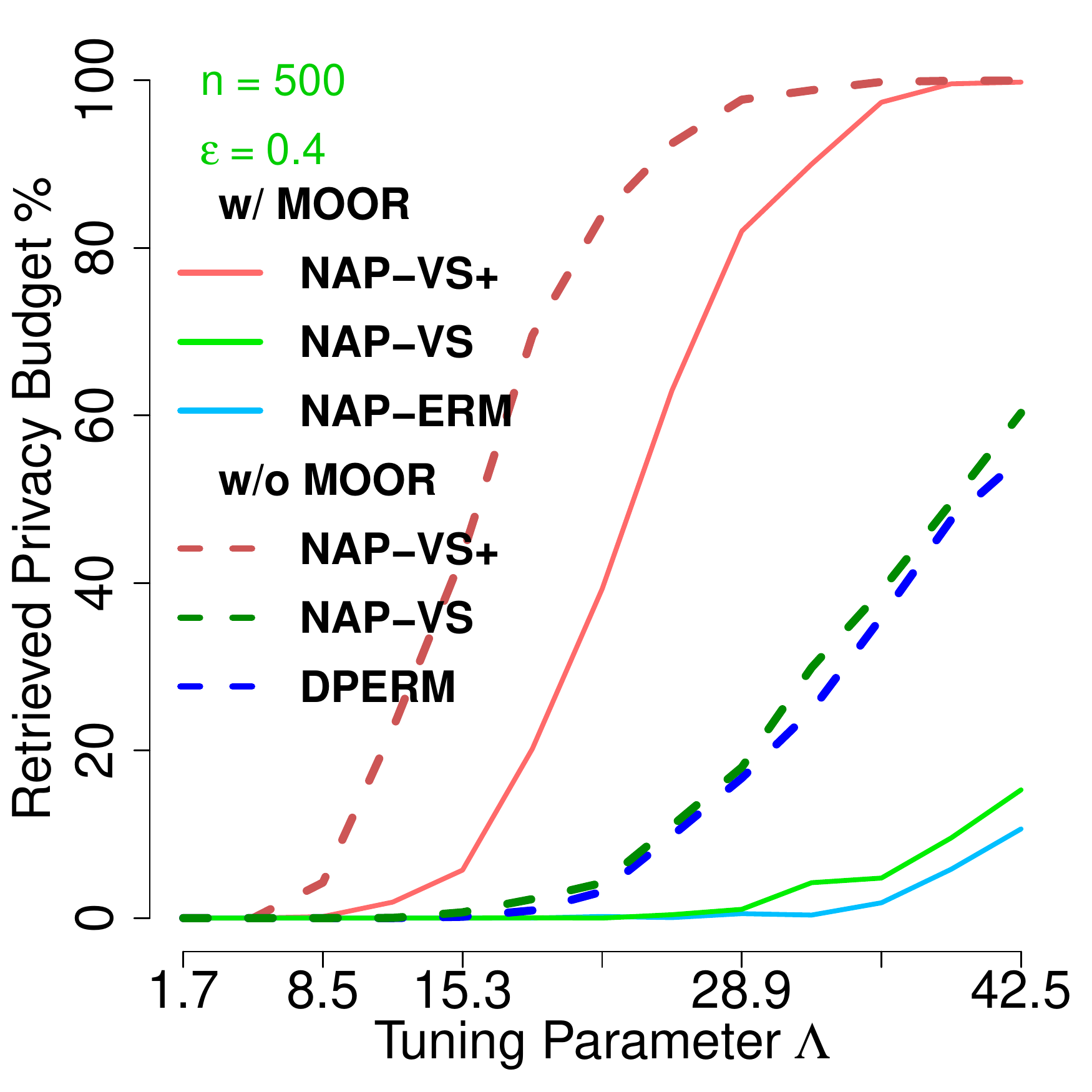}
\includegraphics[width=0.24\linewidth, trim=4pt 9pt 15pt 18pt,clip]{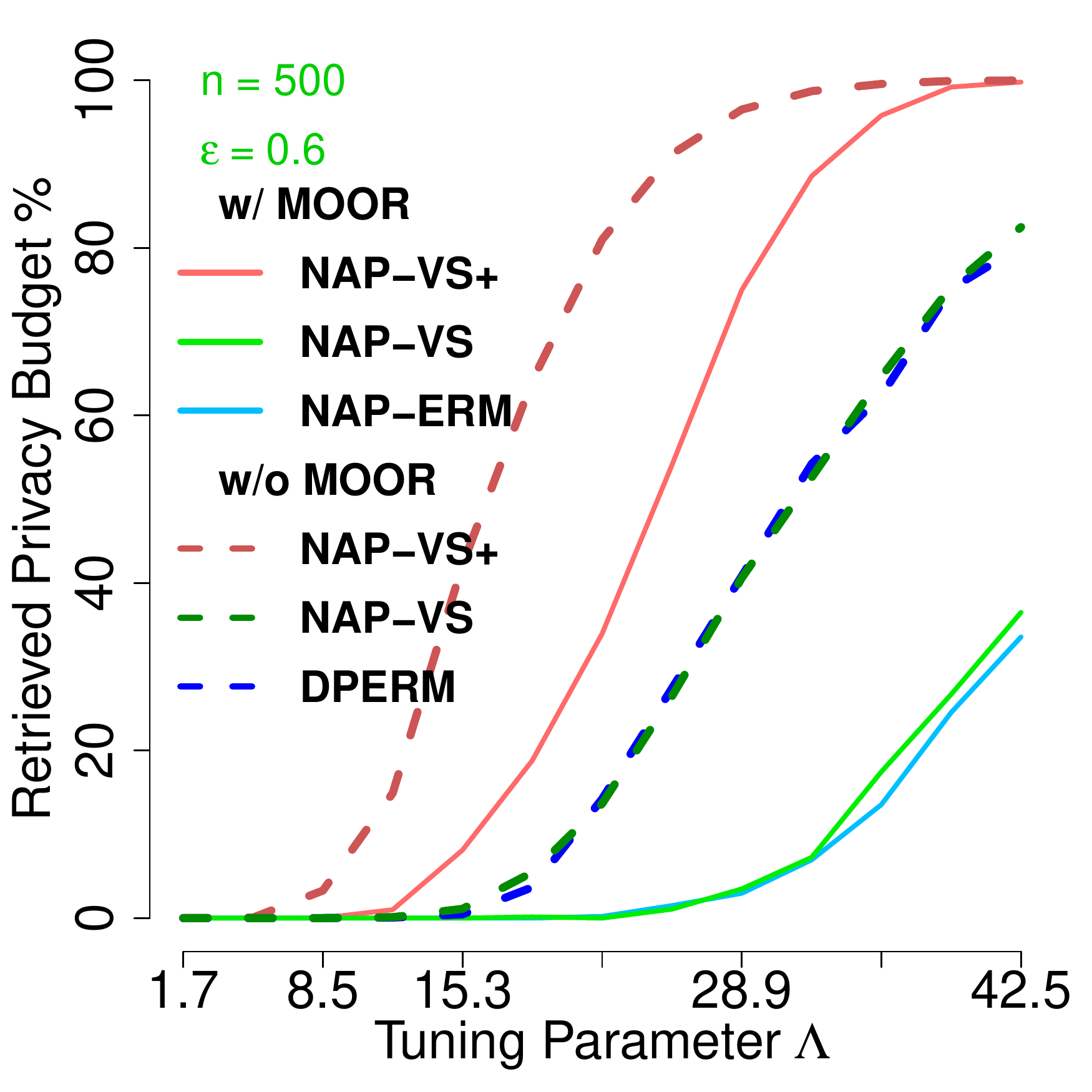}
\includegraphics[width=0.24\linewidth, trim=4pt 9pt 15pt 18pt,clip]{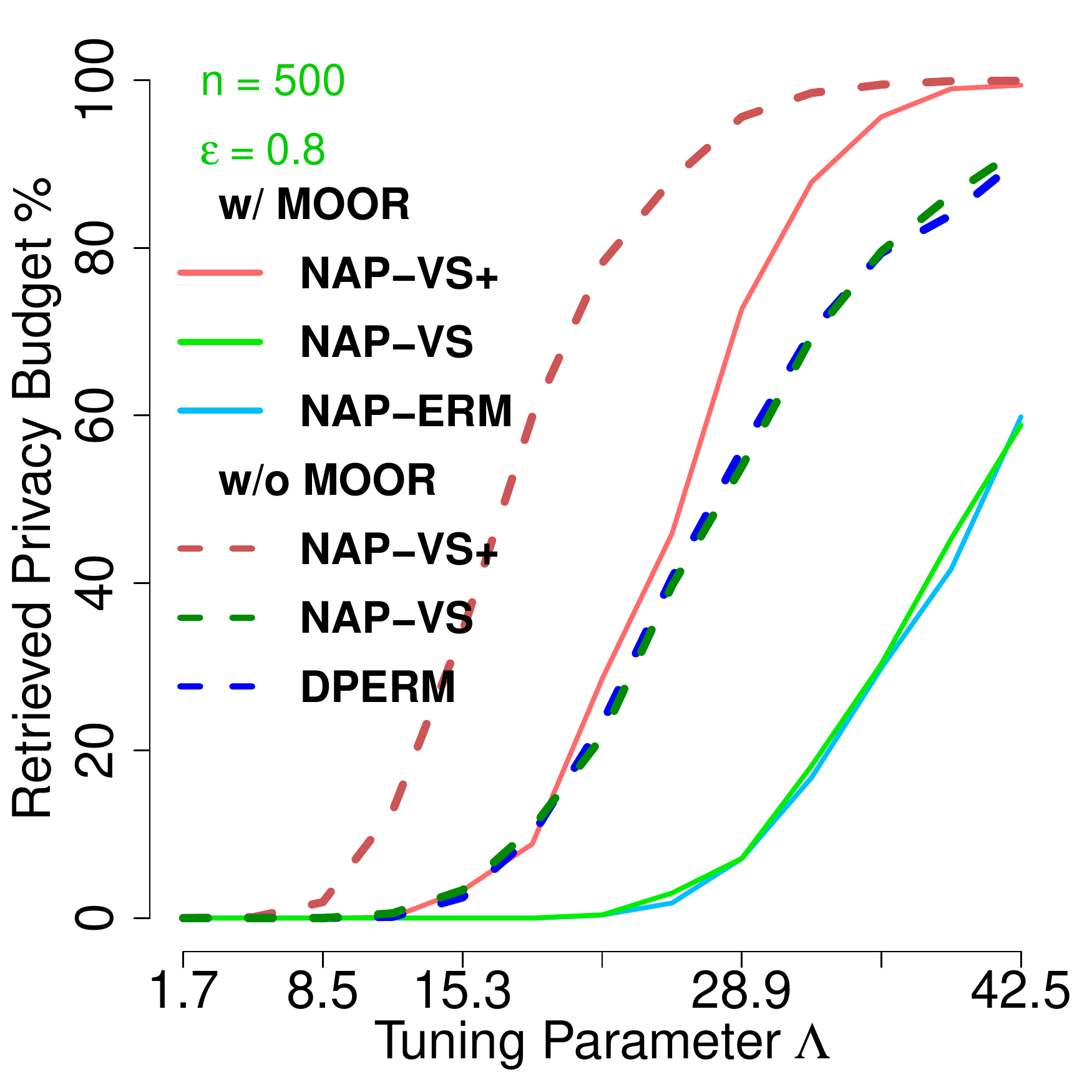}
\includegraphics[width=0.24\linewidth, trim=4pt 9pt 15pt 18pt,clip]{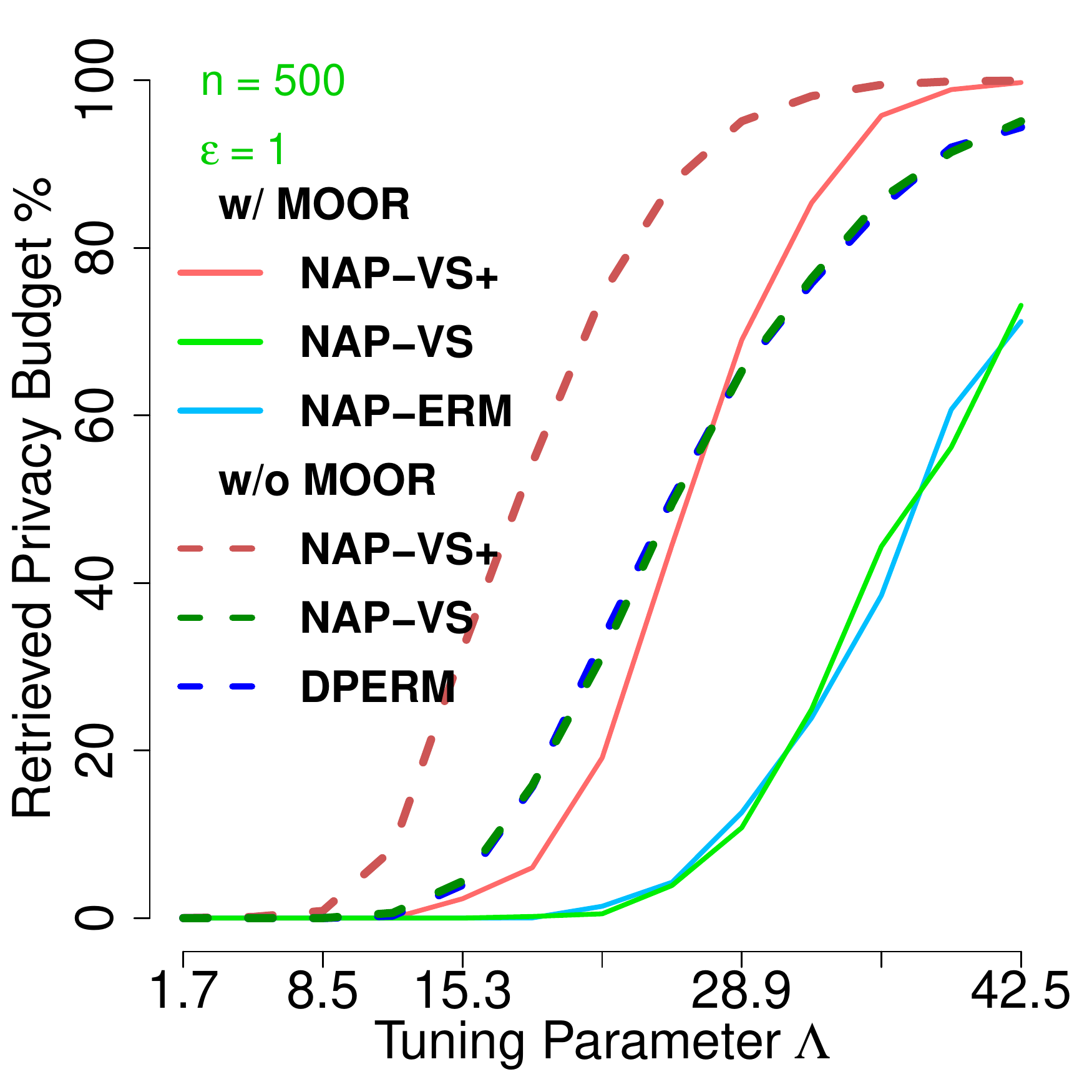}
\end{minipage}

\begin{minipage}{0.08\textwidth}\footnotesize $n=1000$ \end{minipage}
\begin{minipage}{0.91\textwidth}
\includegraphics[width=0.24\linewidth, trim=4pt 9pt 15pt 18pt,clip]{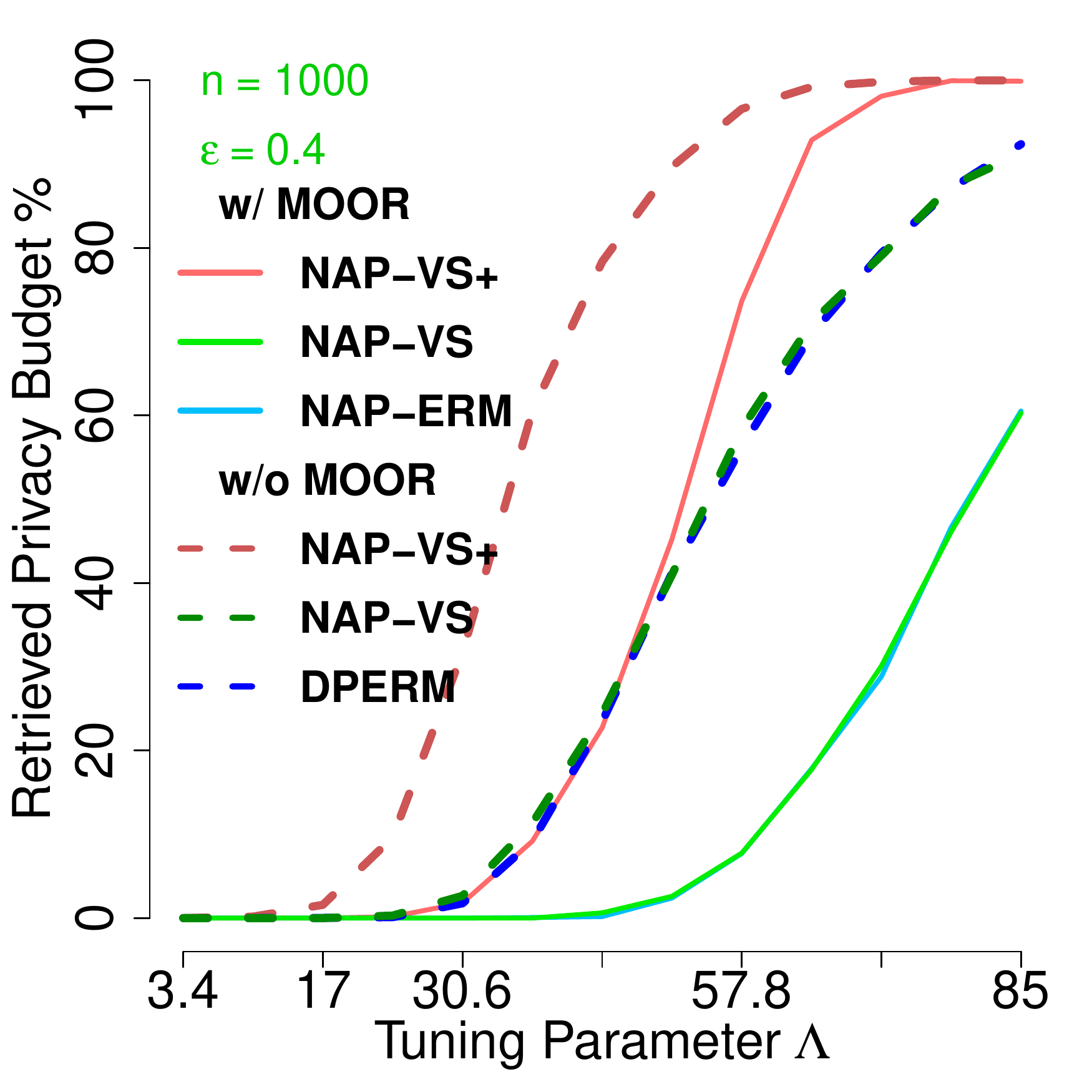}
\includegraphics[width=0.24\linewidth, trim=4pt 9pt 15pt 18pt,clip]{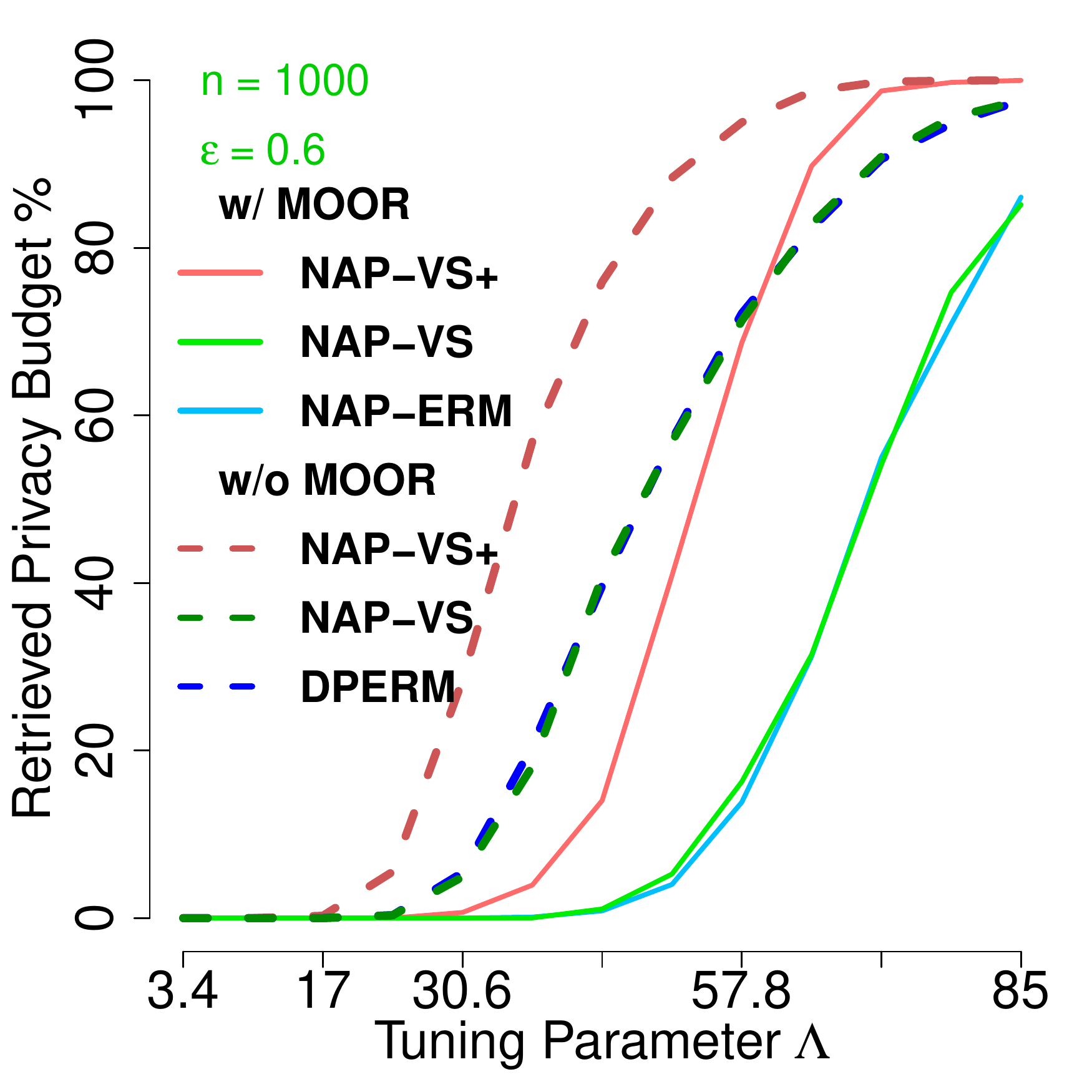}
\includegraphics[width=0.24\linewidth, trim=4pt 9pt 15pt 18pt,clip]{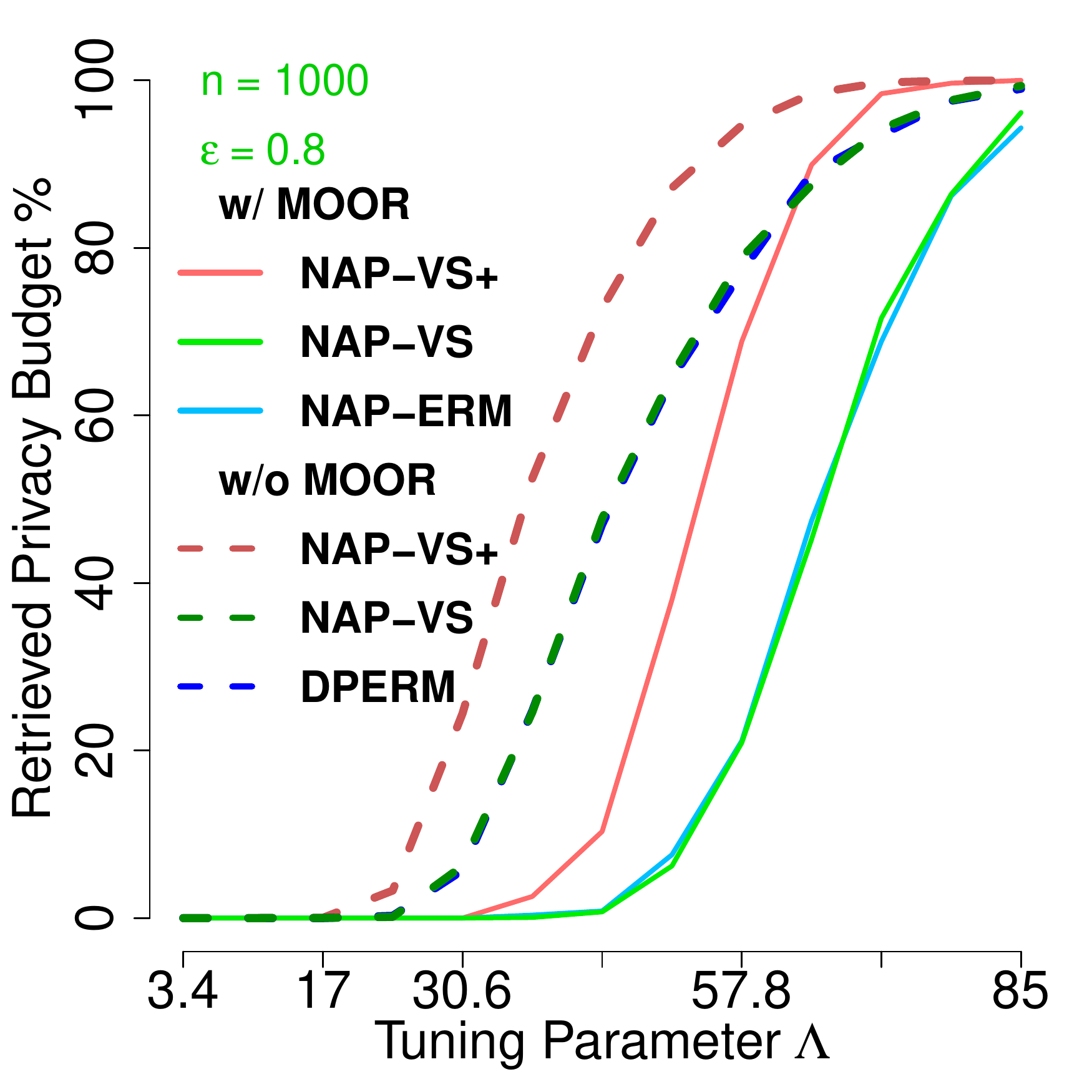}
\includegraphics[width=0.24\linewidth, trim=4pt 9pt 15pt 18pt,clip]{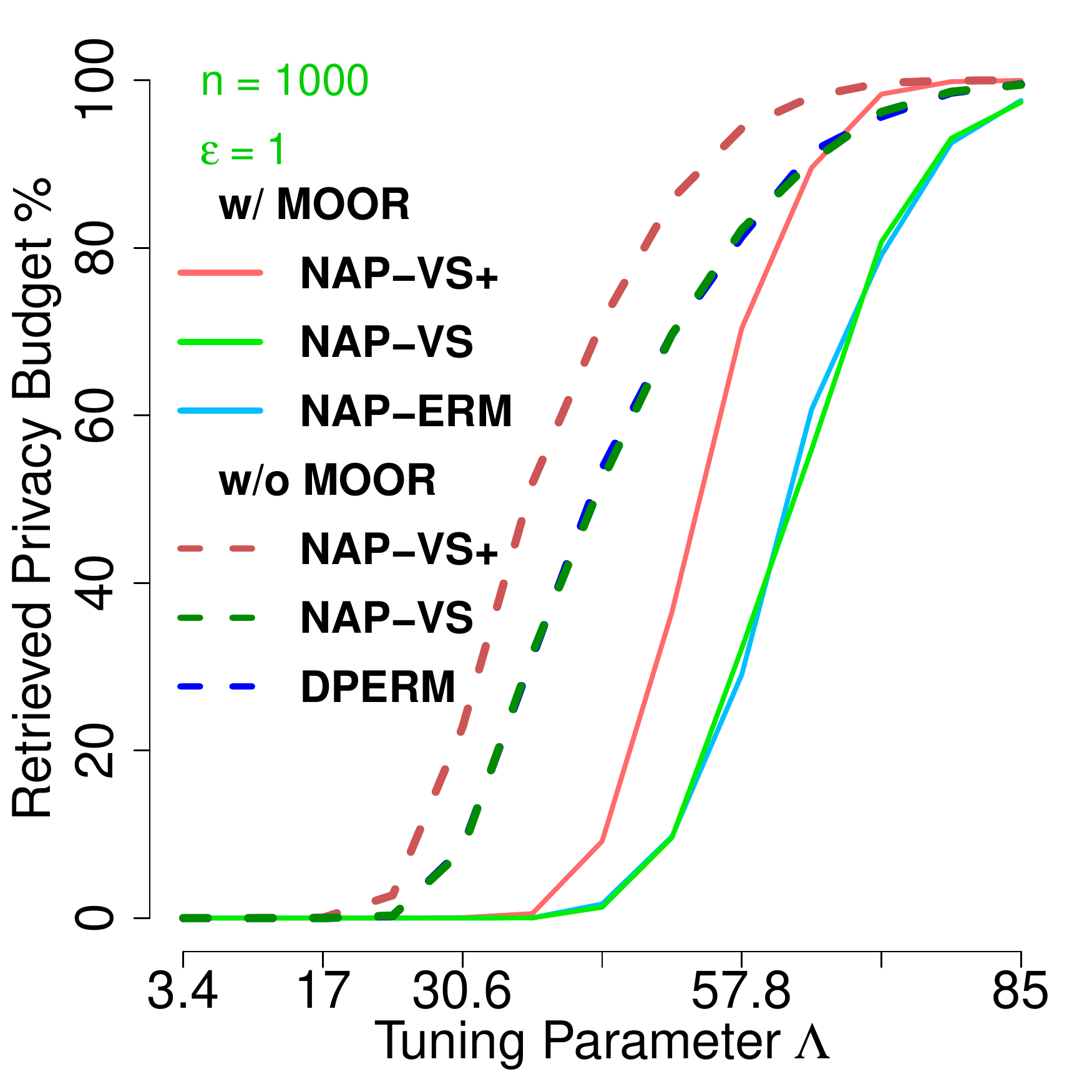}
\end{minipage}

\begin{minipage}{0.08\textwidth}
\footnotesize logistic regression $n=500$ \end{minipage}
\begin{minipage}{0.91\textwidth}
\includegraphics[width=0.24\linewidth, trim=4pt 9pt 15pt 18pt,clip]{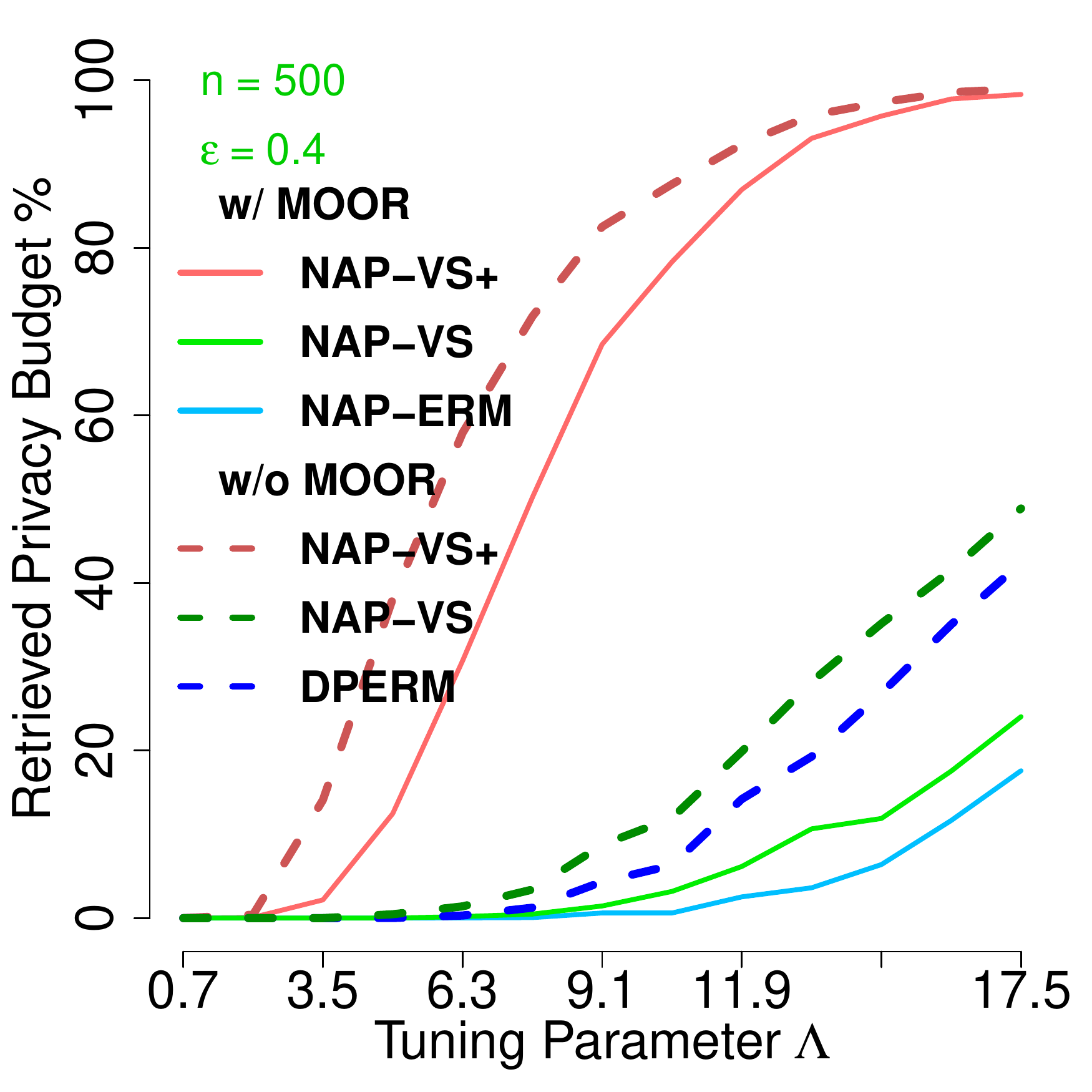}
\includegraphics[width=0.24\linewidth, trim=4pt 9pt 15pt 18pt,clip]{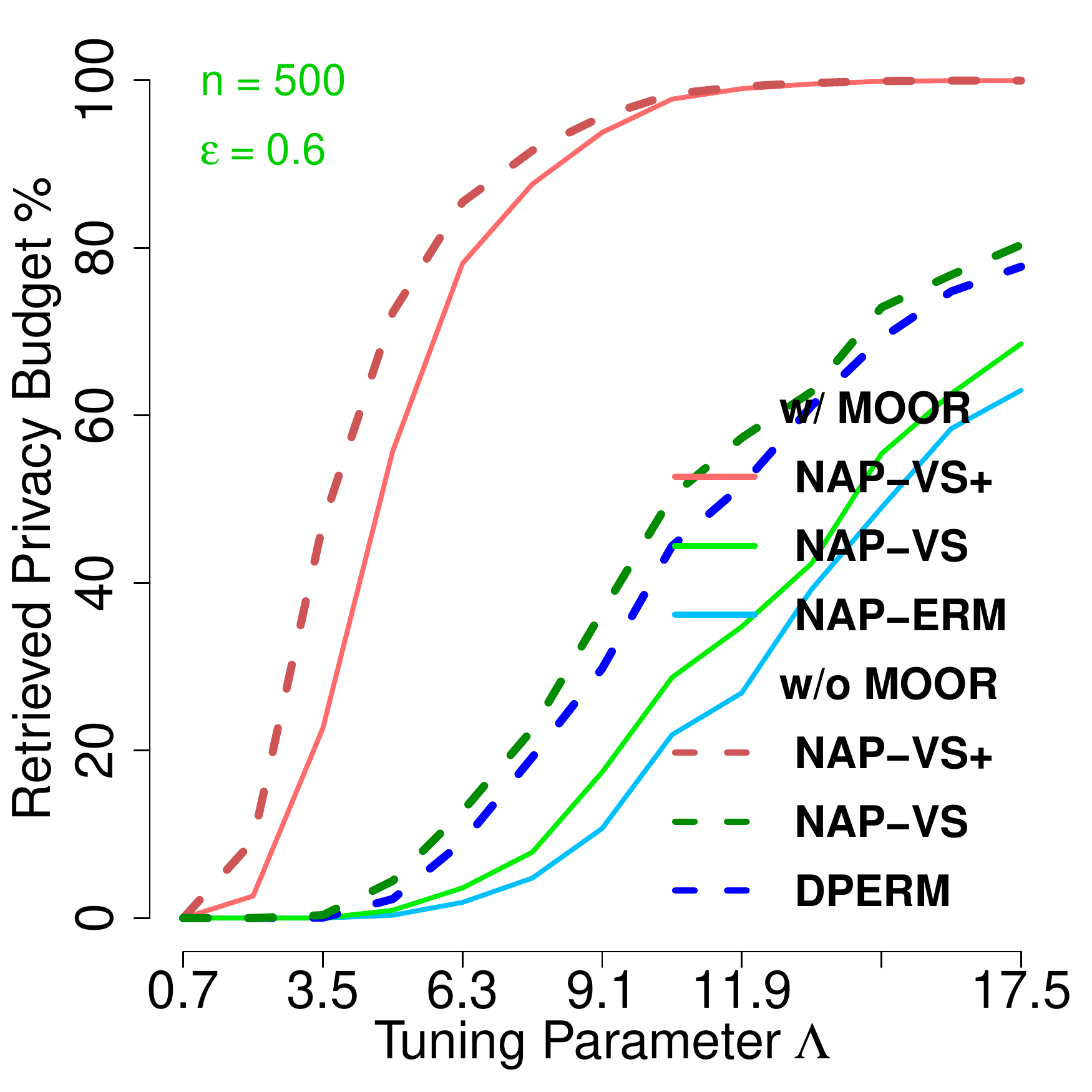}
\includegraphics[width=0.24\linewidth, trim=4pt 9pt 15pt 18pt,clip]{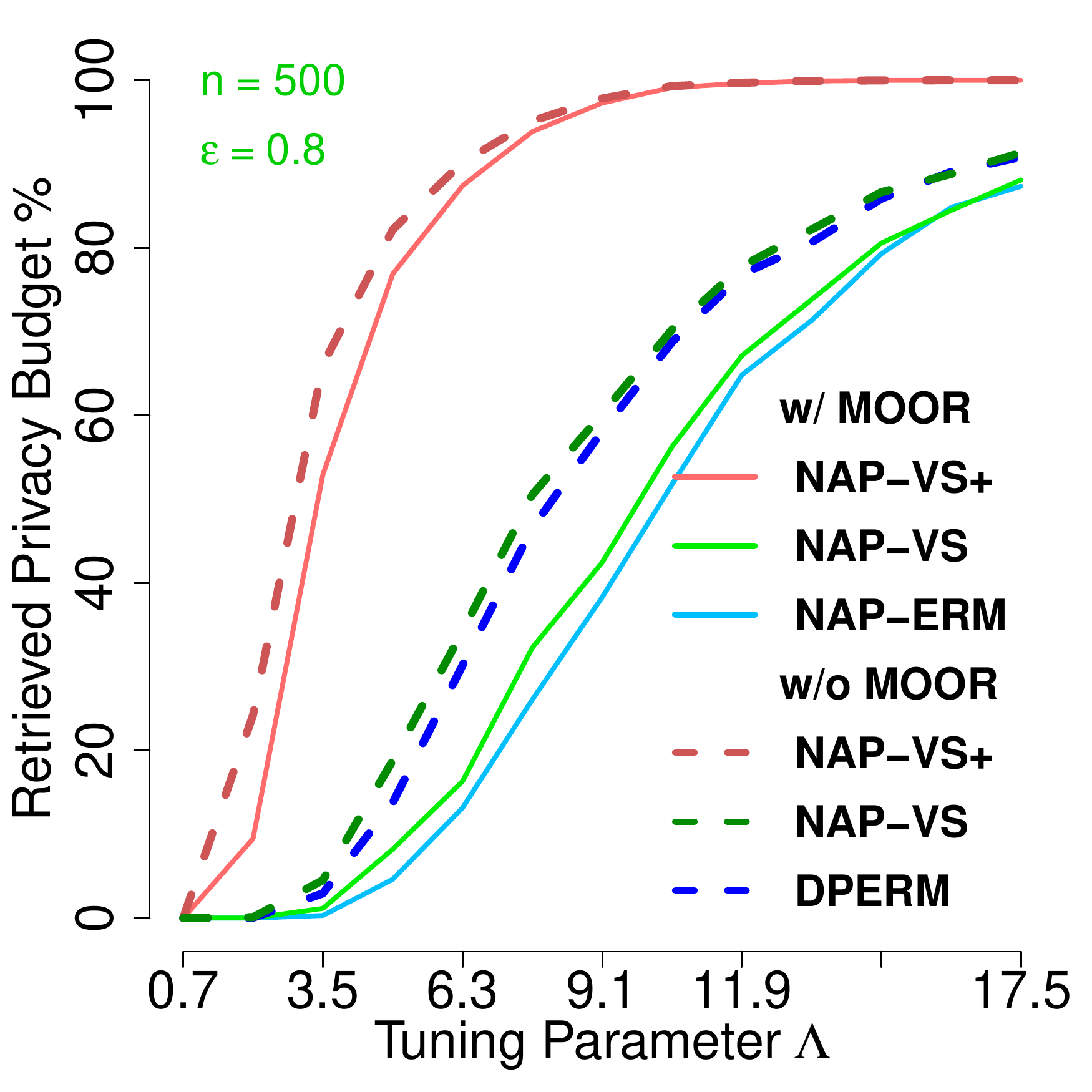}
\includegraphics[width=0.24\linewidth, trim=4pt 9pt 15pt 18pt,clip]{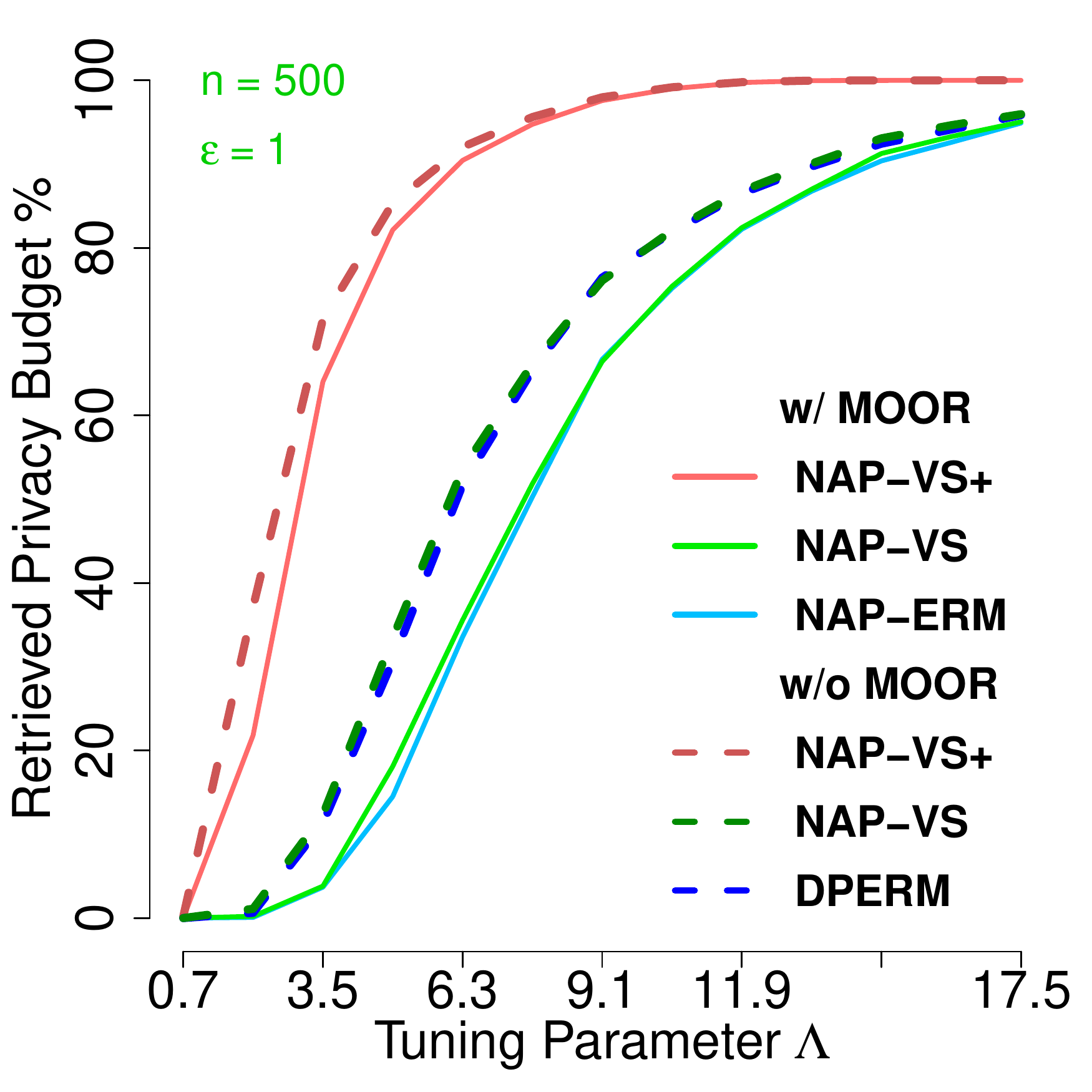}
\end{minipage}

\begin{minipage}{0.08\textwidth}\footnotesize $n=1000$ \end{minipage}
\begin{minipage}{0.91\textwidth}
\includegraphics[width=0.24\linewidth, trim=4pt 9pt 15pt 18pt,clip]{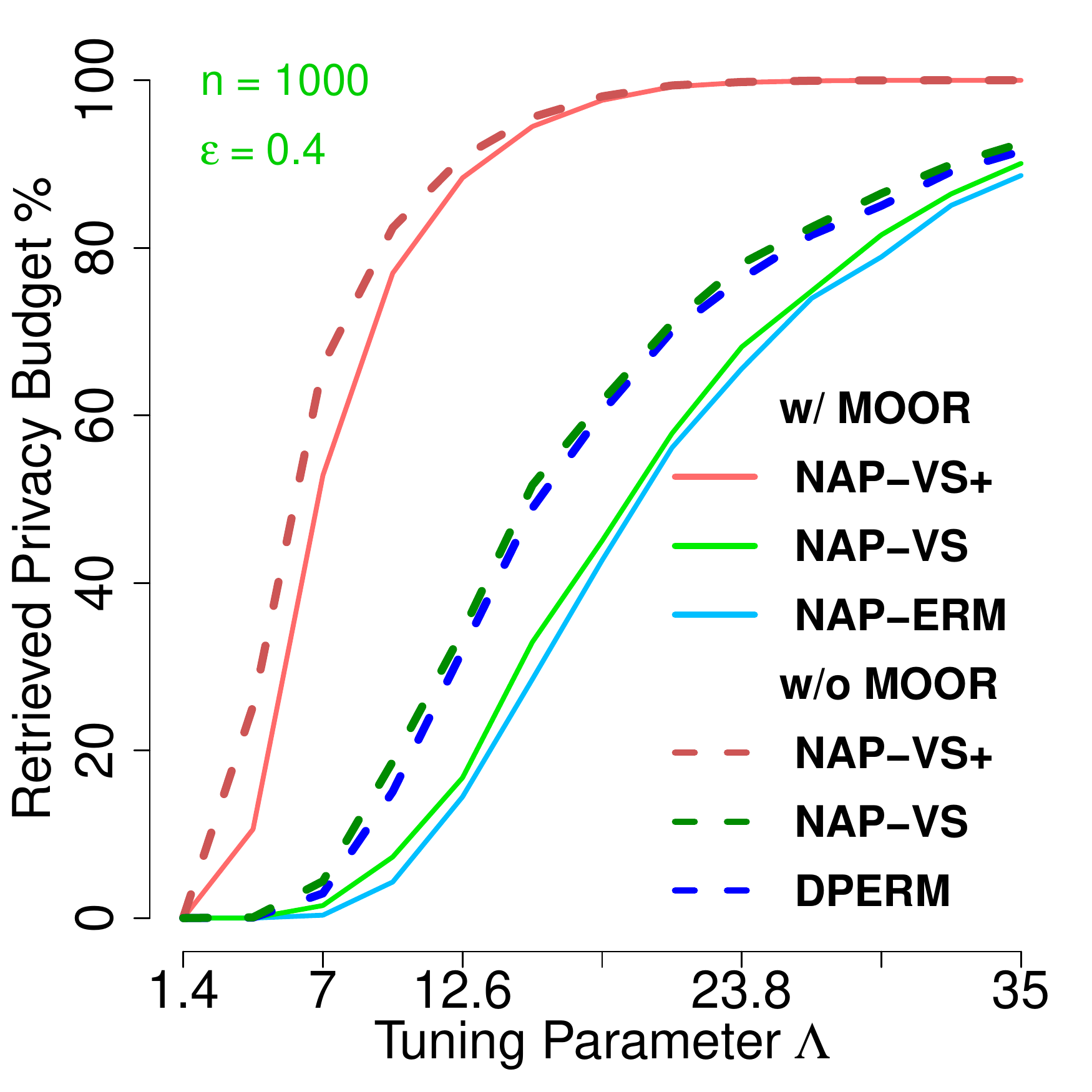}
\includegraphics[width=0.24\linewidth, trim=4pt 9pt 15pt 18pt,clip]{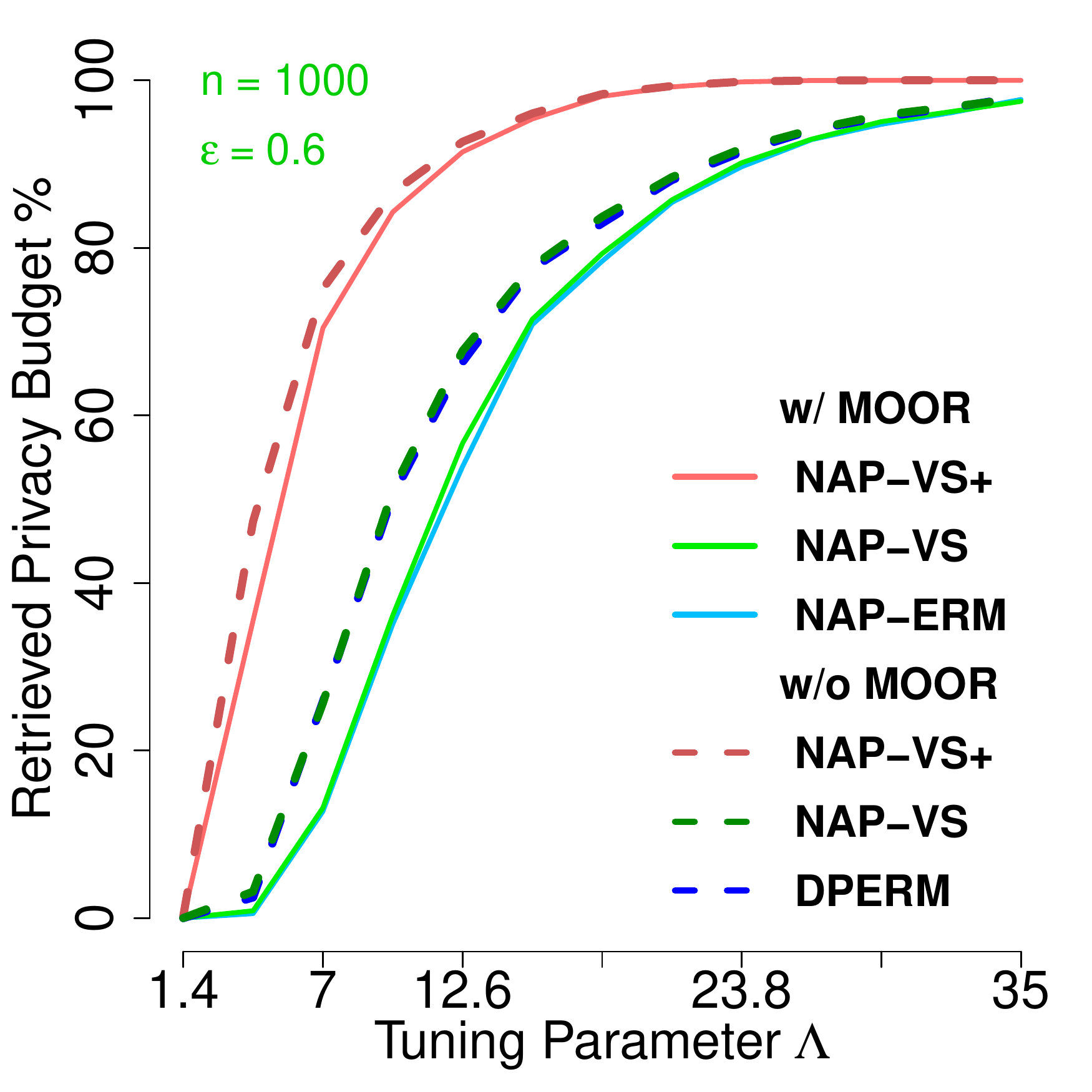}
\includegraphics[width=0.24\linewidth, trim=4pt 9pt 15pt 18pt,clip]{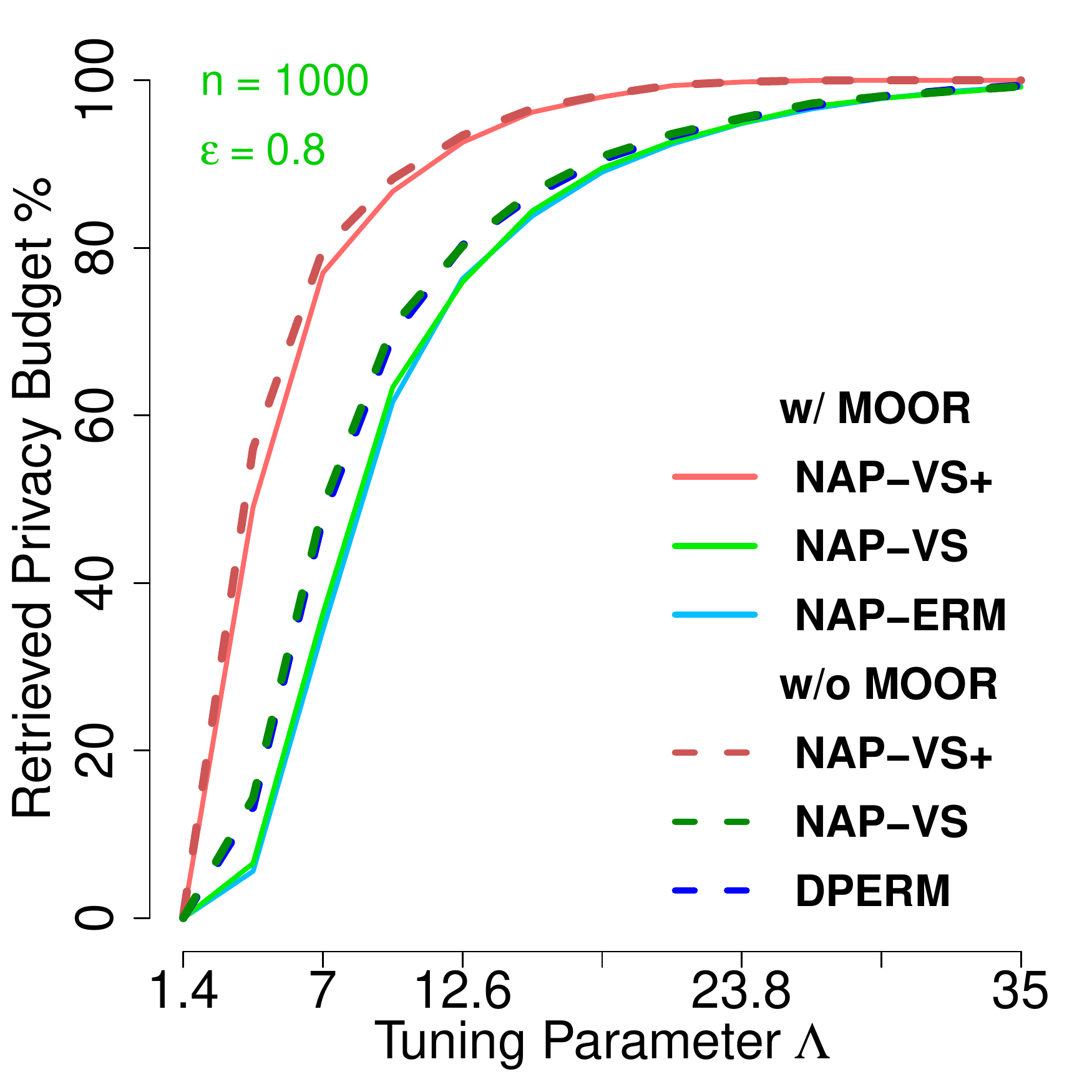}
\includegraphics[width=0.24\linewidth, trim=4pt 9pt 15pt 18pt,clip]{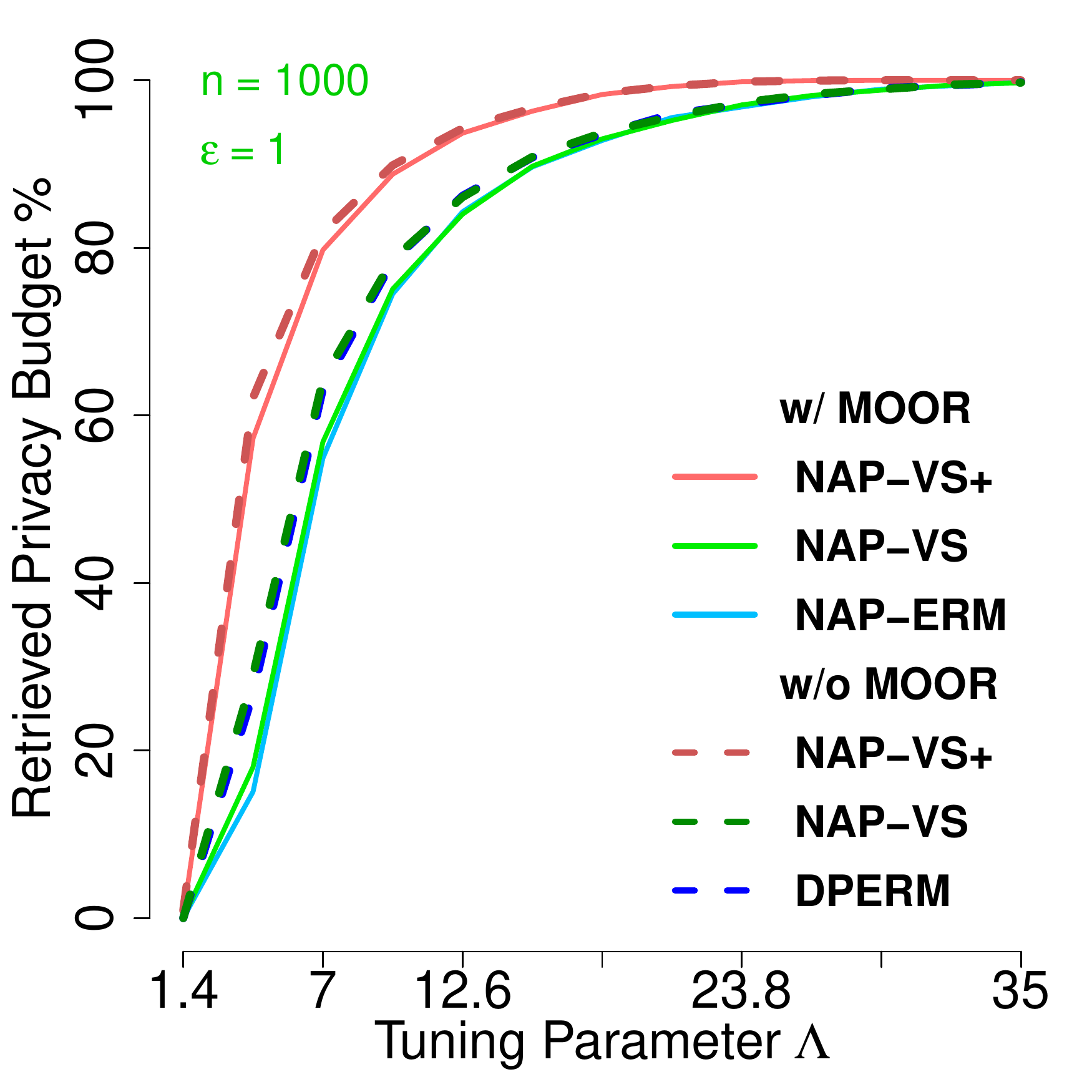}
\end{minipage}\vspace{-6pt}
\caption{Retrieved Portion of the privacy budget allocated to bounding the Jacobian ratio}  
\label{fig:retrieval1}
\end{figure}


\begin{figure}[!htb]
\begin{minipage}{0.08\textwidth}
\footnotesize  variable selection ROC curve via lasso $n=200$ \end{minipage}
\begin{minipage}{0.90\textwidth}
\includegraphics[width=0.24\linewidth, trim=4pt 9pt 15pt 18pt,clip]{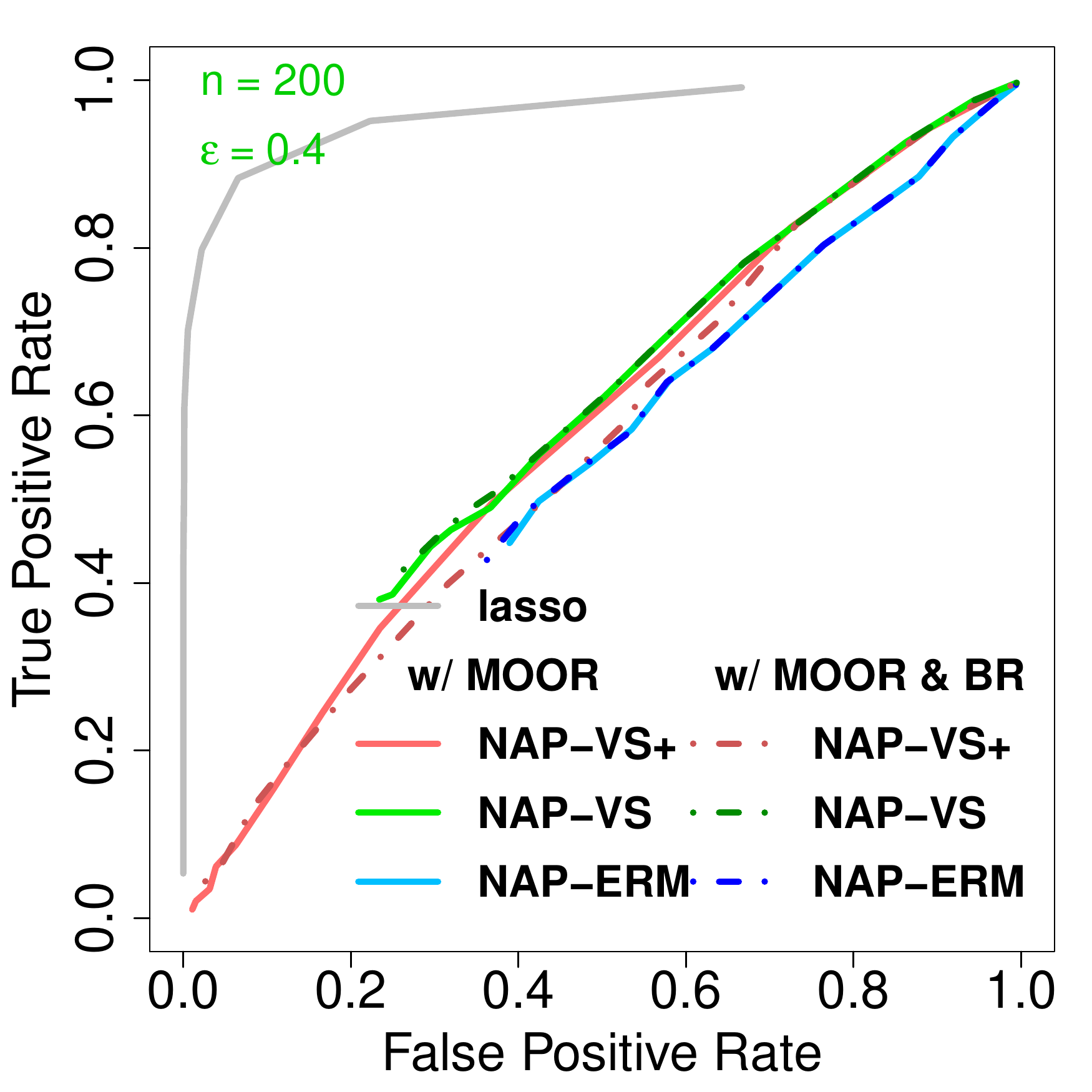}
\includegraphics[width=0.24\linewidth, trim=4pt 9pt 15pt 18pt,clip]{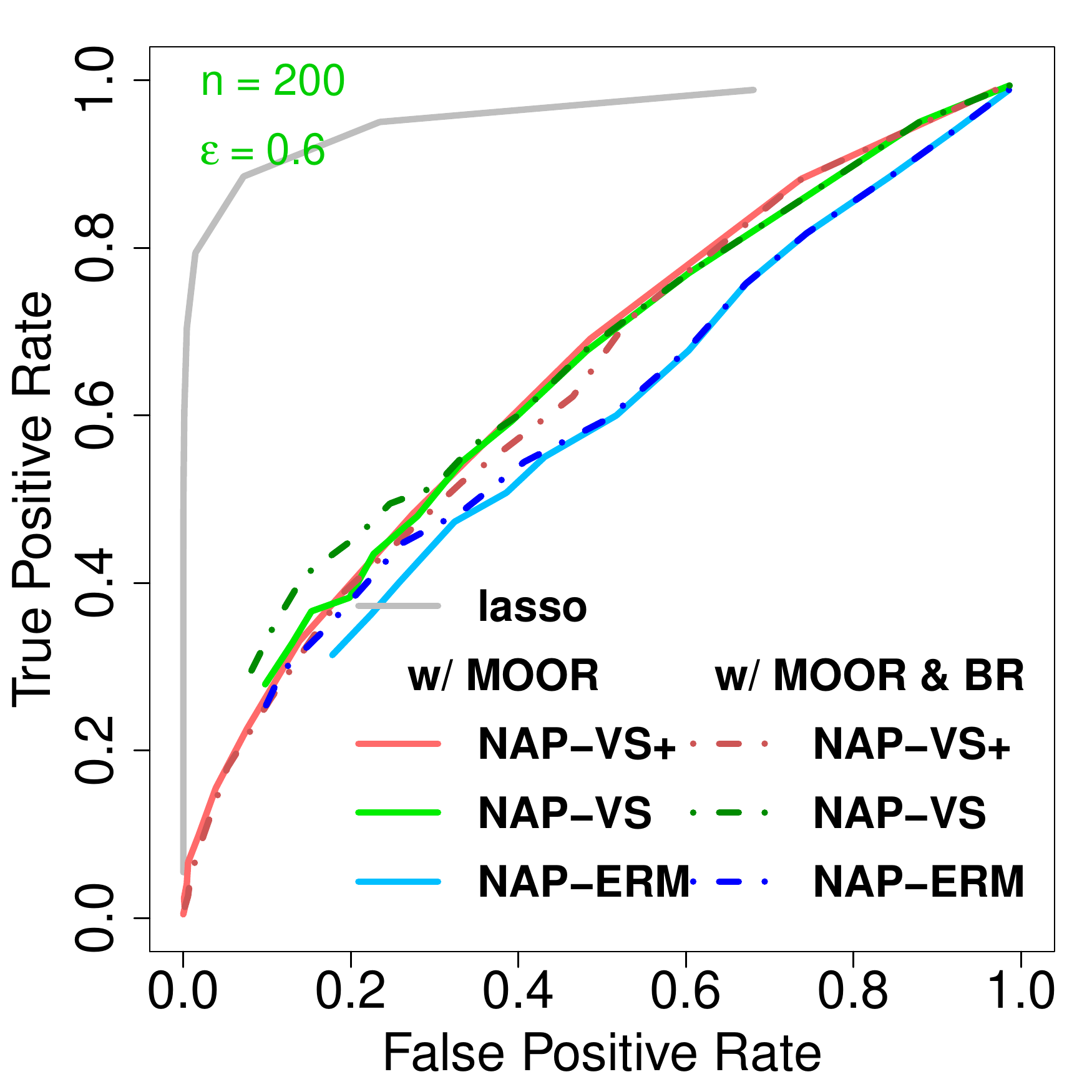}
\includegraphics[width=0.24\linewidth, trim=4pt 9pt 15pt 18pt,clip]{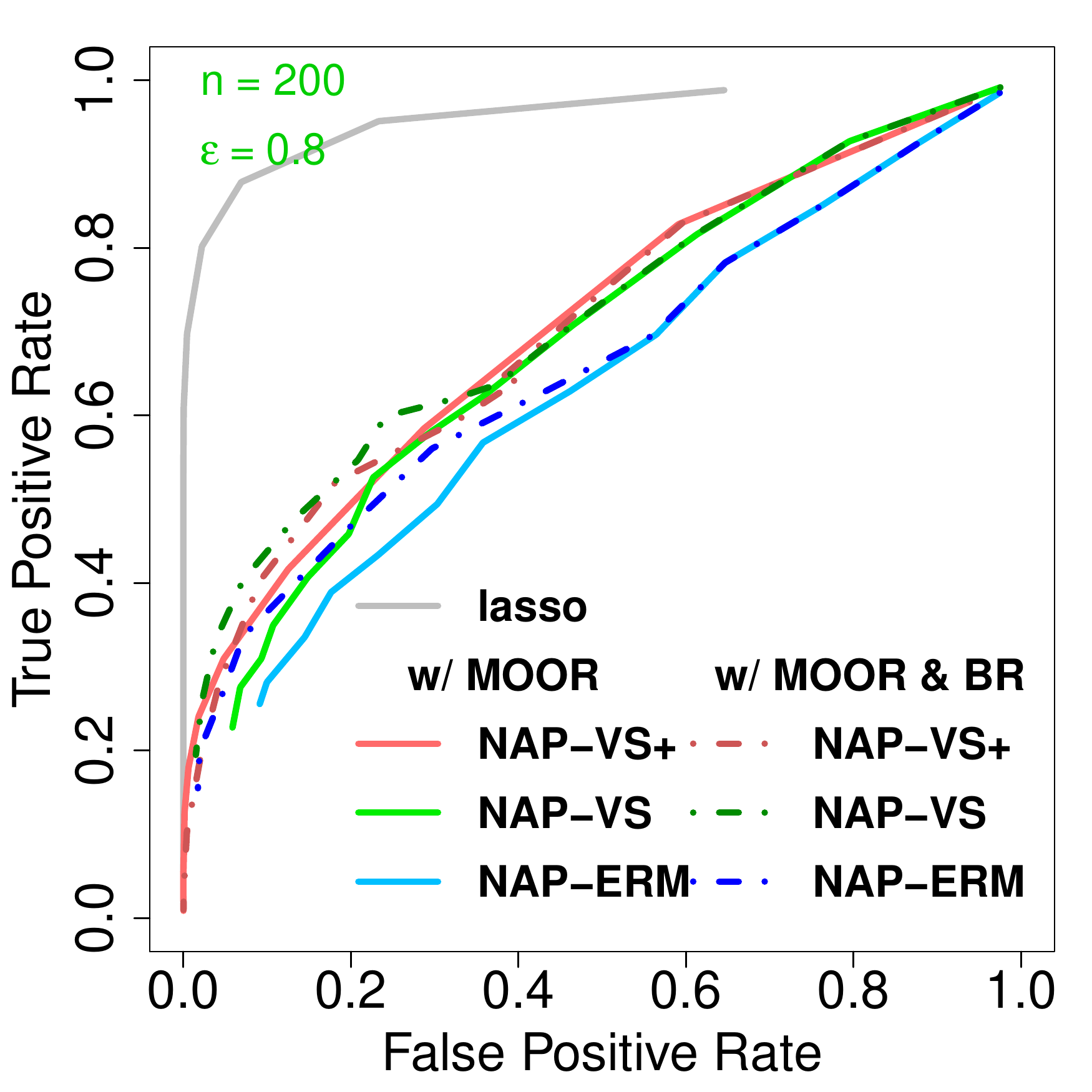}
\includegraphics[width=0.24\linewidth, trim=4pt 9pt 15pt 18pt,clip]{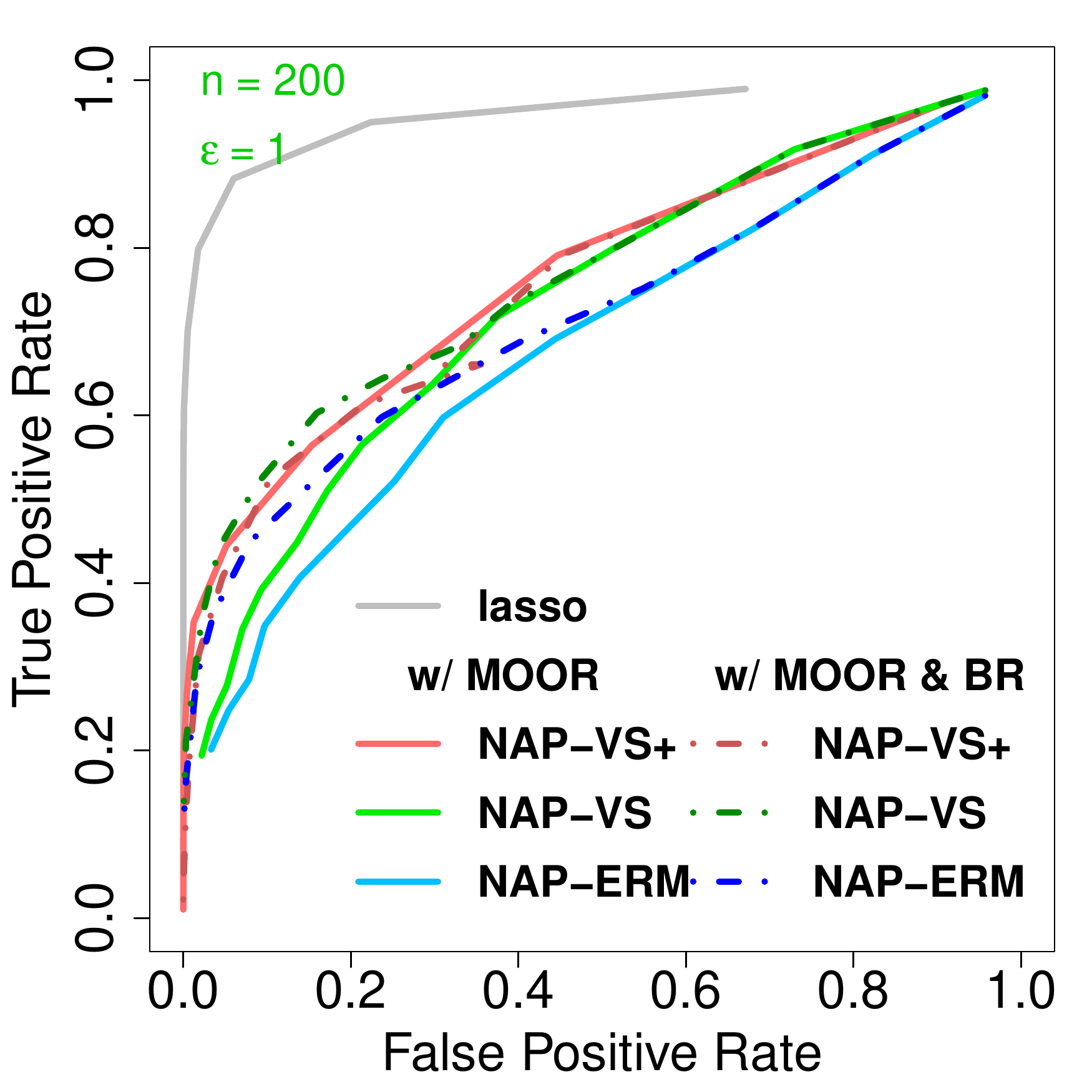}
\end{minipage}
\begin{minipage}{0.08\textwidth}\footnotesize $n=500$ \end{minipage}
\begin{minipage}{0.90\textwidth}
\includegraphics[width=0.24\linewidth, trim=4pt 9pt 15pt 18pt,clip]{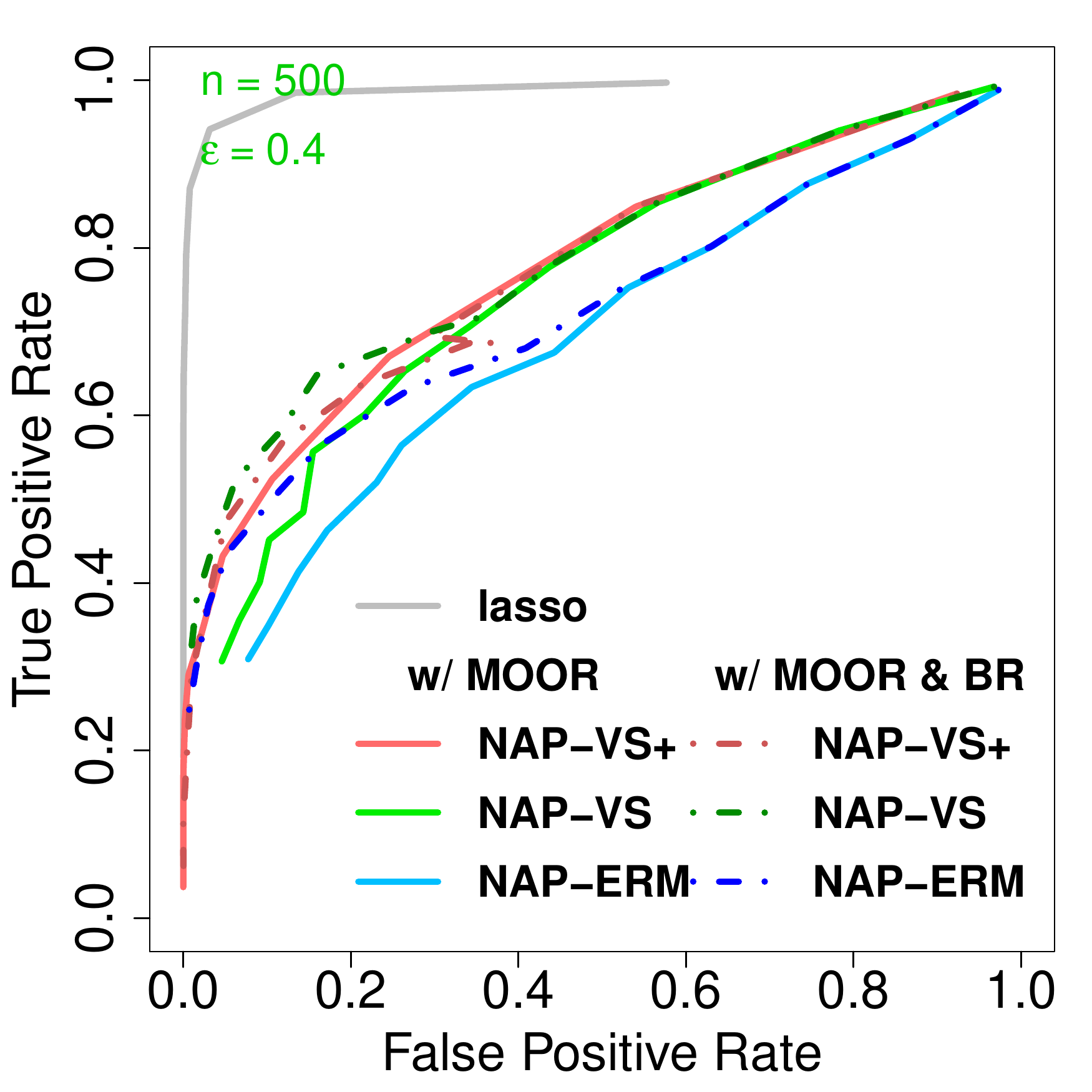}
\includegraphics[width=0.24\linewidth, trim=4pt 9pt 15pt 18pt,clip]{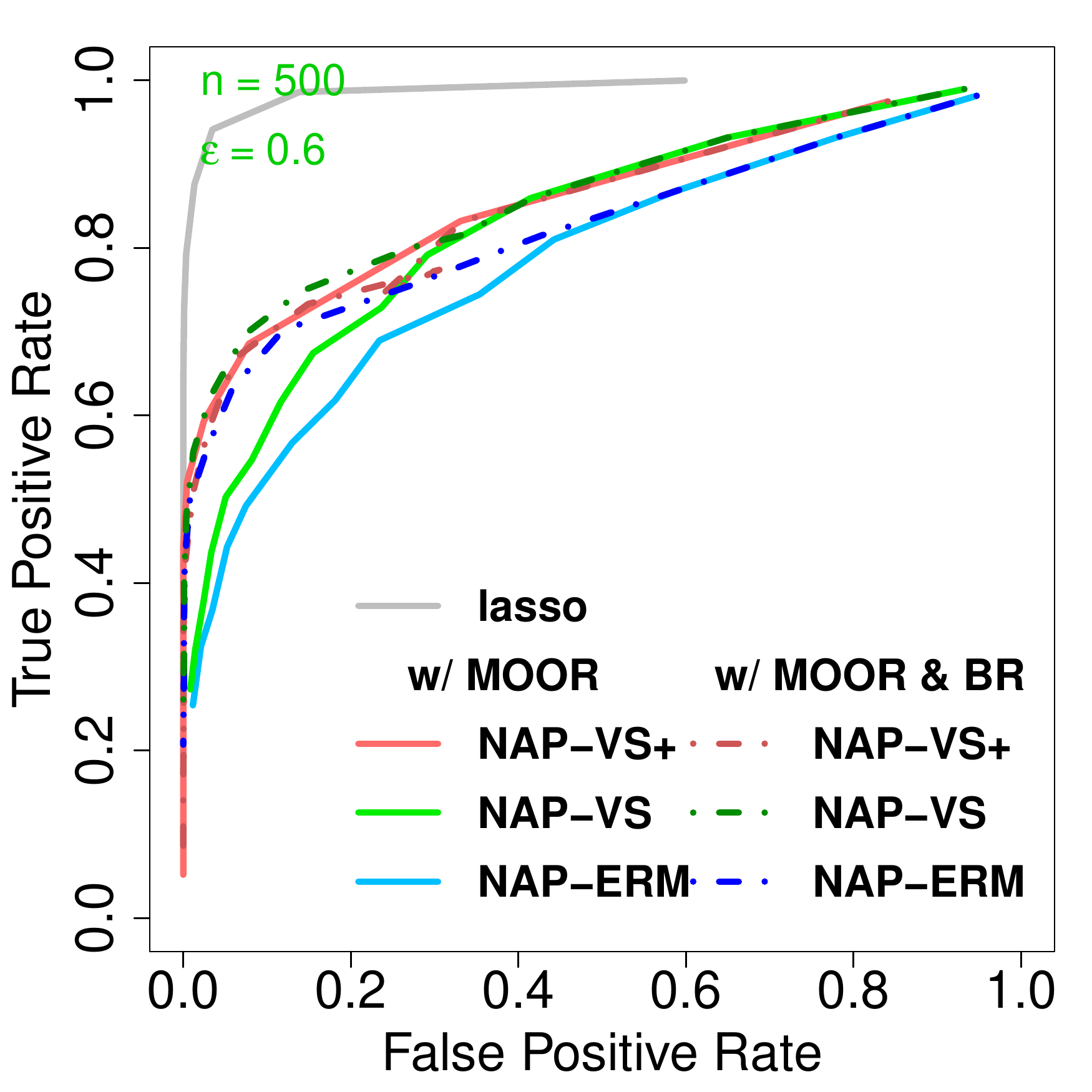}
\includegraphics[width=0.24\linewidth, trim=4pt 9pt 15pt 18pt,clip]{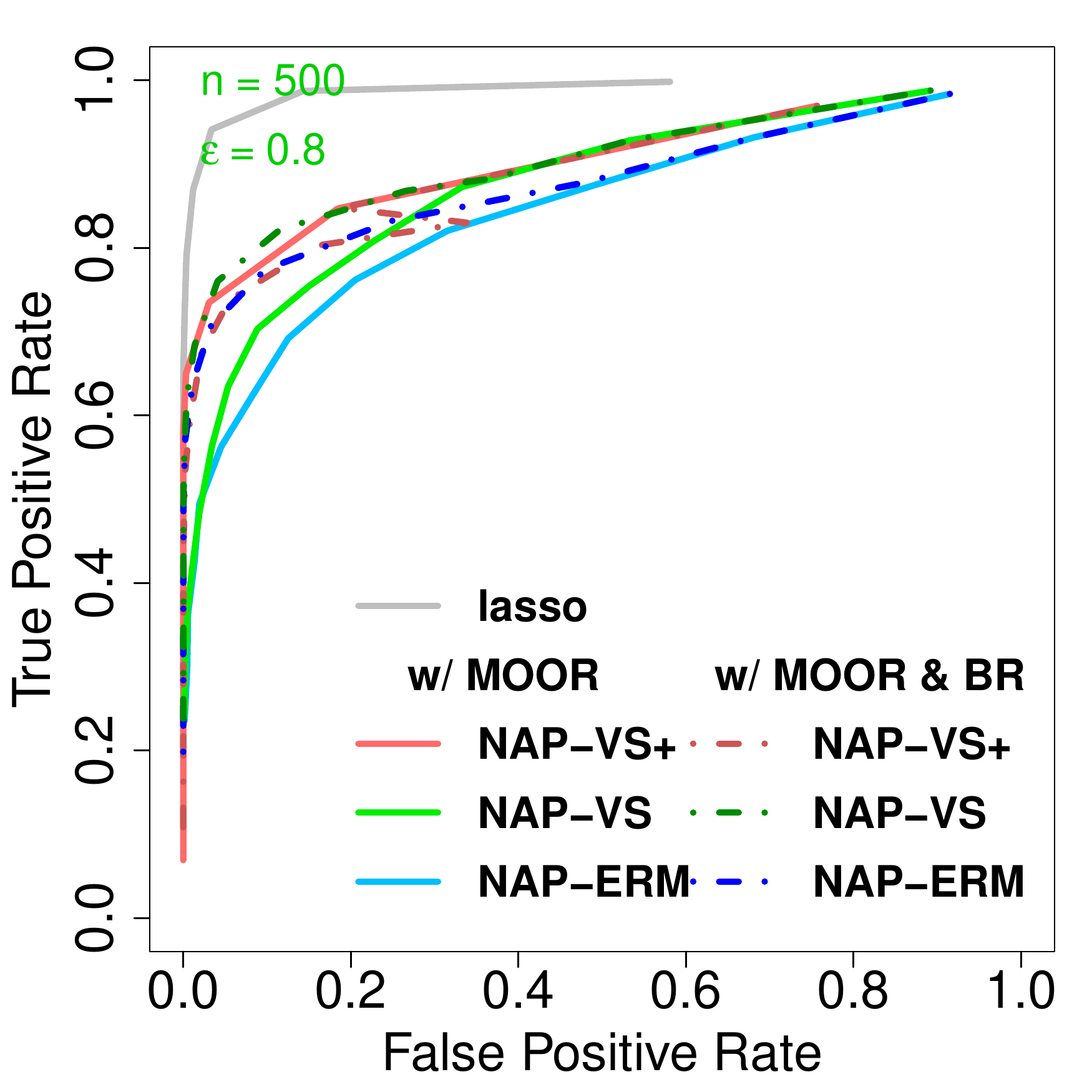}
\includegraphics[width=0.24\linewidth, trim=4pt 9pt 15pt 18pt,clip]{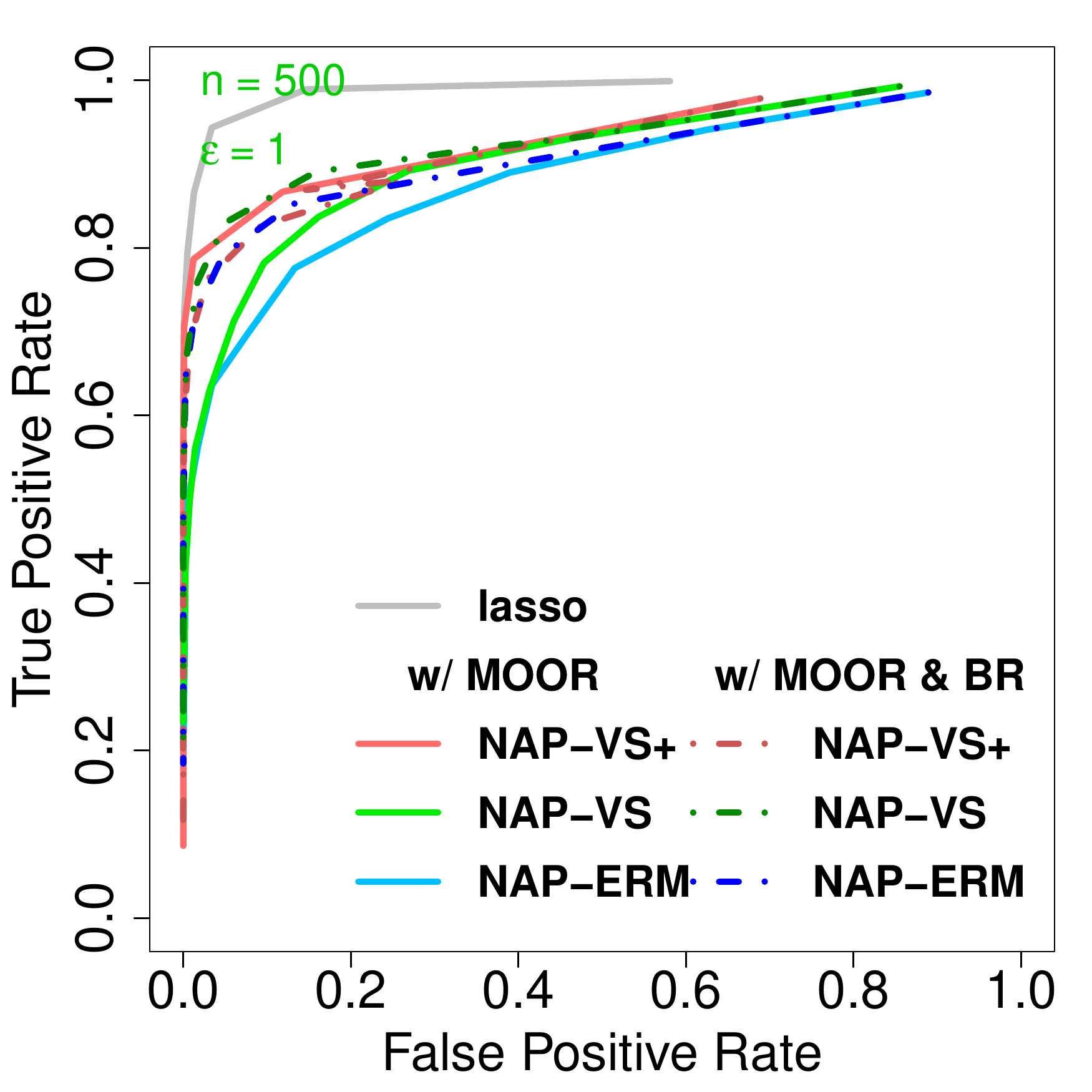} 
\end{minipage}
\begin{minipage}{0.09\textwidth}
\footnotesize outcome prediction MSE $n=200$ \end{minipage}
\begin{minipage}{0.9\textwidth}
\includegraphics[width=0.24\linewidth, trim=4pt 9pt 15pt 18pt,clip]{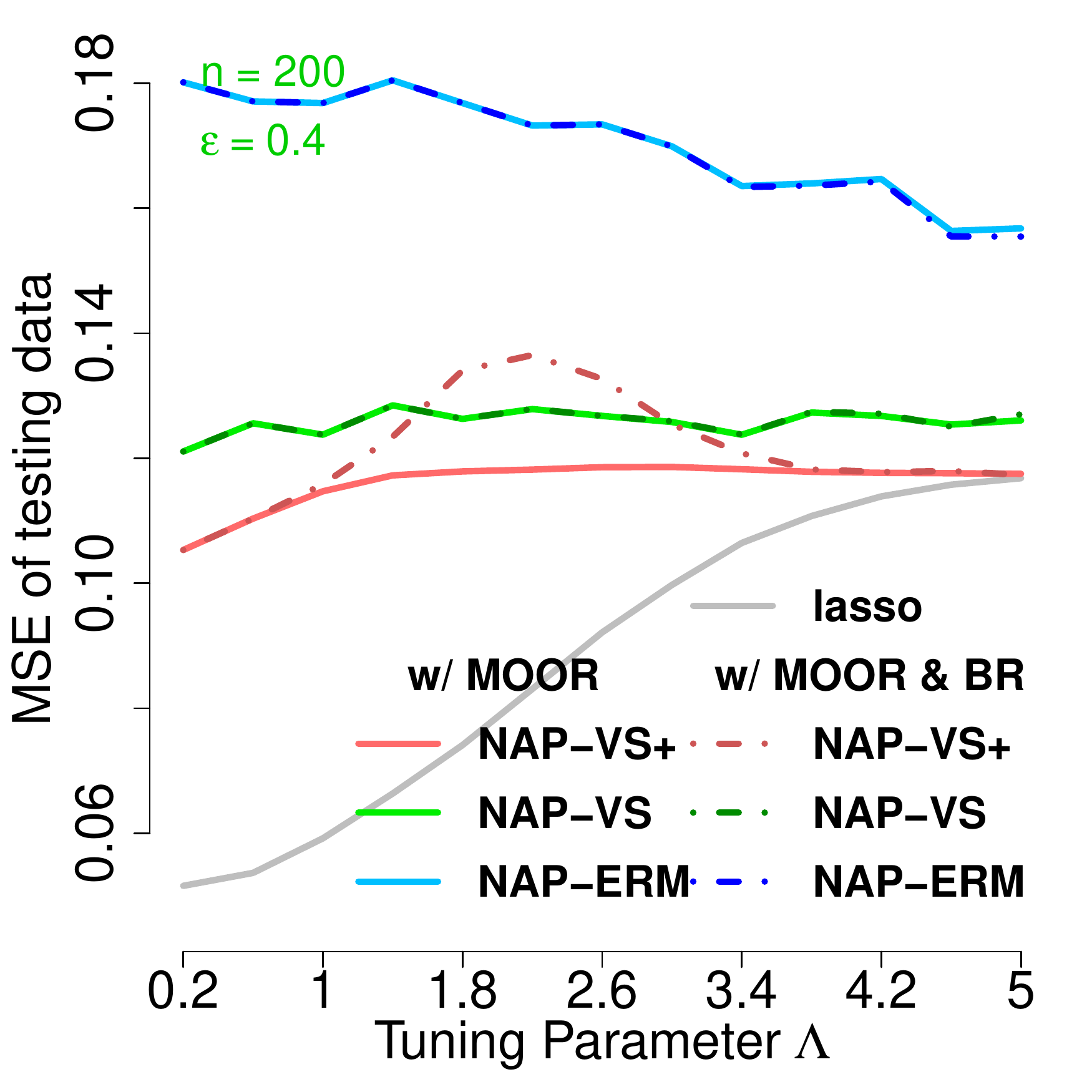}
\includegraphics[width=0.24\linewidth, trim=4pt 9pt 15pt 18pt,clip]{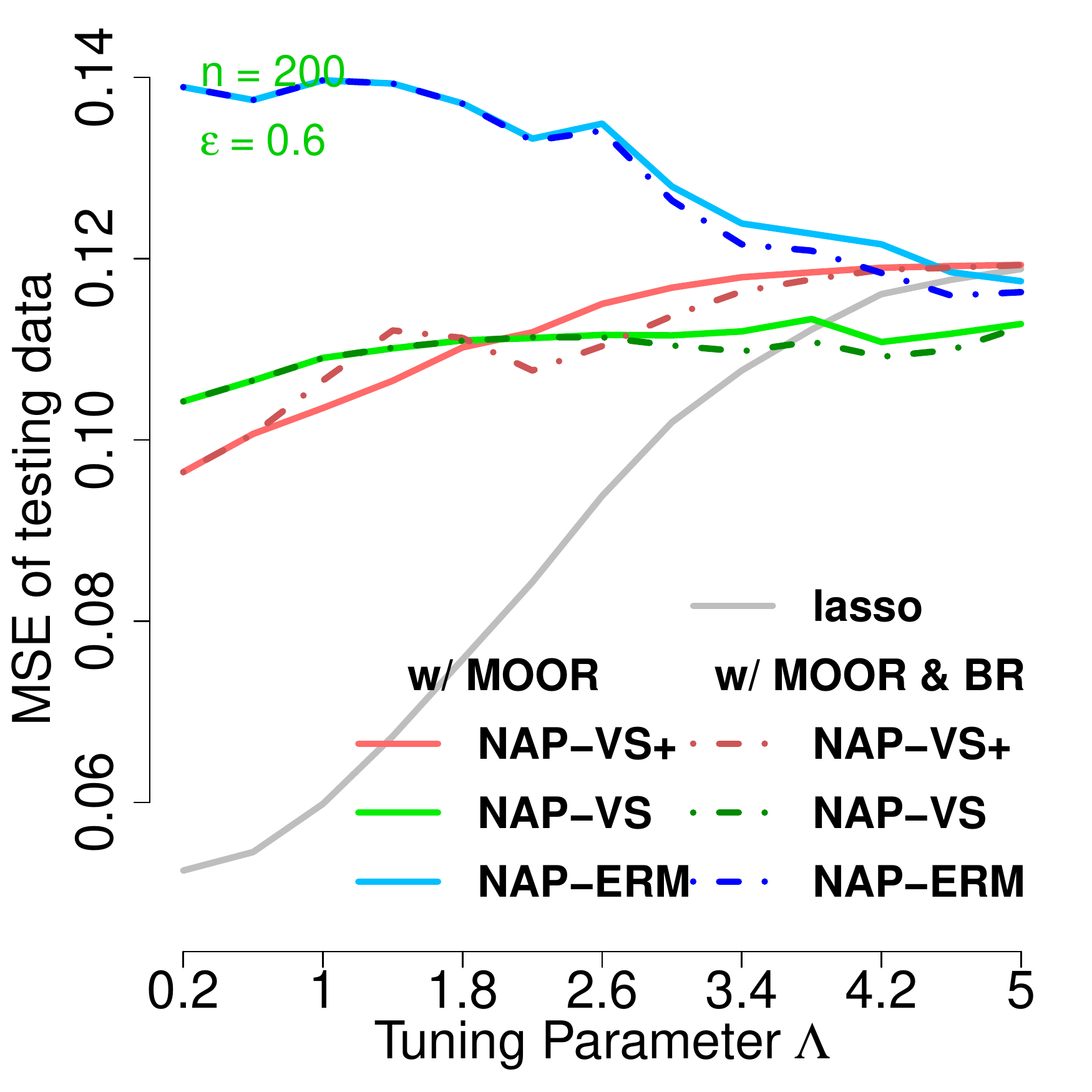}
\includegraphics[width=0.24\linewidth, trim=4pt 9pt 15pt 18pt,clip]{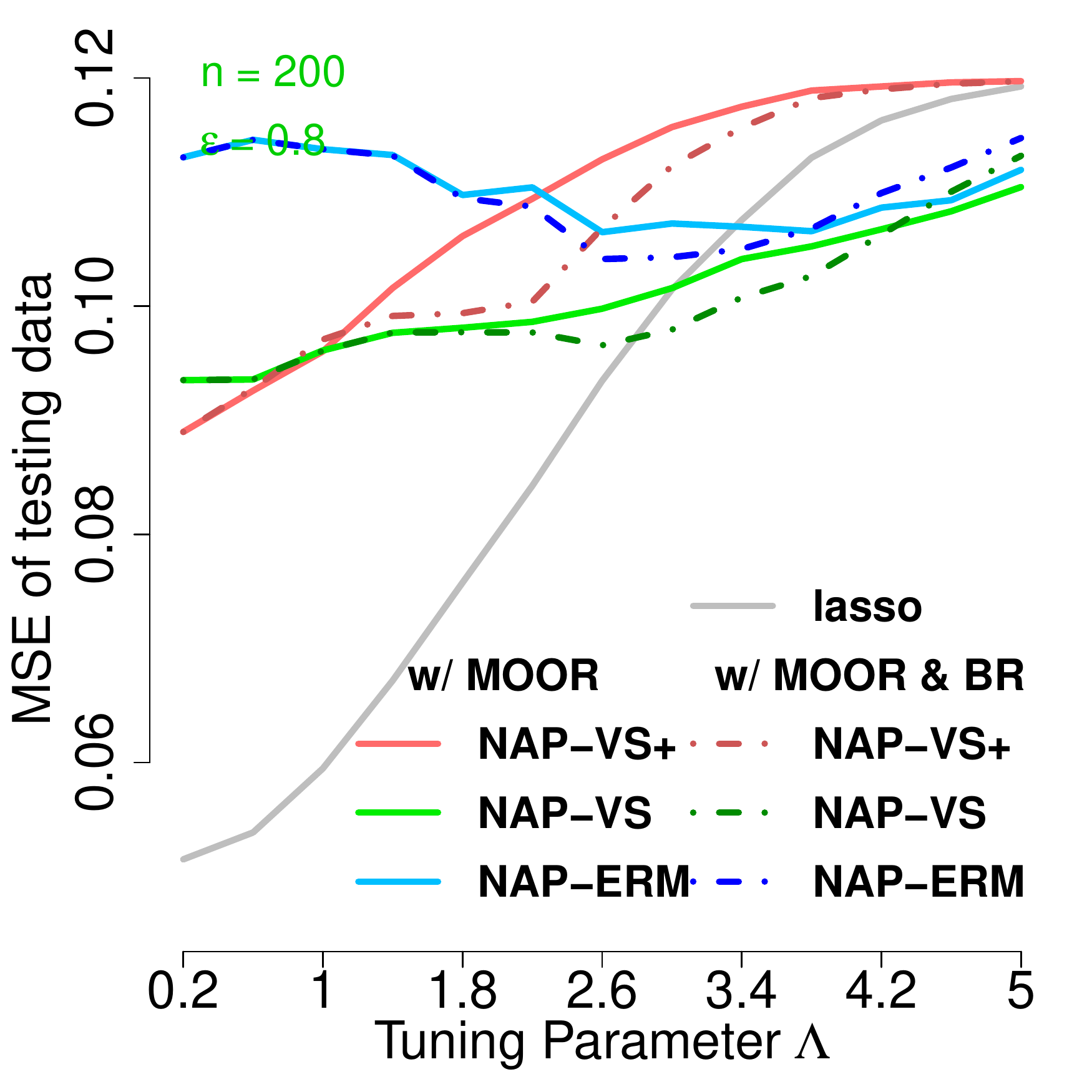}
\includegraphics[width=0.24\linewidth, trim=4pt 9pt 15pt 18pt,clip]{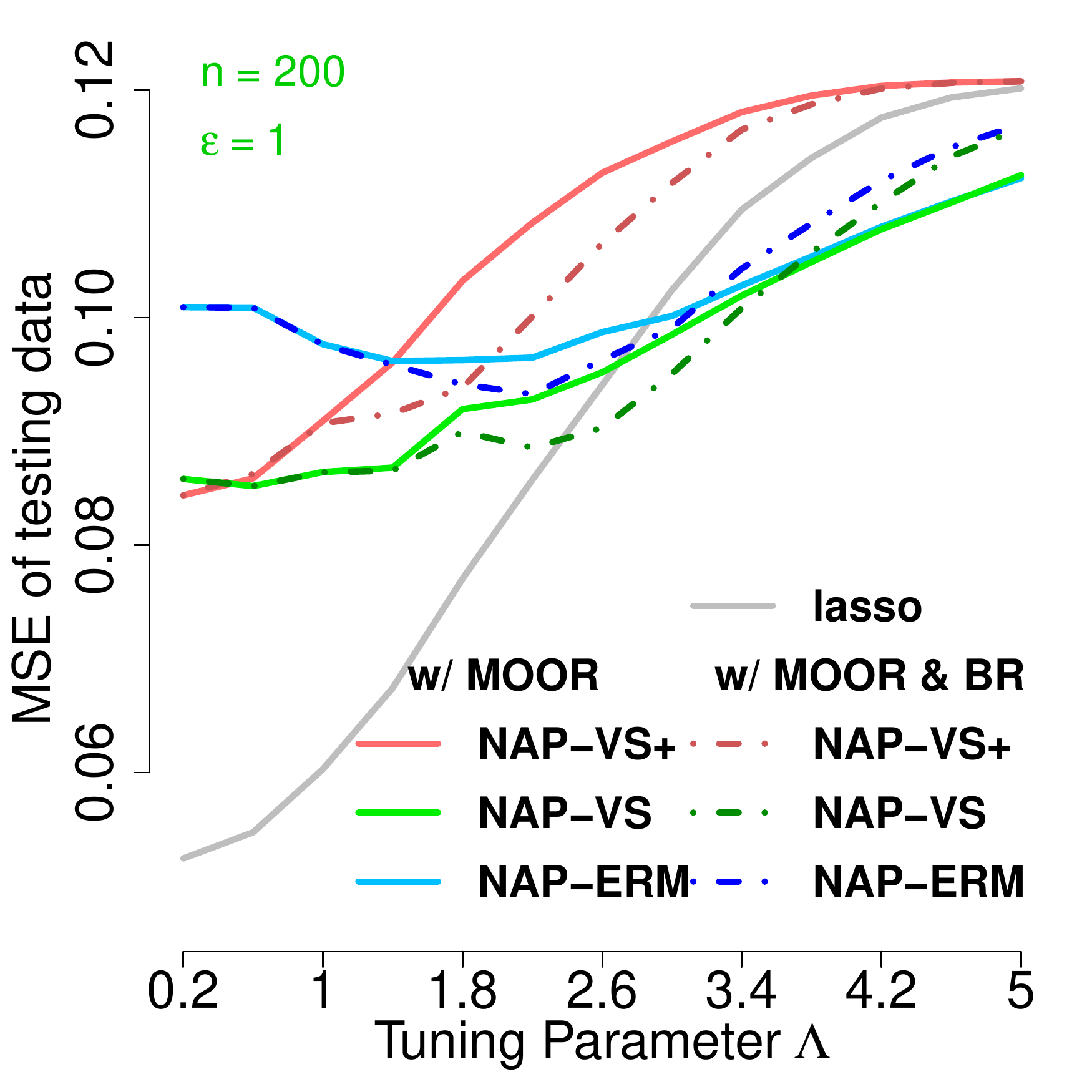}
\end{minipage}
\begin{minipage}{0.09\textwidth}\footnotesize $n=500$ \end{minipage}
\begin{minipage}{0.9\textwidth}
\includegraphics[width=0.24\linewidth, trim=4pt 9pt 15pt 18pt,clip]{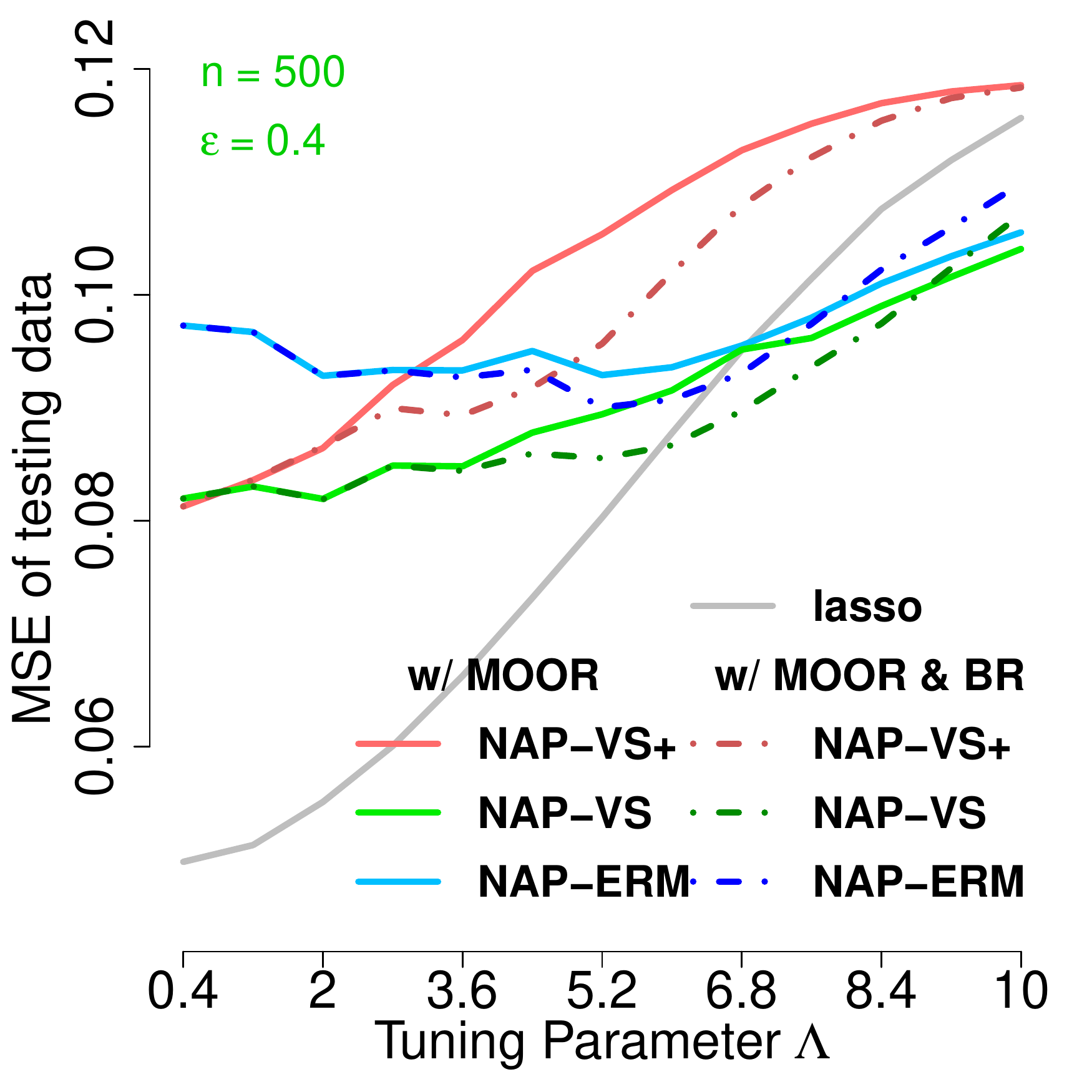}
\includegraphics[width=0.24\linewidth, trim=4pt 9pt 15pt 18pt,clip]{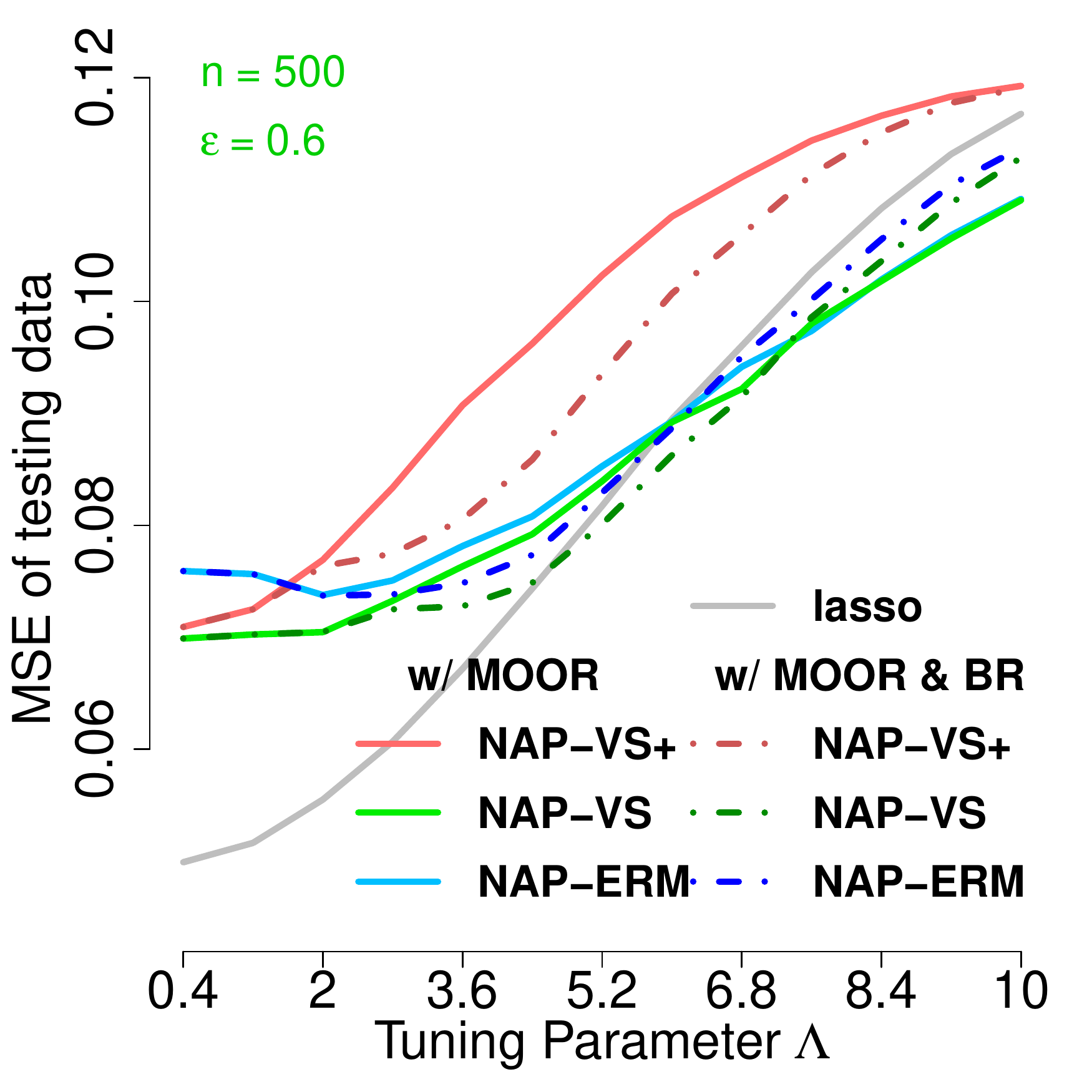}
\includegraphics[width=0.24\linewidth, trim=4pt 9pt 15pt 18pt,clip]{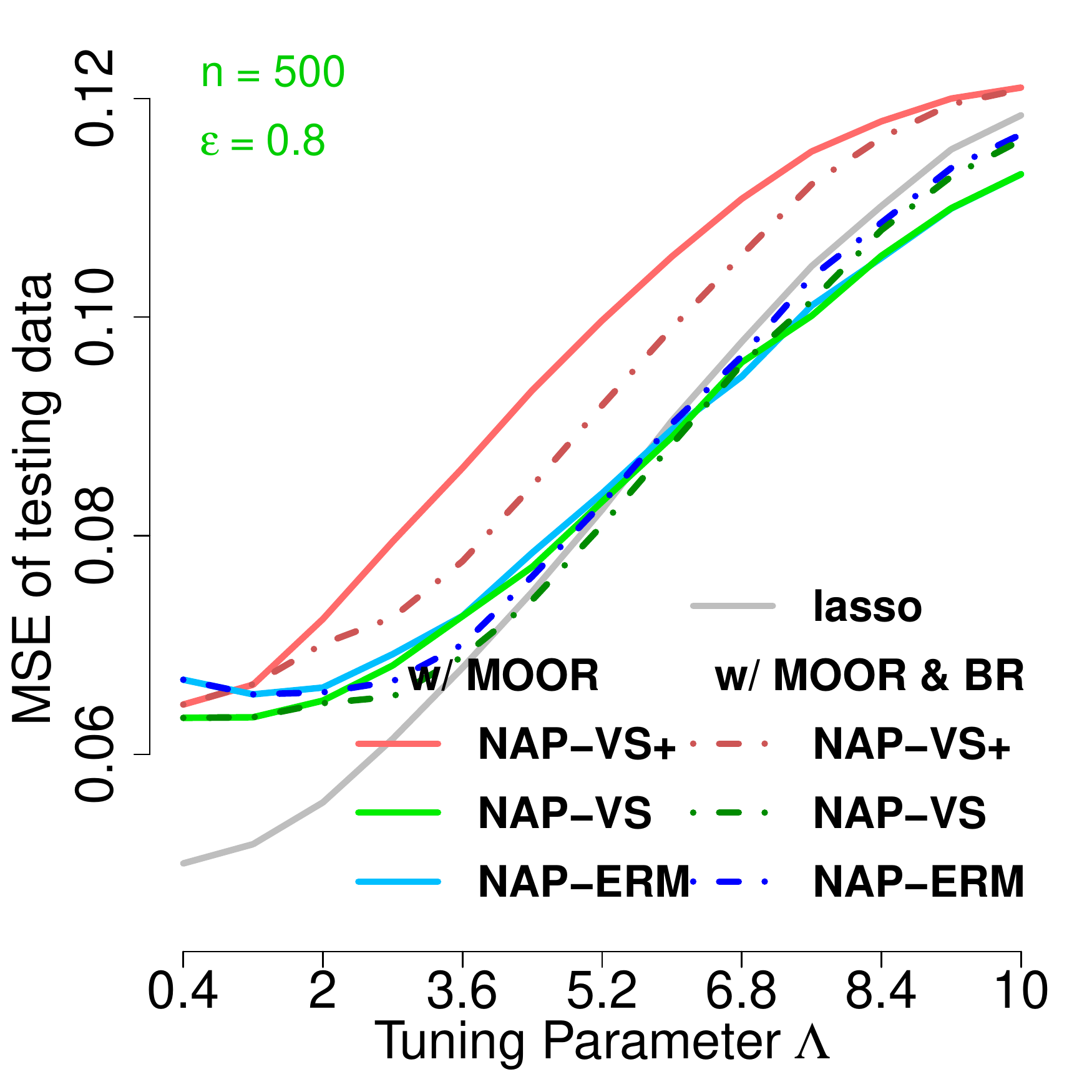}
\includegraphics[width=0.24\linewidth, trim=4pt 9pt 15pt 18pt,clip]{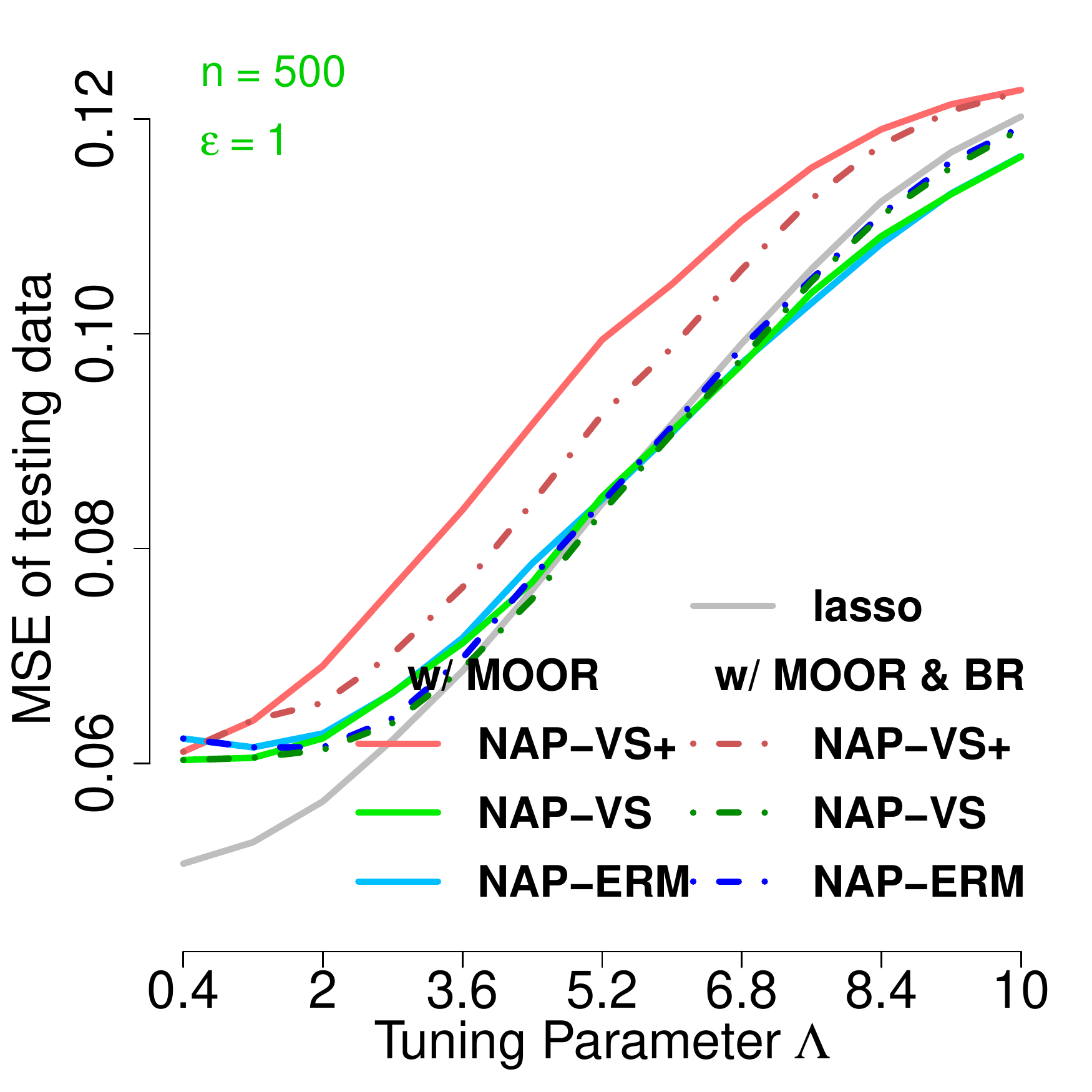}
\end{minipage} \vspace{-6pt}
\caption{Variable selection ROC curves and outcome prediction MSE in the testing data in linear regression with lasso via NAP with vs without recycled privacy budget (BR in the legends stands for Budget Recycling)} \label{fig:recyling1} \vspace{-12pt}
\end{figure}

\end{document}